\documentclass[12pt]{report}
\usepackage{upennstyle} 

\usepackage[square,sort&compress,comma,numbers,nonamebreak]{natbib}
\usepackage{bibentry}
\nobibliography*
\usepackage{graphicx}
\usepackage{bm}
\usepackage{soul}
\usepackage{amsmath}
\usepackage{amssymb}
\usepackage{amsfonts}
\usepackage{tabularx} 
\usepackage{multicol}
\usepackage{multirow}
\usepackage{siunitx}
\usepackage{enumitem}
\usepackage{amssymb} 
\usepackage{pifont}

\usepackage{multirow}
\usepackage{subcaption}
\usepackage{pdfpages}
\usepackage{float}

\usepackage{caption}

\usepackage{booktabs}
\usepackage[table,xcdraw]{xcolor}
\usepackage{longtable}
\usepackage{array}     
\usepackage{makecell}  

\usepackage{libertinus}

\usepackage{listings}

\definecolor{codegreen}{rgb}{0, 0, 1}
\definecolor{codegray}{rgb}{1, 1, 1}
\definecolor{codepurple}{rgb}{1, 1, 1}
\definecolor{backcolour}{rgb}{1, 1, 1}

\lstdefinestyle{mystyle}{
    backgroundcolor=\color{backcolour},   
    commentstyle=\color{codegreen},
    keywordstyle=\color{magenta},
    numberstyle=\tiny\color{codegray},
    stringstyle=\color{codepurple},
    basicstyle=\ttfamily\footnotesize,
    breakatwhitespace=false,         
    breaklines=true,                 
    captionpos=b,                    
    keepspaces=true,                 
    numbers=left,                    
    numbersep=5pt,                  
    showspaces=false,                
    showstringspaces=false,
    showtabs=false,                  
    tabsize=2
}

\lstdefinelanguage{profilelog}{
  basicstyle=\ttfamily\scriptsize,   
  breaklines=true,                   
  breakatwhitespace=true,            
  frame=tb,                          
  columns=fixed,                     
  keepspaces=true,                   
  showstringspaces=false             
}

\newcommand{\revision}[1]{\color{black}#1 \color{black}}
\newcommand{\bl}[1]{\boldsymbol{#1}}

\usepackage{etoolbox}
\AtBeginDocument{
  \renewcommand{\chapterautorefname}{Chapter}

}

\title{Real Time Local Wind Inference for Robust Autonomous Navigation}
\author{Spencer Folk}
\gradgroup{Mechanical Engineering and Applied Mechanics} 
\date{\revision{2026}} 
\cosupervisor{Mark Yim}
\cosupervisortitle{Asa Whitney Professor, Mechanical Engineering and Applied Mechanics}
\supervisor{Vijay Kumar}
\supervisortitle{Nemirovsky Family Dean, School of Engineering and Applied Science}
\gradchair{Jordan R. Raney}
\gradchairtitle{Associate Professor, Mechanical Engineering and Applied Mechanics}
\committee{M. Ani Hsieh, Associate Professor, Mechanical Engineering and Applied Mechanics}
\committee{Bruce Kothmann, Senior Lecturer, Mechanical Engineering and Applied Mechanics}
\committee{John Melton, Aerospace Engineer, NASA Ames Research Center}

\authorlegal{Spencer Ryan Folk} 

\dedication{\revision{To my dad, for investing in my education and fostering my scientific curiosity since day one.}} 

\acknowledgement{\revision{

This thesis simply would not have been successful without the support of so many family members, friends, colleagues, teachers, and mentors throughout the years. 
I would have never imagined arriving at this point if it were not for their help. 

I want to first thank my advisors, Mark Yim and Vijay Kumar, not just for their guidance, intuition, and expertise, but also for giving me the time and intellectual freedom to explore ideas I thought worth pursuing. 
Many of those ideas later turned out to be dead ends, but the lessons I learned along the way defined me. 
I am also grateful for my dissertation committee members Ani Hsieh and Bruce Kothmann, whose feedback consistently raised the bar, taking my work to new heights. 

John Melton, my committee member and mentor, deserves a special mention for bringing me under his wing at NASA.
Like many engineers, I always dreamed of one day working at NASA.
John was the one who made that dream a reality, and for that reason I am forever grateful. 
During my brief time at NASA I had the immense privilege of seeing some of the greatest landmarks in aerospace history, from the iconic desert views at Armstrong Flight Research Center to the formidable wind tunnels at Ames Research Center.  
Bearing witness to these places was inspirational, no doubt, but it was John's mentorship and his advocacy for the WindShaper facility that made me feel like I was playing an actual part in this history. 
Likewise I want to thank Ben Margolis for his feedback early on in my involvement with NASA. 
My weekly meetings with John and Ben helped sculpt the projects that would later define my thesis.
My experience at NASA taught me the value of government service and government research, especially during a time when federal employees were under such scrutiny. 
On that note, I am also grateful for the public funding we received in support of the work for this thesis, in particular the Convergent Aeronautics Solutions Project under NASA’s Transformative Aeronautics Concepts Program.

I want to acknowledge the contributions of other mentors that influenced me prior to my time at Penn. 
First, I would like to thank the mechanical engineering faculty at Lafayette College for giving me such a solid engineering foundation to build off of. 
I am especially grateful for Alex Brown, since he was the one who introduced me not just to robotics but also academic research.
Next I want to recognize John Gerdes from the US Army Research Laboratory.
I can confidently say I wouldn't be working with drones today if it weren't for John. 
During my multiple internships under his supervision, he exposed me to the magic of autonomous flight which evidently struck quite the chord with me. 
Ever since, all I've wanted to do is work on aerial robots.
As frustrating as they may be to work with sometimes, there is nothing quite like it when they work.

Next I want to thank my close colleagues at Penn, many of whom I am fortunate to consider lifelong friends. 
Between study sessions, happy hours, free lunches, and various other social events both in MEAM and GRASP, the transition to Penn and Philadelphia was seamless because of these communities. 
Special thanks go out to Greg Campbell, Jake Welde, and Parker LaMascus for their companionship especially during the early years of my PhD. 
I am also deeply grateful for the friendship of Kashish Garg, Fernando Cladera, and Jason Hughes who not only made my last year in the lab an absolute blast, but also genuinely made me a better software developer and engineer in almost every way with all the tools they shared with me. 
My Penn collaborators James Paulos, Anusha Srikanthan, Hersh Sanghvi, Thomas Zhang, and Anne Somalwar all taught me valuable lessons in collaboration that helped me mature as a researcher. 
Early discussions with James Svacha helped me narrow down on my research interests and craft my expertise. 
Lastly, I would not have made it to this point without the terrific support staff in MEAM and GRASP: Maryeileen Griffith, Peter Litt, Charity Payne, Jeremy Wang, Alex Zhou, and Britny Major. 

I would like to thank my family for their love and support.
My parents heavily supported my education and gave me the independence to pave my own path in life. 
My dad has always motivated me to aim higher and think critically with an engineering mindset. 
Meanwhile, my mom taught me how to live life a little less seriously and enjoy the little things in life.
Both encouraged me not to be afraid to break the status quo.
My older brothers, through their poking and prodding over the years, helped shape my resilience and sense of perspective. 
The never-ending \textit{so what?} questioning is what kept my research approachable and at least somewhat grounded in reality. 
Beyond my own family, I also want to thank the Shults family for their unending kindness and encouragement--from the very beginning they made me feel at home and one of their own. 

Finally, I thank my partner, Julia, for coming into my life during such a transformative time. 
We had our first date just five days after my PhD qualifying exam in $2020$. 
Our first few months together unfolded during the height of the COVID lockdown, but as it turns out, social distancing and multiple quarantines were not enough to keep us apart.
Since then, Julia has been a constant source of joy and adventure--our long walks around the neighborhood, exploring the Philadelphia restaurant scene, grabbing coffee drinks on the weekend, watching fireworks from our roof, not to mention our two-week road trip through the American West from Yellowstone to the Grand Canyon. 
All of these moments, and so many more, helped me endure even the toughest moments during the PhD, and for that she will forever have my heart. 
Our time in Philly has been unforgettable, and while I am sad to leave the city where we met, I know that with her by my side, anywhere we go will be home.

}} 

\abstract{
Urban air mobility and autonomous package delivery represent promising avenues for integrating aerial robots into everyday life.
However, operating these systems safely and efficiently in windy urban environments remains a major challenge due to the complexity of urban wind flow fields.
Existing methods for predicting and navigating wind fields rely on precise environmental knowledge, distributed sensor networks, or extensive exploration.
On the contrary, this thesis presents a solution that enables aerial robots to reason about surrounding wind flow fields in real time using on board sensors and embedded flight hardware.

The core novelty of this research is the fusion of range measurements with sparse in situ wind measurements to predict surrounding flow fields.
We aim to address two fundamental questions: first, the sufficiency of topographical data for accurate wind prediction in dense urban environments; and second, the utility of learned wind models for motion planning with an emphasis on energy efficiency and obstacle avoidance.
Drawing on tools from deep learning, fluid mechanics, and optimal control, we establish a framework for local wind prediction using navigational LiDAR, and then incorporate local wind model priors into a receding-horizon optimal controller to study how local wind knowledge affects energy use and \revision{robustness} during autonomous navigation.

Through simulated demonstrations in diverse urban wind scenarios we evaluate the predictive capabilities of the wind predictor, and quantify improvements to autonomous urban navigation in terms of crash rates and energy consumption when local wind information is integrated into the motion planning.
Sub-scale free flight experiments in an open-air wind tunnel demonstrate that these algorithms can run in real time on an embedded flight computer with sufficient bandwidth for stable control of a small aerial robot. 
Philosophically, this thesis contributes a new paradigm for localized wind inference and motion planning in unknown windy environments.
By enabling robots to rapidly assess local wind conditions without prior environmental knowledge, this research \revision{accelerates} the introduction of aerial robots into increasingly challenging environments.
} 

\begin{document}
\maketitle 
\setcounter{page}{2}

\makecopyright 

\makededication 

\makeacknowledgement 

\makeabstract
\tableofcontents

\clearpage \phantomsection \addcontentsline{toc}{chapter}{LIST OF TABLES} \begin{singlespacing} \listoftables \end{singlespacing}

\clearpage \phantomsection \addcontentsline{toc}{chapter}{LIST OF ILLUSTRATIONS} \begin{singlespacing} \listoffigures \end{singlespacing}


\begin{mainf} 

\chapter{Introduction}\label{ch:introduction}

Urban Air Mobility (UAM) has attracted considerable attention from investors, researchers, and the public in recent years. 
Fueled by advancements in electric propulsion, batteries, rotor noise reduction, air traffic control, and aerial autonomy, UAM promises to usher in a renaissance for commercial aviation. 
In many ways, the renaissance is already here--Fortune Business Insights estimates that the UAM market is expected to grow from \$3.01 billion in 2021 to \$8.91 billion in 2028 \cite{fortunebusinessinsights2020uam}; air taxi and drone delivery companies are rapidly approaching deployment; government agencies such as the Federal Aviation Administration (FAA) \cite{faa2023conceptofoperations} and National Aeronautics and Space Administration \revision{(NASA)} \cite{goyal2019uamstudy} are preparing \revision{regulation strategies} for large scale deployments of autonomous air vehicles in national urban airspaces.

Before visions of flying cars and autonomous aerial package delivery can become reality, there are still substantial hurdles that must be overcome \revision{before} widespread adoption in the cities of tomorrow.
\revision{One of the primary barriers is the uncertainty imposed on aircraft in dense urban environments by the} weather \cite{bauranov2021airspace, sharma2023quantifying}. 
Wind in the so-called urban canopy layer is particularly challenging to predict, owing to the high density of heterogeneous structures that divert and accelerate the air mass between buildings, under bridges, and through pockets of vegetation. 
For instance, clusters of buildings can create channels that strongly accelerate local flows \revision{leading to large spatial gradients increasing} both the costs and risks \revision{associated with} aerial operations.
\revision{As cities evolve and skylines are redefined by taller, more complex high-rises, so too does the behavior of the wind.}
\revision{More so, urban winds can change on much faster timescales; for example, a calm morning may transition into a gusty afternoon due to a newly formed storm front.}
\revision{Because urban winds are such dynamic processes, autonomous aerial systems will need access to dense real-time wind predictions so that their routing can adapt to the changing winds on the fly.}

\revision{Taking inspiration from the dynamic soaring literature, this thesis} takes the perspective that \revision{while wind is a serious yet unavoidable threat to autonomous operations, it also} presents plenty of opportunities for energy savings.
Since the late 19\revision{\textsuperscript{th}} century with Lord Rayleigh's \textit{The Soaring of Birds} \cite{rayleigh1883dynamicsoaring}, it has been well known that birds have somehow evolved to identify and exploit wind gradients near mountain ridges and over oceans to maximize flight endurance. 
In \autoref{sec:introduction:background:planning} to follow, we review research that has demonstrated similar capabilities for both fixed- and rotary-wing uncrewed aerial vehicles (UAVs), theoretically and experimentally showing that UAVs can benefit significantly in terms of both \revision{robustness} and \revision{energy} efficiency.
However, a primary concern expressed in these works is the level of assumed knowledge about the wind. 
Approaches thus far to obtaining this information have their limitations: some place strict assumptions on the underlying model of the fluid flow field, limiting their generalizability in novel settings; others assume no model instead opting to employ costly exploration of the environment; still others require a known, static map of the environment to make predictions based on pre-computed simulations.

It is these shortcomings, \revision{at the intersection of} wind prediction and wind-aware motion planning, that motivate the work presented in this thesis\revision{--to} augment existing autonomous UAVs with heightened situational awareness of complex wind flow fields while requiring very little \textit{a priori} knowledge about the operating environment.
\revision{More concretely, the} objective is to develop a framework for predicting the wind from on board \revision{sensors,} essentially gifting the robot with the ability to ``see'' the \revision{wind,} without excessive computational burden. 
\revision{This new capability is then incorporated} into \revision{an} autonomous navigation stack to pave a path towards \textit{proactive} flight in unknown cluttered and windy environments. 
\revision{Along the way, this thesis takes a deep dive into the mechanics and challenges associated with flight in the urban canopy layer.}

\section{Background and Related Work}\label{sec:introduction:background}

This thesis bridges \revision{numerous domains} ranging from fluid mechanics and aerodynamics to deep learning and optimal control.
\revision{In} the context of autonomous navigation in windy environments, the most important \revision{context to consider is:} 1) \textit{how do we understand and model complex urban winds?} 2) \textit{how does wind affect the dynamics and aerodynamics of UAVs?} and 3) \textit{to what extent have we already demonstrated wind-aware navigation?}

\subsection{Urban Wind Flow Fields}\label{sec:introduction:background:urbanwinds}

\revision{Urban winds are forged in the intricate fluid dynamics interactions that arise as the Earth's atmospheric boundary layer (ABL)} is forced between and around buildings, bridges, vegetation, and \revision{other urban infrastructure.}
\revision{These interactions result in complex spatio-temporal flow features within the urban canopy that present a serious concern for any and all potential urban air mobility operations.}
\revision{As Mikhailuta \textit{et al.} \cite{mikhailuta2017urbanwindsstudy} describes,} there are broadly three classes of tools that scientists use to understand how urbanization \revision{affects the wind:} physical models in wind tunnels, mathematical models (in other words, simulations), and field measurements. 

Wind tunnel studies, e.g. \cite{zhao2022windtunnelreview, frey2024windtunnelurbanwind}, are timeless experiments that provide privileged views into the \revision{elaborate} patterns that form as a fluid is redirected through a maze of human-made structures. 
Sub-scale models, often in the $1$:$100$ range of length scales, are immersed in a recreation of the atmospheric boundary layer under steady state inlet conditions.
As described by Plate \cite{plate1999windphysicalmodels}, these studies are often interested in wind loading on existing or new buildings, or the mass transport of gases such as car exhaust, with model fidelity varying from simple wooden blocks to detailed 3D printed structures.
A primary focus in these studies is accurately recreating the velocity and turbulence intensity profiles of the atmospheric boundary layer\footnote{Atmospheric boundary layer profiles are indeed their domain of study, which \revision{should be} understood \revision{as having their own uncertainties associated with} physical modeling, simulations, and field measurements.}, \revision{for example} the studies presented by Plate \cite{plate1999windphysicalmodels} and Wang \textit{et al.} \cite{wang1996windtunnelscaling}, which use surface roughness elements and turbulence generators to mimic the atmosphere.
In physical studies like these, flow measurements are often obtained via hot-wire probes \revision{which provide somewhat unobstructed high frequency point-wise measurements.} 
Alternatively, the wind can be visualized with particle velocimetry methods, or even simpler particle erosion techniques that provide qualitative visualization of the relative winds and shear layers. 
In many cases, however, both techniques are used \revision{simultaneously} to better understand the physics of fluid flow around urban infrastructure.

With increasing access to cheap compute \revision{and cloud-based high performance computing clusters as a service,} computational fluid dynamics (CFD) is \revision{now one of the most popular approaches to} urban wind studies. 
We refer the reader to Toparlar \textit{et al.} \cite{toparlar2017cfdreview} and especially Blocken \textit{et al.} \cite{blocken2014fiftyyears} for \revision{comprehensive} reviews of these methods.
CFD offers a more accessible alternative to physical wind tunnel studies in that it does not require the development and maintenance of expensive scientific equipment and instrumentation--simulations can be designed, carried out, and studied in a fraction of the time and cost.
Further, one might argue CFD provides more insight into wind flow fields, \revision{provided their results can be validated for accuracy,} because the simulations can be saved, replayed, probed again, and quantities such as velocity and pressure can be visualized with \revision{extremely fine} spatial and temporal resolutions for deeper analysis.  
The current state of the art \revision{algorithm} for urban CFD \revision{is} Large Eddy \revision{Simulation} (LES) \cite{murakami1992les} which can model fine-scale turbulence without the computational intractability \revision{that plagues} direct numerical simulations.  
As Plate points out, even the current state of the art in CFD falls victim to model simplifications of the flow, most often in the definition of turbulence closure models.
Also, solutions are highly influenced by the choice of \revision{other} boundary conditions, such as the inlet profile and surface conditions. 
\revision{It is important to approach urban wind CFD studies with extreme caution, as these modeling choices can influence the results and lead to erroneous conclusions.}

Field measurements provide the most accurate data for scientists to understand urban flow phenomena simply because they are direct in situ observations of the \revision{full-scale} fluid flows--no fussing with models or boundary condition tuning, just pure measurements. 
Sensors such as scientific weathervanes, ultrasonic anemometers, and even more recently wind Light Detection and Ranging sensors (LiDARs) offer different ways to probe and visualize urban wind flow fields. 
Works like \revision{Gadian \textit{et al.} \cite{gadian2004directionalpersistence} and Lo Brano \textit{et al.} \cite{lobrano2011fieldmodeling}} use time series wind data from weather stations to build statistical models of the wind flow through urban environments.
Taking this idea to the extreme, Mikhailuta \textit{et al.} \cite{mikhailuta2017urbanwindsstudy} presents analyses on $15$ years of data collected from urban monitoring stations. 
\revision{Historical data on such large scales can be analyze to discover trends in urban flow fields for particular regions.}
However, because \revision{cities evolve over time and these measurements are sparse in nature, the} claims and insight from these studies are \revision{inherently} limited to the geographical location and time period with which they were recorded.
\revision{Ultimately,} in situ measurements only give a partial view of the underlying flow field.

\revision{While wind tunnel experiments, CFD, and field measurements provide invaluable scientific insights into the fundamental mechanics of urban aerodynamics, they possess inherent limitations when translated to an engineering domain like robotics or air traffic control.} 
\revision{The primary limitation of the aforementioned methods is the time it takes to generate predictions and insight.}
\revision{Physical modeling requires labor-intensive fabrication, CFD demands significant computational overhead that often precludes real-time execution, and field measurements are spatially sparse and tied to fixed infrastructure.}
\revision{Consequently, these approaches serve as excellent tools for retroactive analysis or long-term urban planning, but they are ill-suited for the dynamic on board requirements of a robotic system.}

\revision{For an autonomous UAV to robustly navigate the urban canopy, it cannot wait for CFD to converge to a solution or rely on a distant weather station's historical data.}
\revision{Instead, there is a critical need for methods that can synthesize these high-fidelity insights into lightweight, predictive models capable of running in real time.}
\revision{The following sections explore how the robotics community has sought to bridge this gap, moving from exhaustive atmospheric modeling toward agile, on board wind estimation and localized flow prediction.}

\subsection{Center-of-Mass (CoM) Wind Estimation in Robotics}\label{sec:introduction:background:comestimation}

Being able to measure the wind velocity at the UAV's immediate location can provide unbiased, albeit sparse, in situ samples of the local wind flow field \revision{much like the weather stations used in field campaigns, but this time without being constrained to a single location.}
Broadly speaking, approaches to wind estimation can be categorized as either \textit{direct} or \textit{indirect} \cite{tomic2022modelbasedwindestimation}. 
Direct wind sensing techniques employ dedicated sensors to measure the wind, traditionally using pitot-static tubes, ultrasonic sensors, and hot-wire anemometers \cite{abichandani2020sensingreview}, but also more recently bio-inspired whiskers \cite{kim2020whiskerprecursor, tagliabue2020touch, thomas2025whisker} and novel lightweight MEMS sensors \cite{simon2023flowdrone}. 
The primary challenge, especially for rotary-wing UAVs, is placing the sensor(s) far enough away from the rotors' region of influence so as to not corrupt the signal with flow induced by the rotors.
For \revision{context}, Prudden \textit{et al.} \cite{prudden2016flyinganemometer} determined that the region of influence of their rotors was about two and a half times the rotor diameter, implying that a dedicated sensor would need to be placed at least that far away to avoid coupling with the rotors. 
Works like \cite{tagliabue2020touch} take a different approach--accepting that the rotors influence their whisker measurements, they attempt to model these affects using various techniques including deep learning, with the intent of filtering out these effects in post processing.

Nonetheless, direct wind sensing requires sensor infrastructure to work, which may be prohibitively heavy on particularly size-, weight-, and power-constrained UAVs. 
The literature has bypassed this issue by engineering solutions that get rid of the dedicated sensor entirely.
This is known as \textit{indirect} wind sensing, and in this class of algorithms the local wind vector is inferred by fusing inertial and odometry measurements--information already required for navigation and control--with an appropriate model of the aerodynamics that describes how the wind affects these signals.
We direct the reader to \cite{tomic2022modelbasedwindestimation, chao2021windestimationsurvey} for more detailed surveys of these techniques, but in the vast majority of cases indirect wind sensing algorithms manifest as specialized forms of state estimation algorithms (e.g. Kalman filters). 
\revision{The biggest challenge with indirect wind estimation is accurately modeling the aerodynamic forces which can be quite complex for rotary-wing UAVs.
These modeling errors often contribute the largest bias in the resulting predictions.}

The single \revision{most important} limitation of CoM wind estimation can be spotted in the name: these techniques only provide a means of estimating the wind at the UAV's \textit{current} location.
These estimates in isolation help with short term feedback \revision{tracking} control, e.g. \revision{\cite{svacha2017improving, simon2023flowdrone},} but provide very little information about the spatial and temporal dynamics of the wind around the UAV--aspects which we will see are \revision{critical} to autonomous navigation and higher level decision-making.

\subsection{\revision{Flow} Field Prediction and Estimation \revision{for Navigation}}\label{sec:background:fieldestimation}

Building in complexity, we now consider existing methods for predicting \revision{flow \textit{fields}.}
\revision{The traditional} CFD and wind tunnel methods described in \autoref{sec:introduction:background:urbanwinds} require too many resources \revision{(time, energy, compute, privileged knowledge, or all of the above)} to be applied on scales that are reasonable for robotics, because robots have limited time to think and react to their environment and their energy is a precious resource.
\revision{To that end, there are numerous other} approaches \revision{from different communities in the literature that focus on} \revision{modeling flow fields} \revision{in a} more computationally tractable \revision{way.}

\revision{Methods} like super-resolution \cite{oettershagen2019realtimewindprediction, fukami2023super} offer ways to \revision{quickly} up sample coarse numerical weather simulations \revision{that could in thory be made} available to robots \revision{connected to cloud computing infrastructure.}
Although super-resolution has \revision{so far} been demonstrated \revision{in} larger scale mountain regions, \revision{its} application to urban wind flow fields \revision{is questionable because the flow features occur on much smaller length scales.
This remains an active area of research for super-resolution.}
Alternatively, a large number of CFD simulations \revision{computed on particular urban domains} with varying boundary conditions \revision{can constitute} a database of labeled wind simulations. 
\revision{This type of database has been} used directly \revision{in the literature} as a lookup table \cite{gianfelice2022lookup}.
\revision{CFD databases have also been} encoded in reduced order \revision{models} using methods like proper orthogonal decomposition \cite{ebert2023gappy} \revision{and} Gaussian Process Regression \cite{patrikar2020particle}.
In situ measurements, either from a few strategically-located sensors or from the robot itself (\autoref{sec:introduction:background:comestimation}) \revision{can} be used as inputs to the reduced-order \revision{models} to recover the larger flow field.

In the aforementioned works, the \revision{numerical weather simulations and} CFD databases were \revision{specifically} tailored to a specific city or map, which may be ineffective in situations where the simulations become outdated. 
\revision{While urban expansion gradually alters city skylines, natural disasters can accelerate these structural changes instantaneously, rendering existing CFD databases obsolete.} 
\revision{In this light, an appropriate urban flow field modeling technique needs to evolve with cities and it is questionable whether methods like CFD can adapt fast enough.}
\revision{Fortunately there are more} generalized and versatile \revision{approaches} to \revision{flow field modeling that leverage} nonparametric techniques.
For instance, \revision{Gaussian Process Regression (GPR) has been used to model spatio-temporal flow fields from on board measurements in oceanic \cite{hansen2018gaussianprocessesoceans, lee2019gaussianprocessoceans, goncalves2019gaussianprocessoceans, berlinghieri2023gaussianprocessoceans} and atmospheric \cite{lawrance2010simultaneous, lawrance2011soaring} contexts.}
\revision{Similarly, adaptive sampling methods using transfer operators (in particular the Koopman operator \cite{koopman1931koopmantheory}) approach the problem of estimating complex time-varying flow fields by identifying modes \cite{mezic2013koopmanforfluidflows} or coherent sets \cite{salam2022learning, salam2023l4dc, li2024enkode} in the flow.
By using the trajectories of Lagrangian particles (e.g. a robot navigating the environment) as measurements, these methods can discover the underlying linear operators that govern the evolution of the flow. 
This is particularly advantageous for time-varying predictions, as the Koopman operator allows for the forward-propagation of the flow state in a computationally efficient, linear manner while maintaining the physical interpretability of the lifted state \cite{li2024enkode}.}

Although these \revision{nonparametric} models \revision{are relatively fast, flexible, can be learned using online data collection, and provide predictions for regions not yet visited,} they still have associated structure and inductive biases that limit their generalizability \revision{particularly in} urban contexts.
An interesting perspective on this facet comes from Ware (\cite{ware2016thesis}, page 22), in which he describes why kernel-based methods in particular might not be effective for urban winds. 
Ware argues that GPR is ill-conditioned in environments with extremely sparse measurements, implying that these techniques may require large networks of sensors, or significant excursions in direct opposition to a robot's primary objective (e.g., \revision{package delivery)} that may put the robot in unforeseen danger.
Further, Ware explains that the typical kernel functions used \revision{with} these modeling techniques come with spatial smoothness and correlation assumptions, which are often violated by the sharp discontinuities in the wind flow field caused by urban infrastructure.\footnote{As an illustrative example, Ware explains that GPR assumes \revision{a certain} correlation between two points in space, but the winds on the windward and leeward sides of buildings may rarely be correlated at all.}

More recently, methods leveraging deep learning have been developed to distill large amounts of data from many CFD simulations on arbitrary maps to produce fast and generalizable \revision{flow field} prediction.
For instance, Achermann \textit{et al.} \cite{achermann2024windseer} trained a deep convolutional neural network to predict 3D wind over randomly-generated mountain ridges using sparse wind measurements.
In urban contexts, researchers have deployed generative adversarial networks \cite{kastner2023gansurrogate} and graph neural networks for predicting wind flow through both single maps \cite{gao2023GNNwindpredictor} and random city layouts \cite{shao2023pignncfd}. 
In contrast to the previously mentioned kernel-based methods, there are no explicit assumptions about the spatial smoothness or correlation, however deep neural networks are still subject to inductive biases, among many other issues, that may affect their performance in urban settings. 
Existing works that utilize deep learning to predict wind require privileged information about the environment, such as an up-to-date top-down view of the city or a topographic map, which as already discussed may be infeasible in robotics contexts.
Nevertheless, deep learning offers a brute force way to distill meaningful relationships from a large database of simulations.
\revision{To address the black box nature of the deep learning approaches above, hybrid techniques under the domain of Physics-Informed Machine Learning (PIML) \cite{raissi2019pinns} have emerged as methods to improve interpretability and sample efficiency at the cost of additional bias.}
\revision{Besides having the capacity to learn very complex flow features characteristic of urban environments, the} primary benefit of deep learning methods is that the \revision{\textit{trained}} model is very fast to evaluate at run time, which lends itself to real time implementation on a robotic platform especially given the wide availability of dedicated GPU compute. 

\subsection{Rotary-wing UAV Energy Models}\label{sec:introduction:background:energymodels}

In this thesis, we are primarily interested in finding energy efficient trajectories in windy urban environments.
It is therefore a prerequisite that we have reasonable models for energy consumption in the pursuit of a definition for ``energy optimality''. 
While there is a wide breadth of energy modeling techniques for fixed-wing UAVs, this work is most concerned with the rotary-wing counterparts. 

Probably the simplest approach to approximating energy consumption is to assume that the propulsive energy expended is proportional to the thrust force multiplied by the distance covered \cite{rienecker2023planning}, and we can approximate the thrust force as being equal to the drag force in steady level flight. 
Therefore the energy cost between two positions in space can be approximated, assuming an appropriate drag model and knowledge of the airspeed and ground speed of the vehicle.
Another slightly more complex approach uses traditional modeling techniques from the rotorcraft literature (\cite{leishman2006principles, johnson2012helicopter} for example) for a slightly better approximation of energy usage in forward flight. 
An example of such model is presented by Davoudi \textit{et al.} \cite{davoudi2020quadrotorflightsimulation}. 
Assuming steady state forward flight conditions, a rotor's power consumption can be thought of as the sum of power required to overcome different drag forces--lift-induced drag, viscous drag on each individual rotor, and parasitic drag on the UAV frame. 
Depending on the relative impact of each of the drag-like effects just listed, power consumption can vary non-monotonically with airspeed, owing to the idea of an optimal speed to fly at to minimize energy consumption. 
This model can be expanded to include empirical models of other maneuvers such as acceleration, deceleration, and turning \cite{shivgan2020geneticpathplanning, ding2018morepowermodels}, or even consider the effects of payloads \cite{tagliabue2019modelfreeenergy}.
Energetics for small scale rotary-wing UAVs is reviewed by Karydis \textit{et al.} \textit{et al.} \cite{karydis2017energetics}, where besides the aforementioned models they also mention predicting energy consumption strictly as functions of the rotor speeds. 
However, predicting what the rotor speeds will be in the presence of wind is nontrivial since the aerodynamic loads on the rotor vary dramatically.

\subsection{UAV Motion Planning with Wind}\label{sec:introduction:background:planning}

Langelaan \textit{et al.} produced a series of landmark works demonstrating that fixed-wing UAVs could extract energy from wind flow fields much like birds \cite{langelaan2009gust, langelaan2011prediction}. 
This energy extraction principle, known as dynamic soaring, can be used to improve endurance for fixed-wing UAVs considerably and has been studied thoroughly \cite{mir2018dynamicsoaring}. 
Approaches to this problem typically use kinematic trees (as in Langelaan \textit{et al.}) or RRT-like sampling-based planners such as Lawrance and Sukkarieh \cite{lawrance2011soaring}, and can leverage simplified models due to the fact that planning can occur over distances on the order of hundreds of meters. 
\revision{These seminal works represent some of the earliest attempts at autonomous wind-aware navigation to improve flight endurance which exclusively occurred in mountainous regions.}

In the context of autonomous \textit{urban} navigation, \revision{the length scales are much smaller and typical dynamic soaring features like thermals or ridge lifts are not present.}
\revision{Instead, robustness and situational awareness are often key motivators} and a crucial aspect \revision{to improving these factors} is \revision{the identification of} hazardous wind conditions. 
For instance, Orr \textit{et al.} \cite{orr2005fixedwingmodelinurbanenv} and Galway \textit{et al.} \cite{galway2011control} develop \revision{urban} flight simulations of fixed- and rotary-wing UAVs, respectively, with the purpose of understanding the affects of urban wind flow fields on waypoint tracking. 
Galway \textit{et al.} claim that assuming a constant wind flow field is not sufficient in predicting the flight path of a rotary-wing UAV in the presence of buildings, and that there must be a way to identify dangerous and costly regions in the flow field that should be avoided.
\revision{These flight simulations help quantify the degradation of control authority in the presence of these unique flight hazards.}

Ware and Roy \cite{ware2016canopy} developed a graph-based planning framework that optimized UAV trajectories taking into consideration hazardous aspects of the wind field, mostly in terms of avoiding energy-draining hazards like strong wind tunnel effects manifesting in between buildings.
Other graph search-based approaches include \cite{kularatne2016time, baskar2020planning, ebert2023gappy,  takemura2023energyperceptionawareplanning, chan2023graphwindawarenavigation, aiello2022realtimeaware, chakrabarty2010graphsearch} where the algorithmic structure and focus on energy remains the same, but the specific approaches to modeling energy consumption differ. 
For instance, \revision{Kularatne \textit{et al.} \cite{kularatne2016time}} constructs an energy cost based on infinitesimal displacements, \revision{whereas}  Rienecker \textit{et al.} \cite{rienecker2023windgraphplanning} instead \revision{computes} the energy consumption by integrating a drag force along the route. 
While these methods are capable of finding globally optimal paths,\footnote{Global optimality in this context is subject to the resolution of their search space. For instance, if the spatial resolution of the graph is too coarse, small yet opportune shortcuts in the environment may be missed.} in practice \revision{they rely} on privileged \textit{a priori} information regarding both the map and the wind field.
This limitation affects their viability in dynamic and partially known environments. 

Another class of approaches to wind-aware motion planning uses \revision{classical} kinematic tree \cite{chakrabarty2013kinematictree} or sampling-based motion planning algorithms \cite{lawrance2009planner, lawrance2010simultaneous, hao2022bamdp} which are similar to the graph search-based planners in that they use kinematic representations of the UAV dynamics. 
However, real time implementation in dynamic winds is much more feasible with these online approaches because the motions are selected from a predefined set of discretized actions on a receding horizon, and optimal trajectories can be selected based on current estimates of the wind flow field which can be updated with in situ measurements as in \cite{lawrance2010simultaneous}. 
\revision{In fact, these works provide strong evidence towards the notion that decomposing the motion planning problem onto a receding horizon is a viable approach towards achieving real time capability.}

More recently, there have been examples of energy-aware trajectory optimization \cite{salzmann2024learning, yacef2020energytrajopt} and reinforcement learning (RL) \cite{biferale2019zermelo, liu2024teevtol, banerjee2024rlenergy} within complex flow environments. 
In the former, the flow field is incorporated into a nonlinear optimization program and solved using an off-the-shelf solver. 
In the latter, RL agents learn point-to-point navigation in complex flows while optimizing time and energy.
Both instances rely on global knowledge of the flow environment, but they are closer towards the idea of operating on continuous state and action spaces. 

The literature provides substantial evidence towards the claim that model priors for the wind field can provide vital information to on board UAV path planning algorithms, helping to create routes between desired trajectory waypoints that 1) minimize the collision risk with objects in the sensed environments; 2) avoid expected regions of large wind gradients or excessive wind accelerations; and 3) avoid regions of high uncertainty or variability, such as those caused by unsteady turbulence in building wakes.

\subsection{Free Flight UAV Wind Tunnel Studies}

A \revision{crucial} aspect of this thesis is the evaluation of algorithms using free flight testing of a UAV in a wind tunnel \revision{with the purpose of validating the proposed algorithms with respect to their real-time application.} 
Traditionally, UAVs have been tested in wind tunnels using stings--testing fixtures that are used to mount the UAV statically at various orientations. 
This serves two purposes: 1) stings are by definition instrumented with force sensors to measure the aerodynamic forces on the model; and 2) the rigid mounting provides an additional layer of safety in order to protect the wind tunnel equipment. 
However, hard mounting the UAV adds confounding variables because the sting disrupts not just the freestream velocity (although this is typically explicitly considered in the sting's design and calibration), but also potentially the rotor wakes, thus affecting the net aerodynamic loads on the UAV. 
For \revision{this reason,} free flight testing becomes a desirable testing configuration that can complement the sting-mounted testing, although safety requirements for most wind tunnel facilities preclude the adoption of this method. 

Nevertheless, there have been instances in the \revision{past} where free flight testing has been demonstrated \revision{using} small UAVs. 
\revision{In fact, as Chambers \cite{chambers2010modeling} discusses in his book \textit{Modeling Flight: The Role of Dynamically Scaled Free-Flight Models in Support of NASA's Aerospace Programs} \cite{chambers2010modeling}, free flight testing played an important yet often untold role throughout the 20\textsuperscript{th} century during the development of many historic aircraft.}
\revision{Otherwise in the literature, free flight} experiments have been used \revision{somewhat sparingly} to validate a combination of wind estimation algorithms \cite{mohamed2016freeflight, tomic2022modelbasedwindestimation, kristner2024windtunnel}, power consumption models \cite{ware2016canopy}, or flight dynamics models and controllers \cite{bannwarth2018freeflight, kubo2018freeflight, foster2020freeflightuav}.
\revision{Nowadays, free} flight testing is \revision{trending} because of a series of new open-air wind tunnels developed by \revision{the Swiss company} WindShape.\footnote{\url{https://www.windshape.com/}}
\revision{Their flagship product,} called the WindShaper, was first introduced to the academic community \revision{in a paper by} Noca \textit{et al.} \cite{noca2019windshapeintro}.
Since then it has been used in numerous studies ranging from UAV aeroacoustics \cite{putzu2020aeroacoustic}, power modeling \cite{olejnik2022powerconsumption}, atmospheric turbulence simulations \cite{walpen2023windshapereplication}, and even propulsion performance under icing conditions \cite{catry2021windshapeicing}. 
The absence of walls introduces additional complexities to the flow as it is able to expand into the surrounding air, but this can be addressed through closed loop inlet tuning \cite{walpen2024windshapeautomation}. 
Because the WindShaper is a self-contained system, it can be placed in any room, including motion capture spaces, and alleviate much of the safety considerations that previously inhibited \revision{substantive} free flight testing of UAVs in more traditional wind tunnel facilities.

\section{Outline and Contributions}\label{sec:introduction:contributions}

\revision{The literature discussed here spans multiple key domains, from fluid dynamics to state estimation and motion planning, and they provide an invaluable set of tools to tackle the problem of real-time wind prediction and robust autonomous navigation within the urban canopy.}

The chapters to follow will explore three key themes--wind prediction, wind-aware navigation, and free flight testing--all in the context of  autonomous aerial operations in urban environments. 
Starting with \autoref{ch:dynamics}, we provide a primer on the fundamental models that establish the scope and fidelity of our approaches in subsequent chapters.
\autoref{ch:dynamics} includes thorough descriptions of the rigid body dynamics, aerodynamics, power consumption, and wind models used throughout the thesis \revision{which also helps inform the limitations of the current study.} 
\autoref{ch:filtering} and \autoref{ch:prediction} discuss ways for UAVs to estimate the wind using on board \revision{sensing, beginning with the derivation of a model-based indirect wind estimator and subsequently expanding on the idea with a novel deep-learning-based wind flow field prediction algorithm.} 
Then, \autoref{ch:planning} addresses a hallmark question in engineering--\textit{why is this useful?}--by incorporating the aforementioned wind prediction methods into a motion planning algorithm \revision{resulting in a novel wind-aware autonomous navigation stack.}
All of these concepts are \revision{brought to bear on} real hardware \revision{with actual embedded compute resources} through \revision{novel} sub-scale testing discussed in \autoref{ch:realworld}. 
This thesis ends with \autoref{ch:conclusion}, which has concluding remarks on the \revision{contributions and} lessons learned, limitations of the present study, and \revision{new opportunities} for future research.

\subsection{Local Wind Predictions}

The literature described \revision{in this chapter} provides a strong foundation for wind estimation using UAVs and on board sensing. 
Researchers have developed ways not just to measure the wind with dedicated sensors mounted on the UAV, but also to estimate the wind in a statistically rigorous way using sensing and information already in use for navigation and control, thus unlocking wind estimation even for the smallest of UAVs. 
However, these methods only provide information about the wind at the UAV's immediate location, which is helpful for feedback control but doesn't lend itself to more proactive motion planning. 
To that end, the literature has also thoroughly investigated various ways to predict flow fields, ranging from instrumented wind tunnel testing and computational fluid dynamics simulations to model-free deep neural network architectures. 
The primary limitation of these works is that to date they rely on precise maps of the \revision{environment--this might} prove disadvantageous in cases where the map is outdated, low-resolution, or outright unavailable. 

In this thesis, we bridge the gap between the point-mass wind estimation and learning-based wind flow field prediction strategies to enable truly \textit{proactive} wind prediction while also limiting the amount of \textit{a priori} or privileged information required. 
Starting with \autoref{ch:filtering} we derive a model-based wind estimator for small UAVs which provides \revision{real-time} in situ measurements of the wind flow field without dedicated anemometry.
Then, in \autoref{ch:prediction}, we explore a novel set of technologies and methods for inferring the wind flow field in a region around the UAV by fusing range sensors from a navigational LiDAR with sparse measurements of the wind provided by the wind estimator, allowing UAVs to ``imagine'' the surrounding wind from information available online.
\revision{In contrast to the current state of the art, our methods have minimal computational requirements and require no environmental priors besides what can be immediately measured using on board sensors.}

\subsection{Wind-Aware Autonomous Navigation}

In the literature, wind flow field predictions have been explored thoroughly in the context of motion planning. 
Many approaches have demonstrated that incorporating wind information into decision making can lead to \revision{more robust and} energy-efficient flight even in very complex urban settings.  
\revision{However, one limitation common among many of the works in this domain is the computational requirements--very few wind-aware motion planning algorithms have been demonstrated running in real time and on real hardware, and none have been shown in the context of urban navigation where the spatial and temporal scales of the wind require faster than $1$ Hz planning.}

To that end, in \autoref{ch:planning} the wind estimation schemes from subsequent chapters are incorporated into a receding horizon stochastic optimal controller, which continuously approximates the optimal trajectory towards a goal position while also reasoning with local information of the wind. 
In doing so, the contribution from this chapter is a novel wind-aware motion planning framework that adapts to changing wind conditions on the fly using on board sensing, leading to more proactive behavior exploiting favorable wind conditions while avoiding dangerous ones. 
\revision{The core novelty with this framework is that it runs in real time and can be deployed in novel regions with no prior information.}

\subsection{Novel Free Flight Testing for Urban Studies}

Lastly, this thesis explores the utility of free flight testing in the context of autonomous aerial navigation in windy urban environments. 
\autoref{ch:realworld} explores the thoughtful design and execution of laboratory experiments in a highly instrumented sub-scale environment. 
These tests are, to the best of the author's knowledge, the first ever free flight wind tunnel studies of UAVs with \textit{real} obstacles immersed in the flow to create urban-like flow fields, \revision{which presents an exciting advancement in the use of wind tunnels to bridge the gap between the wind engineering and aerospace communities.} 
Further, the algorithms from \autoref{ch:filtering}, \autoref{ch:prediction}, and \autoref{ch:planning} are all implemented on a \revision{commercially-available, flight-ready} single board computer to test the \revision{computational burden} of these algorithms in the real world. 
\revision{With} these experiments, we validate that \revision{the} algorithms \revision{presented in this thesis} can run \revision{on today's existing computational resources} at sufficient frequencies for the real time control of a small scale UAV. 
Along the way, \autoref{ch:realworld} provides valuable data to the scientific community through the aerodynamic characterization of a popular UAV \revision{platform} as well as numerous highly documented instances of flights at the limits of control authority for UAVs operating in urban-like winds. 
We believe this data will serve as an important preliminary dataset \revision{and a helpful guideline} for continued efforts in \revision{understanding the flight mechanics of autonomous UAVs operating in windy urban environments.}

Through the contributions of this thesis just outlined, aerial robots \revision{are now able} to reason about complex surrounding wind flow patterns \revision{like never before:} in real time using on board sensing. 
The simulations to follow demonstrate that this elevated situational awareness can lead to less energy consumption and fewer crashes during autonomous urban aerial operations, and hardware experiments validate that these algorithms are not too computationally demanding to run on existing embedded \revision{computers.} 
Philosophically, this thesis contributes a new way of thinking about wind prediction and wind-aware navigation.
In cluttered environments, visual information combined with sparse wind measurements can provide the necessary cues to infer wind formations, and this knowledge fits neatly into \revision{state-of-the-art} receding horizon optimal control frameworks to deliver \revision{efficient and robust} autonomous flight in the cities of tomorrow. 
\chapter{Dynamics and Energetics of Rotary-Wing Aerial Vehicles}\label{ch:dynamics}

\begin{contribution}
    This chapter includes material from the non-archival work: \bibentry{folk2023rotorpy}. The author of this thesis contributed to the software design, modeling choices, validation experiments, analysis, and writing of the original manuscript.
\end{contribution}

Modeling plays a \revision{fundamental} role in all aspects of engineering, anchoring theory and implementation to an established set of rules. 
All engineering practitioners must clearly define both the scope \textit{(what phenomena is modeled)} and fidelity \textit{(how \revision{accurately} the phenomena is captured)} of their models. 
\revision{Because models would otherwise be an infinite endeavor,} decisions \revision{on scope and fidelity} more often than not are informed by constraints on model accuracy, \revision{uncertainty,} interpretability, and complexity (or more practically: computational burden). 
However, scope and fidelity choices have considerable implications on downstream algorithmic synthesis, experimental design, and subsequent analyses and conclusions. 

When considering the focus of this thesis--aerial navigation through windy urban environments--the sentiments above cannot be overstated. 
Urban wind flow fields are extremely \revision{complicated} dynamical systems evolving on seemingly infinite temporal and spatial length scales.
The evolution of urban wind fields depends on a variety of factors, including not just large-scale solid structures (human-made or not) and meteorological processes, but also meter-scale pressure and temperature gradients.
Neglecting urban winds for a moment, the aerodynamic forces and torques that arise in interactions between the atmosphere and the static and dynamic components of UAVs pose additional modeling challenges.
At some point, a boundary must be established and model choices need to be made to make the problem feasible.

To that end, this chapter will establish the ``rule book'' via the derivation of rotary-wing rigid body dynamics, aerodynamics, energetics, and environmental models utilized throughout this thesis, with the intent of providing the reader with the tools necessary to understand the scope and corresponding limitations of the results and conclusions in the remaining chapters. 

\begin{figure}
    \centering
    \includegraphics[width=1.0\linewidth]{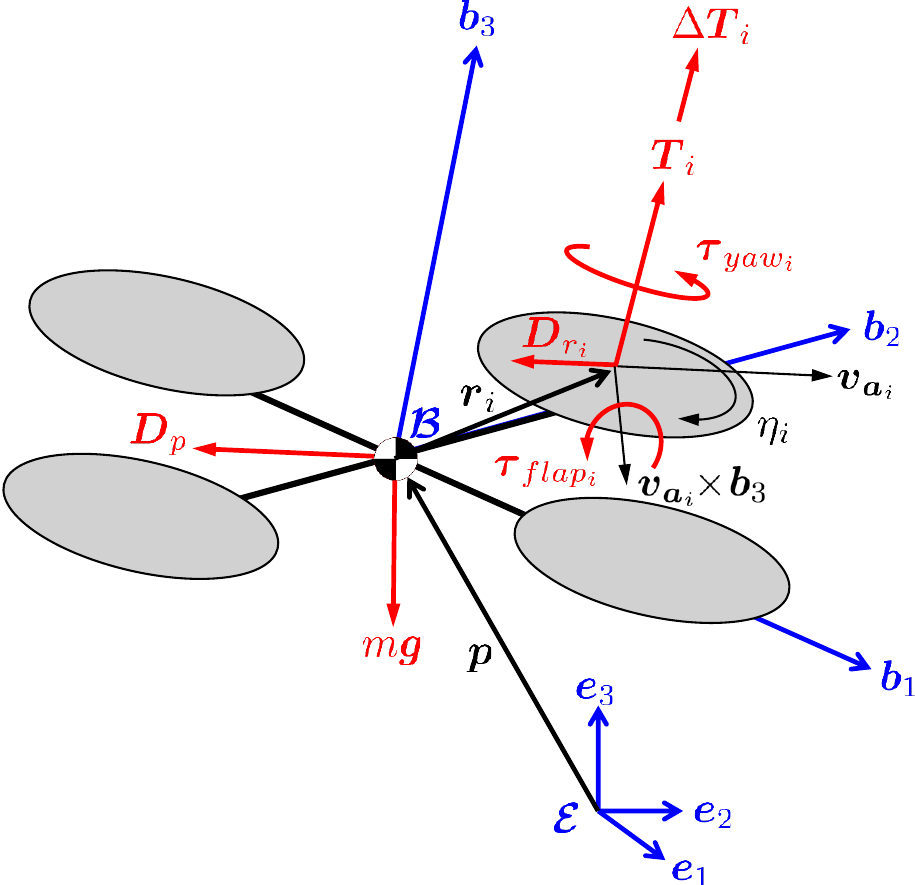}
    \caption{Free body diagram of a rotary-wing UAV subjected to forces and torques generated by gravity and aerodynamic interactions with the air. The blue, red, and black arrows indicate coordinate frames, wrenches, and positional vectors, respectively.}
    \label{fig:quadrotor}
\end{figure}

\subsubsection{A Note on Notation}

Throughout this thesis, vector quantities are written in bold italicized font (e.g. $\bl{x}$) which represent magnitude and direction in an $n$-dimensional Euclidean space. 
Vectors can be resolved into frames denoted as bold calligraphic characters (e.g. $\bl{\mathcal{F}}$), which can be regarded as a set of orthogonal basis vectors $\{ \bl{f}_1 , ... , \bl{f}_n \}$. 
When vectors are resolved in this way, they are indicated with a \revision{prepended} superscript ($^{\bl{\mathcal{F}}}\bl{x}$) and at this point can be thought of as an array of numbers of length $n$, where the $i$'th term corresponds to the projection of the vector $\bl{x}$ onto the $i$'th basis vector $\bl{f}_i$.
Matrices and tensors are written using capital italicized letters, such as the rotation matrix $R$.
Finally, constants and scalar values, e.g. $m$ for mass, are written using lowercase italicized letters. 

\section{Rigid Body Dynamics}\label{sec:dynamics:rigidbodydynamics}

We begin with description of the rigid body dynamics of rotary-wing UAVs, whose temporal evolution can be described using the Newton-Euler equations: 
\begin{align}
    ^{\bl{\mathcal{E}}}\dot{\bl{p}} &= {^{\bl{\mathcal{E}}}\bl{v}} \label{eq:pos_kinematics} \\
    ^{\bl{\mathcal{E}}}\dot{\bl{v}} &= \frac{1}{m}R({^{\bl{\mathcal{B}}}\bl{f}_{c}} + {^{\bl{\mathcal{B}}}\bl{f}_{a}}) -  g\:{^{\bl{\mathcal{E}}}}\bl{e}_3 \label{eq:pos_dynamics} \\
    \dot{R} &= R {^{\bl{\mathcal{B}}}\bl{\Omega}}_{\times} \label{eq:att_kinematics} \\
     ^{\bl{\mathcal{B}}}\dot{\bl{\Omega}} &= J^{-1}( {^{\bl{\mathcal{B}}}\bl{\tau}_{c}} + {^{\bl{\mathcal{B}}}\bl{\tau}_{a}} - {^{\bl{\mathcal{B}}}\bl{\Omega}} \times J\:{^{\bl{\mathcal{B}}}\bl{\Omega}}) \label{eq:att_dynamics} 
\end{align}
where $\bl{p} \in \mathbb{R}^3$ and $\bl{v} \in \mathbb{R}^3$ are respectively the position and velocity vectors; $R \in SO(3)$ describes the rotation from the body frame $\bl{\mathcal{B}}:= \{ \bl{b}_1 , \bl{b}_2, \bl{b}_3 \}$, which is collocated with the center of mass without loss of generality, to a fixed world frame $\bl{\mathcal{E}} := \{ \bl{e}_1 , \bl{e}_2, \bl{e}_3 \}$; $\bl{\Omega} \in \mathbb{R}^3$ is the angular velocity of the body frame and the $(\cdot)_\times$ notation in \autoref{eq:att_kinematics} indicates the skew-symmetric matrix form of $\bl{\Omega}$; $m$ is the total mass of the UAV, $J$ is the inertia tensor resolved into the body frame, and $g$ is the magnitude of the acceleration due to gravity. 
\revision{The dynamics equations above neglect the curvature of the Earth.}

The forces $\bl{f}_{(\cdot)}$ and torques $\bl{\tau}_{(\cdot)}$ acting on the rigid body, visualized in \autoref{fig:quadrotor}, describe the net wrench on the rigid body ultimately dictating the motion and behavior of the UAV in response to its actuators and external disturbances.
The terms $\bl{f}_c \in \mathbb{R}^3$ and $\bl{\tau}_c \in \mathbb{R}^3$ constitute the \textit{nominal} control wrench generated by the $N$ rotors. 
The control wrench is modeled as follows: 

\begin{align}
    \bl{f}_c &= \sum_{i=1}^N \bl{T}_i = k_\eta \sum_{i=1}^{N}\eta_i^2 \label{eq:f_control} \bl{b}_3 \\
    \bl{\tau}_c &= k_\tau \sum_{i=1}^{N} \epsilon_i \eta_i^2 \bl{b}_3 + \sum_{i=1}^n \bl{r}_i \times \bl{f}_{c_i}. \label{eq:m_control}
\end{align}
Here, $\eta_i$ is the $i$'th rotor speed,  $\bl{r}_i$  is the vector from the center of mass to the rotor hub, and $\epsilon_i \in \{-1, 1\}$ is the rotor's direction of rotation. 
These terms represent the nominal thrust and torque (in the absence of wind) produced by all of the rotors. 
Without loss of generality, it is assumed that each rotor has the same static thrust, $k_\eta$, and drag torque, $k_\tau$, coefficients which can be identified using static thrust stand testing, and also that each rotor is aligned with the $\bl{b}_3$ axis which is typical for most rotary-wing UAVs such as quadrotors. 

\subsection{Aerodynamic Wrenches} \label{sec:dynamics:aero}

The equations defined up to this point are adequate for \revision{low-speed} flight in calm conditions\revision{; but} if winds are to be estimated and compensated for, explicit models of the effects of wind on the rigid body must be described. 
The effect of relative motion through the air is captured in the terms $\bl{f}_a \in \mathbb{R}^3$ and $\bl{\tau}_a \in \mathbb{R}^3$, which define models for the aerodynamic forces and torques, i.e. the aerodynamic wrench.
There are numerous physical phenomena that produce aerodynamic wrenches on the UAV.  
As demonstrated by \cite{bangura2012nonlinear}, and more recently in works like \cite{svacha2017improving} and \cite{tomic2016flying}, many of these aerodynamic effects can be effectively captured and compensated for using \revision{physically-informed} lumped-parameter models that are characteristically functions on the relative airspeed, which is defined using the \textit{wind triangle}:
\begin{equation}
    \bl{v}_a := \bl{v} - \bl{w}
\end{equation}
where $\bl{w}\in\mathbb{R}^3$ represents the wind at the UAV's location. 
As it turns out, another important state variable for modeling the aerodynamics of a rotary-wing UAV are the rotor speeds $\eta_i$.  
Following the literature, we make the conscious decision to lump the aerodynamic wrenches into distinct components which will be detailed in the next couple sections.

\subsubsection{Parasitic drag}
Parasitic drag models the combined effects of skin friction and pressure drag acting on the airframe of the UAV, including any payloads or otherwise large structures rigidly attached \revision{to} the body. 
It is characteristically proportional to the airspeed squared: 
\begin{equation}
    \label{eq:parasitic}
    \bl{D}_p = -C \vert \bl{v}_a \vert \bl{v}_a
\end{equation}
where $C = \text{diag}(c_{Dx}, c_{Dy}, c_{Dz})$ is a matrix of parasitic drag coefficients corresponding to each axis of the body frame $\bl{\mathcal{B}}$ and $|\cdot|$ indicates the magnitude of the vector.
\revision{Notice here that we assume parasitic drag is independent of the attitude of the UAV relative to the direction of motion.}

\subsubsection{Rotor drag}
In contrast to parasitic drag, which is often only considered at higher airspeeds, rotor drag (sometimes referred to as \textit{induced drag}), is another source of aerodynamic drag that is prevalent on small UAVs at moderate or even low airspeeds \cite{svacha2017improving}.
The physical phenomenon responsible for rotor drag is the dissymmetry of lift produced by a rotor in forward flight, whereby the advancing blade experiences a higher airspeed than the retreating blade producing an imbalance of forces on the rotor.
Rotor drag has been well understood by rotary-wing \revision{aerodynamicists} since the invention of the helicopter, but the robotics community has only recently shown interest in this effect to \revision{improve tracking control} \cite{bristeau2009rotordrag, martin2010feedback}, and sometimes it is modeled simply as linear in airspeed as in \cite{faessler2018rotordragdiffflat} \revision{where the implicit assumption is that the rotor speeds are approximately constant across the flight envelope.} 
We adopt the rotor drag model (and notation) used \revision{by} Svacha \cite{svacha2020inertia} in which drag is bilinear in the airspeed and the rotor speed \revision{proportional to} lumped parameters. 
\begin{equation}
    \label{eq:rotor_drag}
    \bl{D}_{r_i} = -K \eta_i \bl{v}_{a_i} 
\end{equation}
where $K = \text{diag}(k_d, k_d, k_z)$ is a matrix of rotor drag coefficients\footnote{As noted in \cite{svacha2020inertia}, the $k_z$ term isn't actually a source of drag, but rather a linear approximation of loss of thrust due to change in inflow. However, it resembles an effective drag on the body z axis. This relationship is experimentally shown in \autoref{ch:realworld}.} corresponding to each axis in the body frame $\bl{\mathcal{B}}$. 
Rotor drag is a complex function of the spanwise chord and twist distributions (see \cite{leishman2006principles} ch. 3), but experiments on small UAVs, including the ones in this thesis, provide evidence that the simplification above holds in many cases.

\subsubsection{Translational Lift}

Translational lift is an effect characteristically linear in airspeed. 
Conceptually, the effect can be described as an increase in lift (i.e. thrust) generation by rotors in forward flight.
Relative airspeed through the atmosphere results in rotor wakes being carried away from beneath the rotor. 
The absence of these wakes within the rotor's region of influence is associated with lower induced velocities at the rotor plane, effectively increasing the local angle of attack of each rotor through the air, and thus the thrust produced by the rotor is increased. 
In this work, we adopt the following lumped parameter model for the translational lift of the $i$'th rotor: 
\begin{equation}\label{eq:dynamics:translational_lift}
    \Delta \bl{T}_{i} = k_h V_{h_i}^2 \bl{b}_3
\end{equation}
where $k_h$ is a lumped parameter identified from flight testing and $V_{h_i}$ is the projection of the horizontal airspeed onto the rotor plane--since the rotors are assumed to be aligned with the $\bl{b}_3$ axis, this corresponds to projecting $\bl{v}_a$ onto the plane spanned by $\bl{b}_1$ and $\bl{b}_2$. 
Both \cite{svacha2017improving} and \cite{faessler2018rotordragdiffflat} explicitly incorporate translational lift into their \revision{models,} remarking on its importance in more accurately modeling the thrust produced by each rotor. 
Note that the model presented here neglects induced velocity effects beyond second-order airspeed terms.

\subsubsection{Blade flapping}
Dissymmetry of lift at the advancing and retreating sides of the rotor will also cause the rotor blades to deflect up and down as they revolve in a flapping motion. 
The magnitude and phase of the deflection is a function of the material stiffness of the rotor.
Svacha \textit{et al.} \cite{svacha2020inertia} provides empirical evidence for flapping moments for multirotor UAVs because rotors at this scale are relatively stiff and lack any articulation that would otherwise prevent loads from the rotor being transferred to the airframe. 
This is a very complex phenomenon that can produce both longitudinal and lateral torques on the body depending on the rigidity of the blades \cite{allen1946flapping}. 
In the models in this thesis, blade flapping is limited to a longitudinal moment:
\begin{equation}
    \label{eq:flapping_moment}
    \bl{\tau}_{flap_i} = -k_{flap} \eta_i \bl{v}_{a_i} \times \bl{b}_3
\end{equation} 
with $k_{flap}$ being the flapping coefficient identified empirically. 
The simplification above is justified by the fact that UAVs typically have an even number of counter rotating rotors which are arranged in pairs such that the lateral torques are canceled out. 

The total aerodynamic force in the body frame is $\bl{f}_a = \bl{D}_p + \sum_{i=1}^N (\bl{D}_{r_i} + \Delta \bl{T}_i)$ combining parasitic drag, rotor drag, and translational lift.
The total aerodynamic torque is $\bl{\tau}_a = \sum_{i=1}^N (\bl{\tau}_{flap_i} + \bl{r}_i \times \bl{D}_{r_i})$ which captures blade flapping and any additional torques caused by rotor drag. 

\subsection{Actuator dynamics}

Even for very small UAVs, the rotors take time to settle to a commanded speed. 
Capturing this delay has proven to be important especially for applications in reinforcement learning \cite{panerati2021pybullet} as well as state estimation \cite{svacha2019imu}, the latter being relevant in subsequent chapters. 
Actuator delay on small UAV motors can be modeled using a first order process on the motor speeds:
\begin{equation} \label{eq:actuator}
    \dot{\bl{\eta}} = \frac{1}{\tau_m}(\bl{\eta}_c - \bl{\eta})
\end{equation}
where $\bl{\eta}, \bl{\eta}_c \in \mathbb{R}^N$ are the actual and commanded rotor speeds, respectively, and $\tau_m$ is the motor time constant.
The actuator dynamics can be identified using classical system identification techniques from linear control theory and analyzing responses to step or sinusoidal inputs \revision{on a thrust stand.}


\section{Rotary-Wing Energetics}\label{sec:dynamics:power}

The majority of the energy consumed by a rotorcraft comes from the propulsion system, in other words: the rotors.
This claim is substantiated by empirical evidence both for full scale helicopters (\cite{leishman2006principles} ch. 5 pg. 227) and perhaps even more so in small scale UAVs \cite{mulgaonkar2014power}, where the latter reference reports that rotors contribute to a surprising $94$\% of the average power consumption on small quadrotors. 
Modeling power consumption for rotary-wing UAVs is somewhat nuanced because there are multiple modeling approaches at varying levels of abstraction. 
A review of the various approaches to modeling rotary-wing UAV energy consumption is provided in \autoref{sec:introduction:background:energymodels} painting a complicated picture of how the literature has approached this topic. 

In this thesis, we take a balanced empirically-driven approach to energy modeling, which follows from related energy-aware motion planning \revision{literature} \cite{ware2016canopy, ebert2023gappy, shivgan2020geneticpathplanning, tagliabue2019modelfreeenergy, wu2022modelfreeenergy}, effectively treating the UAV as a point mass where power consumption is a function of the magnitude of the airspeed. 
The motivation behind the point-mass approximation is two-fold: 1) by remaining at the point-mass abstraction, the motion planning problem (\autoref{ch:planning}) is simplified because the decision variables need not include the UAV's attitude; and 2) the corresponding solution to this motion planning problem is abstract enough to apply to UAVs of arbitrary length scales.
The latter point is demonstrated later on in the thesis, where the same motion planning algorithm is used for both full scale simulations and sub-scale hardware experiments in autonomous navigation. 
Although, recent empirical studies by Herz \textit{et al.} \cite{herz2025powervsattitude} and Wu \textit{et al.} \cite{wu2022modelfreeenergy} suggest that thoughtful consideration of attitude (via the yaw angle) in forward flight may lead to measurable improvements in energy consumption especially for UAVs with large  asymmetric payloads.
\revision{This is a promising aspect to consider but the impact seems second-order in nature, and therefore is currently left} to future work. 

While the power model in this thesis is found empirically with wind tunnel testing, we now present some theory that informs the choice of parameterization. 
This theory comes from Leishman (\cite{leishman2006principles} ch. 5) and considers three primary sources of power consumption for rotary-wing vehicles: profile power $P_{pro}$, induced power $P_{i}$, and parasitic power $P_{par}$. 
For illustrative purposes, let $V := \vert \bl{v} \vert$, $W := \text{sign}(\bl{v}\cdot\bl{w})\vert \bl{w} \vert$, and from the wind triangle $V_a = V - W$. 
In these scalar terms, the total power $P_{tot}$ is defined as:
\begin{equation}\label{eq:dynamics:total_power}
    P_{tot}(V_a) = P_{pro}(V_a) + P_i(V_a) + P_{par}(V_a)
\end{equation}
Note that our adaptation of Leishman's model ignores power requirements for climb or descent, as well as any power due to rapid accelerations. 
In other words, this is a steady state forward flight model at a constant altitude. 

     
     

\begin{figure}
    \centering
    \includegraphics[width=0.9\textwidth]{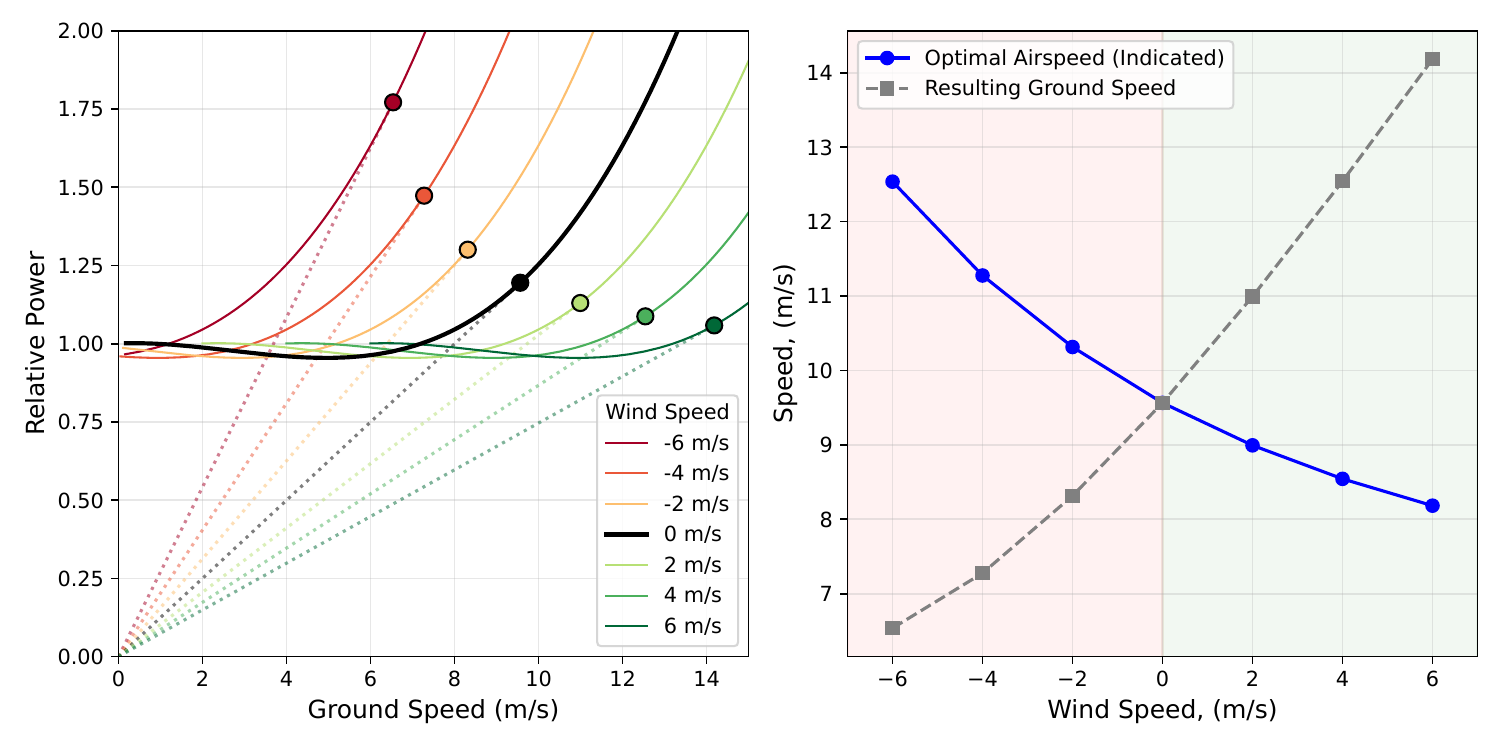}
    \caption{Empirically-driven power curves for a rotary-wing UAV as a function of ground speed and wind speed. On the left, power curves are plotted for varying headwinds (-) and tailwinds (+) with the max range indicated for each. On the right, the optimal range airspeeds and ground speeds are plotted versus the wind speed.}
    \label{fig:powermodel_withwind}
\end{figure}

Induced power is the dominant contributor to rotor power consumption. 
It can be thought of as the power required for the rotor to generate thrust by imparting downward momentum to the surrounding air.
Using momentum theory, induced power can be approximated as
\begin{equation}\label{eq:dynamics:induced_power}
    P_i(V_a) = Tv_i(V_a) \approx \frac{T^2}{2\rho A V_a}
\end{equation}
where $v_i$ is the induced velocity at the rotor plane which is a direct function of the airspeed, $T$ is the thrust produced by the rotor (which is assumed to be constant), $\rho$ is the air density, and $A$ is the swept-area of the rotor. 
The right side of \autoref{eq:dynamics:induced_power} is known as Glauert's high speed approximation, which is a simplification of the induced velocity at the rotor plane used here only for illustrative purposes. 
Induced power will decrease with airspeed, asymptotically approaching zero as the airspeed tends towards infinity. 

Parasitic power \revision{is another significant contributor to} power consumption, and it is understood as the power required to overcome the drag forces on the UAV's \textit{airframe} rather than the rotor(s).
This term captures pressure and skin friction drag contributions from the aspects like the airframe, motor arms, or any external payloads carried by the UAV, \revision{but} neglects any lifting surfaces which are ignored in the present work.
Parasitic power is defined by the following relationship:
\begin{equation}\label{eq:dynamics:parasitic_power}
    P_{par}(V_a) = D_{par}(V_a) V_a = \frac{1}{2}\rho C_D S V_a^3
\end{equation}
where $D_{par}$ is the parasitic drag, quadratic in the airspeed and proportional to the air density, a non-dimensional drag coefficient $C_D$ which describes the shape of the airframe, and a reference area of the airframe $S$. 
Hence, the form of \autoref{eq:dynamics:parasitic_power} indicates that parasitic power is cubic in the airspeed \revision{which explains why it has a dominant influence at higher airspeeds.} 

The profile power is the power required to overcome viscous losses (skin friction, pressure drag) specifically at the rotor itself. 
Profile power is a function of the rotor airfoil distribution, solidity, and rotation speed.  
\revision{For illustrative purposes only, we consider the profile power to be linear in airspeed:
\begin{equation}
    P_{pro}(V_a) = P_0 + \revision{m_{pro} V_a}
\end{equation}
Compared to the other sources of power just described, profile power typically contributes much less to overall UAV power consumption.}

Combining these definitions \revision{illustrates the complexities associated with rotary-wing power consumption,} but empirical data suggests that the total power consumption follows a cubic relationship mostly because \revision{parasitic power dominates the overall power consumption} at high speeds. 
However, parasitic power is \revision{typically} small compared to induced power at low to moderate airspeeds, and induced power decreases with increasing airspeed. 
This means that total power curves for rotary-wing UAVs have a ``dip'', or an absolute minimum power consumption at a nonzero airspeed. 
The magnitude of this ``dip'' depends on the relative contributions of parasitic power and induced power, but its presence means that there are nonzero optimal airspeeds associated with maximum range and maximum endurance. 
Notional power curves with these characteristics are visualized in \autoref{fig:powermodel_withwind}. 
In \autoref{ch:realworld}, we validate such a model using real data collected from free flight testing of a small UAV in a wind tunnel.

\subsection{Accounting for Wind}

As \autoref{fig:powermodel_withwind} suggests, the speed for maximum range (and therefore optimal energy efficiency) depends on the direction and magnitude of the wind--whether or not the UAV is flying \textit{with} (tailwind) or \textit{against} (headwind) the wind. 
To further understand the effect of wind on energy consumption, consider the fact that maximizing energy efficiency for something like a package delivery task is the same as minimizing the amount of total energy $E_{tot}$ consumed per unit \textit{ground} distance covered, $\Delta x$. 
This can be derived in terms of the power curve: 
\begin{equation*}
    \frac{E_{tot}}{\Delta x} = \frac{P_{tot}(V_a)\Delta t}{\Delta x} = \frac{P_{tot}(V_a)}{V}
\end{equation*}
Graphically speaking, this corresponds to the point of tangency from the origin to the power curve (\cite{leishman2006principles}, ch. 5). 
Mathematically, the max range speed can be computed using the following expression:
\begin{equation}
    V_{R} = \arg\min_{V} \frac{P_{tot}(V_a)}{V}
    \label{eq:wind_opt}
\end{equation}
Since $V_a = V-W$, wind can be thought of as a change of coordinates that shifts the power curve thus changing the location of the point of tangency.
This effect is visualized in \autoref{fig:powermodel_withwind}, where adding wind shifts the power curve left or right and the optimal range speed shifts in accordance. 

The main takeaway from this discussion is that, perhaps unsurprisingly, the optimal speed for energy efficiency in steady state forward flight is deeply connected with the wind speed and direction relative to the direction of travel \revision{even in this abstracted formulation.} 
Accounting for this relationship in the planning layer can help improve the energy efficiency considerably--on the contrary, neglecting the effect of wind in favor of tracking a fixed trajectory could quickly lead to actuator saturation and catastrophe--and prior works in the literature have leveraged this abstracted model for similar applications. 


\section{Urban Wind Field Modeling}\label{sec:intro:urbanwindmodeling}

Accurately representing wind in the urban canopy is \revision{an important} yet challenging task in the context of modeling and control of UAVs, because interactions between the wind and buildings can generate extremely complex and hazardous conditions significantly impacting flight. 
A brief discussion about methods from the literature for urban wind field modeling can be found in \autoref{sec:introduction:background:urbanwinds}, where the general sentiment is that simulation plays a key role in understanding the complexities of urban wind fields, especially in the context of aerial navigation. 
For more perspectives on the matter \revision{in the context of robotics,} we refer the reader to \cite{ware2016thesis}, which further discusses the specific challenges associated with modeling winds in the urban canopy. 

Mathematically, wind is a fluid, and so its dynamics are governed by the Navier–Stokes equations:
\begin{align}
\frac{\partial \bl{w}}{\partial t} + (\bl{w} \cdot \nabla)\bl{w} &= -\frac{1}{\rho}\nabla p + \nu \nabla^2 \bl{w} + \bl{f}_{ext} \label{eq:dynamics:navier_stokes} \\
\nabla \cdot \bl{w} &= 0 \label{eq:dynamics:incompressible}
\end{align}
Here, $\bl{w}(\bl{x},t): \mathbb{R}^n \times \mathbb{R} \rightarrow \mathbb{R}^n$ is temporarily treated as a vector \textit{field} \revision{rather than a single vector,} representing the wind parameterized across position $\bl{x}\in\mathbb{R}^n$ and time $t\in\mathbb{R}$. 
The term $p(\bl{x},t): \mathbb{R}^n \times \mathbb{R} \rightarrow \mathbb{R}$ is the pressure scalar field, $\rho$ and $\nu$ are density and kinematic viscosity scalars, and $\bl{f}_{ext}(\bl{x}, t): \mathbb{R}^n \times \mathbb{R} \rightarrow \mathbb{R}^n$ represents an external body force field acting on the fluid, such as gravity.
\autoref{eq:dynamics:incompressible} is the divergence-free condition that enforces conservation of mass. 
At the wind speeds relevant to small UAVs, the atmosphere can be treated as an incompressible fluid since the Mach numbers safely remain in the subsonic region, hence why density is considered a constant in this model.

In the case of urban canyons, the characteristic lengths (building widths, for instance) are quite large compared to the flow velocities corresponding to large Reynolds numbers, suggesting that the viscous term $\nu \nabla^2 \bl{w}$ is small relative to inertial and pressure terms, allowing further simplification to the Euler equations:
\begin{align}
\frac{\partial \bl{w}}{\partial t} + (\bl{w} \cdot \nabla)\bl{w} &= -\frac{1}{\rho}\nabla p + \bl{f}_{ext} \label{eq:dynamics:euler} \\
\nabla \cdot \bl{w} &= 0
\end{align}
The Euler equations describe the motion of an ideal, inviscid, incompressible fluid and form the theoretical model behind the wind simulations presented in this thesis. 
These equations capture the fundamental mechanisms of advection and pressure-driven flow that dominate wind behavior at the scales of interest.
\revision{There also exist algorithms for solving these equations quickly while remaining numerically stable.
In this thesis, a dedicated implementation of Stam's \textit{Stable Fluids} algorithm \cite{stam2023stable} forms the backbone of wind simulations used for training and evaluating the methods to follow.}

However, from \autoref{eq:dynamics:euler} alone, it's not clear how urban objects (buildings, bridges, etc.) affect the flow field.
Objects influence the flow through boundary conditions, typically by imposing no-penetration (and sometimes no-slip) constraints on the surfaces of solid obstacles.
The wind fields used in this thesis \revision{incorporate} the effects of urban geometry as no-penetration boundary conditions on the \revision{flow, resulting in the hazardous} urban wind conditions \revision{(e.g. vortices and corner accelerations) that are a keen interest for autonomous aerial navigation.}
The resulting velocity fields from these numerical simulations provide both the spatial and temporal evolution of the wind, which is then sampled at the UAV’s location to compute the relative airspeed $\bl{v}_a = \bl{v} - \bl{w}$ used in the aerodynamic models described in \autoref{sec:dynamics:aero}. 
The specific numerical simulation algorithm is described in \autoref{sec:prediction:euler_sim}, and these simulations are employed in \autoref{ch:prediction} and \autoref{ch:planning} to evaluate UAV performance and robustness in complex wind environments.

\revision{

\subsection{Dimensionality and Flow Regimes in Urban Canyons}\label{sec:modeling:2dvs3d}

Driven primarily by computational limitations, the urban wind simulations in this thesis are restricted to 2D specifically on the ``horizontal'' or $x$-$y$ motion of the wind. 
However, we acknowledge that urban wind fields are inherently 3D with numerous updraft and downdraft regions that also potentially impact aerial navigation.
In this section, we provide some context into what influences the vertical motion of the wind to help establish a boundary on where our quasi-2D assumptions hold.
The three-dimensionality of the wind is caused by many factors, in part by the interaction of the nonuniform atmospheric boundary layer with the buildings but also flow separation over the roof and sides of the buildings that create multiple horseshoe vortices. 
These horseshoe vortices that develop in the wake of buildings are a large contributor to the vertical motion of the air.
The relationship between wake geometry and building geometry is very complex, often requiring bespoke CFD or wind tunnel studies, like those referenced in \autoref{sec:introduction:background:urbanwinds}, with hand-tuned boundary layer profiles and accurate building geometry to understand how specific flow features develop.
Nevertheless, if we consider a building with ``simple'' enough geometry, like a cuboid for example, this allows for some rules of thumb to be established. 

For illustrative purposes, consider cuboid buildings defined by a characteristic length ($L$), width ($W$), and height ($H$), spaced apart by a street width ($S$).
To quantify the flow around an isolated building, Martinuzzi and Tropea \cite{martinuzzi1993pristmaticfluidstudy} ran a series of water tunnel experiments on isolated cuboid obstacles with varying $W/H$ aspect ratios.
One of their primary findings was that the vertical component of the flow is negligible only for very wide obstacles where $W/H > 6$, with the explanation that the influence of the corner vortices remains confined to the edges.

In the case of multiple buildings, within the urban canyon so-to-speak, vertical fluid motion is governed more so by the canyon aspect ratio, defined as the ratio of building height to street width ($H/S$).
This is because the wind-inducing horseshoe vortices from one building are disrupted by other buildings downwind.
Based on historical empirical analyses, Oke \cite{oke1988streetcanopy} identifies three distinct flow regimes in the street canyon based on this ratio:
\begin{enumerate}
    \item \textbf{Isolated Roughness ($H/S < 0.3$):} Buildings are widely spaced, and the flow reattaches to the ground between obstacles.
    \item \textbf{Wake Interference ($0.3 < H/S < 0.7$):} The wake of one building impinges directly on the next, creating complex, turbulent interactions with measurable vertical gusts.
    \item \textbf{Skimming Flow ($H/S > 0.7$):} The buildings are closely spaced such that the bulk of the boundary layer flow ``skims'' over the rooftops. Within the canyon itself, a stable, recirculating vortex is trapped, driven by the shear layer at the roof level.
\end{enumerate}
In the \textit{Skimming Flow} regime, Oke describes that the vertical momentum transport is confined largely to the shear layer at the roofline ($z \approx H$). 
Deep within the canyon ($z < 0.7H$), the mean flow is dominated by channeling effects aligned with the street geometry.
These effects still induce vertical motion, but they are smaller than that caused by the horseshoe vortices. 

} 

\section{Model Limitations}

The aerodynamic, energetic, and wind models described in this chapter have limitations that are important to outline. 
For one, the lumped parameter aerodynamic models in \autoref{sec:dynamics:aero} \revision{simplify} the aerodynamic \revision{forces acting on the UAV, neglecting transient effects while also ignoring any} interactions between the rotor, airframe, and wind.
\revision{These effects are second-order in nature, but the assumption} that there are \revision{neither} \revision{lifting} surfaces \revision{nor} rotor-airframe interactions \revision{may lead to significant differences between simulation and reality.} 
\revision{Lifting} surfaces will likely be \revision{necessary to be economically-viable,} and \revision{payloads} may incur nontrivial interactions with the wakes generated by the \revision{rotors, so future work should strive to include these in simulated models.} 
Also, the energetics model in \autoref{sec:dynamics:power} assumes constant speed and altitude. 
In \revision{reality, flights in the urban canopy will entail} other transient flight modes like \revision{accelerations,} climbing, and \revision{descent, all consuming} additional power \revision{not predicted by our} energetics model. 

\revision{On} the wind modeling side, \revision{there are a few critical assumptions that the reader should consider for subsequent chapters.} 
\revision{For one,} an important assumption is that the spatial features of the wind \revision{fields} occur on a length scale larger than the \revision{UAV, such} that the wind is assumed to be constant across all of the rotors. 
\revision{This means that there are no rolling or pitching moments induced by wind spatial gradients in simulation.}
\revision{Also, the quasi-2D assumption should be considered carefully given the 3D nature of urban wind flow fields. 
Based on \cite{oke1988streetcanopy} and \cite{martinuzzi1993pristmaticfluidstudy}, the validity of the quasi-2D flow assumption depends heavily on the density of the urban environment.
For isolated slim high-rises ($W/H < 6$) or sparse suburban layouts ($H/S < 0.3$), the flow is dominated by 3D horseshoe vortices and roof downwash, making a 2D planar model insufficient.
However, in the dense city centers, the close proximity of buildings ($H/S > 0.7$) suppresses these isolated wake structures, forcing the flow into the \textit{Skimming Flow} regime.
In this regime, the flow is predominately channeled horizontally along the streets rather than wrapping vertically around individual structures.
Therefore, provided the UAV operates deep within the canyon ($z < 0.7H$), avoiding the immediate vicinity of rooftop shear layers or ground recirculation zones, the vertical component of the wind velocity is negligible compared to the horizontal components.
In this context, we assume that the simulated UAV is operating in this regime, bearing in mind that the analysis provided above is only strictly valid for cuboid buildings and exact threshold values may vary for more complicated geometry. 
}
\chapter{Model-Based Indirect Wind Estimation}\label{ch:filtering}

\begin{contribution}
    This chapter includes material from the non-archival work: \bibentry{folk2023rotorpy}. The author of this thesis contributed to the algorithmic design, modeling choices, experiments, and writing of the original manuscript.
\end{contribution}

Wind estimation from on board sensors is \revision{a technology that is foundational} to the contributions of this thesis, as it provides in situ measurements of the local wind field at the UAV’s position. 
The various approaches to wind estimation are reviewed in depth in \autoref{sec:introduction:background:comestimation}. 
This chapter provides the formulation of \textit{Wind UKF}: a filtering approach to predicting the wind at the UAV's current location. 
Rather than relying on a dedicated wind sensor, e.g. an ultrasonic anemometer, \textit{Wind UKF} follows more recent indirect methods, e.g. \cite{svacha2019thesis, tomic2022modelbasedwindestimation}, by using inertial sensing paired with an appropriate aerodynamic model to infer the wind vector. 
This is especially relevant for smaller UAVs, such as the one used in subsequent hardware experiments, that do not have the payload capacity for dedicated wind sensors. 

Following the definitions and notations defined in \autoref{ch:dynamics}, \textit{Wind UKF} estimates the following quantities: the UAV's orientation $R \in SO(3)$, angular velocity $\bl{\Omega} \in \mathbb{R}^3$, inertial velocity $\bl{v} \in \mathbb{R}^3$, wind velocity at the center of mass $\bl{w} \in \mathbb{R}^3$, and the rotor speeds $\bl{\eta} \in \mathbb{R}^N$. 
The assumed ``known'' inputs to the filter are the \textit{commanded} rotor speeds $\bl{\eta}_c$.

\section{Transcription}

The filter state explains \textit{what} is being estimated, but this abstract definition is not immediately suitable for implementation on a real UAV.
The following sections outline the steps taken to transcribe the states, process models, and measurement models. 

\subsection{Body Frame Resolution of Velocities}

First, the vectors $\bl{\Omega}$, $\bl{v}$, and $\bl{w}$ corresponding to the angular, inertial, and wind velocities are resolved into the body frame $\bl{\mathcal{B}}$.
The choice to resolve the linear velocities in the body frame was made because the aerodynamic models are \revision{also} defined in this frame (see \autoref{sec:dynamics:aero}), so the number of rotation operations, and therefore the computational burden, is minimized by keeping the filter states in the body frame of reference. 

\subsection{Orientation Representation}

Second, the robot's orientation with respect to the world frame, $R$, is represented using Euler angles. 
This decision was primarily made to reduce computational complexity. 
Euler angles evolve in Euclidean space ($\mathbb{R}^3$) rather than directly on the manifold of rotations ($SO(3)$), thus limiting the size of the orientation state to three dimensions and simplifying the overall state space. 
Furthermore, the noise can be added independently to each Euler angle, which while not strictly necessary, is a simpler and more straightforward approach to implement compared to the additive noise processes for other rotation representations such as quaternions or rotation matrices \revision{(see for instance \cite{loianno2016ufkonmanifold}).}

For clarity, the orientation $R \in SO(3)$ can be decomposed into a series of rotations: $R = R_\psi R_\theta R_\phi$
where $\psi$ is the rotation angle about the z axis (yaw), $\theta$ about the y axis (pitch), and $\phi$ about the x axis (roll). 
Each entry of the rotation matrix can be defined from these angles:
\begin{equation}
    R(\psi, \theta, \phi) = \begin{bmatrix} R_{11} & R_{12} & R_{13}\\ R_{21} & R_{22} & R_{23} \\ R_{31} & R_{32} & R_{33} \end{bmatrix} = \begin{bmatrix} c_\psi c_\theta & c_\psi s_\theta s_\phi - s_\psi c_\phi & c_\psi s_\theta c_\phi + s_\psi s_\phi \\ s_\psi c_\theta & s_\psi s_\theta s_\phi + c_\psi c_\phi  & s_\psi s_\theta s_\phi - c_\psi s_\phi \\ -s_\theta  &  c_\theta s_\phi & c_\theta c_\phi \end{bmatrix}
\end{equation}
where $s_{(\cdot)} = \sin (\cdot)$ and $c_{(\cdot)} = \cos (\cdot)$. 
From this definition, the vector of \revision{the so-called} Z-Y-X Euler angles, $\bl{\Theta}$, can be extracted from the rotation matrix via:
\begin{align}
    \bl{\Theta} := \begin{bmatrix} \psi \\ \theta \\ \phi \end{bmatrix} = \begin{bmatrix} \mathrm{arctan2} (R_{21}, R_{11} ) \\
    \mathrm{arcsin}(R_{31}) \\ 
    \mathrm{arctan2} ( R_{32}, R_{33} ) \end{bmatrix}
\end{align}
It is here where we note a critical limitation of Euler angles, which is that a singularity exists when $\cos(\theta) = 0$, i.e. when $\theta = \pm\pi/2$. 
This is known \revision{colloquially} as gimbal lock, and in this context physically corresponds to the UAV pitching up or down 90 degrees. 
In this work, while we certainly do not make any small angle or hover approximations, we do assume that the UAV does not reach the pitch singularity \revision{and gimbal lock is not a concern.}

\subsection{Filter State and Input Vectors}\label{sec:filtering:stateinputdef}

The result of the transcriptions above is the final, slightly revised, filter state vector:
\begin{equation}
    \bl{\mathrm{x}} := \begin{bmatrix} \bl{\Theta} \\ ^{\bl{\mathcal{B}}}\bl{\Omega} \\ ^{\bl{\mathcal{B}}}\bl{v} \\ ^{\bl{\mathcal{B}}}\bl{w} \\ \bl{\eta} \end{bmatrix} \in \mathbb{R}^{12+N}
\end{equation}
The reader may notice that now the filter state evolves strictly in Euclidean space, making it suitable for standard filtering techniques. 

The input vector to \textit{Wind UKF} is defined as the commanded motor speeds: 
\begin{equation}
    \bl{\mathrm{u}} := \bl{\eta}_c
\end{equation}
which is a practical consideration to be discussed in further detail in \autoref{ch:realworld}. 

\subsection{Process Model}\label{sec:filtering:process_model}

The process model, or state transition model, captures how the filter state evolves over time. 
The filter state dynamics are modeled in continuous time as a nonlinear differential equation: 
\begin{equation}
    \dot{\bl{\mathrm{x}}} = \bl{f}(\bl{\mathrm{x}}, \bl{\mathrm{u}})
\end{equation}

\subsubsection{Attitude Dynamics}

For attitude, there is a closed form nonlinear expression describing the time evolution of Euler angles due to angular velocities in the body frame: 
\begin{equation}
    \dot{\bl{\Theta}} = \begin{bmatrix} 0 & \frac{\sin(\phi)}{\cos(\theta)} & \frac{\cos(\phi)}{\cos(\theta)} \\ 0 & \cos(\phi) & -\sin(\phi) \\ 1 & 0 & 0 \end{bmatrix} {^{\bl{\mathcal{B}}}\bl{\Omega}}
\end{equation}

\subsubsection{Linear Velocity Dynamics}

The process model for the linear velocity term is derived using Newton's second law and follows from \autoref{sec:dynamics:rigidbodydynamics}: 
\begin{equation}\label{eq:filtering:vdot}
    \dot{\bl{v}} = \frac{1}{m}\bl{f}_{net} = g\bl{e}_3 + \frac{1}{m} \left(  \sum_{i=1}^N k_\eta \eta_i^2 + k_hV_h^2 \right)\bl{b}_3 - \frac{1}{m} K\sum_{i=1}^N\eta_i \bl{v}_a
\end{equation} 
Note that the \textit{filter's} dynamics ignore parasitic drag, implying that the rotor drag and translational lift terms are the dominant aerodynamic forces acting on the UAV. 
This implication is in line with similar UAV wind estimation implementations (e.g. \cite{svacha2019thesis} ch. 7) where winds are in the low to moderate speed ranges. 
To transcribe this, all the vector components above must be resolved in the body frame which also necessitates handling the transport (or Coriolis) term associated with taking the time derivative of a rotating frame (${^{\bl{\mathcal{E}}}\dot{\bl{v}}} = {^{\bl{\mathcal{B}}}\dot{\bl{v}}} + {^{\bl{\mathcal{B}}}\bl{v}}\times{^{\bl{\mathcal{B}}}\bl{\Omega}}$): 
\begin{equation}\label{eq:filtering:process_linearvel}
    ^{\bl{\mathcal{B}}}\dot{\bl{v}} = \frac{1}{m}  \sum_{i=1}^N \left( k_\eta \eta_i^2 + k_h V_h^2 \right) \begin{bmatrix} 0 \\ 0 \\ 1\end{bmatrix} - \frac{1}{m} K \left( \sum_{i=1}^N \eta_i \right) \left( ^{\bl{\mathcal{B}}}\bl{v} - {^{\bl{\mathcal{B}}}\bl{w}} \right) + R^\top \begin{bmatrix} 0 \\ 0 \\ g \end{bmatrix} - {^{\bl{\mathcal{B}}}\bl{\Omega}} \times {^{\bl{\mathcal{B}}}\bl{v}}
\end{equation}
Note that $m$, $g$, $k_\eta$, $k_h$, and $K$ are assumed to be known \textit{a priori}. 
System identification techniques for the aerodynamic parameters are discussed in \autoref{ch:realworld}. 

\subsubsection{Actuator Dynamics}

The rotor speeds are assumed to follow a first order process:
\begin{equation}
    \dot{\bl{\eta}} = \frac{1}{\tau_m}(\bl{\eta}_c - \bl{\eta})
\end{equation}
where $\tau_m$ is the motor time constant which must be identified \textit{a priori} using bench top testing. 

\subsubsection{Static States}

For the remaining filter states, the process model assumes that these variables are fixed in time: \begin{equation}\label{eq:filtering:static}
{^{\bl{\mathcal{B}}}\dot{\bl{\Omega}}} = {^{\bl{\mathcal{B}}}\dot{\bl{w}}} = 0 \end{equation}
The reader should note that this does not mean that these filter states remain constant for all time.
Rather, the time evolution of the angular body rates and wind vector are dictated by the interplay between the process model above and the measurements models below, alongside their respective noise parameters. 

\subsection{Measurement Models}

Measurement models\revision{, $\bl{h}_{(\cdot)}$,} provide partial, noisy observations of the filter state $\bl{\mathrm{x}}$. 
This section outlines the available measurements and their corresponding sensor models.

\subsubsection{Inertial measurement unit}
A general model of the \revision{measurements provided by the} inertial measurement unit (IMU), which includes an accelerometer and gyroscope, is given by: 
\begin{equation} \label{eq:filtering:imu}
    \bl{h}_{IMU} = \begin{bmatrix} R_\mathcal{I}^\mathcal{B}(\dot{\bl{v}} + g\bl{e}_3) + \bl{a}_{IMU} + \bl{b}_a \\ \bl{\Omega} + \bl{b}_{g} \end{bmatrix}
\end{equation}
where $\dot{\bl{v}}$ is given by \autoref{eq:filtering:vdot} above, $\bl{a}_{IMU} = \bl{\Omega}\times(\bl{\Omega}\times\bl{r}_{IMU})$, $\bl{r}_{IMU}$ and $R_\mathcal{I}^\mathcal{B}$ are the position and orientation of the sensor in the body frame. 
The biases are denoted as $\bl{b}_{(\cdot)}$. 
Notably, the acceleration term in \autoref{eq:filtering:imu} is the only signal associated with the wind, which is captured via the airspeed term in \autoref{eq:filtering:vdot}. 

\subsubsection{External motion capture}

The experiments in this thesis \revision{were run} in an indoor motion capture space. 
The external motion capture sensor is meant to supplement the role of the Global Positioning System (GPS) and the Attitude and Heading Reference System \revision{(AHRS), or perhaps even vision-based odometry,} that would typically be available in the outdoor setting.
These sensors provide information about the ground velocity and orientation of the UAV. 
\begin{equation}\label{eq:filtering:mocap}
    \bl{h}_{MC} = \begin{bmatrix} \bl{v} \\ \bl{\Theta} \end{bmatrix}
\end{equation}
Note that practically the ground velocity is provided in the world frame, $\bl{\mathcal{E}}$. 
A minor preprocessing step in the actual filter implementation involves rotating this velocity into the body frame using the orientation $\bl{\Theta}$ \revision{which incurs some additional computational processing.}

\section{Nonlinear Observability Analysis}

In this section, we briefly show that the \textit{Wind UKF} just described is observable. 
This system is nonlinear owing to the aerodynamic and Coriolis terms in \autoref{eq:filtering:process_linearvel}. 
Rather than linearizing this system, we instead opt to follow the lead of Svacha \cite{svacha2019thesis} and use nonlinear observability analysis techniques described in \cite{hermann2003nonlinearobservabilityanalysis} to understand the observability of the filter. 

The process model is first rewritten in control affine form
\begin{equation}
    \dot{\bl{\mathrm{x}}} = \bl{f}_0(\bl{\mathrm{x}}) + \sum_i \bl{f}_i(\bl{\mathrm{x}})\mathrm{u}_i
\end{equation}
where
\begin{equation}
    \bl{f}_0 = \begin{bmatrix} \begin{bmatrix} 0 & \frac{\sin(\phi)}{\cos(\theta)} & \frac{\cos(\phi)}{\cos(\theta)} \\ 0 & \cos(\phi) & -\sin(\phi) \\ 1 & 0 & 0 \end{bmatrix} {^{\bl{\mathcal{B}}}\bl{\Omega}} \\ 
    \bl{0}_{3} \\ 
    \frac{1}{m}  \sum_{i=1}^N \left( k_\eta \eta_i^2 + k_h V_h^2 \right) \begin{bmatrix} 0 \\ 0 \\ 1\end{bmatrix} - \frac{1}{m} K \left( \sum_{i=1}^N \eta_i \right) {^{\bl{\mathcal{B}}}\bl{v}_a}  + R^\top \begin{bmatrix} 0 \\ 0 \\ g \end{bmatrix} - {^{\bl{\mathcal{B}}}\bl{\Omega}} \times {^{\bl{\mathcal{B}}}\bl{v}} \\ 
    \bl{0}_{3} \\ 
    -\frac{1}{\tau_m} \bl{\eta}
    \end{bmatrix}
\end{equation}
and
\begin{equation}
    \bl{f}_i = \begin{bmatrix} \bl{0}_{12} \\ 
    \frac{1}{\tau_m} \bl{z}_i
    \end{bmatrix} \quad i=1,..., N
\end{equation}
where $\bl{z}_i$ is an indexing vector where the $i$'th element is 1 and the rest are 0. 

The measurement vector is the combination of the models presented in the prior section:
\begin{equation}
    \bl{h} = \begin{bmatrix}  R_\mathcal{I}^\mathcal{B}(\dot{\bl{v}} + g\bl{e}_3) + \bl{a}_{IMU} + \bl{b}_a \\ 
    \bl{\Omega} + \bl{b}_g \\ 
    \bl{v} \\ 
    \bl{\Theta} \\ 
    \bl{\eta}
    \end{bmatrix}
\end{equation}
For the purposes of this observability analysis, the following assumptions are made: $\bl{b}_a = \bl{b}_g = 0$, $R_{\mathcal{I}}^{\mathcal{B}} = I_3$, and $\bl{r}_{IMU} = 0$. 
These assumptions physically correspond to the IMU being unbiased and collocated with the body frame of the UAV. 
In practice, the biases are identified via a brief calibration procedure at the beginning of each \revision{flight, although previous work \cite{svacha2019thesis} demonstrated that it is possible to simultaneously estimate the accelerometer biases.}\newline\newline

The next step after constructing $\bl{f}_0$, $\bl{f}_i$, and $\bl{h}$ as above is to construct the nonlinear observability matrix $\mathcal{O}$:
\begin{equation}\label{eq:filtering:observability_matrix}
    \mathcal{O} := \begin{bmatrix}
        \bl{h} \\ 
        L_{\bl{f}_0} \bl{h} \\ 
        L_{\bl{f}_1} \bl{h} \\ 
        \vdots \\ 
        L_{\bl{f}_N} \bl{h} 
    \end{bmatrix}
\end{equation}

According to \cite{hermann2003nonlinearobservabilityanalysis}, for a filter state $\bl{x}\in\mathbb{R}^n$, if $\text{rank}\ \mathcal{O}(\bl{x}) = n$ then the system is \textit{locally weakly observable}.
Intuitively, this means that small changes in the filter state produce distinguishable variations in the measurements over time, and therefore the state can be reconstructed locally. 

The full observability analysis was conducted for a quadrotor ($N=4$) using Mathematica and is available publicly on Github.\footnote{\url{https://github.com/spencerfolk/wind_aware_ros/tree/main/wind_aware_ukf/analysis}}
The rank of $\mathcal{O}$ was found to be $16$, which is equal to the length of $\bl{\mathrm{x}}$ for a quadrotor, thus satisfying the condition for a system to be locally weakly observable. 

\section{Unscented Kalman Filtering}

Due to the nonlinear aerodynamic models used in this thesis, the state dynamics and measurement equations, an unscented Kalman filter (UKF) is adopted for the wind estimation framework. 
The UKF is a well-established nonlinear filtering technique introduced in \cite{julier1997ukf} that avoids linearization of the process and measurement models via Jacobians--a requirement \revision{and pain point} of the Extended Kalman Filter (EKF). 
The UKF, and its advantages over the EKF, are presented in \cite{wan2000ukf}. 
For completeness, we briefly review the UKF and provide the equations relevant for implementation and reproducibility. 

\subsection{Unscented Transform}

The unscented transform (UT) is a method for calculating the posteriori statistics of a random variable undergoing a nonlinear transformation accurate to a third order for Gaussian priors \cite{wan2000ukf}.
Consider an $n$-dimensional Gaussian random variable $\bl{x}\in\mathbb{R}^n$ with mean $\bar{\bl{x}}$ and covariance $P$ that is propagated through an arbitrary nonlinear function: 
\begin{equation}
    \bl{y} = \bl{g}(\bl{x})
\end{equation}
The UT approximates the statistics of $\bl{y}$ using a set of $2n+1$ \textit{sigma} points, $\mathcal{X}_i$, with associated weights $W_i^{(m)}$ and $W_i^{(c)}$, chosen so that the sample mean and covariance of the sigma points match $\bar{\bl{x}}$ and $P_x$ up to a second order. 
Following \cite{wan2000ukf}:
\begin{align}
    \mathcal{X}_0 &= \bar{\bl{x}} \\  \label{eq:filtering:sigmapoints}
    \mathcal{X}_i &= \bar{\bl{x}} + \left[\sqrt{(n + \lambda)P_x}\right]_i \ \ ,  \ \ i = 1, ... , n \\ 
    \mathcal{X}_i &= \bar{\bl{x}} - \left[\sqrt{(n + \lambda)P_x}\right]_i \ \ ,  \ \ i = n+1, ... , 2n \\ 
    W_0^{(m)} &= \frac{\lambda}{n+\lambda} \\ 
    W_0^{(c)} &= \frac{\lambda}{n+\lambda} + (1-\alpha^2 + \beta) \\ 
    W_i^{(m)} &= W_i^{(c)} = \frac{1}{2(n+\lambda)} \ \ , \ \ i = 1, ..., 2n \label{eq:filtering:sigmaweights}
\end{align}
where $\lambda = \alpha^2(n+\kappa) - n$ is a scaling parameter, $\alpha$ determines the degree of spread of the sigma points around $\bar{\bl{x}}$, $\kappa$ is a secondary scaling parameter, and $\beta$ is used to incorporate prior knowledge of the distribution of the random variable $\bl{x}$.
In this context, $\sqrt{(\cdot)}$ indicates the matrix square root and $[\cdot]_i$ is the $i$'th row of that matrix.  
The matrix square root is typically implemented using the Cholesky decomposition to ensure numerical stability.

The sigma points are propagated through the full nonlinear function: 
\begin{equation}
    \mathcal{Y}_i = \bl{g}(\mathcal{X}_i), \ \ i = 0, ... , 2n
\end{equation}
From here the mean and covariance can be approximated: 
\begin{align}
    \bar{\bl{y}} &\approx \sum_{i=0}^{2n} W_i^{(m)}\mathcal{Y}_i \label{eq:filtering:sigma_mean} \\
    P_y &\approx \sum_{i=0}^{2n} W_i^{(c)} \left( \mathcal{Y}_i - \bar{\bl{y}} \right) \left( \mathcal{Y}_i - \bar{\bl{y}} \right) ^\top \label{eq:filtering:sigma_covariance}
\end{align}

\subsection{Prediction Step}

The prediction step propagates the filter state forward in time using the process model defined in \autoref{sec:filtering:process_model}. 
The UKF uses a discrete approximation of this transformation, $\bl{f}_d$, which \revision{is obtained} through numerical integration (e.g. Euler or Runge-Kutta) and a fixed timestep $\Delta t$:
\begin{equation}
    \bl{\mathrm{x}}_{k+1} = \bl{f}_d(\bl{\mathrm{x}}_k, \Delta t) + \bl{\mathrm{w}}_k 
\end{equation}
where $\bl{\mathrm{w}_k} \sim \mathcal{N}(0, Q_F)$ represents process noise, whose covariance $Q_F$ is a user-defined parameter accounting for unmodeled dynamics and external disturbances. 

Given the previous estimate of the state $\hat{\bl{\mathrm{x}}}_{k-1|k-1}$, uncertainty $P_{k-1|k-1}$, the sigma points and weights are generated following \autoref{eq:filtering:sigmapoints} through \autoref{eq:filtering:sigmaweights}. 
Then, each sigma \revision{point} is propagated through the discrete nonlinear dynamics: 
\begin{equation}\label{eq:filtering:predict_sigma}
    \left[\mathcal{X}_{k|k-1}\right]_i = \bl{f}_d \left( \left[\mathcal{X}_{k-1|k-1} \right]_i, \Delta t \right)
\end{equation}
and the mean and covariance are updated following \autoref{eq:filtering:sigma_mean} and \autoref{eq:filtering:sigma_covariance}. 
\begin{align}
    \bl{\mathrm{x}}_{k|k-1} &= \sum_{i=0}^{2n} W_i^{(m)} \left[\mathcal{X}_{k|k-1} \right]_i \\ 
    P_{k|k-1} &= Q_F + \sum_{i=0}^{2n} W_i^{(c)} \left( \left[\mathcal{X}_{k|k-1} \right]_i - \bl{\mathrm{x}}_{k|k-1} \right) \left( \left[\mathcal{X}_{k|k-1} \right]_i - \bl{\mathrm{x}}_{k|k-1} \right) ^\top
\end{align}

Note that unlike the EKF, the Jacobian of \revision{the process model} $\bl{f}_d$ need not be computed\revision{, easing} the burden on the practitioner and \revision{leaving} less room for error. 
However, both the computational burden and numerical conditioning scales poorly with $n$ due to having to calculate the square root of $P_{k|k}$ in the sigma point generation.
Computational complexity with large filter state dimensions is one of the primary drawbacks of the UKF. 

\subsection{Measurement Update Step}

During this step, the filter state is corrected whenever a new measurement, $\bl{\mathrm{y}}_k$ is received. 
First, using the following measurement model:   
\begin{equation}
    \bl{\mathrm{z}}_k = \bl{h}(\bl{\mathrm{x}}_{k}) + \bl{\mathrm{v}}_k
\end{equation}
where $\bl{\mathrm{v}_k} \sim \mathcal{N}(0, R_F)$ represents the measurement noise.
The user-defined parameter $R_F$ is a covariance matrix reflecting the degree of uncertainty in each measurement. 
The sigma points generated in \autoref{eq:filtering:predict_sigma} are then transformed into measurement beliefs using the above measurement models: 
\begin{equation}
    \left[\mathcal{Z}_{k|k-1}\right]_i = \bl{h}\left( \left[\mathcal{X}_{k|k-1}\right]_i \right)
\end{equation}
The predicted measurement mean and covariance are computed once again following \autoref{eq:filtering:sigma_mean} and \autoref{eq:filtering:sigma_covariance}. 
\begin{align}
    \bl{\mathrm{z}}_{k|k-1} &= \sum_{i=0}^{2n} W_i^{(m)} \left[\mathcal{Z}_{k|k-1} \right]_i \\ 
    P_{zz} &= R_F + \sum_{i=0}^{2n} W_i^{(c)} \left( \left[\mathcal{Z}_{k|k-1} \right]_i - \bl{\mathrm{z}}_{k|k-1} \right) \left( \left[\mathcal{Z}_{k|k-1} \right]_i - \bl{\mathrm{z}}_{k|k-1} \right) ^\top
\end{align}
The cross-covariance between the state and measurement sigma points is
\begin{equation}
    P_{xz} = \sum_{i=0}^{2n} W_i^{(c)}\left( \left[ \mathcal{X}_{k|k-1} \right]_i - \bl{\mathrm{x}}_{k|k-1} \right) \left( \left[ \mathcal{Z}_{k|k-1} \right]_i - \bl{\mathrm{z}}_{k|k-1} \right)^\top
\end{equation}
The Kalman gain is then
\begin{equation}
    K_k = P_{xz}P_{zz}^{-1}
\end{equation}
With the Kalman gain computed, the state estimate and covariance can be updated as follows:
\begin{align}
    \bl{\mathrm{x}}_{k|k} &= \bl{\mathrm{x}}_{k|k-1} + K_k\left( \bl{\mathrm{y}}_k - \bl{\mathrm{z}}_{k|k-1} \right) \label{eq:filtering:measurement_update_state} \\ 
    P_{k|k} &= P_{k|k-1} - K_k P_{zz} K_k^\top \label{eq:filtering:measurement_update_cov}
\end{align}
An interesting observation is that \autoref{eq:filtering:measurement_update_state} and \autoref{eq:filtering:measurement_update_cov} are equivalent to the standard update equations of an EKF but the computation of $\bl{\mathrm{z}}_{k|k-1}$ and $P_{zz}$ uses the sigma points rather than Jacobians (linear approximations) of the measurement model. 

\section{Simulation Validation}\label{sec:filtering:simulationresults}

The wind estimator was first validated in simulation using \textit{RotorPy} \cite{folk2023rotorpy}. 
The unscented Kalman filter implementation was provided by the \textit{FilterPy} library\footnote{\url{https://github.com/rlabbe/filterpy}} and the following tuning parameters were used: $\alpha=0.001$, $\beta = 2$, $\kappa=-1$.
\textit{RotorPy} provided simulated noisy measurements of the inertial measurement unit, motion capture system, and motor speeds. 
In simulation, all the measurements were provided synchronized at a rate of 100 Hz. 

In order to test the performance of the filter across a variety of possible UAV configurations, the mass and aerodynamic coefficients were randomly sampled from uniform distributions with bounds reported in \autoref{tab:filtering:montecarlosimulations}.
These ranges were centered around the nominal identified values for an Astec Hummingbird \cite{svacha2017improving, neely2015hummingbirdparams}. 
For this preliminary validation study, $50$ randomized UAV parameters were selected. 
A calibration procedure was performed prior to each evaluation whereby the simulated UAV was flown at varying speeds along each axis and the aerodynamic coefficients were fitted from this data using least squares regression. 
In other words, the filter was given the ground truth UAV mass, but the drag coefficients were fitted to simulated flight data to mimic a real world setting. 
In each evaluation trial, the UAV was subjected to winds generated using an implementation of the Dryden gust model\footnote{\url{https://github.com/spencerfolk/rotorpy/blob/main/rotorpy/wind/dryden_winds.py}} with average magnitude $w_{avg}$ and turbulence intensity $\sigma_w$ randomly sampled following \autoref{tab:filtering:montecarlosimulations}. 

\begin{table}[]
\caption{Randomized UAV and wind parameters for the Monte Carlo evaluation of the wind filter. 
The symbols are consistent with those in \autoref{ch:dynamics}.}
\label{tab:filtering:montecarlosimulations}
\begin{center}
\begin{tabular}{ c | c c }
 \textbf{Parameter} & \textbf{Unit} & \textbf{Range} \textbf{(min-max)} \\ 
 \hline
 $m$  & kg & 0.375--0.9375 \\  
 $c_{Dx}$ & N-(m/s)$^{-2}$ & 0--1($10^{-3}$)  \\
 $c_{Dy}$ & N-(m/s)$^{-2}$ & 0--1($10^{-3}$) \\
 $c_{Dz}$ & N-(m/s)$^{-2}$ & 0--2($10^{-2}$) \\
 $k_d$ & N-rad-m-s$^{-2}$ & 0--1.19($10^{-3}$)\\
 $k_z$ & N-rad-m-s$^{-2}$ & 0--2.32($10^{-3}$) \\
 \hline 
 $w_{avg}$ & m-s$^{-1}$ & -3--3 \\ 
 $\sigma_w$ & m-s$^{-1}$ & 30--60
\end{tabular}
\end{center}
\end{table}

\begin{figure}[b!]
    \centering
    \includegraphics[width=0.5\columnwidth]{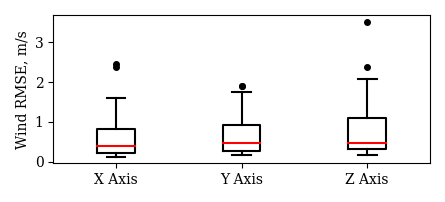}
    \caption{Monte Carlo evaluation of the \textit{Wind UKF} over $50$ simulations; each instance has randomized mass, drag coefficients, and average wind magnitudes.}
    \label{fig:filtering:simulationmonte}
\end{figure}

\autoref{fig:filtering:simulationmonte} summarizes the average root mean square error (RMSE) over the aforementioned randomized evaluations.
In half of the trials, the filter's RMSE falls around or under $0.5$ m/s; however, performance is poor in cases where the calibration procedure fails to find good drag coefficients, like when the ground truth drag coefficients are small. 
Whether or not the calibration procedure is successful, small drag coefficients adversely affect the filter's performance.
This is because according to an analysis presented in \autoref{appendix:uncertaintymodeling}, the \textit{a priori} measurement uncertainty associated with the wind is proportional to the ratio of the mass to the drag coefficient on a given axis. 

\begin{figure}[t]
    \centering
    \includegraphics[width=0.85\columnwidth]{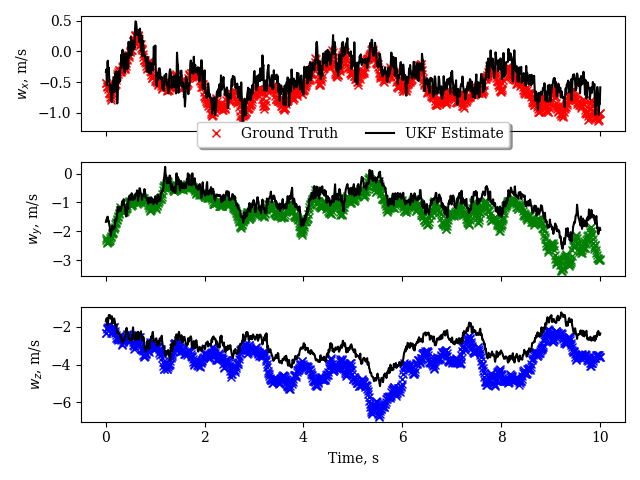}
    \caption{A simulated instance of the unscented Kalman filter estimating the local wind velocity vector for a UAV subject to Dryden wind gusts.}
    \label{fig:filtering:simulation_casestudy}
\end{figure}

\autoref{fig:filtering:simulation_casestudy} visualizes one instance of the $50$ trials, comparing the actual wind components to that estimated from the filter. 
The uncertainty of the prediction is omitted in these plots for clarity. 
Qualitatively speaking, there is good agreement between the predicted and actual wind on each axis, although there is bias \revision{in each axis} that is \revision{especially} apparent on the z axis.
\revision{The z axis bias remains a challenge for model-based wind estimation algorithms on rotary-wing UAVs because of modeling errors in predicting the accelerations due to rotor thrusts.}

\section{\revision{Summary}}

In this chapter, we derive a model-based wind estimation algorithm, \textit{Wind UKF}, that infers the wind at the UAV's immediate location.
The estimator jointly estimates the UAV's orientation, angular body rates, linear ground and wind velocities, and the motor speeds using measurements from the on board inertial measurement unit and an external motion capture for velocity and orientation\revision{; although outside of the lab the motion capture system could be replaced by GPS or existing visual odometry algorithms.} 
A nonlinear observability analysis demonstrates the theoretical conditions for observability that enable this filter. 
Monte Carlo evaluations of the filter in simulation on randomized UAV parameters provided validation of the filter and indicates estimation errors of around $0.5$ m/s on average. 
These simulation results provided the confidence for implementing the filter on a real platform.

The key contribution of this chapter is a means for estimating the wind at the UAV's immediate location without a dedicated wind sensor. 
This significance of this capability will become apparent in \autoref{ch:prediction} and \autoref{ch:realworld}, where this wind measurement will be an essential input for the local wind flow field prediction on a platform with extremely limited payload capacity. 
\chapter{Learned Local Wind Flow Field Prediction from Range Sensing}\label{ch:prediction}

\begin{contribution}
    This chapter includes material from the journal publication: \bibentry{folk2024flowdecoding}. The author of this thesis led the conceptualization, experimental design, analysis, and writing of the original manuscript.
\end{contribution}

Urban airspaces are a prime target for the deployment of autonomous aerial vehicles at scale. 
However, as discussed in \autoref{ch:introduction}, \revision{wind is a clear and present danger} to urban flight \revision{operations that are difficult to model.}
Buildings act as bluff bodies immersed in the wind, which result in wake structures with large turbulent fluctuations. 
Clusters of buildings create accelerated regions of the wind that act like artificial wind tunnels.
The sharp corners found on buildings create unique corner accelerations with strong shear layers which would pose a challenge even for a skilled pilot. 
The upstream and downstream regions of an immersed building have very little spatial correlation, making it difficult to apply kernel-based models to predict the wind. 

\revision{Thus,} it is \revision{important for UAVs capable of predicting urban winds, but} existing methods discussed in \autoref{sec:background:fieldestimation} suffer from requiring either privileged information about the map, a distributed network of sensors, or significant exploration of the environment.
This last point is important since there are many dangers that pose a risk to aerial operations in urban environments. 
In safety-critical situations it is imperative that we have methods to predict dangerous situations caused by the wind before the robot navigates into one. 

\begin{figure}[!ht]
    \centering
    \includegraphics[width=0.90\linewidth]{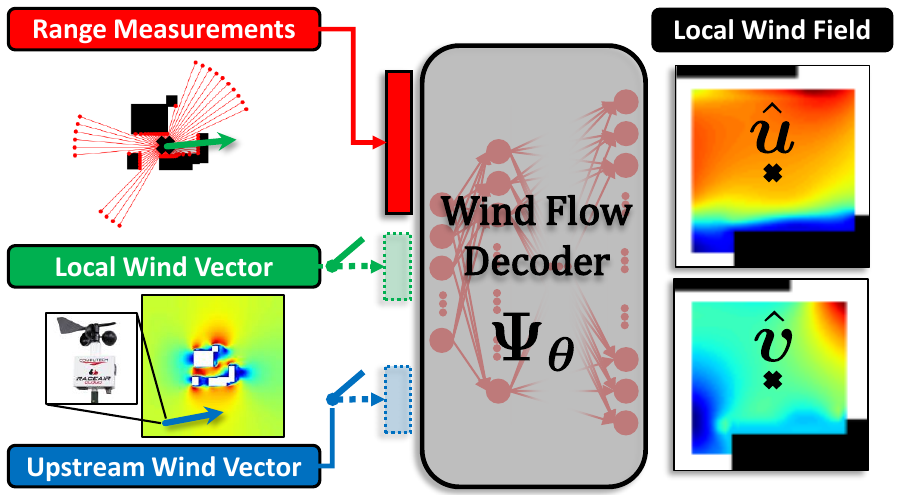}
    \caption{Method overview for predicting local wind vector fields in real time without a map of the environment. Buildings are sensed using range sensors and fed through a deep neural network alongside sparse wind measurements to predict the components of the time-averaged wind flow field.}
    \label{fig:prediction:method_diagram}
\end{figure}

Recognizing the shortcomings of current approaches to wind field estimation in urban environments, this chapter contributes the following: 
\begin{itemize}
    \item A new method for predicting surrounding wind flow vector fields without \textit{a priori} map knowledge or global navigation systems, capable of achieving real-time\footnote{200 Hz inference on an Intel(R) Xeon(R) CPU E5-2698 v4 @ 2.20GHz} performance with on board sensors and compute. 
    \item Validation, benchmarking, and applications of a Python-based staggered grid fluid solver for scalable dataset generation of 2D urban winds.
\end{itemize}
Our method, seen in \autoref{fig:prediction:method_diagram}, uses a deep neural network (DNN) to synthesize sparse wind measurements from an upstream weather station \textit{and/or} on board \revision{wind} sensor with a scan of nearby buildings provided by a range sensor, \revision{where the task is to} to predict the wind vector field on a grid centered at the robot.
A fast unsteady 2D fluid solver was implemented and validated to generate thousands of physically plausible wind patterns around randomized urban environments, which were used to train a model that generalizes to previously unseen maps.

This approach is an important step forward towards real-time reasoning of surrounding wind flow fields utilizing information about immediate geometry, all while remaining generalizable to arbitrary urban environments.
Herein we present extensive studies in the form of Monte Carlo simulations to understand the performance and limitations of our localized wind prediction approach, which will provide a useful scaffolding for future applications to UAV planning and control in windy urban environments.

\section{Unsteady Staggered Grid Fluid Simulations}\label{sec:prediction:euler_sim}

To generate a large database of urban winds, we implemented a 2D staggered grid \revision{solver in Python\footnote{\url{https://github.com/spencerfolk/pystaggerflow}}} following the algorithm pioneered by Stam \cite{stam2023stable} and accelerated by \texttt{Numba} to rapidly simulate mass- and momentum-conserving winds between arbitrary building arrangements. 
This approach to unsteady fluid simulation has seen widespread adoption in the computer graphics community because of its speed, robustness, and computational stability.
We were motivated to implement this algorithm because:
\begin{enumerate}
    \item There is currently a dearth of unsteady wind simulators that natively integrate with existing aerial robot dynamics simulators, especially in Python \cite{dimmig2023survey}.
    \item A fast solver capable of representing arbitrary obstacles accelerates the creation of large domain-randomized training datasets and promotes experimentation. In our solver, buildings are represented using a cell masking scheme, which enables automated 2D mesh generation for arbitrary building planform layouts. 
    \item The stability and robustness of Stam's algorithm eliminates any need for individual case tuning, which greatly accelerates the creation of training data.
\end{enumerate}

\begin{table}[b!]
\caption{Run times for our fluid solver on varying grid sizes compared to our network on an Intel Core i5 @ 2.60 GHz.}
\large
\label{tab:prediction:cpu_timing}
\resizebox{\columnwidth}{!}{%
\begin{tabular}{p{0.5\columnwidth}cc}
\hline
\textbf{Map} & \textbf{CPU Time Per Timestep (s)} & \textbf{Real Time Factor} \\ \hline
Training Map \\ ($500$$\times$$500$ m) & 0.378 & 0.66 \\
Evaluation Map \\ ($1{\small,}200$$\times$$1{\small,}200$ m) & 1.606 & 0.16 \\ \hline
Prediction Domain \\ ($25$$\times$$25$ m) & 0.019 & 12.78 \\ \hline
Wind Flow Decoder \\ ($25$$\times$$25$ m) & 0.005* & - \\ \hline
\end{tabular}%
}
\begin{minipage}{\columnwidth}
\small
 
\raggedright*The average time for one forward pass through our network.
\end{minipage}
\end{table}

In \autoref{tab:prediction:cpu_timing}, we report single-core run times on a laptop for the staggered grid solver for both training and validation maps (see \autoref{sec:prediction:learning_algorithm} for representative examples). 
On training maps with a timestep of $\Delta t = 0.25$ m and 1 m grid resolution, the solver has a real time factor of 0.66, which is approximately 200 times faster than STAR-CCM+\footnote{https://plm.sw.siemens.com/en-US/simcenter/fluids-thermal-simulation/star-ccm/} with default settings.
These real time factors, coupled with parallelization on an \texttt{NVIDIA DGX}, enabled the generation of 53 hours of wind simulations in around 6 hours of wall time.

For validation, we compared flow around rectangular blocks at varying incidence angles with published experimental data. 
\autoref{fig:prediction:euler_strouhalangle} compares the trends in Strouhal number, computed from a Fourier analysis of the rectangles' vertical force histories, for rectangles of aspect ratio $1$$\times$$1$, $2$$\times$$1$, and $3$$\times$$1$ at angles of attack between 0$^{\circ}$ and 90$^{\circ}$ against experimental data (\cite{norberg1993strouhal}, Figure 6).    
The staggered grid solver accurately captures the symmetric peaks at low and high angles, and a plateau for moderate angles of incidence.
\autoref{fig:prediction:euler_strouhalaspect} shows a comparison of Strouhal numbers for a wide variety of rectangle aspect ratios at 0$^{\circ}$ angle of attack compared with experimental data from Nakaguchi \textit{et al} \cite{nakaguchi1968experimentalcylinder}.  
The trend of decreasing Strouhal number with increasing aspect ratio and the break that occurs for aspect ratios greater than 2.0 are both represented in our solver.
Lastly, qualitative comparisons to STAR-CCM+ for both isolated and grouped rectangles are found in \autoref{fig:prediction:euler_unsteadyvortices}, finding good agreement in both unsteady and time-averaged simulations.

\begin{figure}[t!]
  \centering
  \begin{subfigure}{0.32\textwidth}
    \centering
    \includegraphics[width=\textwidth]{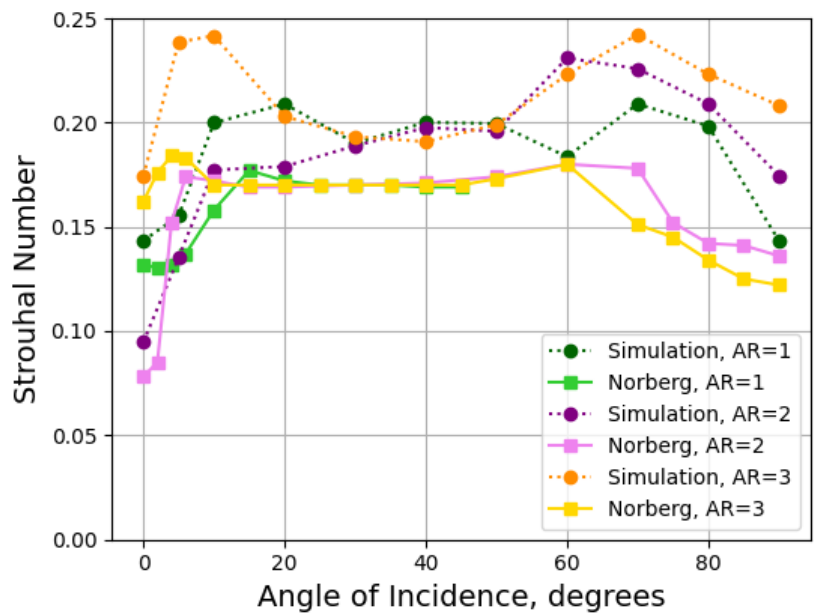}
    \caption{}
    \label{fig:prediction:euler_strouhalangle}
  \end{subfigure}
  \begin{subfigure}{0.32\textwidth} 
   \centering
    \includegraphics[width=\textwidth]{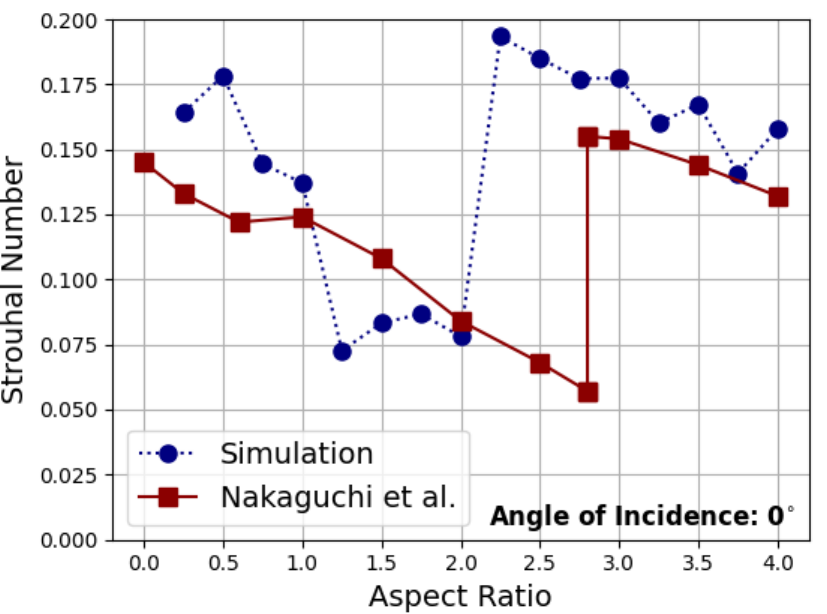}
    \caption{}
    \label{fig:prediction:euler_strouhalaspect}
  \end{subfigure}
  \begin{subfigure}{0.32\textwidth}
    \centering
    \includegraphics[width=0.9\textwidth]{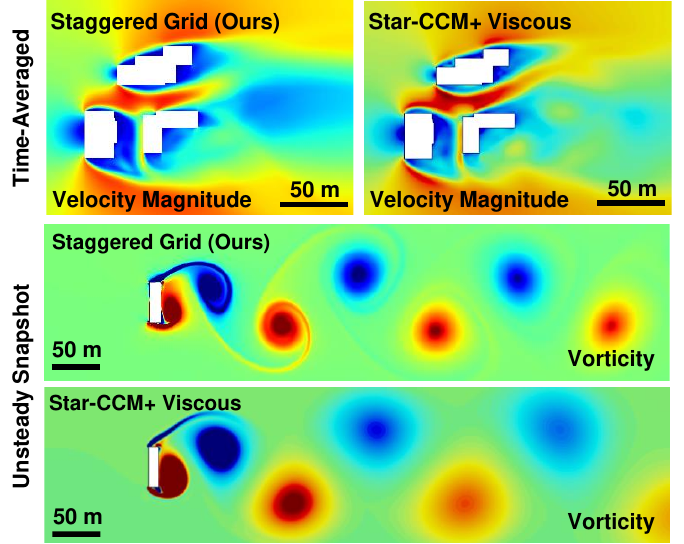}
    \caption{}
    \label{fig:prediction:euler_unsteadyvortices}
  \end{subfigure}
  \caption{(a) Strouhal numbers for isolated rectangles of various shapes and onset flow angles compared with experimental data \cite{nakaguchi1968experimentalcylinder, norberg1993strouhal}. (b) Strouhal numbers for various aspect ratios compared with experimental data \cite{nakaguchi1968experimentalcylinder, norberg1993strouhal}. (c) Qualitative comparisons of our fluid solver against commercial solutions for isolated (bottom) and grouped (top) rectangles.}
\end{figure}

These comparisons indicate that fluid flows from our simulator are not too dissimilar from those observed experimentally for immersed rectangles, which are reasonable 2D models of buildings subject to wind.
\revision{The buildings in our simulations were nominally around $25$ m in size with freestream velocities around $5$ m/s.}
\revision{This means that a nominal Strouhal number of $0.15$ corresponds to vortex shedding frequencies of around $3$ Hz or $12$ simulation timesteps.}
\revision{Combined with the fact that our solver runs $300$ inner solver iterations \textit{within} each timestep, we are confident that the solver is able to converge within the time associated with primary flow unsteadiness.}

\begin{figure*}[!ht]
  \centering
  \includegraphics[width=0.95\textwidth]{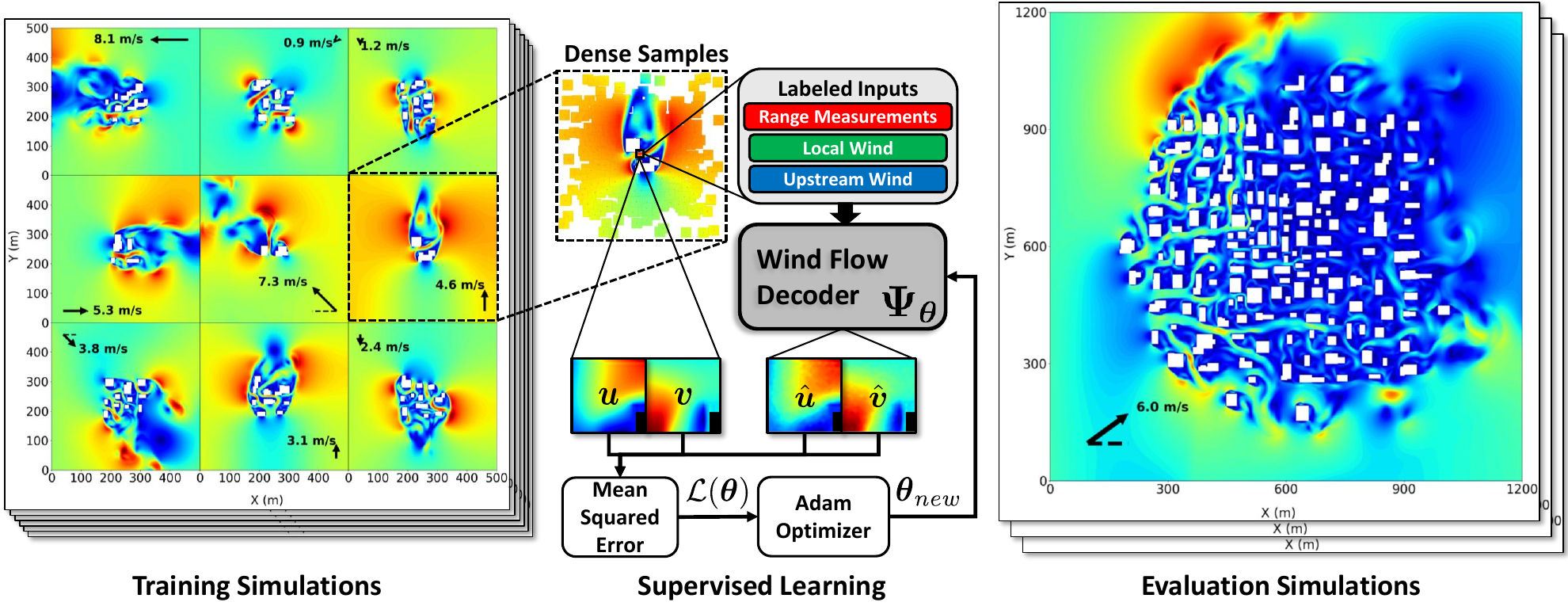}
  \caption{An overview of the data acquisition, training, and evaluation of a wind flow decoder model. Thousands of wind simulations on small maps are densely sampled (left), and the labeled pairs are used to train the model via supervised learning (middle). The model is evaluated on randomly sampled locations in much larger maps (right) to measure generalization.}
  \label{fig:prediction:training_diagram}
\end{figure*}

\section{Deep Learning-Based Local Wind Prediction}\label{sec:prediction:learning_algorithm}

Our approach synthesizes nearby contextual information provided by range measurements with sparse measurements of the wind to predict the wind velocity vector field in a region around the robot.
The algorithm is depicted in \autoref{fig:prediction:method_diagram}. 
A robot equipped with a LiDAR scanner is provided distance measurements to nearby buildings in a 360$^\circ$ field of view. 
The slice of the LiDAR scan at the robot's altitude is denoted as $\boldsymbol{D} \in \mathbb{R}^{n_r}, D_{min} \leq D_i \leq D_{max}$ where $n_r$ is the number of range measurements determined by the LiDAR's field of view and angular resolution. 
The 2D wind flow decoding task is defined as finding a mapping $\boldsymbol{\Psi}$ such that
\begin{equation}\label{eq:prediction:mapping}
    \begin{bmatrix} \bar{\boldsymbol{u}} \\ \bar{\boldsymbol{v}} \end{bmatrix} = \boldsymbol{\Psi}(\boldsymbol{D}, \boldsymbol{W}_{\!\!up}, \boldsymbol{W}_{\!\!robot})
\end{equation}
where $\boldsymbol{W}_{\!\!up} \in \mathbb{R}^2$ is a single upstream measurement of the freestream wind; $\boldsymbol{W}_{\!\!robot}\in \mathbb{R}^2$ is a measurement of the wind at the robot's location (as in \autoref{sec:introduction:background:comestimation}); $\bar{\boldsymbol{u}}$ and $\bar{\boldsymbol{v}}$ are discretized time-averaged velocity fields for the $x$ and $y$ directions, respectively.

\begin{figure}[t!]
    \centering
    \includegraphics[width=0.99\textwidth]{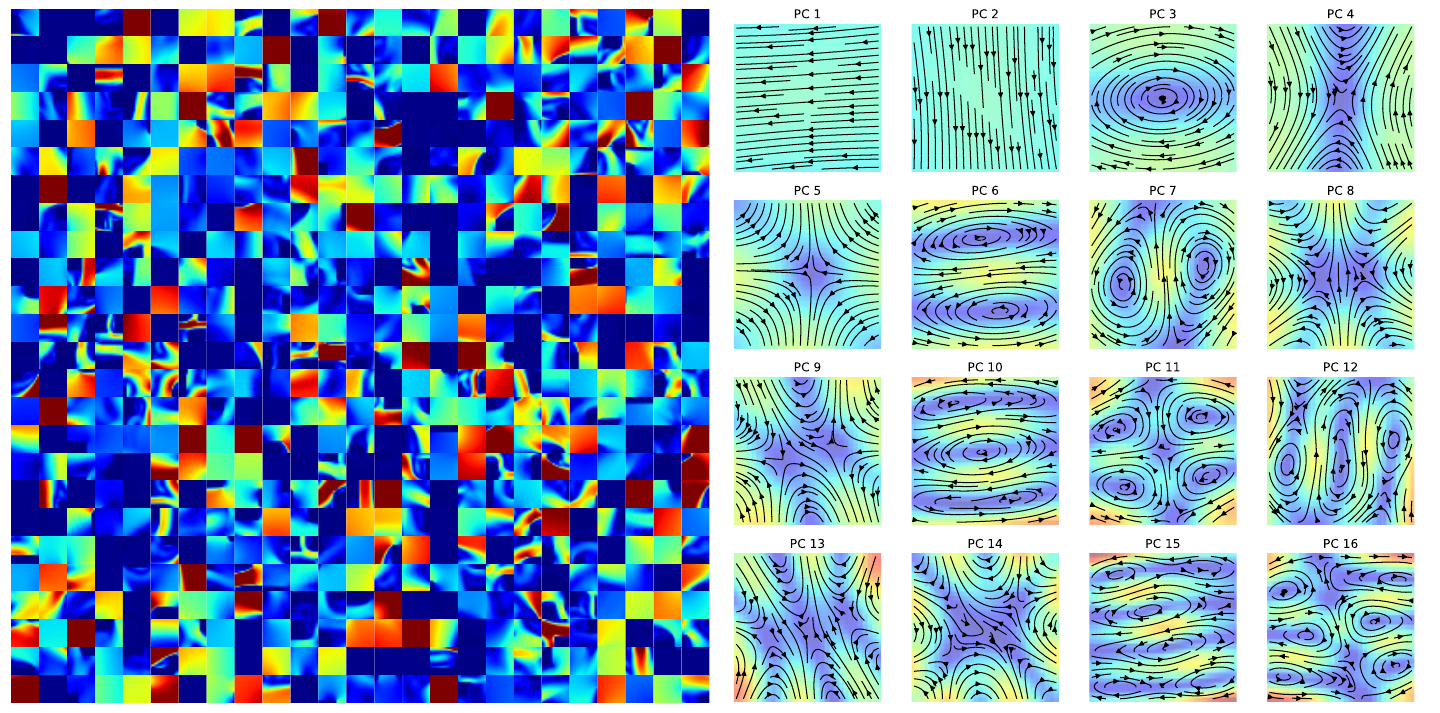}
    \caption{Because of the large spatial scale of urban flow features, local croppings of urban flow fields can actually be thought of as evolving on a much lower dimensional manifold composed of elementary flows. The $30$$\times$$30$ m local flow fields on the left can be reproduced by unique linear combinations of the principal components from the right, some of which mimic fundamental flows in fluid mechanics such as uniform flow, vortex pairs, and shear flow.}
    \label{fig:prediction:flow_pca_components}
\end{figure}
\begin{figure}[t!]
    \centering
    \includegraphics[width=0.7\textwidth]{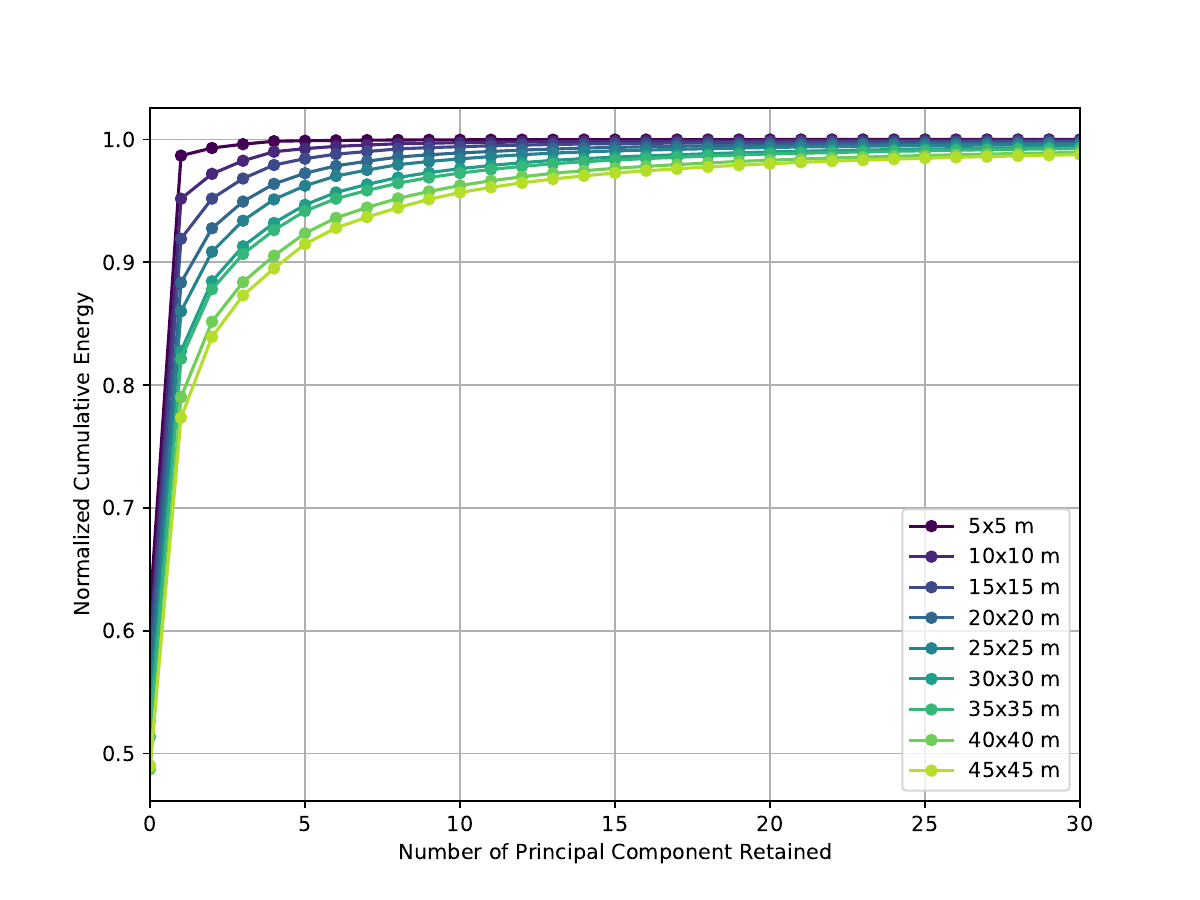}
    \caption{Cumulative explained variance as a function of the number of components retained from principal component analysis. As the region size grows, the local flow fields become more complex and therefore more components are required to recover the training data statistics.}
    \label{fig:prediction:pca_explained_variance}
\end{figure}

We refer to this mapping $(\boldsymbol{\Psi})$ as a \textit{wind flow decoder} because we liken the combination of LiDAR and sparse wind data \revision{as constituting} a lower-dimensional encoding of the local wind field.
The intuition behind this is the fact that buildings impose conservation of mass and momentum conditions on the wind, and the sparse wind measurements inform the approximate direction and magnitude of the wind flow field. 
In dense urban environments, the boundary conditions imposed by buildings promote predictable structures in the local winds like corner accelerations and channel flows.
In fact, the existence of predictable flow structures mean that the local flow fields can be thought of as evolving on a lower dimensional manifold composed of corner accelerations, channel flows, and more. 
This idea is better illustrated visually in \autoref{fig:prediction:flow_pca_components}, where a principal component analysis on the local flow field training data reveals fundamental structures in the flow. 
A small sample of the $30$$\times$$30$ m local flow fields used in the principal component analysis are shown on the left while the resulting ``Eigen flows'' are shown on the right. 

Principal component analysis was repeated for region sizes ranging from $5$$\times$$5$ m up to $45$$\times$$45$ m, and similar elementary flow structures were recovered. 
As shown in \autoref{fig:prediction:pca_explained_variance}, \revision{as little as} $16$ components explain at least $98\%$ of the variance in the training data--even less are required as the size of the prediction region shrinks.
These components represent orthogonal modes in the fluid flow space that can be linearly combined to reconstruct any of the flow fields on the left.
This analysis provides partial insight into why the learning problem in \autoref{eq:prediction:mapping} is perhaps feasible: the learning target (local flow fields) evolves on a surprisingly low dimensional space.

\subsection{Dataset Generation}\label{sec:prediction:dataset}

Using the solver from \autoref{sec:prediction:euler_sim}, we simulated winds through $64$ randomly generated maps: $32$ on a $500$$\times$$500$ m domain and $32$ on a $300$$\times$$300$ m domain.
The number of buildings in a map varied between $1$ and $10$, and generally the spacing between each building was at least $10$ m to represent streets.
For each map we simulated inlet winds of $[0.1, 2.575, 5.05, 7.525, 10.0]$ m/s, each coming from the cardinal (N, S, W, E) and intercardinal (NE, NW, SW, SE) directions, totaling $40$ inlet conditions per map. 
The wind was simulated on a $1$ m grid with a $0.25$ \revision{sec} timestep for $100$ initial timesteps for flow development, and then the wind was averaged over a further $200$ timesteps representing $50$ \revision{sec} of wind. 
The \revision{discretized time-averaged arrays generated from the flow solver} were then spatially interpolated using \texttt{SciPy RectBivariateSpline} \revision{when sampling training data.}

Examples of the training simulations can be seen in \autoref{fig:prediction:training_diagram}.
For each simulated instance, we randomly placed the robot in $2{\small,}048$ locations around the buildings and collected noisy range measurements $\tilde{\boldsymbol{D}}$, the ground truth freestream velocity $\boldsymbol{W}_{\!up}$, the ground truth local velocity $\boldsymbol{W}_{\!robot}$, and the local time-averaged $u$- and $v$-components of the wind on a grid around the robot.
The noisy range measurements were constructed by casting rays in a $360$$^{\circ}$ field of view at $1$$^{\circ}$ angular resolution, clipping the measurements between [$0.25$, $100$] m and corrupting them with Gaussian noise ($\sim\mathcal{N}(0, 0.35)$ m).

We also generated an entirely separate evaluation dataset consisting of just $10$ maps, but each map's domain is $1.2$$\times$$1.2$ km in size and has upwards of $200$ or more buildings (e.g. \autoref{fig:prediction:training_diagram}, right).
One evaluation map was separated from the dataset and used to compute a validation loss during training.
The function of the evaluation dataset is two-fold: to test the model on wind flow fields more similar to dense urban environments, and to conduct an aggressive distribution shift from the training data (smaller maps with far fewer buildings) to challenge the performance and generalization of our approach while highlighting any overfitting or training biases that may occur. 
For the evaluation maps, we simulated $8$ random inlet conditions using the same simulation setup and recorded the time-averaged wind at $2{\small,}048$ random locations around buildings, totaling $147{\small,}456$ labeled evaluation samples with which we measure the performance of our models.

\subsection{Model Selection}\label{sec:prediction:model_selection}

The wind flow decoder was parameterized by a set of weights, $\boldsymbol{\theta}$, in a fully-connected deep neural network, $\boldsymbol{\Psi}_{\boldsymbol{\theta}}$.
Deep neural networks were chosen because 1) they constitute an extremely powerful class of hypothesis functions; 2) they are memory efficient and fast to run inference; and 3) they excel at extracting complex features and patterns from high dimensional data such as LiDAR. 
Also, we observed that networks were able to predict flow around the corner of buildings--in such a situation, the LiDAR scan is occluded by the building, but the network leverages patterns from the training dataset to infer what wind flow around it may look like.
This is an advantage over an analytic model of $\boldsymbol{\Psi}$, which would need some heuristic to handle these cases. 

In order to represent \autoref{eq:prediction:mapping} with a DNN we discretize $\bar{\boldsymbol{u}}$ and $\bar{\boldsymbol{v}}$ on a square grid centered at the robot with length $H$ and spatial resolution $\Delta$.
The ratio of the prediction area to the LiDAR's perception area is: 
\begin{equation}\label{eq:prediction:area_ratio}
    \frac{A_{pred}}{A_{total}} = \frac{1}{\pi} \left ( \frac{H}{D_{max}} \right)^2
\end{equation}
Unless otherwise noted, our method uses $\Delta=1$ m, chosen to balance potential downstream motion planning tasks with prediction accuracy.
To prevent overfitting and limit compute and memory requirements, we elected to keep the model relatively small at three hidden layers, each with 2048 neurons. 

\subsubsection{Training and loss function}\label{sec:prediction:training_loss}
The unsteady wind simulations (\autoref{sec:prediction:euler_sim}) provided labeled data for supervised learning using \texttt{PyTorch}. 
Mini-batch training was employed with a batch size of $8{\small,}192$ for favorable training stability and more efficient use of available GPU resources. 
The inputs were scaled to have a magnitude near $1$ using a constant normalization factor. 
After each training epoch, the models were validated on the holdout validation map.
All models were trained for $10$ epochs at which point the validation losses no longer decreased, taking an average of $2.2$ hours each on an \texttt{NVIDIA Tesla V100-SXM2} with $32$ \texttt{GB} of RAM.

The loss function was the mean squared error (MSE) with weight regularization parameterized by $\lambda$: 
\begin{equation}
    \label{eq:prediction:loss}
    \mathcal{L}(\boldsymbol{\theta}) := \frac{1}{N_b} \sum_{i=1}^{N_b} ||\boldsymbol{z}_i - \hat{\boldsymbol{z}}_i||_2^2 + \frac{\lambda}{2}||\boldsymbol{\theta}||_2^2
\end{equation}
where $\boldsymbol{z}_i = [\texttt{flatten}(\bar{\boldsymbol{u}}_i), \texttt{flatten}(\bar{\boldsymbol{v}}_i)]^\top$ is the flattened ground truth velocity component fields sampled on a grid around the robot, $\hat{\boldsymbol{z}}$ (defined similarly) is the network prediction, and $N_b$ is the batch size.
In this work, we set $\lambda = 0.002$.
During preliminary investigations we considered a loss on the divergence of the predicted flow field; however, we observed that models trained on \autoref{eq:prediction:loss} already produced low-divergence velocity fields, and computing the divergence of each prediction slowed training.

\subsection{Baseline Local Wind Models}\label{sec:prediction:baselines}

For points of comparison in the results to follow, we consider two baseline models for local wind prediction. 
In both baselines, it is assumed that the \textit{local} surrounding wind flow field is completely uniform. 
\textbf{Upstream Baseline} uses $\boldsymbol{W}_{\!up}$, the time-averaged measurement of the wind from the upstream weather station, whereas \textbf{Local Baseline} uses $\boldsymbol{W}_{\!robot}$, the time-averaged measurement of the wind at the robot's location, which in practice would be obtained via methods discussed in \autoref{sec:introduction:background:comestimation}, as the uniform wind velocity.

\begin{figure*}[!ht]
    \centering
    \includegraphics[width=0.86\textwidth]{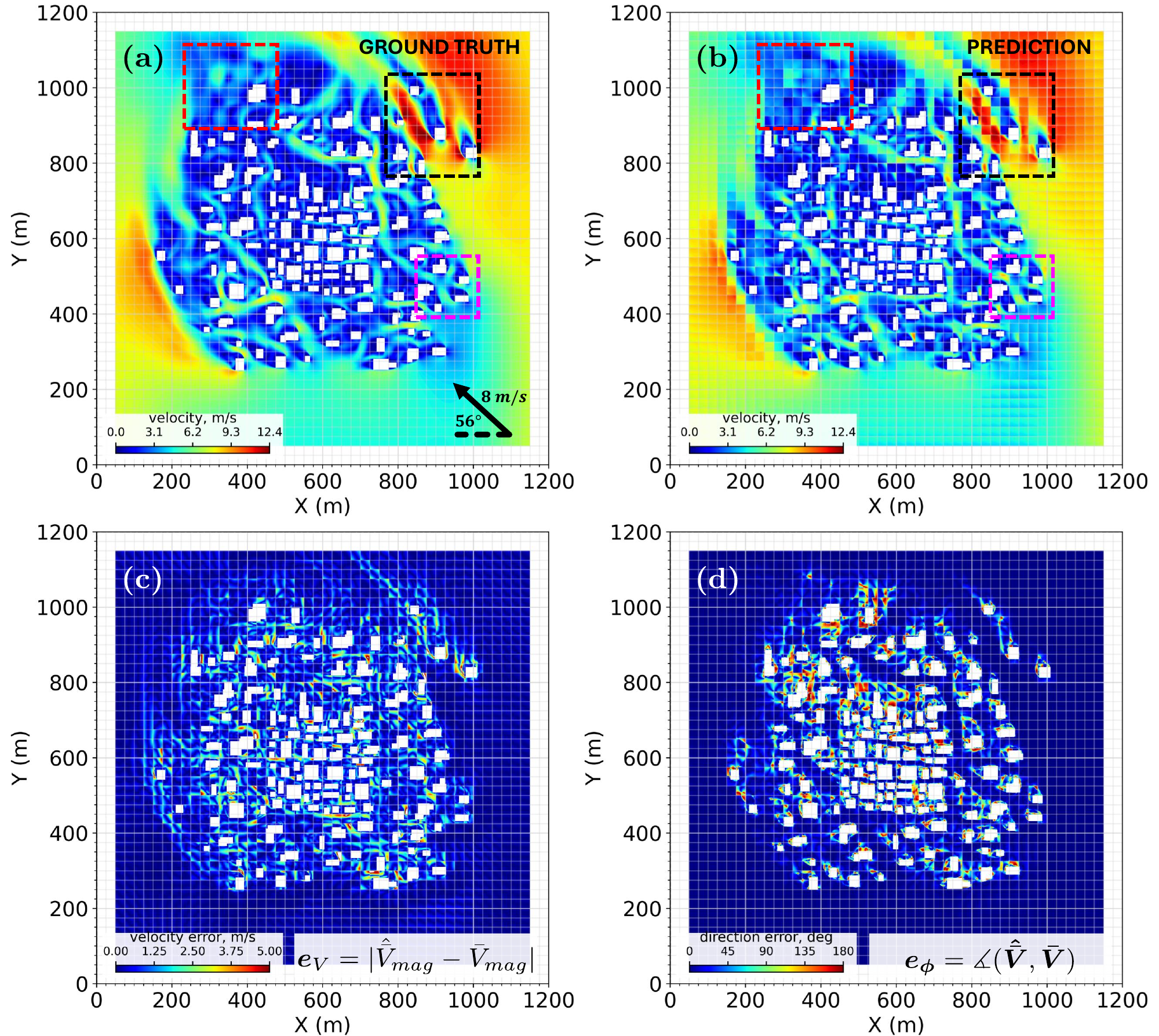}
    \caption{A global reconstruction of an urban wind flow field time-averaged over $50$ seconds, generated by translating the robot in and around an evaluation map and overlapping $25$$\times$$25$ m predictions: (a) ground truth velocity magnitude; (b) reconstructed velocity magnitude; (c) velocity magnitude error; (d) velocity direction error. Points of interest for comparison are marked in dashed boxes highlighting strong shears (black), city wakes (red), and intersecting (pink) channel flows.}
    \label{fig:prediction:mosaic}
\end{figure*}

\section{Experimental Results}\label{sec:prediction:experiments}

\subsection{Global Reconstruction}\label{sec:prediction:global_reconstruction}

In this scenario, a robot equipped with a simulated range sensor moves around one of the evaluation maps producing predictions on a regular interval. 
As the robot traverses, it captures noisy scans of nearby building geometry and it is provided ground truth measures of the wind both upstream ($\boldsymbol{W}_{\!up}$) and at the robot ($\boldsymbol{W}_{\!robot}$). The wind flow decoder with $H=25$ m outputs predictions of the $u$- and $v$-components of the local wind flow field around the robot with $1$ m spatial resolution. 
We emphasize that during training, the loss function (\autoref{eq:prediction:loss}) only applies to the $25$$\times$$25$ m \textit{local} velocity prediction, so there is no encouragement during training to produce predictions that ``stitch'' together in this manner.

In \autoref{fig:prediction:mosaic} the local predictions for one of the evaluation maps are tiled on the map and compared to the ground truth wind flow field. 
Interestingly, despite the model being trained to produce \textit{local} measurements, \autoref{fig:prediction:mosaic}b indicates that the model can produce coherent \textit{global} structures of the wind through the city that extend outside the $25$$\times$$25$ m prediction region.
In other words, larger phenomena such as the outer layer shears (black box), city wakes (red box), and intersecting channels (pink box) are predicted by the model despite having \textit{no knowledge} of the buildings beyond the robot's perception radius ($100$ m). 
These visualizations highlight the consistent spatial representation our model learns for the time-averaged wind around buildings, such that independent local predictions can be composed to predict larger flow structures.

\autoref{fig:prediction:mosaic}c shows the absolute error in velocity magnitude ($V$), which is computed as $e_{V} = |\hat{\bar{\bl{V}}} - \bar{\bl{V}}|$, overlaid on the map. 
Most of the prediction errors fall within $2$ m/s and are \revision{localized} either in building wakes or near the strong shear layers on \revision{the outskirts} of the city.

\autoref{fig:prediction:mosaic}d shows the direction error, computed as the angle between the actual wind vector field and the predicted vector field, $e_\phi = \measuredangle (\boldsymbol{\hat{\bar{V}}}, \boldsymbol{\bar{V}}) = \texttt{arccos}(\boldsymbol{\hat{\bar{V}}}\cdot \boldsymbol{\bar{V}}/||\boldsymbol{\hat{\bar{V}}}|| ||\boldsymbol{\bar{V}}||)$. 
Generally the network's prediction degrades in building wakes--an extremely challenging region where vortices cause sharp transient redirection of the wind--but the prediction is better in sparser regions of the city that are mostly dominated by channel flows and shear layers. 

\begin{figure}[b!]
    \centering
    \includegraphics[width=0.8\textwidth]{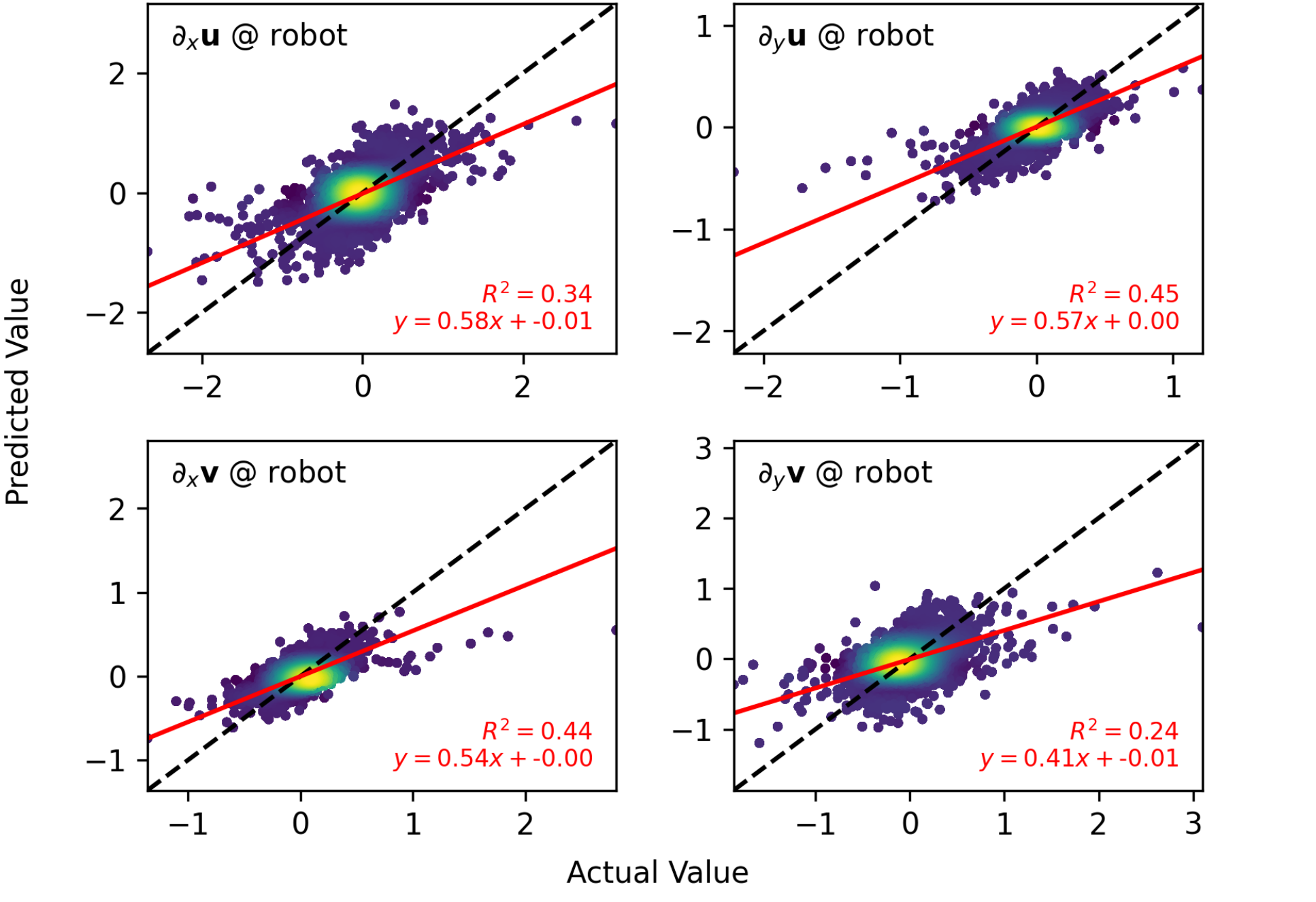}
    \caption{Density scatter plots comparing between the predicted and actual spatial gradients of the wind flow field at the robot's location for a model with $H=25$ m given both local and upstream wind measurements.}
    \label{fig:prediction:gradients}
\end{figure}

\subsection{Gradient Reconstruction}\label{sec:prediction:local_reconstruction}

Spatial gradients ($\partial_x \boldsymbol{u}$, $\partial_y \boldsymbol{u}$, $\partial_x \boldsymbol{v}$, $\partial_y \boldsymbol{v}$) may be important quantities for downstream tasks, such as motion planning or control.
\autoref{fig:prediction:gradients} compares predicted velocity gradients at the robot's location to their labels for the same model as above.
A least squares linear regression is used to compare the predicted and true values.
The coefficient of determination, $R^2$, is also reported to measure the quality of the fit.
The gradients were computed by using central differences on the predicted $\bar{\boldsymbol{u}}$- and $\bar{\boldsymbol{v}}$-components, which may amplify noise.
Also, outliers were not ignored in computing these fits.

\begin{figure}[b!]
    \centering 
    \includegraphics[width=0.7\textwidth]{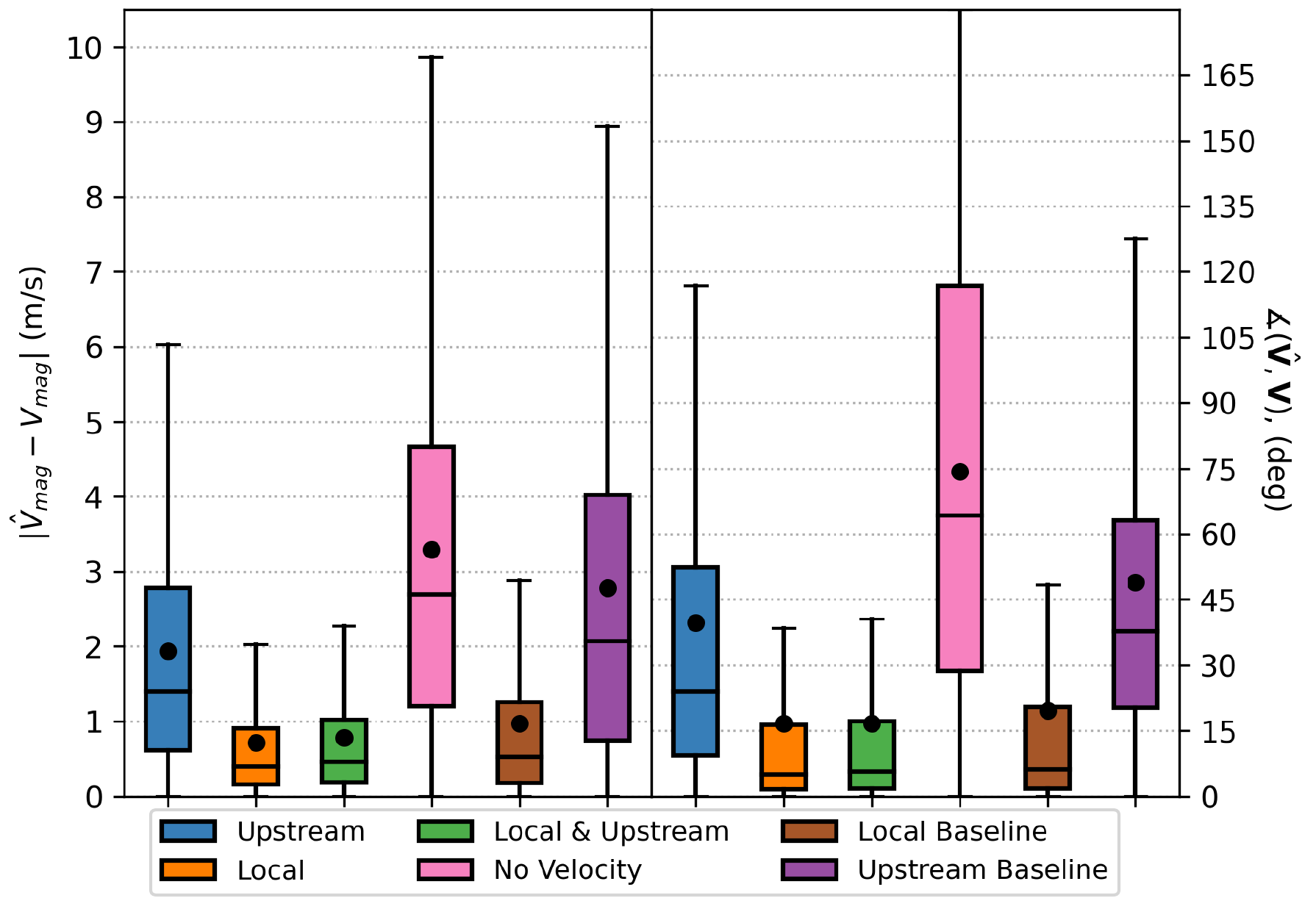}
    \caption{The pointwise absolute error distributions for velocity magnitude (left) and direction (right) for $H=25$ m grid size models with different feature inputs, compared to baselines that assume a uniform flow field. Dots indicate the mean.}
    \label{fig:prediction:ablation_study}
\end{figure}

\subsection{Ablation Study}\label{sec:prediction:ablation_study}

The wind flow decoder \revision{as demonstrated here} utilizes two velocity \revision{sources--local and upstream measurements--which} prompts the question: how important are these measurements?
To answer this, we trained three additional models all with $H=25$ m and $1$ m resolution grid sizes; however, these models omit one or both measurements.
The resulting performance on the evaluation dataset, summarized in \autoref{fig:prediction:ablation_study}, clearly demonstrate that the local velocity measurement is a necessary component for accurate prediction, reducing the average error by over $100$\% compared to the model only given an upstream measurement.
On average, our best performing model outperforms the \textbf{Local Baseline} and \textbf{Upstream Baseline} by $23$\% and $71$\%, respectively, with an average velocity error of $0.77$ m/s and average direction error of $15$$^\circ$.
The network with no velocity information is easily the worst performing network, justifying the importance of the sparse velocity measurements.

\subsection{Prediction Size Study}\label{sec:prediction:size_study}

The task defined in \autoref{eq:prediction:mapping} applies generally to vector fields of arbitrary size, but in practice the grid size needs to be selected based on the desired performance and downstream tasks. 
To investigate how our method's performance varies with the prediction size, four additional models were trained to predict the wind vector field on $5$$\times$$5$, $15$$\times$$15$, $35$$\times$$35$, and $45$$\times$$45$ m grid sizes with a fixed $1$ m resolution.
All models had access to both velocity measurements.
In \autoref{fig:prediction:size_study}, the average absolute velocity magnitude direction errors on the evaluation dataset are reported against the ratio of the prediction area to the LiDAR perception area, $A_{pred}/A_{total}$ (\autoref{eq:prediction:area_ratio}).
In the limit of small grid sizes, our learned model is comparable to \textbf{Local Baseline}, but with increasing grid sizes the network outperforms it. For example, at $H=25$ m the neural network is $26$\% and $22$\% more accurate for direction and magnitude on average.

\begin{figure}[t!]
    \centering
    \includegraphics[width=0.7\textwidth]{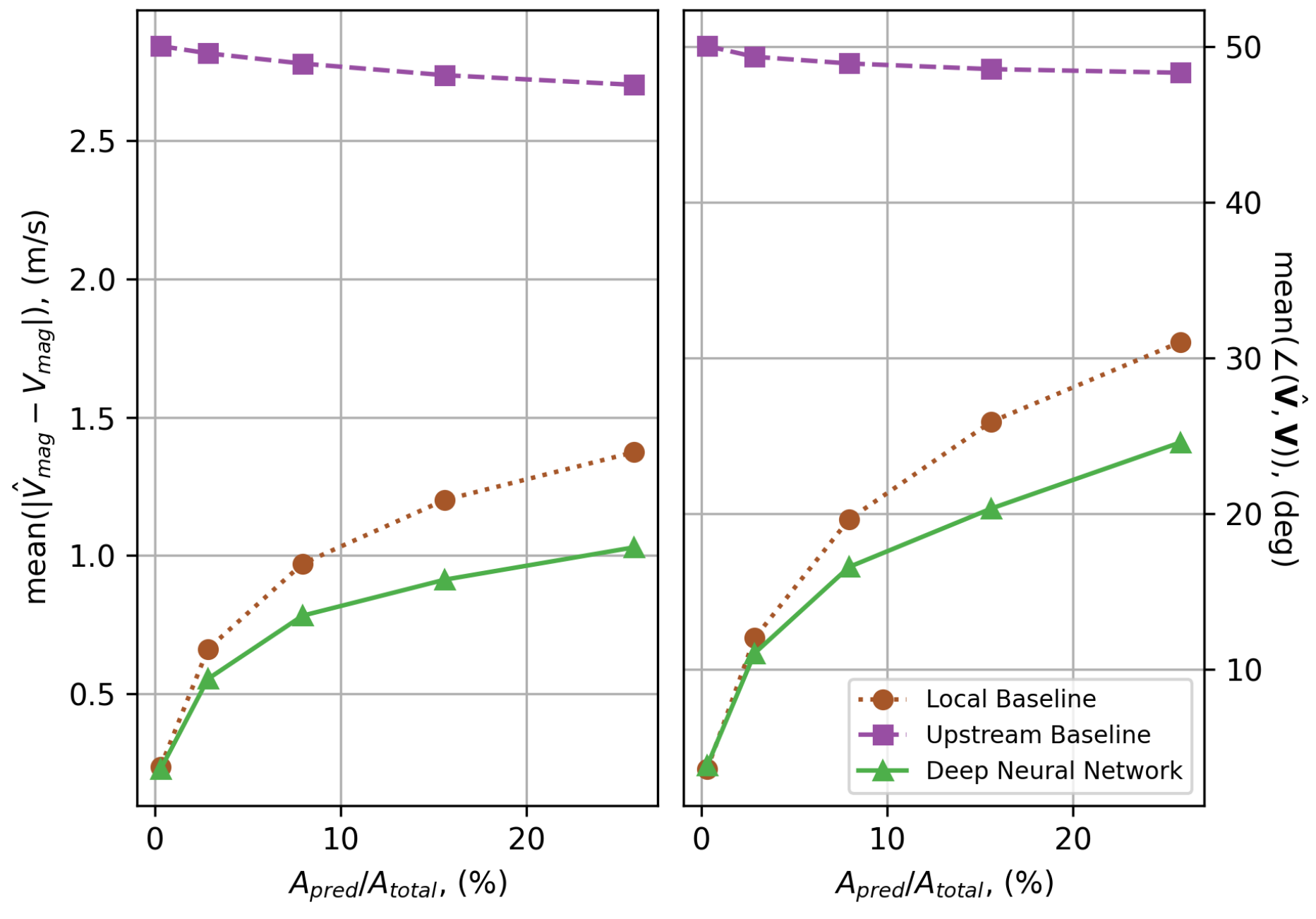}
    \caption{The average absolute error in velocity magnitude (left) and direction (right) for models with varying prediction grid sizes ($H$), reported as the ratio of the prediction area to the maximum LiDAR perception region.}
    \label{fig:prediction:size_study}
\end{figure} 

\section{Discussion}\label{sec:prediction:discussion}

This study presents evidence that it is feasible to infer local wind vector fields from range measurements.
Our rigorous experiments demonstrate that our deep learning-based model is robust to robot translation and capable of learning wind structures larger than its prediction grid size--desirable qualities for downstream motion planning.
However, one caveat is that the network tends to smooth out spatial gradients at the robot's location, which may be an artifact of using the MSE loss thus motivating \revision{instead using} a ``balanced'' MSE loss \cite{achermann2024windseer}.
Another idea is to train separate networks each focusing on different features of the wind like spatial gradients, but also maximum expected winds, vorticity, turbulence statistics, and more. 

The ablation study in \autoref{sec:prediction:ablation_study} revealed that including the local velocity measurement improves performance by over 100\% compared to the model with only upstream measurements. 
This may be evidence of the idea that in some cases the local wind field is decorrelated with the farfield velocity.

Finally, the grid size study in \autoref{sec:prediction:size_study} quantifies how the errors grow with the prediction window size (and therefore the task difficulty).
In the context of our uniform baselines, the results are consistent with the intuition that the flow field will look more uniform as the prediction window shrinks. 
Nevertheless, even for prediction windows as large as $45$$\times$$45$ m (a practical limitation of our computational resources) the average velocity magnitude error is around $1$ m/s which is comparable to other learning-based methods.

These results open up many exciting research questions.
The present algorithm neglects temporal aspects of the flow, but can this be addressed by operating in the frequency domain?
Also, while currently the loss is simply the mean squared error, a divergence loss was considered--would other physics-informed loss terms improve sample efficiency and generalization of the network?

\section{\revision{Summary}}\label{sec:prediction:conclusion}

In this chapter, we present a novel method for predicting surrounding wind flow fields in real time with applications to UAV and UAM path planning. 
Our method uses range measurements to construct a representation of nearby physical structures, which inherently impose boundary conditions on the wind flow field. 
A fast staggered grid fluid solver based on computer graphics code enabled rapid and unsupervised training data collection at a scale necessary for deep learning.
Experiments include numerous evaluations of eight models in total, each with different levels of knowledge or varying prediction grid sizes, on randomly generated city districts.
We find evidence to suggest that sparse wind measurements, specifically the wind at the robot's location, combined with a LiDAR scan of urban environments provides sufficient information for a data-driven model to characterize the average wind flow field around the robot. 
We found that for a prediction grid size of $25$$\times$$25$ m, the best performing model was able to achieve an average velocity error of $0.77$ m/s and direction error of $15$$^\circ$.
\chapter{Wind-Aware Motion Planning and Control}\label{ch:planning}

\begin{contribution}
    This chapter includes material from the conference publication: \bibentry{folk2025windplanning}. The author of this thesis led the conceptualization, experimental design, analysis, and writing of the original manuscript.
\end{contribution}

\begin{figure}[b!]
    \centering
    \includegraphics[width=0.78\columnwidth]{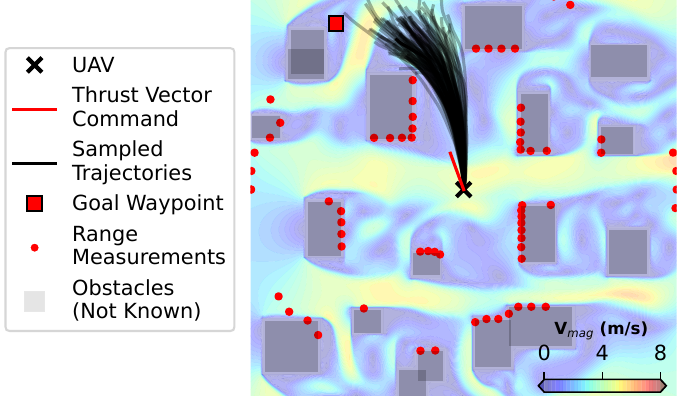}
     
    \caption{Energy-aware autonomous navigation through an urban environment with dynamic wind conditions. 
    The controller adapts in real time using on board wind flow predictions, balancing energy consumption and obstacle avoidance to reach the goal destination \revision{more robustly and with improved energy efficiency.}}
     
    \label{fig:overview}
\end{figure}

In the not-so-distant future, urban airspaces will be bustling with autonomous unmanned aerial vehicles (UAVs) performing package delivery, law enforcement, emergency medical services, infrastructure inspection, and other tasks. 
Autonomous navigation will be necessary as services expand and airspaces become increasingly crowded. 
Winds pose a major hazard in the urban canopy, with a high density of aerodynamically bluff bodies (e.g. buildings, bridges, etc.) generating complex spatio-temporal wind flow fields that surpass the ability of small UAVs to compensate.  
For this reason, future UAVs operating in urban airspaces need to accurately predict local wind conditions and account for them in decision making, motion planning, and control.

\autoref{ch:prediction} presents a novel method for predicting wind flow fields in real time by leveraging information provided by a navigational LiDAR.
This included a neural network to correlate surrounding building topography signatures with local winds, consistent with the mass- and momentum-conserving boundary conditions enforced by obstacles in the flow. 
While the previous chapter was focused on the accuracy of local wind flow field predictions, this chapter explores the application of this method towards real-time wind-aware motion planning. 

Here, we demonstrate a motion planning framework illustrated in \autoref{fig:overview} that combines local wind predictions with local range sensing to produce \revision{collision-free energy-efficient trajectories in windy urban environments.} 

The contributions of this chapter can be summarized as follows:
\begin{itemize}
    \item The formulation of a Model Predictive Path Integral (MPPI) optimal controller capable of autonomous navigation of a UAV in cluttered environments, using point cloud data directly for obstacle avoidance. 
    \item Novel extensions of \revision{the MPPI controller} using local wind information and a cost function associated with power consumption for energy-efficient flight and fewer collisions. 
    \item Evaluation of the method using a realistic UAV simulator including aerodynamic forces and torques across unsteady wind flow fields, \revision{alongside} comparisons to an existing global wind-aware path planner.   
\end{itemize}

Our use of MPPI over gradient-based receding horizon optimal control is motivated by how it scales to high dimensional problems and its ability to robustly handle stochastic disturbances, modeling errors, and local minima induced by nonconvex cost functions, making it suitable for real-time UAV navigation in complex and unpredictable urban wind conditions. 

\section{Problem Formulation}\label{sec:planning:problemformulation}

In this chapter, a UAV is tasked with navigating through an environment $\mathcal{P}$ subject to an unsteady wind flow field $\mathcal{W}$. 
Within this environment, there are numerous obstacles--buildings in an urban setting--which define inaccessible regions of the space $\mathcal{X}_{occ}\subset\mathcal{P}$.
The occupied space $\mathcal{X}_{occ}$ is unknown to the UAV.
Instead, $\mathcal{X}_{occ}$ is partially observed at time $t$ via UAV range sensors. 

The wind field is also only partially observable at a given instant. 
The partial observability is motivated by the contributions of the previous chapter, which provides a prediction of the time-averaged wind flow field in a region around the UAV in real time using the same range sensor used for obstacle detection. 
The current local region $\mathcal{L}_t \subset \mathcal{P}$ is the set of all positions, $\boldsymbol{x}$, within a hypercube with edge length $H$ centered at the UAV's current location $\boldsymbol{p}(t)$:
\begin{equation}\label{eq:local_region_def}
    \mathcal{L}_t := \left\{ \boldsymbol{x} \in \mathcal{P} \:\Big\vert ||\boldsymbol{x} - \boldsymbol{p}(t)||_\infty \leq \frac{H}{2} \right\}
\end{equation}
The partial observability of the obstacles and wind encourages solutions implemented in real time, where an unknown environment and limited on board compute makes ascertaining the entire wind flow field intractable. 

\subsection{UAV Dynamics and Unsteady Winds}\label{sec:fulldynamics}

The full 6 degree-of-freedom (DoF) rigid body dynamics for a rotary-wing UAV are simulated including models of the motor dynamics, parasitic drag forces, rotor drag, and blade flapping effects using \textit{RotorPy} \cite{folk2023rotorpy}.

The UAV's motion is disturbed by an unsteady wind flow field,
\begin{equation}\label{eq:windfield}
    \boldsymbol{w} = \mathcal{W}(\boldsymbol{p}, t)
\end{equation}
Wind flow fields around obstacles are simulated using the 2D staggered grid fluid solver discussed in \autoref{ch:prediction}. 

\subsection{UAV Energetics}\label{sec:planning:energetics}

To evaluate the motion planner, the UAV's energy usage is approximated by integrating power, $P$, at each timestep, 
\begin{equation}\label{eq:problemformulation_energycost}
    E(t) = \int_0^t P(t)dt
\end{equation}
Following \autoref{sec:dynamics:power}, we use an empirically-driven power model relating the UAV's instantaneous power consumption to its airspeed.
This model (\autoref{fig:powermodel_withwind}) considers the effect of relative motion of rotors through a fluid medium derived from wind tunnel test data \cite{ware2016canopy}.
While the model assumes steady state forward flight, it serves as a lower bound of the actual power consumption.  

\section{Energy-Aware MPPI Control}\label{sec:planning:methods}

\begin{figure*}[t!]
    \centering
    \includegraphics[width=0.99\textwidth]{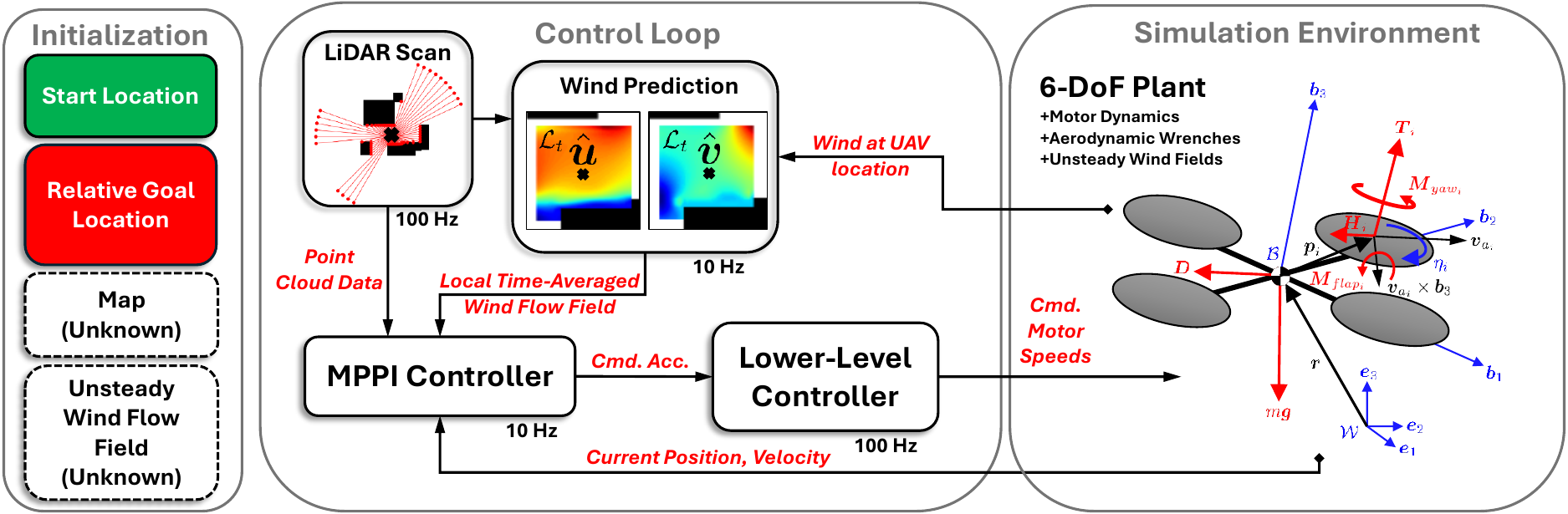}
    \caption{Wind-aware navigation through unknown windy environments.
    A LiDAR Scan measures distances to nearby obstacles. 
    The wind prediction module synthesizes a measurement of the local wind with the LiDAR scan to predict the time-averaged wind flow field in a local region, $\mathcal{L}_t$, around the UAV.
    The stochastic receding horizon controller (MPPI Controller) produces a low frequency acceleration reference, which is then tracked with a higher frequency lower level controller (Lower-Level Controller).
    The simulation environment includes full rigid body dynamics and unsteady wind flow fields.}
    \label{fig:method_diagram}
     
\end{figure*}

The high level methodology is illustrated in \autoref{fig:method_diagram}.
The motion planner in this work is based on the Model Predictive Path Integral control framework--a stochastic receding horizon control method introduced in \cite{williams2016mppiintro} combining ideas from Model Predictive Control and Path Integral Control. 

The MPPI algorithm constructs a collection of trajectories originating from the current state, $\{\boldsymbol{\tau}_j \:\vert \:j=0, 1, \dots, N_\tau \}$ (black lines in \autoref{fig:overview}), by randomly sampling sequences of feasible control inputs and forward integrating a model of the system's dynamics over a fixed horizon. 
Each sampled trajectory is then evaluated based on a user-defined cost function, $J(\boldsymbol{\tau})$, which need not be convex nor differentiable.  
The process repeats on a receding horizon, selecting the first optimal action before resetting and re-solving the problem with updated information. 

For brevity, we omit the full derivation of the MPPI control algorithm and refer the reader to \cite{williams2017mppiref} \revision{for a thorough treatment on the topic.} 

\subsection{State Space, Input Space, and Dynamics Model}

In order to balance fidelity, computational tractability, and generalization, the UAV is treated as a fully-actuated point-mass particle.  
The particle's state is defined as: 
\begin{equation}
    \boldsymbol{\mathrm{x}}(t) := \begin{bmatrix} \tilde{\boldsymbol{p}}(t) \\ \boldsymbol{v}(t) \end{bmatrix} \in \mathbb{R}^{2n}
\end{equation}
where $n\in\{2, 3\}$ indicates the dimensionality of the problem (2D or 3D).
$\tilde{\boldsymbol{p}}(t)$ is the position of the UAV \textit{relative} to its position at the start of the planning period, $t_0 \leq t$: $\tilde{\boldsymbol{p}}(t) := \boldsymbol{p}(t) - \boldsymbol{p}(t_0)$.
The ground velocity, $\boldsymbol{v}(t)$, is the velocity measured by a stationary world frame.
According to the well-known wind triangle, the ground velocity is related to the airspeed and local wind vectors\revision{:
\begin{equation}
    \boldsymbol{v}(t) := \boldsymbol{v}_a(t) + \boldsymbol{w}(t) 
\end{equation}}
The dynamics for the point particle are
\begin{equation}\label{eq:planning:continuousdynamics}
    \dot{\boldsymbol{\mathrm{x}}}(t)  := \begin{bmatrix} \dot{\tilde{\boldsymbol{p}}}(t) \\ \dot{\boldsymbol{v}}(t)\end{bmatrix} = \begin{bmatrix} \boldsymbol{v}(t) \\ \boldsymbol{a}(t) + \boldsymbol{d}(t) \end{bmatrix}
\end{equation}
where $\boldsymbol{a}(t) \in \mathbb{R}^n$ is a mass normalized actuator force (e.g. a thrust vector) and $\boldsymbol{d}(t) \in \mathbb{R}^n$ is a mass normalized aerodynamic drag force.
In this work, the drag force is quadratic in airspeed: 
\begin{equation}\label{eq:dragforce}
    \boldsymbol{d}(t) := -C \vert \vert \boldsymbol{v}_a \vert \vert\boldsymbol{v}_a
\end{equation}
where $C$ is a parasitic drag coefficient that is identified \textit{a priori} from flight or wind tunnel testing.

The control input for the MPPI controller is $\boldsymbol{\mathrm{u}}(t) := \boldsymbol{a}(t)$, constrained within predefined limits: $\boldsymbol{\mathrm{u}}(t) \in [a_{min}, a_{max}]$. 
The minimum and maximum accelerations can be directly related to a particular UAV's actuation capabilities as well as mission requirements.

\subsection{Sensing}

The MPPI controller is given instantaneous measurements of the surrounding obstacles via noisy measurements from a range sensor, $\boldsymbol{D} \in \mathbb{R}^{n_r}, D_{min} \leq D_i \leq D_{max}$ where $n_r$ is the number of rays cast. 
These range measurements are converted into a point cloud representation relative to the UAV's location, transforming the LiDAR scan into a set of \textit{relative} positions indicating occupied regions in space: 
\begin{equation}\label{eq:lidarscan}
    \boldsymbol{h}_{lidar} = \{ \tilde{\boldsymbol{p}}_{o}^0, \tilde{\boldsymbol{p}}_{o}^1, ..., \tilde{\boldsymbol{p}}_{o}^{n_r} \} 
\end{equation}
Importantly, the range sensor only provides partial observations of the obstacles, such as one or two faces of a building, unlike related MPPI controllers for UAV navigation that use global information or representations \cite{mohamed2020mppiquad, higgins2023mppiquadrotor}. 

The MPPI controller is also provided an estimation of the time-averaged wind flow field, $\hat{\boldsymbol{w}}$, but only within the local region $\mathcal{L}_t$: 
\begin{equation}\label{eq:wind_measurement}
    \boldsymbol{h}_{wind} = \{\hat{\boldsymbol{w}}(\tilde{\boldsymbol{p}}) \:\vert\: \tilde{\boldsymbol{p}} \in \mathcal{L}_t \} 
\end{equation}
The local wind estimate is constructed by sampling the ground truth wind flow field averaged over $100$ seconds with no noise added.
\revision{In the next chapter,} real-time estimates of the surrounding winds \revision{will instead be provided} directly from in situ wind measurements and the LiDAR scan \revision{using the wind flow decoder network from \autoref{ch:prediction}.} 

\subsection{Cost Functions}

For energy-aware motion planning, there are two primary tasks to consider: 1) making progress towards the goal location without colliding with obstacles, and 2) minimizing energy consumption along the way.

To motivate progress towards the goal, trajectories are scored by integrating distance from the goal.
\begin{equation}\label{eq:cost_progress}
    J_{prog}(\boldsymbol{\tau}_j) := \int_{\boldsymbol{\tau}_j} ||\tilde{\boldsymbol{p}}(\boldsymbol{\tau}_j) - \tilde{\boldsymbol{p}}_g ||_2^2 d\boldsymbol{\tau}_j
\end{equation}
where $\tilde{\boldsymbol{p}}(\boldsymbol{\tau}_j)$ is the relative position of the UAV along the $j$-th sampled trajectory. 

To avoid navigating too close to obstacles in the scene, an additional cost is associated with the closest distance to sensed obstacles. 
\begin{equation}\label{eq:cost_obs}
    J_{obs}(\boldsymbol{\tau}_j) := \min_{\tilde{\boldsymbol{p}}(\boldsymbol{\tau}_j),\: \tilde{\boldsymbol{p}}_o^{0:n_r}}\left[ \frac{1}{ || \tilde{\boldsymbol{p}}(\boldsymbol{\tau}_j) -  \tilde{\boldsymbol{p}}_o^i ||_2^2} \right]
\end{equation}

A final cost term penalizes trajectories with high predicted energy consumption. 
As in related works \cite{ware2016canopy, ebert2023gappy}, a steady state power model as a function of airspeed, $P(\cdot)$, is used as a lower bound. 
The total energy cost of a trajectory is computed by integrating power along the trajectory. 
\begin{equation}\label{eq:cost_energy}
    J_{energy}(\boldsymbol{\tau}_j) := \int_{\boldsymbol{\tau}_j} P(\boldsymbol{v}(\boldsymbol{\tau}_j) - \boldsymbol{w}(\boldsymbol{\tau}_j))d\boldsymbol{\tau}_j
\end{equation}
where $\boldsymbol{v}(\boldsymbol{\tau}_j)$ and $\boldsymbol{w}(\boldsymbol{\tau}_j)$ are the ground and wind velocities. 

Each term in the cost function is weighted by a set of scalars. 
Weighting the obstacle and energy costs relative to the progress cost makes weight selection more interpretable. 
\begin{equation}\label{eq:cost_total}
    J(\boldsymbol{\tau}_j) = J_{prog}(\boldsymbol{\tau}_j) + Q_O J_{obs}(\boldsymbol{\tau}_j) + Q_E J_{energy}(\boldsymbol{\tau}_j) 
\end{equation}
where $Q_O, Q_E \geq 0$ are scalar cost weights corresponding to the importance of obstacle avoidance and energy consumption, respectively, relative to progress towards the goal.

\subsection{Numerical Transcription and Implementation}

A few modifications are necessary for implementation of the MPPI algorithm. 
First, time is discretized into $\Delta t$ time steps, $\boldsymbol{\mathrm{t}} = \{k\Delta t,\:k=0, 1, ..., N_k+1\}$, where $N_k$ describes sequence length of control inputs sampled for each trajectory.

The dynamics are approximated in discrete time using forward Euler integration: 
\begin{equation}\label{eq:planning:discretedynamics}
    \boldsymbol{\mathrm{x}}_{k+1} = \boldsymbol{\mathrm{x}}_{k} + \begin{bmatrix} \boldsymbol{v}_k \\ \boldsymbol{a}_k + \boldsymbol{d}(\boldsymbol{v}_k, \boldsymbol{w}_k)\end{bmatrix} \Delta t
\end{equation}
where $\boldsymbol{d}(\cdot)$ is defined in \autoref{eq:dragforce}.
If wind information is not available or $\tilde{\boldsymbol{p}}_k \notin \mathcal{L}_t$, $\boldsymbol{w}_k = 0$. 
Both $\Delta t$ and $N_k$ must be selected by the designer.

At the beginning of the task, the relative goal location may very likely be far away, which makes the position error cost (\autoref{eq:cost_progress}) non-descriptive. 
To mitigate this, \revision{we found it helped to project the} goal location onto the LiDAR perception radius, $D_{max}$, whenever the distance to goal is beyond this value:
\begin{equation}\label{eq:pos_projection}
    \tilde{\boldsymbol{p}}_g := \begin{cases} 
          \tilde{\boldsymbol{p}}_g & ||\tilde{\boldsymbol{p}}_g|| < D_{max} \\
          D_{max}\frac{\tilde{\boldsymbol{p}}_g}{||\tilde{\boldsymbol{p}}_g||} & ||\tilde{\boldsymbol{p}}_g|| \geq D_{max}
       \end{cases}
\end{equation}

The cost functions must also be transcribed into the discretized setting.
To improve numerical conditioning and the interpretability of the cost weights, we initially non-dimensionalize the trajectories. 
First, the relative positions are normalized by the LiDAR perception radius: 
\begin{align}
    \bar{\boldsymbol{p}} &= \frac{\tilde{\boldsymbol{p}}}{D_{max}} \\ 
    \bar{\boldsymbol{p}}_g &= \frac{\tilde{\boldsymbol{p}}_g}{D_{max}} \\
    \bar{\boldsymbol{p}}_o^i &= \frac{\tilde{\boldsymbol{p}}_o^i}{D_{max}}
\end{align}
The power consumption is normalized by the power required to hover as in \cite{ware2016canopy}. 
\begin{equation}
    \bar{P} := \frac{P}{P_{hover}}
\end{equation}
The cost functions are now defined in discrete time based on their non-dimensionalized values. 
\begin{align}\label{eq:costs_summary_discrete}
    \bar{J}_{prog}(\boldsymbol{\tau}) = \sum_{k=0}^{N_k+1} \left( \bar{\boldsymbol{p}}_k - \bar{\boldsymbol{p}}_{g} \right)^\top \left( \bar{\boldsymbol{p}}_k - \bar{\boldsymbol{p}}_{g} \right) \\ 
    \bar{J}_{obs} = \min_{\substack{\bar{\boldsymbol{p}}_{0:N_k+1},\: \bar{\boldsymbol{p}}_o^{0:n_r}}} \left[ \frac{1}{\left( \bar{\boldsymbol{p}}_k - \bar{\boldsymbol{p}}_{o}^i \right)^\top \left( \bar{\boldsymbol{p}}_k - \bar{\boldsymbol{p}}_{o}^i \right)} \right] \\ 
    \bar{J}_{energy} = \sum_{k=0}^{N_k+1} \bar{P}(\boldsymbol{v}_k, \boldsymbol{w}_k) \Delta t 
\end{align}

At the beginning of each planning step, the previous optimal sequence of control inputs is perturbed by random noise and clipped, $\boldsymbol{\mathrm{u}}^*_{0:N_k} \leftarrow \texttt{clip}(\boldsymbol{\mathrm{u}}^*_{0:N_k} + \delta \boldsymbol{\mathrm{u}}, u_{min}, u_{max})$, with $\delta \boldsymbol{\mathrm{u}} \in \mathbb{R}^{N_k\times N_\tau} \sim \mathcal{N}(0, \sigma_{\mathrm{u}}^2)$.
A Gaussian kernel with variance $\sigma_{gf}^2$ is then applied to $\boldsymbol{\mathrm{u}}_{0:N_k}$ to smooth large changes in control commands.\footnote{\revision{Note that for the hardware implementation of this algorithm in \autoref{ch:realworld}, the preferential Gaussian sampling scheme used here is replaced by simple uniform sampling which is more in line with the formulation in \cite{williams2017mppiref}.}} 

The planner operates at a frequency $f_{plan}$ which is lower than the lower level controller's control frequency to ensure that commanded accelerations can be reasonably tracked.  

\begin{figure}
    \centering
    \includegraphics[width=0.80\columnwidth]{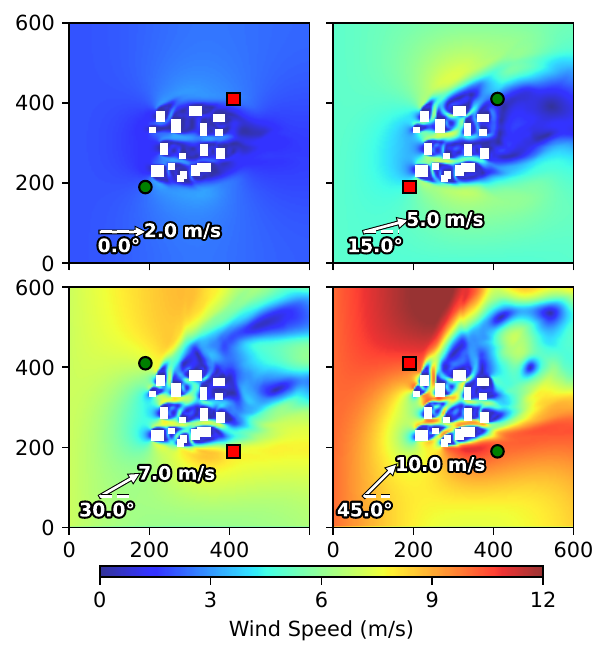}
    \caption{Time-averaged wind flow fields for $4$ out of the $16$ wind scenarios used for experiments \revision{on one of the $16$ maps.}
    Each scenario is labeled with one of the four possible start (green circle) and goal (red square) waypoints.}
    \label{fig:wind_dataset}
     
\end{figure}

\section{Experimental Setup}\label{sec:planning:simulations} 

The simulation experiments were designed to rigorously evaluate the method across a diverse set of wind conditions.
Sixteen urban maps were procedurally generated on a $600$$\times$$600$ m domain by iteratively placing and stacking cuboid primitives.
On each map, there were anywhere from $9$ to $16$ buildings between $10$ and $50$ m in size with minimum separation (alley width) of $15$ m, and $16$ wind scenarios were constructed by simulating fluid flow with inlet magnitudes $[2.0, 5.0, 7.0, 10.0]$ m/s and incidence angles $[0, 15, 30, 45]^{\circ}$ relative to the world $x$ axis. 
The wind was simulated using an unsteady staggered grid fluid solver described in \autoref{sec:prediction:euler_sim}, using a $0.25$ s timestep and $1$ m grid resolution.
Steady state conditions were established for $25$ s, and then wind data was recorded for $100$ s to create the dataset. 
For each wind simulation, $4$ start and goal waypoint combinations were selected: each pair was $282$ m apart, and headwind and tailwind situations were equally represented.
Thus there were 64 different combinations of wind inlet conditions and waypoint pairs per map, totaling $1{\small,}024$ total simulations representing over $28$ hours of unsteady wind simulations.
A subset of the wind scenarios with possible start and goal waypoints is visualized on \revision{one of the evaluation maps} in \autoref{fig:wind_dataset}.\newline\newline

The UAV dynamics, following \autoref{ch:dynamics}, were simulated using \textit{RotorPy} \cite{folk2023rotorpy} with the default \textit{Hummingbird} parameters\footnote{\url{github.com/spencerfolk/rotorpy/blob/main/rotorpy/vehicles/hummingbird_params.py}} and a simulated LiDAR with $D_{max}=100$ m.
At each time step, \textit{RotorPy} queried the unsteady wind dataset to obtain the wind velocity vector at the UAV's current position. 
For obstacle sensing, the LiDAR scan was modified with Gaussian noise with zero mean and standard deviation $0.14$ m.
Flight trials terminated when any of the following conditions were satisfied: 1) the UAV is within $5$ m of the goal destination and $|\boldsymbol{v}|\leq 2$ m/s for at least $5$ s (\textit{success}); 2) the UAV stops making progress towards the goal, $|\boldsymbol{v}|\leq 1$ m/s for at least $5$ s (\textit{hover}); or 3) the UAV collides with an obstacle (\textit{collision}). 

\subsection{MPPI Variants}\label{sec:experiments_algos}

The experiments evaluated the following variations of the energy-aware MPPI controller:
\begin{itemize}
    \item \textbf{MPPI}: The nominal MPPI algorithm with $Q_E=0$. 
    No information about the wind is provided ($\boldsymbol{w}_k=0$), \revision{and the sole focus is on obstacle navigation.}
    \item \textbf{MPPI+W}: The MPPI algorithm with $Q_E=0$ but $\boldsymbol{w}_k$ is known within $\mathcal{L}_t$, \revision{so the controller can account for wind.}
    \item \textbf{EA-MPPI}: The MPPI algorithm with $Q_E>0$. 
    However, the surrounding wind is not known ($\boldsymbol{w}_k=0$). 
    \revision{In theory any energy savings here would be via speed regulation.}
    \item \textbf{EA-MPPI+W}: The fully-featured controller with $Q_E>0$ and known time-averaged winds within $\mathcal{L}_t$. 
\end{itemize}

\subsection{Global Baseline}\label{sec:planning:graph_search_baseline}

To understand the effects of replanning on a local horizon, a global wind-aware graph search algorithm (GS) resembling \cite{ebert2023gappy} with $w=0$ (following their notation) was implemented, although the ground speed is allowed to vary as in \cite{ware2016canopy}. 
The same power model (\autoref{fig:powermodel_withwind}) is used to predict the energy cost based on airspeed, the ground speed is constrained to be $|\boldsymbol{v}| \in [0.5, 20.0]$.
In contrast to the MPPI controller, the GS algorithm has full knowledge of both obstacles and the time-averaged wind field but the states and actions are discretized.
We refer the reader to \cite{ware2016canopy, ebert2023gappy} for more details. 

\begin{table}[h!]
\centering
\caption{Hyperparameters for the simulated nominal MPPI controller. }
\label{tab:mppi_hyperparams}
\begin{tabularx}{0.5\columnwidth}{lXl}
\toprule
\textbf{Parameter} & \textbf{Value} & \textbf{Unit} \\ \midrule
$N_\tau$ & 100 & - \\
$\Delta t$ & 0.25 & s \\
$N_k$ & 187 & - \\
$f_{plan}$ & 10 & Hz \\
$\lambda$ & 0.79 & - \\
$\sigma_\mathrm{u}$ & 5.15 & m/s$^2$ \\
$\sigma_{gf}$ & 5.97 & m/s$^2$ \\
$Q_O$ & 1.25 & - \\ 
$D_{max}$ & 100 & m \\
$P_{hover}$ & 247 & W \\
\bottomrule
\end{tabularx}
 
\end{table}

\subsection{Hyperparameter Tuning}

The performance and behavior of the MPPI controller relies on appropriately chosen values for the hyperparameters. 
\autoref{tab:mppi_hyperparams} summarizes the baseline hyperparameter values, which were auto-tuned with Optuna's default estimator \cite{optuna2019} using scores computed on the intersection case study (see \autoref{fig:case_studies}) without wind.
Once these parameters were selected, they remained fixed across all variants and experiments.

\section{Results}\label{sec:planning:results}

\subsection{Parameter Exploration}\label{sec:results_paramexp}

In \autoref{fig:parameter_study}, the effects of the energy cost weight $Q_E$ and local region size $H$ are visualized. 
Specifically, the percentage reduction in energy consumption relative to \textbf{MPPI} is computed for \textbf{EA-MPPI+W} across different values for $Q_E$ and $H$ and grouped by the magnitude of the velocity inlet condition. 
With respect to $Q_E$, the average reduction in energy consumption is invariant to the magnitude of the freestream velocity--on average this percent improvement is $14.1\%$ over the baseline \textbf{MPPI}. 
However, the variance of the average reduction in energy across the $1{\small,}024$ simulations (the shaded regions in \autoref{fig:parameter_study}) increases with the inlet magnitude. 
The optimal range of values of $Q_E$ is between $3.0$ and $4.0$.
While this is conditioned on the baseline hyperparameters in \autoref{tab:mppi_hyperparams}, it is notable that energy reduction is not strictly monotonic with increasing $Q_E$.

Region size seems to have a subtle effect on the energy reduction indicated by the small differences overall in between no knowledge ($H=0$) and full knowledge ($H=\infty$) of the wind. 
Looking closely, there appears to be an asymptote at around $H=60$ where increasing the size of the local wind knowledge no longer provides any measurable difference in percentage of energy reduction. 
This is most certainly linked to the horizon hyperparameters ($N_k$ and $\Delta t$) associated with the MPPI algorithm. 
It is unlikely that any wind information beyond $H=60$ is helpful due to the short horizon of the MPPI planner. 

\begin{figure}[t!]
    \centering
    \includegraphics[width=0.99\columnwidth]{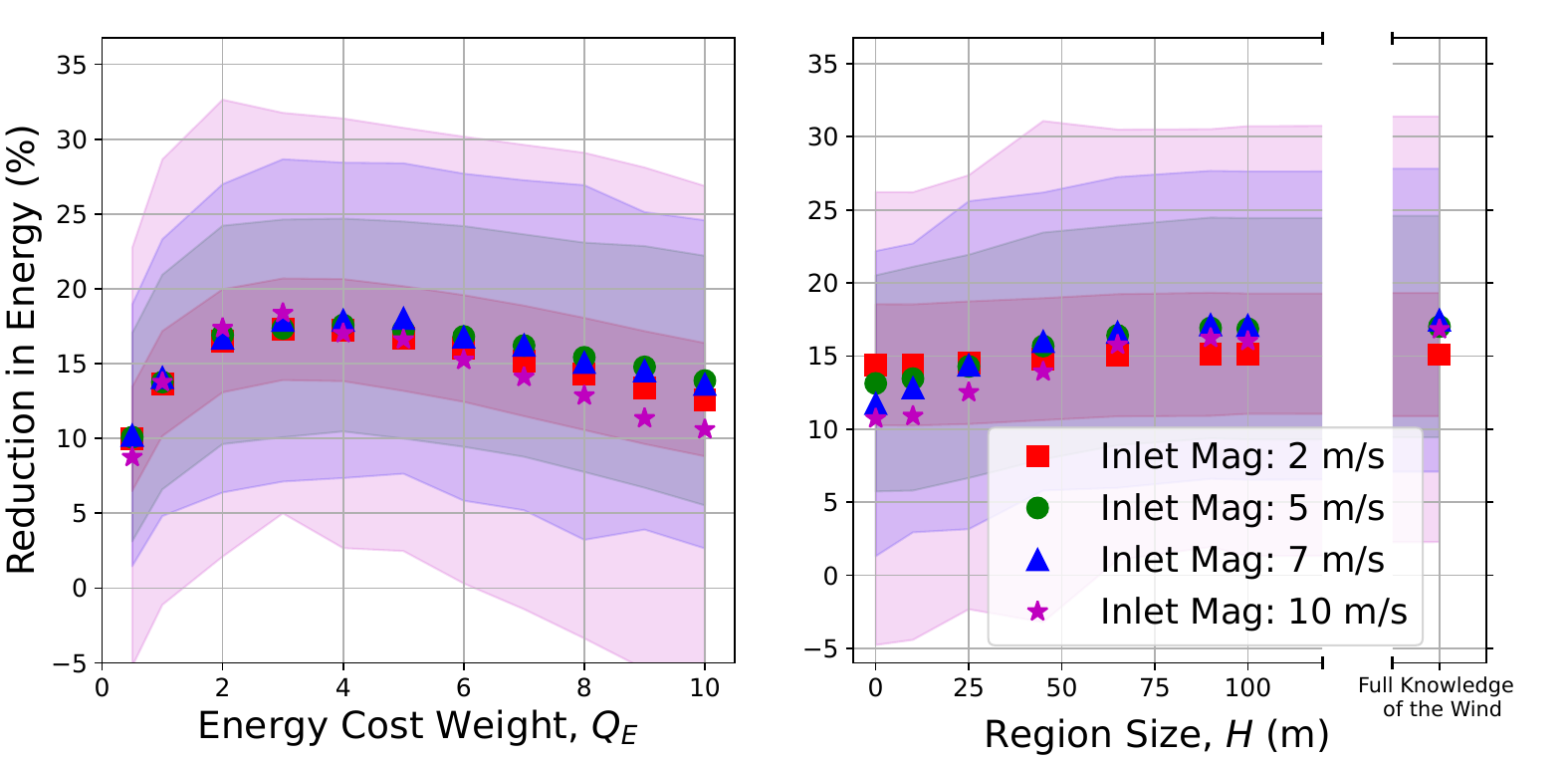}
     
    \caption{Average percent reduction in energy consumption with one standard deviation regions for \textbf{EA-MPPI+W} relative to \textbf{MPPI} against different energy cost weights (left) and wind prediction region sizes (right). Trials where either \textbf{EA-MPPI+W} or \textbf{MPPI} failed are not included.}
    \label{fig:parameter_study}
     
\end{figure}

\subsection{Ablations}\label{sec:results_ablation}

An ablation study was conducted to understand the relative importance of different features of the energy-aware MPPI algorithm. 
The different variants include \textbf{MPPI} ($Q_E=0$, $H=0$), \textbf{MPPI+W} ($Q_E=0$, $H=\infty$), \textbf{EA-MPPI} ($Q_E=3.0, H=0$), and \textbf{EA-MPPI+W} ($Q_E=5.0$, $H=\infty$).
The values for $Q_E$ and $H$ were selected as the best performers from the trials.
Note that $H=\infty$, while not practical, is still considered for this analysis because it represents upper bounds on performance if the full (time-averaged) flow field knowledge was provided.

According to the summary data in \autoref{tab:ablation_results}, wind information (\textbf{+W}) improves the MPPI controller's \revision{robustness.} 
\textbf{MPPI+W} reduces the overall failure rate by $7.5\%$ over \textbf{MPPI}.
This effect is less pronounced when comparing \textbf{EA-MPPI} and \textbf{EA-MPPI+W}, where the percentage improvement in failure rates is only $2.9\%$.
The primary reduction in both cases is in the number of collision events, suggesting that collisions might be due to missing information about the wind.  

Rather than wind information, the energy cost term alone (\textbf{EA-MPPI}) is the primary factor in reducing energy consumption by $18.6\%$ over \textbf{MPPI}. 
Further, the average commanded acceleration is reduced by just under $30\%$ over the nominal controller but wind information does not provide any statistically significant difference in this metric either. 
Looking at time to goal and distance traveled, an important trade-off emerges: the energy-aware (\textbf{EA-}) variants take on average $30\%$ longer to travel approximately the same distance. 
These results indicate that primary energy savings come primarily by regulating speed over long distances. 
\revision{They also connect back to the power model shown in \autoref{fig:powermodel_withwind} where energy optimality comes from flying at a slower ground speed, but that optimal ground speed is a direct function of the surrounding wind.}

\begin{table*}[t!]
\centering
\caption{Performance metrics of different ablations of the MPPI controller across simulated scenarios.}
\label{tab:ablation_results}
\resizebox{\textwidth}{!}{%
\begin{tabular}{@{}lccccc@{}}
\toprule
\textbf{Ablation} & \textbf{Avg. Energy (kJ)} & \textbf{Cmd. Acc. (m/s$^2$)} & \textbf{Time to Goal (s)} & \textbf{Distance Traveled (m)} & \textbf{Failure Rate (\textit{collision}, \textit{hover})* (\%)} \\ \midrule
\textbf{MPPI} & 13.61 & 20.95$\pm$1.48 & 28.29$\pm$4.51 & 329.54$\pm$11.29 & 36.3 (22.4,14.0) \\
\textbf{MPPI+W} & 14.16 & 21.35$\pm$1.85 & 28.41$\pm$4.52 & 329.94$\pm$11.98 & 28.8 (18.9,9.9) \\
\textbf{EA-MPPI} & 11.07 & 14.15$\pm$1.04 & 38.31$\pm$4.90 & 329.15$\pm$9.07 & 18.7 (2.5,16.1) \\
\textbf{EA-MPPI+W}& 10.81 & 14.05$\pm$0.62 & 37.94$\pm$5.63 & 334.18$\pm$11.66 & 15.8 (2.6,13.2) \\
\bottomrule
\end{tabular}%
}
\begin{minipage}{\textwidth}
\small
 
\raggedright*Failure is defined as colliding with an obstacle (\textit{collision}) or no longer making progress towards the goal (\textit{hover}).
\end{minipage}
\end{table*}

\begin{figure}
    \centering
    \includegraphics[width=0.90\textwidth]{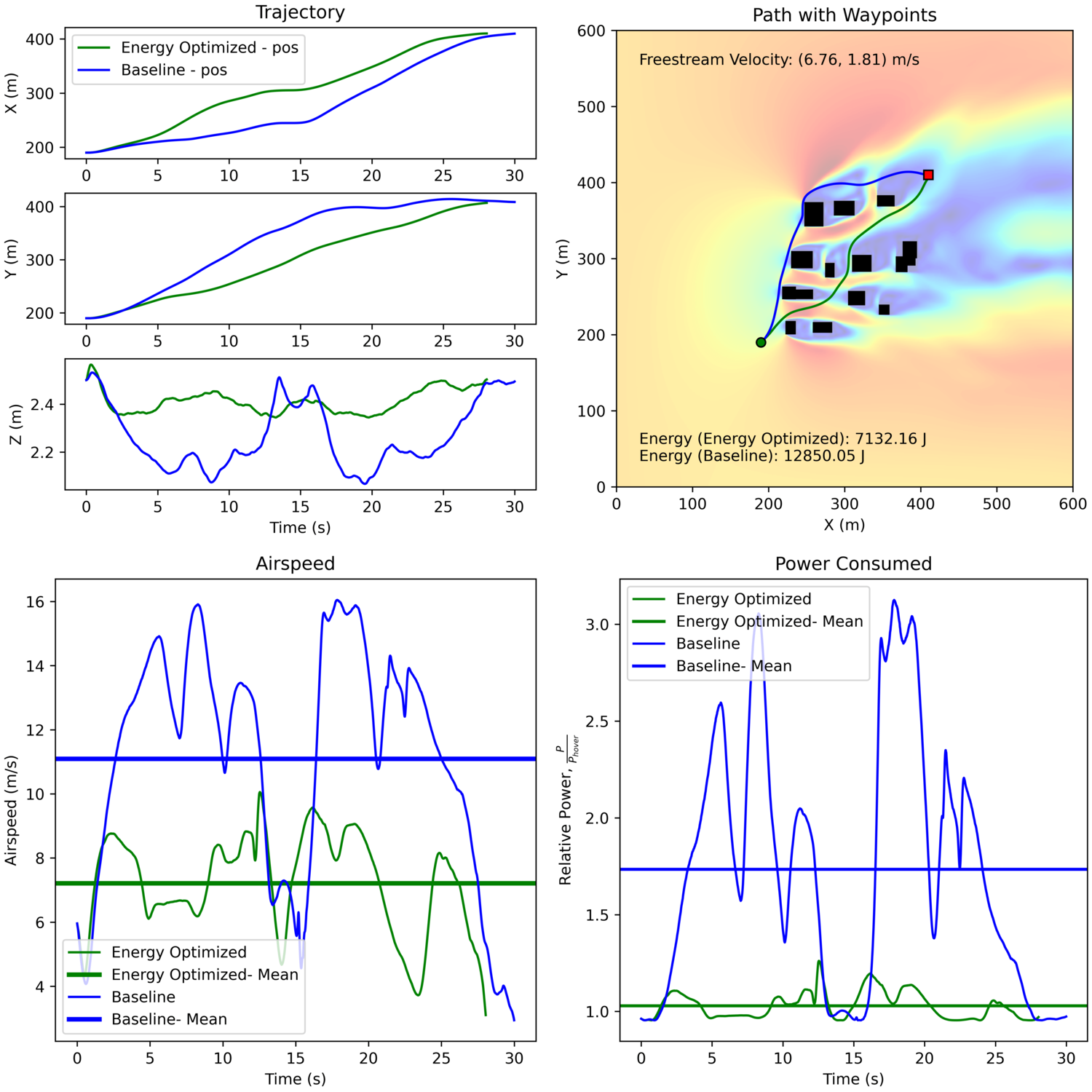}
    \caption{The best trial with respect to energy reduction over the baseline from the simulated wind scenarios. The blue lines correspond to \textbf{MPPI} whereas the green lines correspond to \textbf{EA-MPPI+W}. The wind flow field is seen averaged over time in the top right alongside the paths each ablation took. Airspeed and instantaneous power consumption are shown in the bottom two plots.}
    \label{fig:planning:best_trial}
\end{figure}

\begin{figure}
    \centering
    \includegraphics[width=0.90\textwidth]{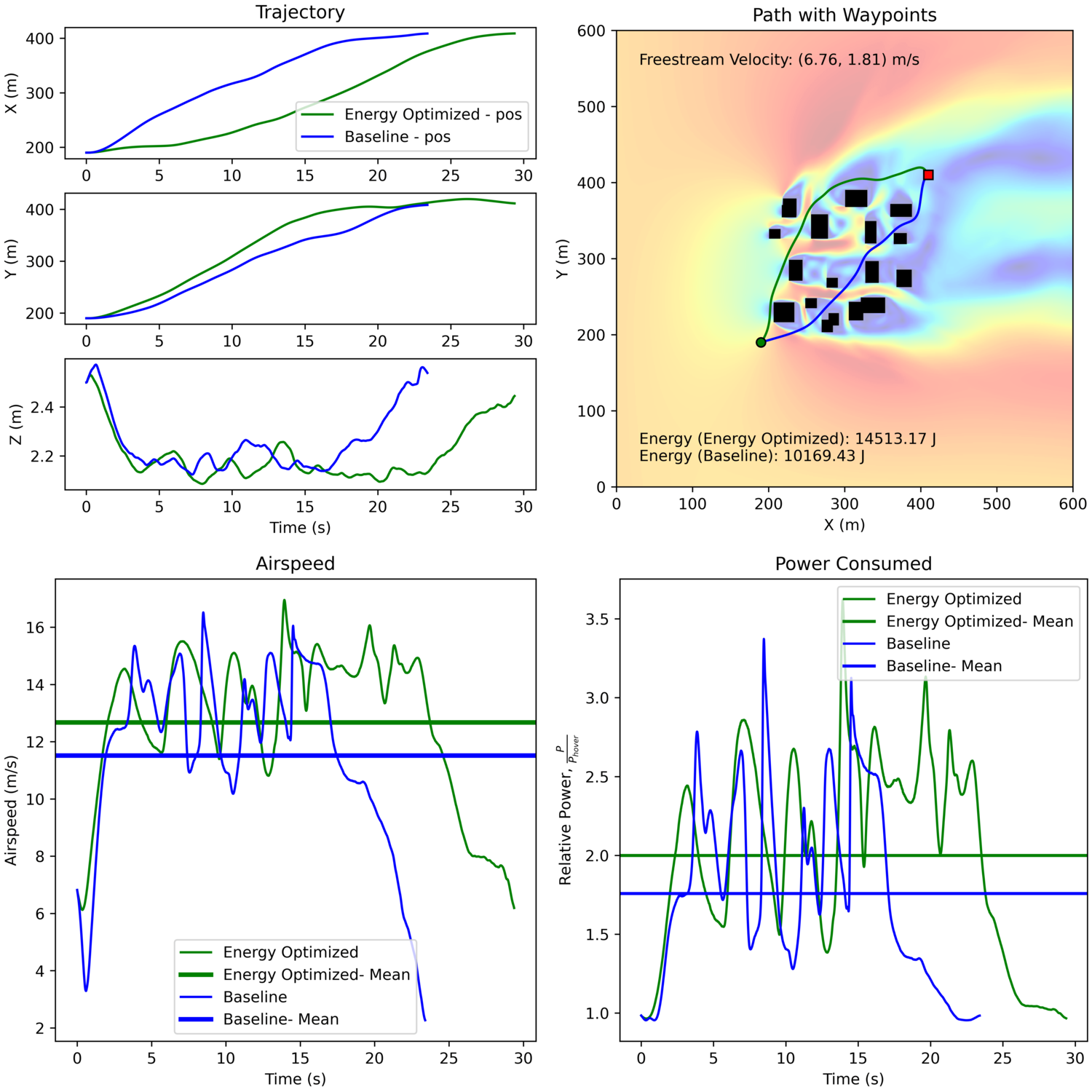}
    \caption{The worst trial with respect to energy reduction over the baseline from the simulated wind scenarios. The blue lines correspond to \textbf{MPPI} whereas the green lines correspond to \textbf{EA-MPPI+W}. The wind flow field is seen averaged over time in the top right alongside the paths each ablation took. Airspeed and instantaneous power consumption are shown in the bottom two plots.}
    \label{fig:planning:worst_trial}
\end{figure}

\subsection{Case Studies}

\subsubsection{Best and Worst Cases}

Across the $1{\small,}024$ simulated wind scenarios, the trials with the best and worst performance of \textbf{EA-MPPI+W} ($Q_E=5, H=\infty$) over \textbf{MPPI} ($Q_E=H=0$) were selected for illustrative purposes and shown in \autoref{fig:planning:best_trial} and \autoref{fig:planning:worst_trial}, respectively. 
The trials were selected only out of those where neither ablation failed due to collision or hover timeout. 
In each trial, the \textbf{MPPI} and \textbf{EA-MPPI+W} ablations are given the same start (green circle) and goal (red square) waypoints and have to operate through the same unsteady wind flow fields.
The wind flow field is visualized in its time-averaged form in the top right figure to give a sense of what coherent flow features exist in the trial. 

Looking at \autoref{fig:planning:best_trial}, \textbf{EA-MPPI+W} consumes an impressive $45\%$ less energy than the baseline \textbf{MPPI}. 
There are two apparent explanations for the energy savings: 1) \textbf{EA-MPPI+W} took a shorter more direct path to the goal arriving about $3$ seconds before \textbf{MPPI}; and 2) \textbf{EA-MPPI+W} flew at a lower airspeed resulting in substantially lower instantaneous power consumption, up to 3 times lower than \textbf{MPPI} for most of the simulation.
\revision{Contextualizing these findings within} the full simulation results, the concept of speed regulation corroborates with \autoref{tab:ablation_results}, but the fact that \textbf{EA-MPPI+W} found a shorter route seems to be what makes this trial a notable outlier.

\autoref{fig:planning:worst_trial} tells a very different story, but with important lessons.
In the worst case trial from the simulations, \textbf{EA-MPPI+W} consumes $44\%$ \textit{more} energy than \textbf{MPPI} which is in stark opposition to the best case trial.
At first glance the best and worst case trials look very similar, and indeed the freestream velocity and waypoints are precisely the same between both trials. 
However, the primary difference between these two trials is the map, which emphasizes an important fact: the building configuration determines the flow features that can be exploited for energy savings. 
Because of the myopic view of the MPPI algorithm and the \textit{exploration vs exploitation tradeoff}, if these flow features are not identified within the algorithm's horizon early on, they can be ignored and lead to very different outcomes. 
In the best trial, there is an urban river immediately in view that \textbf{EA-MPPI+W} identifies as a favorable path, but in the worst case trial these features are blocked by a large building. 
Had the base MPPI hyperparameters in \autoref{tab:mppi_hyperparams} been tuned more for exploration, perhaps \textbf{EA-MPPI+W} would be more robust to these situations, but this is left to future work. 

\subsubsection{Comparisons to a Global Planner}

Finally, three \revision{hand-designed} case studies were carried out in order to gain insight into the qualitative behavior of the different MPPI ablations in comparison to the global wind-aware planner (GS) described in \autoref{sec:planning:graph_search_baseline}. 
\autoref{fig:case_studies} visualizes the routes and corresponding energy consumption profiles for each algorithm, as well as the ground truth time-averaged wind flow fields as semi-transparent backgrounds.

In the \textit{Intersection} scenario, \textbf{MPPI} and \textbf{EA-MPPI+W} follow similar paths; however the airspeed of \textbf{EA-MPPI+W} is regulated to improve energy consumption.
The wind-aware graph search algorithm (\textbf{GS+W}) finds an alternative route which consumes slightly less energy.
In \textit{Single Building}, both MPPI methods outperform their GS counterparts, which may be due to coarse discretization of the states and controls by the GS methods, however further investigation would be required to confirm this which is left to future work. 
Lastly, in \textit{Evaluation Map}, \textbf{GS+W} and \textbf{EA-MPPI+W} have very similar routes and energy profiles, highlighting MPPI's ability to approximately solve the energy-optimization problem but with a fraction of the compute time. 
More specifically, the MPPI planner is intended to run in real time, whereas the global graph search algorithm takes upwards of $40$ minutes on a laptop CPU to arrive at an optimal solution. 

\begin{figure*}[t!]
    \centering
    \includegraphics[width=0.99\textwidth]{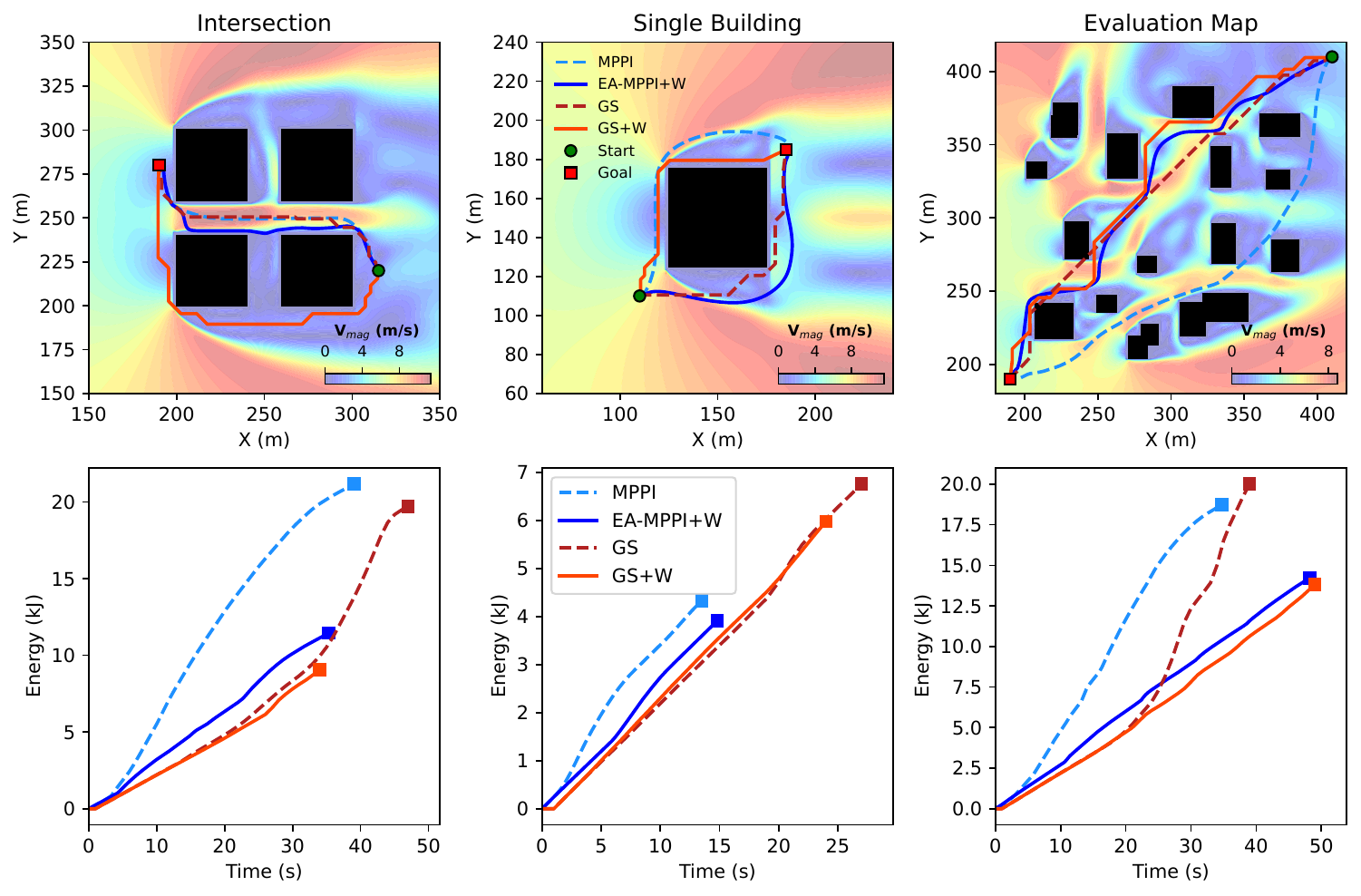}
     
    \caption{
    Case studies demonstrating the qualitative differences between the local receding horizon (MPPI and EA-MPPI+W) and the global graph search (GS and GS+W) methods for navigation in windy urban environments.
    The EA-MPPI+W method has parameters $Q_E=3.0$ and $H=90$.
    The background contours visualize the time-averaged wind magnitude, although in simulation the winds vary over time.
    }
    \label{fig:case_studies}
     
\end{figure*}

\section{\revision{Summary}}\label{sec:planning:conclusions}

In this chapter, a novel energy-aware stochastic receding horizon optimal controller is introduced in the pursuit of real-time \revision{energy-efficient and collision-free navigation} through windy urban environments.
Inspired by the previous chapter, the algorithm exploits limited local information about the wind and obstacles constructed from on board sensors to plan and adjust routes through windy environments in an effort to understand how the wind field prediction network may be useful in motion planning tasks. 

The optimal controller is evaluated on $1{\small,}024$ simulated wind scenarios to provide statistically significant insight as well as \revision{understand} the qualitative performance of the algorithm. 
In these simulations, the energy-aware controller reduces energy consumption by $18\%$ on average, but sometimes by as much as $45\%$ depending on the wind conditions, start and goal waypoints, parameter tunings, and map. 
The cost term associated with energy consumption also lowers commanded accelerations by $30\%$--an important effect that could help improve passenger comfort and protect sensitive packages in future urban air mobility applications. 
Ablation results highlight a fundamental trade-off between energy efficiency and mission duration--energy-aware trajectories tend to take over $30\%$ longer \revision{because it is more efficient to fly at slower airspeeds}, which is an important consideration in urban air mobility contexts. 
Parameter studies and ablations both indicate that providing local wind information only has a minor improvement on energy consumption in the present study, but plays an important role in lowering failure rates, demonstrating that the energy cost term and local wind information each perhaps have complementary roles in \revision{robust and efficient} autonomous navigation. 
Finally, case studies provide insights into the qualitative behavior of the MPPI algorithm, and comparisons to a global wind-aware graph search algorithm reveal that the proposed local algorithm has similar routing behaviors and energy consumption profiles to globally optimal solutions despite limited information about the obstacles and winds.

The contributions of this chapter represent a crucial step towards novel on board motion planning algorithms enabling UAVs to predict and react to changing wind conditions in real time with on board sensors, thus \revision{establishing a new framework for future} autonomous navigation in urban airspaces. 
\chapter{Experimental Validation in Scaled Urban Winds}\label{ch:realworld}

In the penultimate chapter of this thesis, the contributions from the prior three chapters are integrated in a single system to prove out these ideas with real hardware and flight-ready compute. 
Through highly instrumented experiments in a state-of-the-art open-air wind tunnel facility, this chapter addresses an existing data availability and reproducibility gap for urban air mobility studies.
While the focus is on the assessment of the algorithms presented in the previous chapters, in doing so we also contribute a detailed experimental design process and valuable in situ urban flight data to the scientific community interested in experimental validation of methods pertaining to modeling, planning, and control of UAVs in windy urban environments. 

\begin{figure}[!h]
    \centering
    \includegraphics[width=0.99\textwidth]{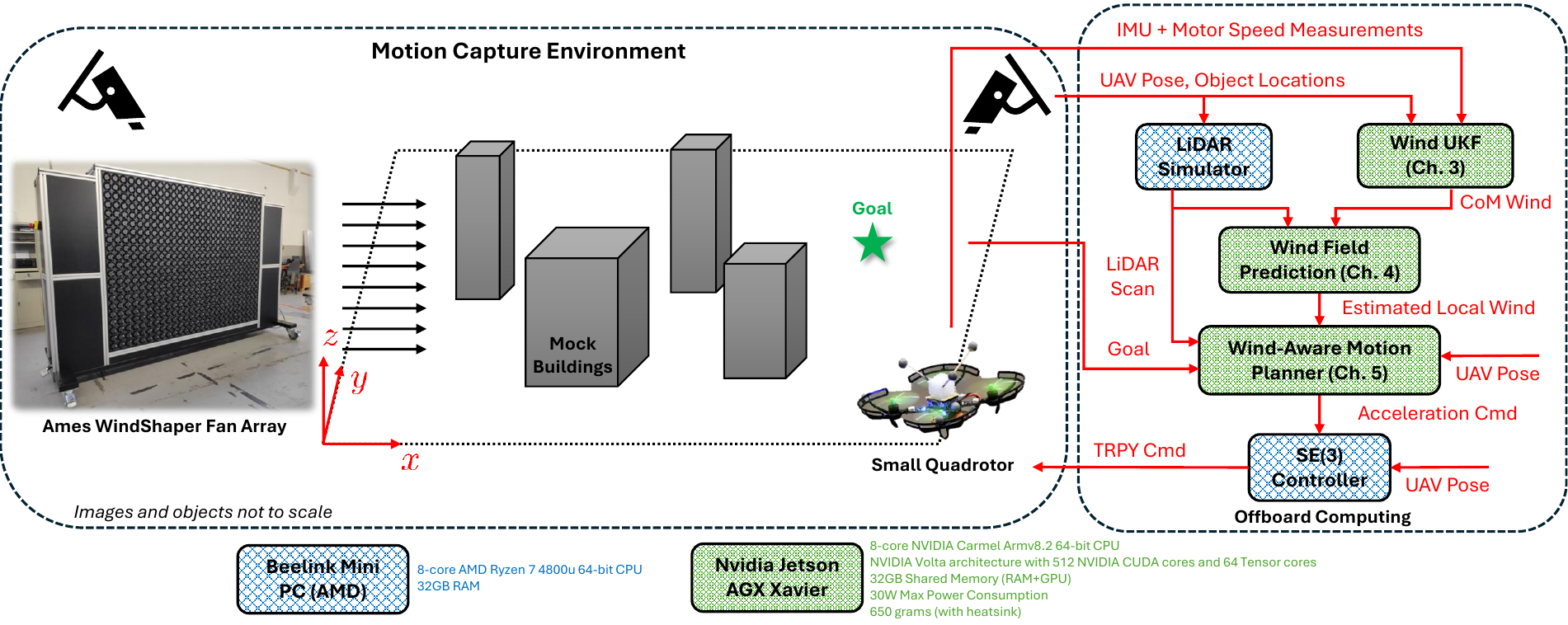}
    \caption{An overview of the experimental design, in which a small UAV is tasked with tracking a dynamic goal position through a small-scale city-like environment subjected to wind generated by the Ames WindShaper fan array.
    The left and right sides of the diagram describe the hardware and software components, respectively. 
    The software components are split between two off board computers: a Beelink Mini PC handles the LiDAR simulation, networking, and lower level control, while a flight-ready Nvidia Jetson AGX Xavier runs the core wind prediction and planning algorithms.}
    \label{fig:realworld:experimentdiagram}
\end{figure}

The experimental design is visualized in \autoref{fig:realworld:experimentdiagram}. 
There are four primary components to these experiments: a small UAV platform, a motion capture space with obstacles mimicking large buildings, an open-air fan array for generating wind, and two off-board computers for high level control of the UAV. 
To follow, each of these components will be described in greater detail.

\begin{figure}[t!]
    \centering
    \includegraphics[width=0.95\textwidth]{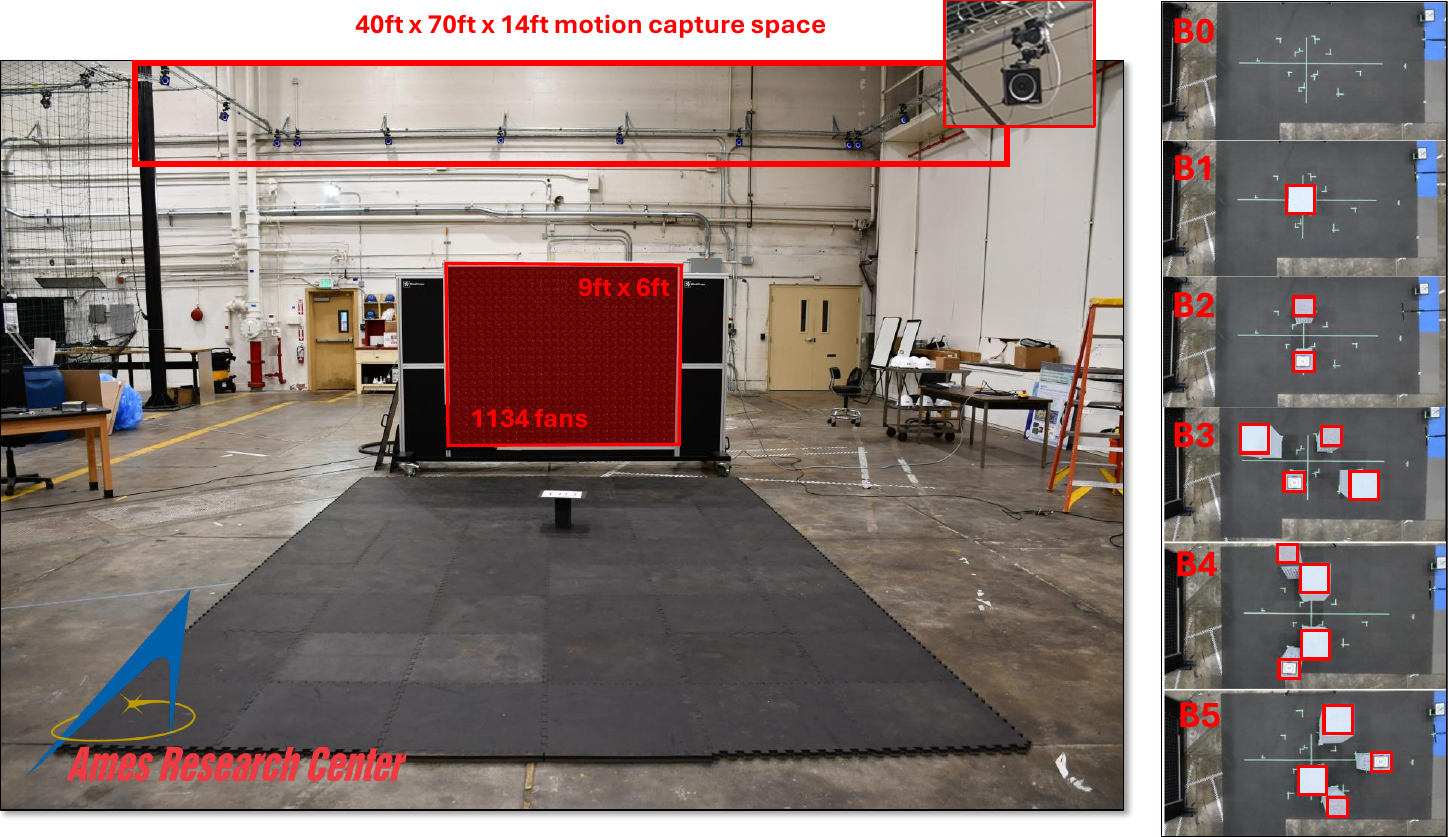}
    \caption{The WindShaper facility at NASA Ames Research Center has a $9$ ft by $6$ ft open air wind tunnel with $1{\small,}134$ individually programmable fans capable of \revision{producing winds up} to $16$ m/s. The wind tunnel is positioned inside a large motion capture space which is used to track the pose of objects with centimeter precision. On the right, the six experiment configurations are visualized with the \revision{sub-scale} buildings highlighted in red. \revision{See \autoref{appendix:realworld:buildingdimensions} for pictures of each obstacle with dimensions.}}
    \label{fig:realworld:facilities}
\end{figure}

\section{Facilities and Instrumentation}\label{sec:realworld:methods}

The experiments in this chapter leverage a newly constructed state-of-the-art laboratory environment at NASA's Ames Research Center, which we refer to as the WindShaper facility for the remainder of this chapter.  
The WindShaper facility boasts an impressive $40$$\times$$70$$\times$$14$ ft OptiTrack\footnote{\texttt{https://optitrack.com/}} motion capture space, providing simultaneous tracking of multiple objects with centimeter-level precision at a rate of $120$ Hz. 
The pose measurements from motion capture are fused with a Kalman filter\footnote{\url{https://github.com/KumarRobotics/motion_capture_system/tree/ros2-humble/mocap_base}} to provide full state (pose and twist) measurements of all the objects in the space. 

The primary feature of the WindShaper facility is the WindShaper fan array.
Pictured in \autoref{fig:realworld:experimentdiagram} and \autoref{fig:realworld:facilities}, the WindShaper is an open-air wind tunnel approximately $9$ ft wide and $6$ ft tall, with $1{\small,}134$ individually programmable fans each capable of generating wind speeds up to $16$ m/s. 
Because the fans are independently driven, it is possible to generate arbitrary spatial and temporal flow patterns across the surface of the fan array. 

The \revision{pairing} of the WindShaper \revision{with} the motion capture system \revision{results in} a truly unique free flight wind tunnel laboratory, where \revision{untethered UAVs can be subjected to spatio-temporal flow fields.}  
The WindShaper facility is the cornerstone of the experimental design in this chapter. 
\revision{Scaled} down urban wind flow fields can be designed, measured, calibrated, and repeated on demand. 

\subsection{Scaled Buildings and Experiment Configurations}\label{sec:realworld:buildings}

In real cities, buildings are the key drivers of wind flow features that are characteristic wind hazards in the urban canyon--accelerations around corners, ``urban rivers'' that channel through the streets, \revision{recirculation zones}, and vortices shedding off of the \revision{buildings--which introduce} high gust factors and shear gradients.
To generate these flow features in the WindShaper facility, scaled representations of \revision{typical mid- or high-rise} buildings were constructed out of $1$ inch thick plywood. 
Four \revision{sub-scale buildings, also referred to as obstacles,} were constructed with the following dimensions: $16$$\times$$16$$\times$$42$ inches, $16$$\times$$16$$\times$$48$ inches, $24$$\times$$24$$\times$$42$ inches, and $24$$\times$$24$$\times$$48$ \revision{inches, corresponding to $W/H$ aspect ratios of $0.38$, $0.33$, $0.57$, and $0.50$, respectively.}
\revision{The obstacle widths} were primarily constrained by the desire to comfortably fit all four buildings within the $9$ ft width of the \revision{WindShaper, while the heights were constrained by the material on hand.}
\revision{Each individual obstacle can be seen with their respective dimensions labeled in \autoref{appendix:realworld:buildingdimensions}, \autoref{fig:appendix:buildingdimensions}.}

\vspace{-5mm}

\revision{In each experiment the buildings were configured to isolate or elicit specific flow features. 
Top-down pictures of the building configurations used for the free flight experiments are provided on} the right side of \autoref{fig:realworld:facilities}.
There were six building configurations tested in total \revision{with a mixture of $1$-, $2$-, and $4$-building layouts.}
\revision{Each configuration was tested at} three different wind speeds generated by the WindShaper \revision{loosely corresponding} to \textit{low}, \textit{medium}, and \textit{high} winds \revision{with respect to the UAV used in testing.} 
\revision{In the multi-building configurations (B2 through B5), no two buildings were spaced further than $38$ inches apart which corresponds to a maximum street canyon aspect ratio ($H/S$) ratio of $1.1$.
Based on \cite{oke1988streetcanopy} and the modeling discussion in \autoref{sec:modeling:2dvs3d}, this $H/S$ value theoretically implies that vertical winds should  be small relative to the horizontal winds in between the obstacles so long as the the UAV flies under $z=0.74$ m in altitude. 
In our experiments, the UAV was commanded to fly at a constant altitude of $z=0.75$ m.
This altitude was chosen for two reasons: 1) higher altitudes decreased the chance of obstacles occluding the UAV from the motion capture cameras which significantly reduced crashes; and 2) preliminary flow visualization using the tuft arrays on one of the obstacles indicated that the streamlines had very little up or down motion at this altitude (\autoref{appendix:realworld:buildingdimensions}).  
While $z=0.75$ m nears the theoretical threshold value for quasi-2D flow, in practice appreciable vertical winds were measured at this altitude for certain regions of the flow, as is shown for configuration B4 in \autoref{sec:realworld:indirectwindestimation} below. 
}

\begin{figure}[t!]
    \centering
    \includegraphics[width=0.95\textwidth]{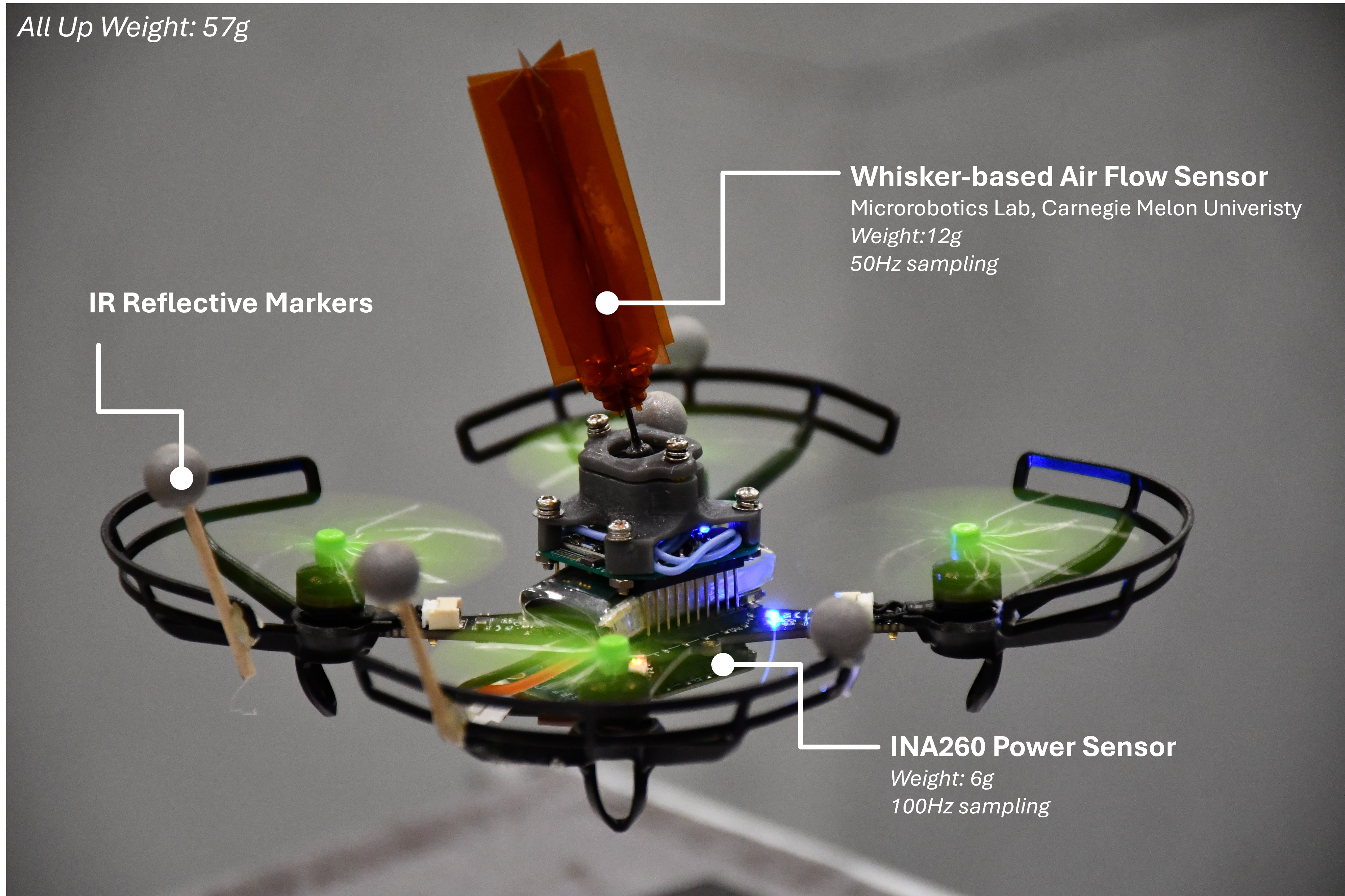}
    \caption{A modified palm-sized Crazyflie 2.1-Brushless UAV designed and manufactured by Bitcraze. The UAV has a rotor diameter of $55$ mm and is modified with IR reflective markers, a dedicated sensor for high frequency power measurements, and a custom whisker-based airflow sensor to measure wind.}
    \label{fig:realworld:crazyflie}
\end{figure}

\subsection{UAV Platform}\label{sec:realworld:robot}

The limited size of the WindShaper and buildings meant that the UAV had to be small enough to roughly match the scaling seen in the real world.
To that end, we chose the Bitcraze Crazyflie 2.1-Brushless,\footnote{\url{https://www.bitcraze.io/2023/10/say-hello-to-the-crazyflie-2-1-brushless/}} referred to as the \revision{Crazyflie, UAV, or flyer} for the remainder of this chapter. 
Seen in its modified form in \autoref{fig:realworld:crazyflie}, this is a new brushless variant of the original Crazyflie UAV platform that had become a well-established tool in aerial robotics laboratories. 

The Crazyflie has a very small footprint with a tip-to-tip width of about $155$ mm and nominal weight of $34$ grams with the battery. 
For these experiments, the Crazyflie is modified with IR reflectors for motion capture tracking and a power sensor for measuring energy consumption in real time.
\revision{The Crazyflie is also equipped} with a whisker-based airflow sensor which will be discussed in further detail below. 
\revision{These additional payloads bring the total mass of the flyer up to $57$ g.}
Based on the uncertainty analysis provided in \autoref{appendix:uncertaintymodeling}, the Crazyflie's low weight relative to its size \revision{provides an advantage to the \textit{Wind UKF} by effectively increasing the signal-to-noise ratio of the accelerations imparted on the flyer by the wind.}

\subsection{Power Measurements}\label{sec:realworld:energy_model}

In order to characterize the power consumption of the UAV, an Adafruit INA $260$ power sensing board was mounted on \revision{the} bottom of the airframe and wired between the battery and the UAV's power distribution board. 
The INA $260$ sensor weighs $6$ grams and measures the voltage and current draw from the battery.
These measurements were communicated over \texttt{I2C} to the Crazyflie; power was calculated in Watts and logged alongside the voltage and current at a rate of $100$ Hz using the Crazyflie's on board microcontroller. 
The INA $260$ sensor was crucial during initial characterization experiments to understand how the Crazyflie's power requirements vary with wind. 

\subsection{Range Sensing}\label{sec:realworld:rangesensing}

Navigational LiDAR is a key enabling sensor for both the wind flow decoder and wind-aware MPPI motion planner presented in previous chapters. 
In principle, LiDAR provides high resolution range measurements of surrounding obstacles in a $360$ degree field of view which are necessary for detecting surrounding buildings in the scene. 
However, existing LiDARs are heavy instruments--according to the datasheet for an Ouster OS1\footnote{\url{https://data.ouster.io/downloads/datasheets/datasheet-rev7-v3p0-os1.pdf}}, a typical LiDAR will weigh about $400$ g, which is over $10$ times the weight of the UAV selected for the experiments. 
While navigational LiDARs are becoming cheaper and smaller year over year, as of \revision{today} it is currently infeasible to put such a sensor on a UAV the size of the Crazyflie.

Fortunately, the motion capture system can simultaneously track the poses of the UAV and obstacles placed in the environment. 
The multi-object tracking capability enables the simulation of a LiDAR sensor on the UAV using ray casting. 
More specifically, each obstacle is fitted with seven motion capture markers--four markers on each of the top corners, one marker at one of the bottom corners, and two markers uniquely arranged on the side faces for identification purposes--in order to track the \revision{extents} of the cuboid. 
These extents are used to mesh each of the obstacles with triangles. 
The meshes are updated with each new measurement provided by motion capture, ensuring that the simulated map is constantly being updated with the current pose of each obstacle. 
Then, a set of rays are generated emanating from the UAV's pose at a fixed resolution and field of view, and any intersections between the rays and the triangle mesh are computed using the M{\"o}ller-Trumbore ray-triangle intersection algorithm \cite{moller1997raycasting}. 
A na{\"i}ve implementation is $O(N_r\times N_t)$ where $N_r$ is the number of rays coming from the UAV and $N_t$ is the number of triangles in the mesh, which presents a challenge for real time implementation.  
However, with some CPU multithreading and clever optimization in minimizing the number of rotations necessary, real time implementation is feasible for environments with $4$ to $6$ obstacles. 

\subsection{Wind Measurements}

For the experiments in this chapter, two sources of wind measurements were considered each serving the purpose of providing measurements of the wind to compare against the \textit{Wind UKF} and wind field prediction algorithms.

\subsubsection{CMU Whisker}

In order to provide direct comparisons of the lateral winds estimated by the \textit{Wind UKF}, the Crazyflie was equipped with a whisker-based flow sensor provided by the Microrobotics Lab\footnote{\url{https://www.cmu.edu/mrl/}} at Carnegie Melon University. 
Originally presented in \cite{kim2020whiskerprecursor}, the \textit{CMU Whisker} works by measuring the deflection of a bio-inspired whisker via a magnetic Hall effect sensor. 
For these experiments, a newer design of this sensor introduced by Thomas \textit{et al.} \cite{thomas2025whisker} is used because it is specifically designed to interface with the Crazyflie ``deck'' peripheral ecosystem. 
The entire sensor payload weighs just $12$ grams and provides measurements at a rate of $50$ Hz. 
The calibration procedure is described in \autoref{appendix:windshaperwhisker} and involves mounting the whisker on top of an ultrasonic anemometer to correlate the magnetic field strength to the airspeed.

Using measurements of the ground speed and sensor orientation from motion capture, the wind speed can be recovered using the well-known wind triangle. 
However, a critical disadvantage of the \textit{CMU Whisker} is that it can only provide direct airspeed measurements along the $x$ and $y$ axes of the UAV's \textit{body} frame.
The full 3D wind vector in the \textit{world} frame can be estimated if it is assumed that the $z$ component of the wind in the world frame is precisely \revision{zero.} 
This assumption introduces uncertainty into the measurements provided by the whisker, but reducing this uncertainty is left to future work. 
One final important note about the whisker sensor is that the raw measurements are quite noisy, so a low pass filter with a cut off frequency of $0.5$ Hz was applied to the signal to improve the signal to noise ratio. 

\subsubsection{WindProbe}

The \textit{WindProbe} is a multi-hole pitot-static probe that is sold alongside the WindShaper.
The datasheet reports a standard velocity range between $2$ and $20$ m/s, a flow velocity error of $\pm2\%$ of the \revision{full-scale} range, and a flow angle error of $\pm2$ degrees.
Raw data is sent at a rate of $100$ Hz from the \textit{WindProbe} to the WindShaper via a USB cable connected to the fan array, and the WindShaper server converts this data into measurements of the 3D velocity vector in the sensor's frame. 
The novelty of the \textit{WindProbe} is that it is designed to be tracked using motion capture, and the WindShaper software uses measurements of the pose from motion capture to associate the \revision{wind} velocity vector measurements with the position\revision{, ground velocity,} and orientation of the probe. 
By scanning the probe around the motion capture space, the \textit{WindProbe} was used to get an unbiased view of the time-averaged flow fields generated by the WindShaper. 
Through this procedure, it was possible to obtain a ``ground truth'' measurement of the wind for analysis with the other biased wind prediction schemes. 
For all of the experiments involving the WindShaper, the \textit{WindProbe} was at a later point meticulously swept through the motion capture space to obtain another reference for the wind flow fields. 

\subsection{Compute}

The Crazyflie has very limited on board compute, and most of the on board resources are taken up handling the lower level state machine and control logic. 
In order to run the algorithms in the preceding chapters, off-board computers are required. 
That said, a critical aspect of this thesis is the development and demonstration of computationally tractable methods for wind prediction and motion planning. 
In recognition of this goal, the experiments in this chapter utilize two off-board computers each with distinct purposes. 

The core wind estimation and motion planning algorithms, which are \revision{ultimately} meant to run on board the UAV, are all implemented on an Nvidia Jetson AGX Xavier. 
This single-board computer has an $8$-core NVIDIA ARM CPU and a dedicated GPU with $512$ CUDA cores and $32$ GB of shared memory. 
The Nvidia Jetson AGX Xavier uses just $30$ Watts at maximum power and weighs around $650$ grams if the heat sink and outer casing are included.  
Because of the low weight and power requirements, Nvidia single-board computers like the Jetson have been used as dedicated flight computers for fully autonomous UAVs in the $5$-$10$ kg range, mostly for applications in computer vision and SLAM \cite{ayoub2021jetsonexample, palmas2022jetsonexample, tao2025halo}.
For this reason, we consider the Nvidia Jetson AGX Xavier to be a ``flight-ready'' computer, in the sense that with only minor modifications this computer can be mounted on a suitable aerial platform and run the exact same algorithms we present in this chapter. 

The second off-board computer is a Beelink AMD Mini PC with an $8$-core AMD Ryzen $7$ $4800$u CPU.
The primary purpose of this supporting computer is 1) to provide the networking and communication infrastructure via Robot Operating System (ROS) between the Crazyflie UAV, motion capture system, and the Jetson, and 2) to simulate the LiDAR sensor, which is a computationally expensive process that would not be necessary on a full-size autonomous platform that already has a dedicated LiDAR on board.
In other words, the simulated LiDAR is an artifact of the experimental environment and therefore its associated computational load is isolated from the Nvidia Jetson. 
\revision{Another purpose for the Beelink is providing necessary and helpful visualizations during operation.}
\autoref{fig:realworld:experimentdiagram} clarifies exactly what algorithms are running on each computer. 

\revision{

\section{Dynamic Similitude and Scaling Laws}\label{sec:realworld:similitude}

A fundamental challenge in sub-scale validation of the wind-aware autonomy stack is ensuring that the physical interactions observed in the laboratory are representative of those found in the full-scale operational domain. 
Dynamic similitude ensures this property by matching the ratios of the forces and torques imparted on the flyer from the fluid medium between the sub- and full-scale environments. 
Full dynamic similitude means that \textit{all} dimensionless quantities are simultaneously matched, but this is seldom possible without specialized pressurized wind tunnels. 
Nevertheless, useful validation of the wind-aware autonomy stack presented herein can still be achieved if the dominant flow characteristics and controller response timescales are consistent between the two scales.
This section explores the scaling laws relevant to flight in urban wind fields and draws careful boundaries on the lessons to be learned from the experiments conducted in this work.

To ground the analysis, we first estimate a length scaling ratio $\lambda$ between the full scale (F.S.) and model scale (M) environments:
\begin{equation}
    L_{F.S} = \lambda L_{M}
\end{equation}
For illustrative purposes only, \autoref{tab:realworld:aircraftcomparison} compares the Crazyflie used in our experiments to a representative eVTOL air taxi: the Joby S4.
Using the rotor diameter as the relevant length scale, we can approximate the length scaling as $\lambda \approx 32$, which will be used to understand how relevant dimensionless quantities scale from the laboratory to the real world. 
For additional context, with this scaling parameter a standard urban canyon width of $10$ m scales to approximately $31$ cm in the WindShaper facility which is consistent with the spacing used in our experiments. 

\begin{table}[t!]
\centering
\caption{A rough comparison between the \revision{sub-scale} flyer (Crazyflie 2.1-Brushless) and the Joby S4 prototype aircraft.}
\label{tab:realworld:aircraftcomparison}
\resizebox{\columnwidth}{!}{%
\begin{tabular}{@{}lllll@{}}
\toprule
Vehicle & Mass (kg) & Rotor Diameter (m) & Wingspan (m) & Max Speed (m/s) \\ \midrule
Crazyflie 2.1-Brushless* & 0.057 & 0.055 & - & - \\
Joby S4 Prototype** & 2000\dag & 1.8 & 10.6 & 89.4 \\ \bottomrule
\end{tabular}%
}
\begin{minipage}{\columnwidth}
\small
\vspace{1mm}
\raggedright*Values taken from \texttt{https://www.bitcraze.io/2023/10/say-hello-to-the-crazyflie-2-1-brushless/}. \\
\raggedright**Values taken from \texttt{https://evtol.news/joby-s4}. \\
\raggedright$^{\dag}$ Rough order of magnitude, a value between the reported empty weight and maximum takeoff weight.
\end{minipage}
\end{table}

\subsection{Reynolds Number Independence in Urban Flows}

The Reynolds number expresses the ratio of inertial forces to viscous forces, defined as 
\begin{equation}\label{eq:experiments:reynolds}
    Re = \frac{UL}{\nu}
\end{equation}
where $U$ is a characteristic flow velocity, $L$ a characteristic length, and $\nu$ the kinematic viscosity of the fluid medium. 
In our experiments, the kinematic viscosity is the same for both the model and the full scale environment because the fluid remains the same. 
Thus, Reynolds similarity requires that 
\begin{equation}\label{eq:experiments:reynolds_sim}
    U_M L_M = U_{F.S.} L_{F.S} \implies U_{M} = \lambda U_{F.S}
\end{equation}
For a nominal full-scale wind speed of $U_{F.S} = 5$ m/s, \autoref{eq:experiments:reynolds_sim} means that the WindShaper would have to produce wind at $160$ m/s.
Even if this was possible, such a high freestream velocity would likely start introducing compressibility effects that would violate other scaling laws. 

Besides this fact, it can be argued that strict Reynolds matching is unnecessary in the case of urban wind fields. 
The reason for this is that dominant urban wind features are generated by flow separation around the many sharp edges that define urban landscapes. 
Unlike streamlined bodies like airfoils, where the separation point shifts with Reynolds number, flow separation on sharp-edged geometry remains fixed at the corners--this can be observed in the various urban wind simulations presented throughout this thesis.
Townsend \cite{townsend1976structure} and Snyder \cite{snyder1981guideline}, alongside others much later in the wind engineering community \cite{uehara2003reynoldsindependence, cui2014reynoldsindependence, chew2018reynoldsindependence, shu2020reynoldsindependence, zhu2025reynoldsindependence}, have studied this notion and its implications in the principle of \textit{Reynolds number independence}: provided the flow is already fully turbulent, and thermal or Coriolis effects can be neglected, the macroscopic topology of building wakes--the size and number of the recirculation zones as well as the location of shear layers--remains approximately invariant with respect to Reynolds number.
This phenomena has been studied theoretically, for instance by Shu \textit{et al.} \cite{shu2020reynoldsindependence}, and measured experimentally for grid-like city layouts with cuboid buildings, although the exact $Re$ for transition depends on the aspect ratios associated with the buildings and street widths.
For example, a more recent study by Zhu \textit{et al.} \cite{zhu2025reynoldsindependence} and references therein suggests the critical Reynolds number can reach an order of magnitude as high as $10^4$ when considering 3D flow around buildings with higher height-to-width aspect ratios. 
In our sub-scale experiments, the Reynolds number associated with the obstacles is estimated around $200{\small,}000$.
Considering that the obstacle surfaces were made of rough plywood and that the WindShaper itself produces a minor amount of turbulence, it is more likely than not that the flow is fully turbulent.
Thus by the principle of Reynolds number independence, it is likely that the spatial structures of the wind fields generated in the sub-scale environment are consistent with full-scale urban flows. 
However, this concept was not a consideration during the design of the obstacles, and so future work on the design of these obstacles should explicitly factor in Reynolds number independence.

\begin{figure}[t!]
    \centering
    \includegraphics[width=0.99\textwidth]{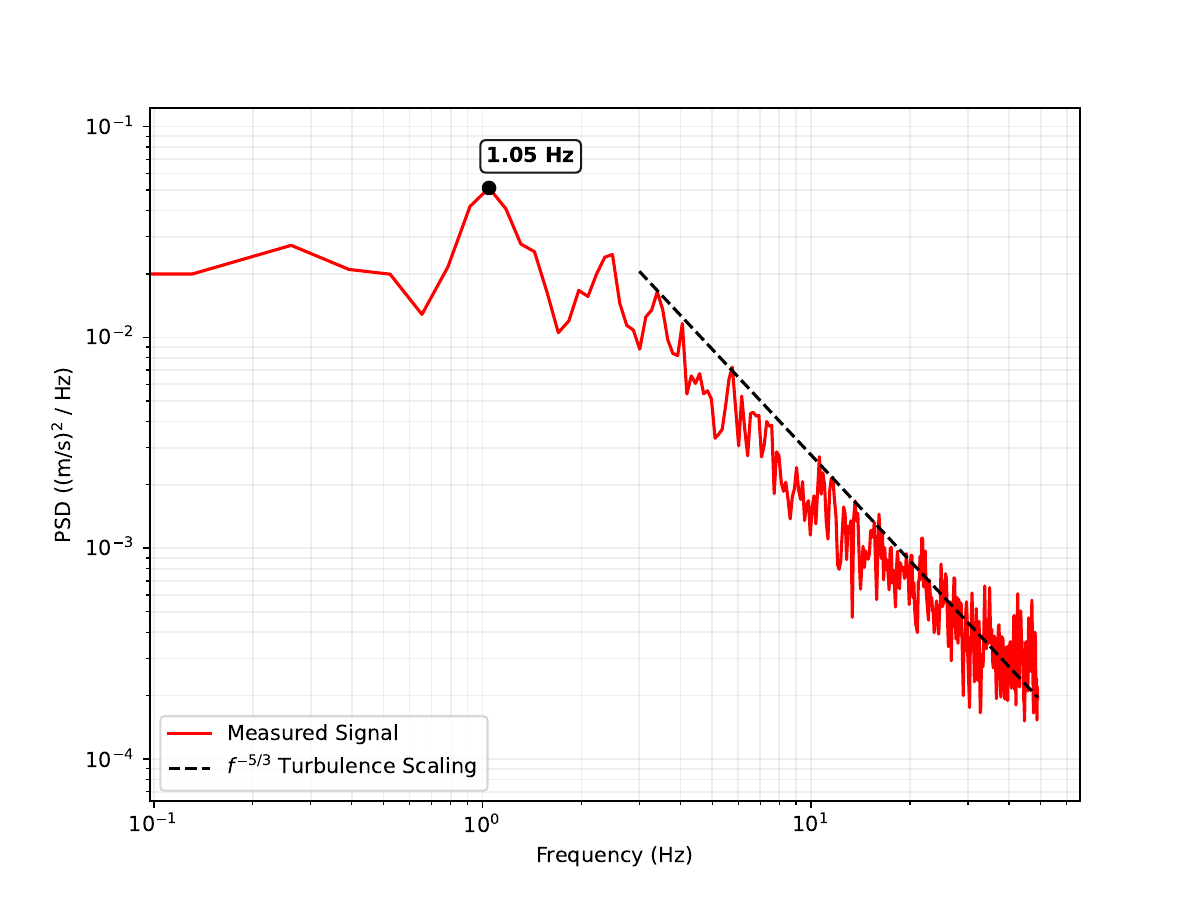}
    \caption{\revision{The power spectral density (PSD) of the wake velocity behind the $16$$\times$$16$ inch obstacle, as measured by the \textit{WindProbe} positioned roughly $0.4$ m downstream from the obstacle's leeward side while the WindShaper operated at $20\%$ power. For additional context, Kolmogorov's -$5/3$ turbulence scaling law \cite{kolmogorov1941equations} is plotted for the higher frequencies.}}
    \label{fig:realworld:psd_20percent}
\end{figure}

\subsection{Timescales and the Strouhal Number}

For autonomous navigation, consistent scaling is critical not just in space but also in time: the relationship between the vortex shedding frequency, $f_s$, and the UAV's planning and control bandwidths, $f_{control}$, should ideally be matched to understand how urban wind disturbances affect aircraft stability.
The unsteady flow phenomena associated with vortex shedding can be characterized by the Strouhal number:
\begin{equation}\label{eq:realworld:strouhal}
    St = \frac{f_s L}{U}
\end{equation}
If the Strouhal number between the sub- and full-scale environments were equal, it would imply that the frequency of gusts scales inversely with length ($f \propto 1/L$).
With this notion we anticipate that sub-scale wake vortices would shed at higher frequencies than those at the full scale.
At the same time, the control bandwidth for UAVs (as measured by the acceleration $a$) also scales inversely with length ($a \propto 1/L$) \cite{wolowicz1979similitude, kushleyev2013towards}, so the ratio of these two timescales should, in theory, be similar going from sub to full scale. 

To investigate this, we measured the frequency spectrum in the wake of one of the obstacles using the \textit{WindProbe}. 
\autoref{fig:realworld:psd_20percent} shows the power spectral density of the wake velocity behind the $16$$\times$$16$ inch obstacle while the WindShaper operated at $20\%$ power.\footnote{\revision{Based on calibration data in \autoref{appendix:complete_windshaper_experiments}, \autoref{fig:appendix:windshapercharacterization}, this power level corresponds to a freestream velocity of approximately $4$ m/s.}} 
The data shows a peak at $1.05$ Hz corresponding to the dominant vortex shedding mode--for this obstacle size and freestream velocity the Strouhal number is $St\approx0.10$, which not only agrees with the Strouhal numbers seen in our full-scale simulations (\autoref{sec:prediction:euler_sim}), but also with Strouhal calculations found in other experimental studies \cite{norberg1993strouhal, nakaguchi1968experimentalcylinder}. 
This analysis substantiates the claim that the Strouhal number remains approximately constant between the WindShaper facility and the full scale simulations carried out in previous chapters. 

As for the control frequency, in the sub-scale environment the UAV receives acceleration commands at approximately $50$ Hz, which indicates a critical time scale separation between the planning loop and unsteady flow dynamics.
The fact that the planning rate is two orders of magnitude higher than the dominant shedding frequency confirms that the sub-scale UAV operates within the necessary bandwidth to react to these unsteady flow features. 
However, the crucial point is that because both the flow frequencies and the vehicle control bandwidths scale inversely with length \cite{kushleyev2013towards}, this specific ratio of timescales ($f_{control}/f_{shed} \approx 50$) is likely also representative of the full-scale problem. 
In other words, while the sub-scale aircraft has higher agility and control authority, the dominant gust frequency building wakes increases similarly such that the relationship between the control and disturbance bandwidths is maintained between the two environments. 

\subsection{Inertial Scaling and the Froude Number}

For subsonic free-flight dynamic model testing, Froude scaling has long since been the standard for using sub-scale models to understand full-scale dynamic behavior \cite{chambers2010modeling}.
The Froude number ($Fr$) is a dimensionless quantity describing the balance between inertial forces and gravitational forces. 
\begin{equation}\label{eq:realworld:froude}
    Fr = \frac{U^2}{gL}
\end{equation}
Matching the Froude number is what ensures that geometric characteristics of the flight trajectories, for instance the displacement of the flyer due to a gust, are similar between the two environments.

Under Froude scaling, velocity scales by the square root of the length ratio ($V_{M} = V_{F.S.} / \sqrt{\lambda}$). 
In the case of $\lambda=32$, the velocity scaling factor is $\sqrt{32} \approx 5.66$. 
The experiments in this work were conducted at wind tunnel speeds ranging from $3$ to $5$ m/s while the UAV was typically flying at a ground speed of $1$ to $2$ m/s. 
Applying Froude scaling, these velocities roughly correspond to full-scale velocities of $17$ to $28$ m/s wind and $5$ to $10$ m/s ground speeds. 
This flight envelope is close to what is expected during the urban maneuvering and approach phase of an eVTOL mission during extremely high winds, which is a critical regime for obstacle avoidance and wind rejection for future urban air mobility systems.

In addition to velocity, Froude similarity imposes requirements on the mass of the sub-scale vehicle to ensure the inertial authority of the platform--its resistance to gusts relative to its size--is representative of the full-scale aircraft.
Chambers \cite{chambers2010modeling} notes that the rule of thumb for maintaining Froude similitude is that mass scales volumetrically with the length ratio cubed ($m \propto \lambda^3$).
Using the reference mass of the Joby S4 ($m_{F.S.} \approx 2000$ kg) and the length scale factor of $\lambda = 32$, the target mass for the sub-scale flyer is $61$ g. 
As mentioned in \autoref{sec:realworld:robot}, the UAV platform used in these experiments weighs approximately $57$ g (including battery and additional sensors), which is remarkably close to the theoretical target.
This coincidental matching of the mass parameter is significant: it implies that the acceleration imparted on the UAV by a wind gust of a given velocity is dynamically consistent with the full-scale prototype. 
Consequently, the disturbance rejection challenges faced by the controller in the wind tunnel are physically representative of those experienced by a full-scale air taxi.

\subsection{Outlook on Similitude}

It is important to reiterate that full dynamic similitude is challenging if not impossible for atmospheric testing at such a small scale because of the Reynolds number. 
However, there are compelling arguments in the literature, substantiated by experimental evidence, that suggest that Reynolds similarity is not the primary quantity of interest when determining what can be learned from our sub scale testing. 
If we instead look at Froude scaling and the timescales associated with the unsteady flow phenomenon relative to control authority, we arrive at the surprising conclusion that many of the dynamic characteristics associated with flight in the urban canyon are approximated in the WindShaper facility. 
As such we argue that the experiments to follow in this chapter are close to satisfying many of the critical requirements for physical validity between the sub- and full-scale urban environments. 
}

\section{Experiments and Results}\label{sec:realworld:experiments}

There were two categories of experiments conducted in the WindShaper facility:
\begin{enumerate}
    \item \textbf{Aerodynamic Characterization}: The UAV was commanded to hover at a particular position while the WindShaper operated at varying speeds. 
    \item \textbf{Flow Traversal}: The UAV was commanded to track the $x$ and $y$ position of a wand, which was hand controlled by a human operator, while holding a constant $z$ position. For these experiments, the WindShaper ran at constant speed. 
\end{enumerate}
The \textbf{Aerodynamic Characterization} experiments were solely used for system identification and tuning purposes.
In the \textbf{Flow Traversal} series of experiments, different building configurations (see the right side of \autoref{fig:realworld:facilities}) with anywhere from $0$ to $4$ buildings were immersed in the wind generated by the WindShaper to create interesting urban-like flow features. 
These experiments were primarily to evaluate the \textit{Wind UKF}, but also to demonstrate the full system (\textit{Wind UKF}, wind decoder network, and wind-aware MPPI) running together in real time.

The full list of experiments is in \autoref{appendix:complete_windshaper_experiments},  \autoref{tab:realworld:experiment_description}.
The remainder of this chapter will describe the insights gained from these experiments.

\subsection{Static Thrust and Motor Characterization}

Having accurate measurements of the rotor speed and static thrust for each rotor is essential to the implementation of the \textit{Wind UKF}.
The reasons for this are simple: the aerodynamic drag models in \autoref{sec:dynamics:aero} depend on rotor speeds for the calculation of the drag forces on the UAV. 
Accelerations due to drag are the only direct signals associated with lateral ($x$, $y$) wind components in the UKF and thus a necessary part of ensuring observability. 
Also, static thrust adds acceleration to the vertical ($z$) axis of the UAV, so being able to subtract that signal from the accelerometer measurement will distinguish the vertical accelerations due to the $z$ wind component.
However, most commercially-available small UAVs including the one used in this thesis do not provide direct rotor speed measurements, and because the UAV is free-flying the thrust cannot be directly measured either. 

In fact, the only signal related to rotor speeds available during flight is the \textit{commanded} pulse-width-modulation (PWM) signal sent to each motor \revision{represented by} a $16$-bit integer. 
Leveraging this limited on board measurement required the Crazyflie UAV to be mounted vertically on a Tyto Series $1520$ Thrust Stand\footnote{\url{https://www.tytorobotics.com/pages/series-1520}} as seen in the inset of \autoref{fig:realworld:thrust_sysid}.
The thrust stand provided measurements of the rotor speeds using an optical RPM probe as well as the net force (in other words, the thrust) via a calibrated load cell. 
Using a bench top power supply, power was supplied to the UAV at constant voltages ranging from $2.7$ V to $4.1$ V, which encompasses the range of battery supply voltages during actual flight. 
Using Bitcraze's Python API and the Crazyflie's on board electronic speed controllers (ESCs), a sequence of PWM commands in the range \revision{$[10{\small,}000, 45{\small,}000]$} with increments of \revision{$5{\small,}000$} were sent to the Crazyflie over radio. 
Each PWM command was held for $1$ second, during which time the command was logged alongside the supply voltage, rotor speed, and force measurement from the thrust stand at rates of $50$ Hz. 

\begin{figure}
    \centering
    \includegraphics[width=0.99\textwidth]{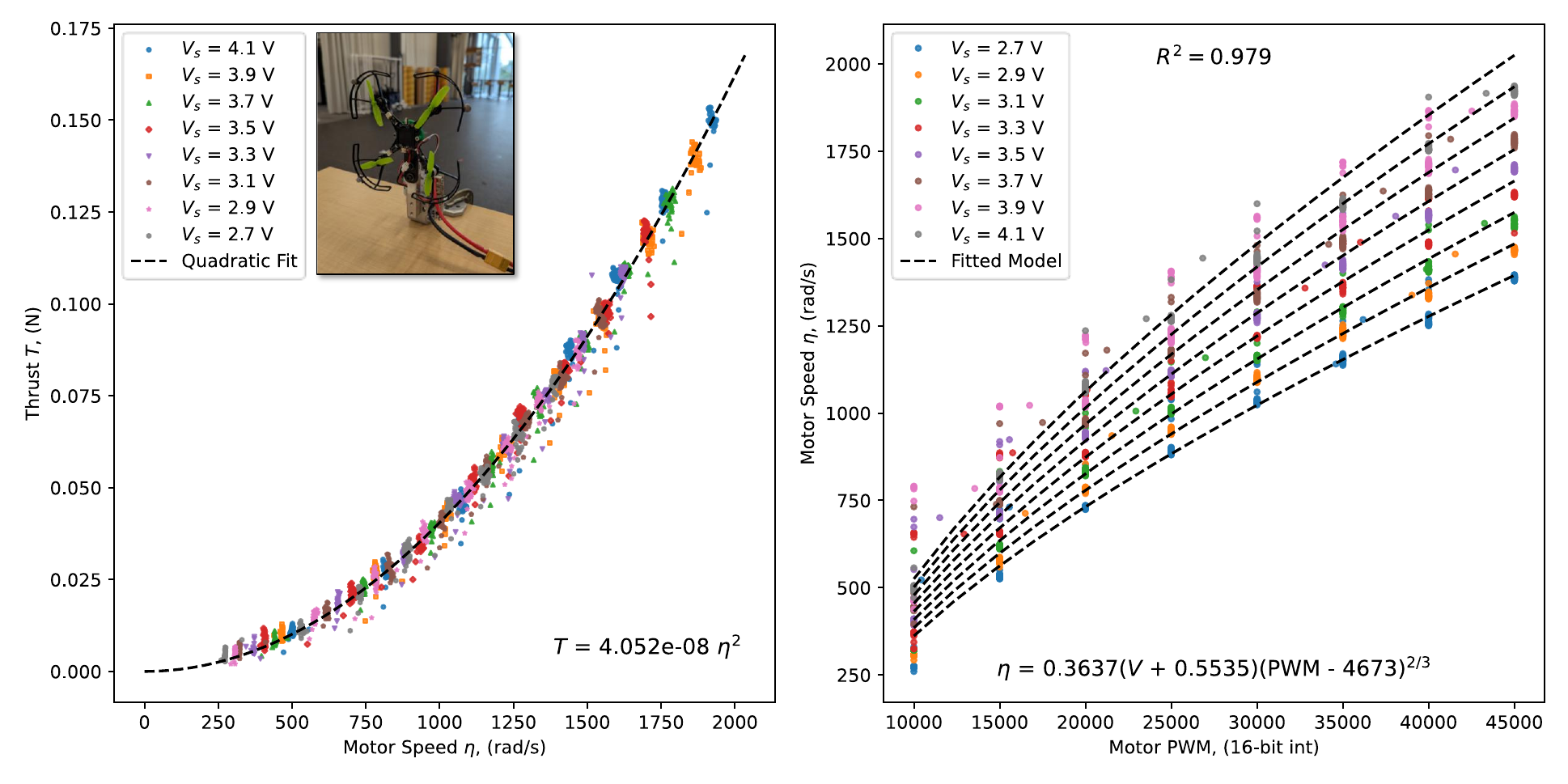}
    \caption{Identification of relationships between thrust, motor speed, input voltage, and commanded pulse width modulation (PWM) for a single rotor of the Crazyflie 2.1-Brushless. The bench top experimental setup is seen in the inset.}
    \label{fig:realworld:thrust_sysid}
\end{figure}

The results from this experiment are summarized in \autoref{fig:realworld:thrust_sysid}. 
On the left side, the thrust of a single rotor is plotted against the rotor speed. 
As expected, the thrust force varies quadratically with the rotor speeds and this relationship does not depend on the supply voltage.
From this data the static thrust coefficient was calculated as $k_\eta = 4.052\cdot10^{-8}$ $N$-$(\text{rad}/s)^{-2}$ using least squares regression. 
On the right side, rotor speed is plotted against the commanded PWM signal for varying supply voltages.
Here it is apparent that for a given PWM command, the supply voltage is an important variable in determining the resulting rotor speed. 
The following model was fitted to this data:
\begin{equation}\label{eq:realworld:rpmmapping}
    \eta = K_v(V_{supply}+V_0)(\text{PWM}-\text{DZ})^{2/3}
\end{equation}
This model is motivated by the model for an ideal DC motor. 
The coefficients $K_v$ and $V_0$ effectively capture the steady state motor constants, while $\text{DZ}$ represents the effective ``dead zone'' of the motor in terms of the PWM command.
The $2/3$ exponent was selected by hand to capture the observed sublinear increase in motor speed as a function of PWM. 
Note that this model is only valid for $\text{PWM} \geq \text{DZ}$.
From the data in \autoref{fig:realworld:thrust_sysid}, the coefficients were determined using least squares regression as $K_v = 0.3637$ $(\text{rad}/s)$-$V^{-1}$, $V_0 = 0.5535$ $V$, and $\text{DZ}=4{\small,}673$, with a coefficient of determination of $0.979$. 

Putting these two models together, we now have a means of generating \textit{pseudo} measurements of the rotor speeds and static thrust force during free flight, leading to more effective isolation of the aerodynamic forces contributing to the accelerometer measurement, and therefore more accurate estimation of the local wind.
\revision{This model could be significantly improved by expanding the thrust measurements to include varying inclinations to different wind speeds generated by the WindShaper, but this is left to future work.}

\subsection{Aerodynamic and Power Characterization}

The brushless Crazyflie is a newer platform that to date has not seen the same level of characterization of its former brushed version \cite{forster2015system}. 
\revision{This} section \revision{presents} data and subsequent system identification \revision{using} the WindShaper facility \revision{alongside on board} inertial measurements from the Crazyflie. 
During the \textit{Aerodynamic Characterization} experiment, the Crazyflie was commanded to hover at a position roughly $2$ meters away from the WindShaper at a height of $1.15$ meters, which \revision{approximately centered the UAV within the} fan array.
The WindShaper \revision{operated at discrete speeds} between $0$\% and $50$\% power\footnote{\revision{This power level corresponds to approximately $8$ m/s wind speed based on \autoref{fig:appendix:windshapercharacterization}.}} with intervals of $10$\%, holding each power level for $15$ seconds at a time. 
During a previous experiment, the wind at the nominal hovering location was recorded using the \textit{WindProbe} for a range of power \revision{levels--the resulting} calibration data shown in \autoref{appendix:complete_windshaper_experiments} provided the wind speed which, when combined with \revision{ground speed obtained from} motion capture, \revision{was sufficient for calculating} the relative airspeed of the \revision{UAV.} 

\begin{figure}[b!]
    \centering
    \includegraphics[width=0.7\textwidth]{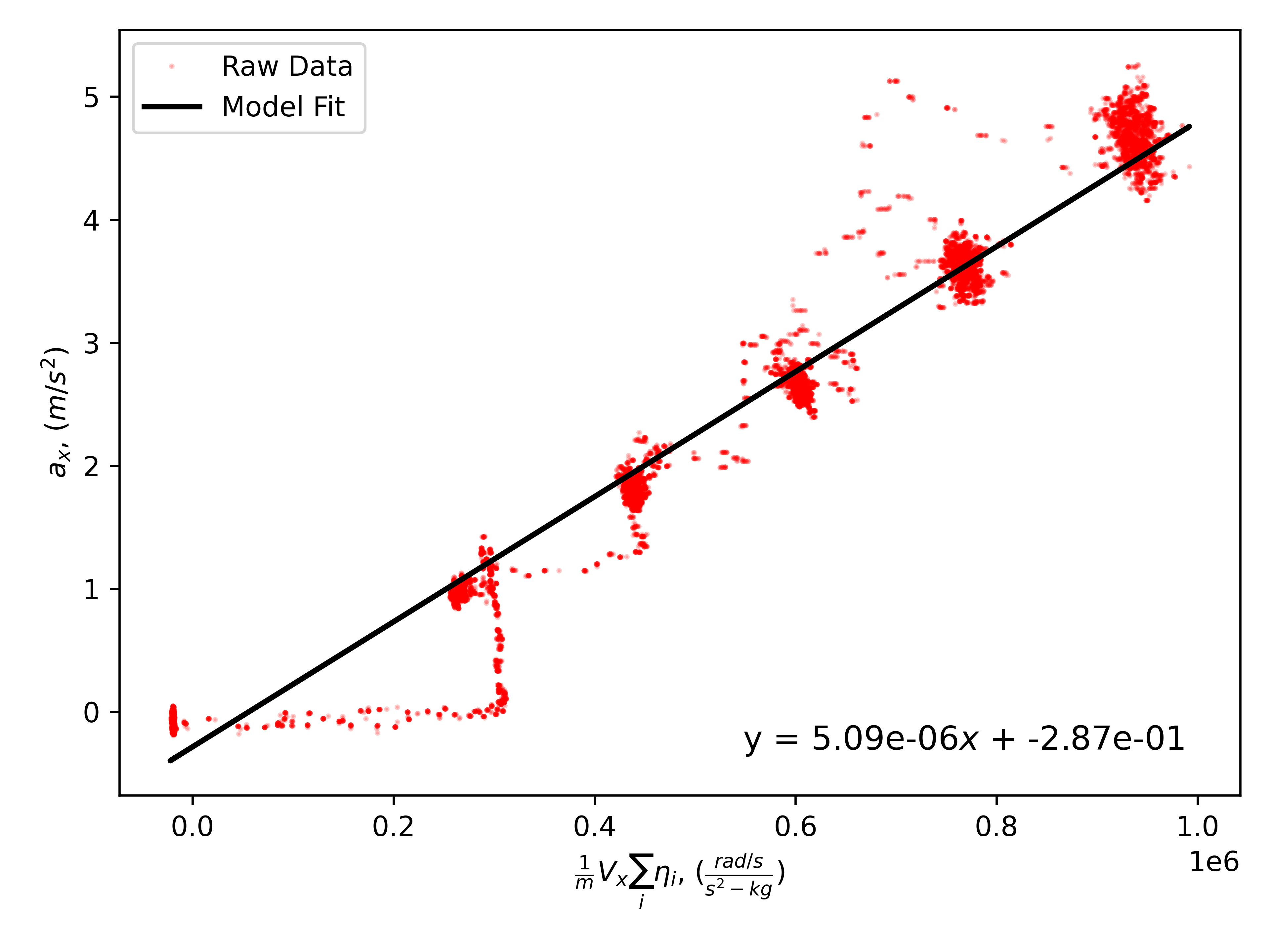}
    \caption{Identification of the induced drag coefficient, $k_d$, for the Crazyflie via linear regression, following the procedure outlined in \cite{svacha2017improving}. }
    \label{fig:realworld:kd_fit}
\end{figure}

The data from \autoref{fig:realworld:kd_fit} was used to validate the rotor drag model described in \autoref{sec:dynamics:aero} following the procedure in \cite{svacha2017improving}. 
In this plot, the lateral accelerations are plotted against the product of the wind speed and sum of the rotor speeds \revision{following the functional form of rotor drag discussed in \autoref{sec:dynamics:aero}.} 
The slope of the linear fit to this data corresponds to the induced drag coefficient, $k_d$, while the intercept is the bias on the accelerometer \revision{on the $x$ body axis.} 
The \revision{observed linear relationship in \autoref{fig:realworld:kd_fit} strongly encourages} that \autoref{eq:rotor_drag} is an appropriate model for the lateral drag forces on the \revision{Crazyflie.}

Next, the data collected during the hover experiment was used to identify the effective vertical drag $k_z$ and translational lift $k_h$ coefficients. 
Following \revision{Svacha \textit{et al.}} \cite{svacha2017improving}, in steady state forward flight the following relationship holds approximately: 
\begin{equation*}
    m a_z = k_\eta \sum_{i=1}^N \eta_i^2 - k_z V_z \sum_{i=1}^N \eta_i + N k_h V_h^2
\end{equation*}
where $a_z$ and $V_z$ are the components of the acceleration and airspeed, respectively, aligned with the $\bl{b}_3$ axis, and $V_h$ is the projection of the airspeed vector onto the $\bl{b}_1$-$\bl{b}_2$ plane. 
Given $M$ measurements, the mass-normalized parameters $\hat{\bl{z}}:=[k_\eta', k_z', k_h']$ can be jointly estimated by solving the following linear system: 
\begin{equation*}
    \underbrace{\begin{bmatrix} \vert & \vert & \vert \\ \left[ \sum_{i=1}^N \eta_i^2 \right]_k & \left[V_z \sum_{i=1}^N \eta_i \right]_k & \left[ \frac{1}{N}V_h^2 \right]_k \\ \vert & \vert &\vert \end{bmatrix}}_{A\in\mathbb{R}^{M\times3}} \hat{\bl{z}} = \underbrace{\begin{bmatrix} \vert \\ \left[ a_z \right]_k\\ \vert \end{bmatrix}}_{\bl{y}\in\mathbb{R}^M}, \quad k=1, ..., M
\end{equation*}
The physical coefficients can be recovered by multiplying $\hat{\bl{z}}$ by the UAV mass. 
Using least squares regression on the data shown in \autoref{fig:realworld:kwkzkh_fit}, the parameters for the Crazyflie were calculated as $k_\eta = 4.26\cdot10^{-8}$ N-(rad/s)$^{-2}$, $k_z = 1.19\cdot10^{-5}$ N-(m-rad)$^{-1}$-s$^{2}$, and $k_h=1.11\cdot10^{-3}$ N-m$^{-2}$-s$^2$.

\begin{figure}[h]
    \centering
    \includegraphics[width=0.99\textwidth]{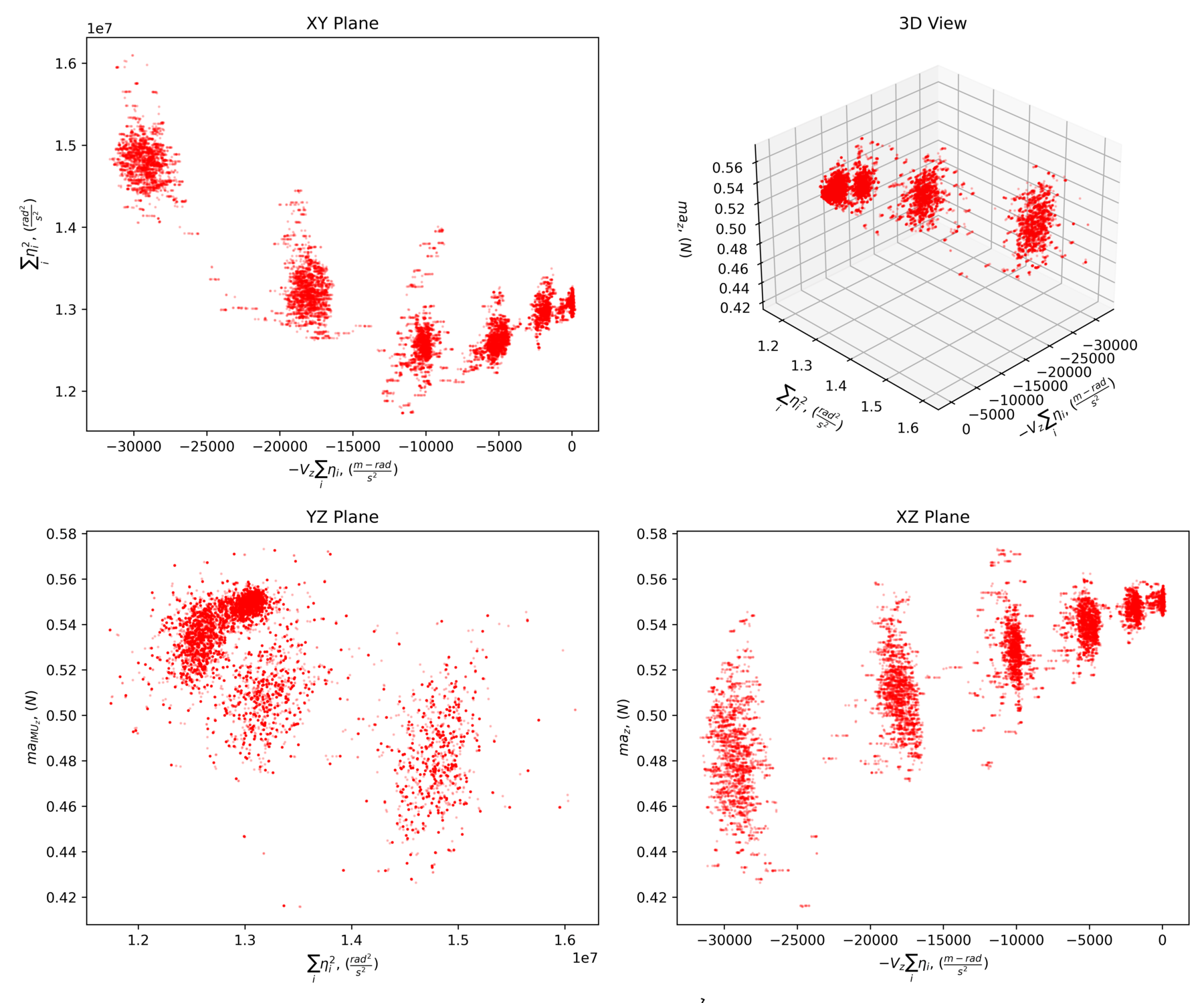}
    \caption{Data used for the identification of the thrust $k_{\eta}$, inflow $k_z$, and translational lift $k_h$ coefficients for the Crazyflie, following the procedure outlined in \cite{svacha2017improving}.}
    \label{fig:realworld:kwkzkh_fit}
\end{figure}

The full set of parameters identified from the hovering experiment are summarized in \autoref{tab:realworld:aerocoeffs}. 
An interesting point of comparison is between the thrust coefficients identified from bench top testing (\autoref{fig:realworld:thrust_sysid}) versus the free flight data in \autoref{fig:realworld:kwkzkh_fit}. 
The thrust coefficients identified from both experiments are comparable, \revision{the small differences can be} attributed to biases in the on board accelerometer measurement or the fact that in the bench top testing the UAV is mounted at a $90$ degree pitch angle, which \revision{changes} how the rotor \revision{wakes} interact with the \revision{rotors.} 

\begin{table}[b!]
\centering
\caption{Lumped parameter aerodynamic coefficients identified using the procedure in \cite{svacha2017improving}. The magnitudes of each coefficient are indicated in the parenthesis.}
\label{tab:realworld:aerocoeffs}{%
\begin{tabular}{@{}cccc@{}}
\toprule
\textbf{Parameter}             & \textbf{Variable} & \textbf{Value}         & \textbf{Unit} \\ \midrule
Thrust Coefficient             & $k_{\eta}$        & $4.26(10^{-8})$ & $\frac{N}{(\text{rad}/s)^2}$  \\
Induced Drag Coefficient       & $k_d$             & $5.09(10^{-6})$ & $\frac{N-s^2}{m-\text{rad}}$  \\
Inflow Coefficient             & $k_z$             & $1.19(10^{-5})$ & $\frac{N-s^2}{m-\text{rad}}$   \\
Translational Lift Coefficient & $k_h$             & $1.11(10^{-3})$ & $\frac{N s^2}{m^2}$      \\ \bottomrule
\end{tabular}%
}
\end{table}

Finally, the \textbf{Aerodynamic Characterization} experiment was used \revision{to characterize the steady state power consumption} as a function of airspeed. 
In \autoref{fig:realworld:power_vs_airspeed}, the total power consumption as measured by the on board power sensor is plotted against the airspeed. 
Following \autoref{sec:dynamics:power}, a cubic model \revision{was fitted} to this data.
The hover power is approximately $10.56$ Watts, \revision{which means} the hover efficiency of the brushless Crazyflie at $0.188$ \revision{W/g.} 
There is \revision{a notable} decrease in power consumption for airspeeds between $2$ m/s and $5$ m/s, with a minimum power of $10.12$ Watts at an airspeed of $4.15$ m/s \revision{establishing} the maximum endurance speed for the Crazyflie.
At this speed, the Crazyflie consumes $4.1\%$ less power than at hover. 
\revision{Also, the no-wind} maximum range speed is calculated \revision{as} $7.73$ \revision{m/s using on the range tangent line method described by Leishman \cite{leishman2006principles}.}

\begin{figure}
    \centering
    \includegraphics[width=0.7\textwidth]{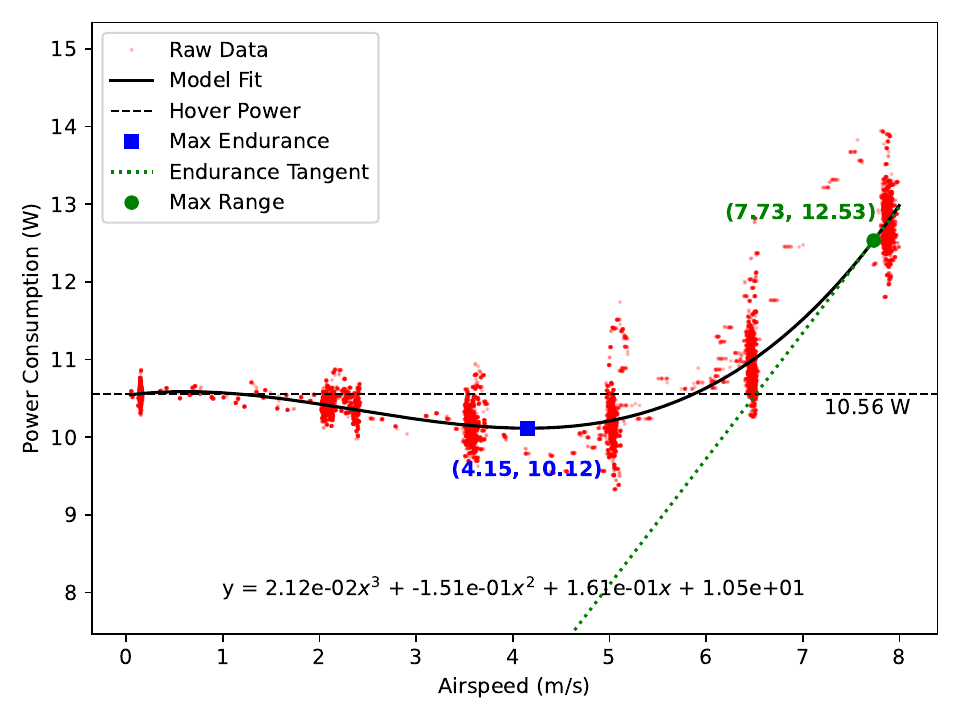}
    \caption{Identification of a cubic power consumption model as a function of airspeed for the Crazyflie. The data is annotated with the hover power and max range and endurance speeds (assuming no wind). The max range speed was determined via the range tangent line, which extends down to \revision{the point (0, 0).}}
    \label{fig:realworld:power_vs_airspeed}
\end{figure}

\revision{
The power curve shown in \autoref{fig:realworld:power_vs_airspeed} exhibits the bucket shape that is characteristic of rotorcraft flight, governed primarily by the onset of translational lift.
In hover, the rotors actively recirculate their own wake which reduces the effective angle of attack of each rotor increasing the power required to produce sufficient thrust.
As the UAV speeds up ($0$ to $4$ m/s), the tip vortices fall trail behind the flyer and their effects on the rotor plane is reduced--this can also be understood as the UAV flying into cleaner and less disturbed air--so the result is more efficient thrust generation from each rotor. 
The observed decrease in total power suggests that the efficiency gains from translational lift are still relevant at this scale, having a small yet discernible impact (about $4\%$ as previously reported) on the power requirements for the Crazyflie despite its small size.
However, this magnitude of the energy reduction due to translational lift is much lower on the Crazyflie compared to full scale helicopters, which sometimes see upwards of $30\%$ or more reduction in power requirements at the max endurance speed \cite{leishman2006principles}. 
Beyond $4.15$ m/s, the parasitic drag of the airframe overcomes these gains, causing the total power consumption to rise.
}

As as point of comparison for this data, Olejnik \textit{et al.} \cite{olejnik2022powerconsumption} ran a similar experiment for the legacy \revision{brushed} version of the Crazyflie and concluded that power was constant with airspeed.
Based on their data, the hover power is estimated to be $8.7$ Watts with a hover efficiency of $0.22$ \revision{W/g.}
In light of \autoref{fig:realworld:power_vs_airspeed}, their analysis that power is constant with airspeed was likely because the maximum tested airspeed ($3.4$ m/s) was too small to see the effect of \revision{translational lift and because of the noise associated with their power measurements.} 

\subsection{Indirect Wind Estimation}\label{sec:realworld:indirectwindestimation}

In this section, data collected from the \textbf{Flow Traversal} series of experiments is used to evaluate the \textit{Wind UKF} on a real UAV subjected to complex spatio-temporal flow fields.  

The \textit{Wind UKF} was implemented in \texttt{C++} and is publicly available on GitHub\footnote{\url{https://github.com/spencerfolk/wind_aware_ros/tree/main/wind_aware_ukf}}--the lumped aerodynamic parameters were identified in the previous experiment. 
In the real world setting, attitude, ground velocity, and body rate measurements were provided by the motion capture system while the acceleration measurements were extracted from the IMU on board the Crazyflie. 
The commanded rotor speeds for each motor were computed with \autoref{eq:realworld:rpmmapping} using the commanded PWM signals and the battery voltage as measured by the on board power sensor.
The measurement covariances were determined based on the expected uncertainty of each measurement source found during calibration, while the process covariances were hand tuned.
\revision{These hyperparameters can also be found on the Github repository linked below.}

\begin{figure}[b!]
    \centering
    \includegraphics[width=0.85\textwidth]{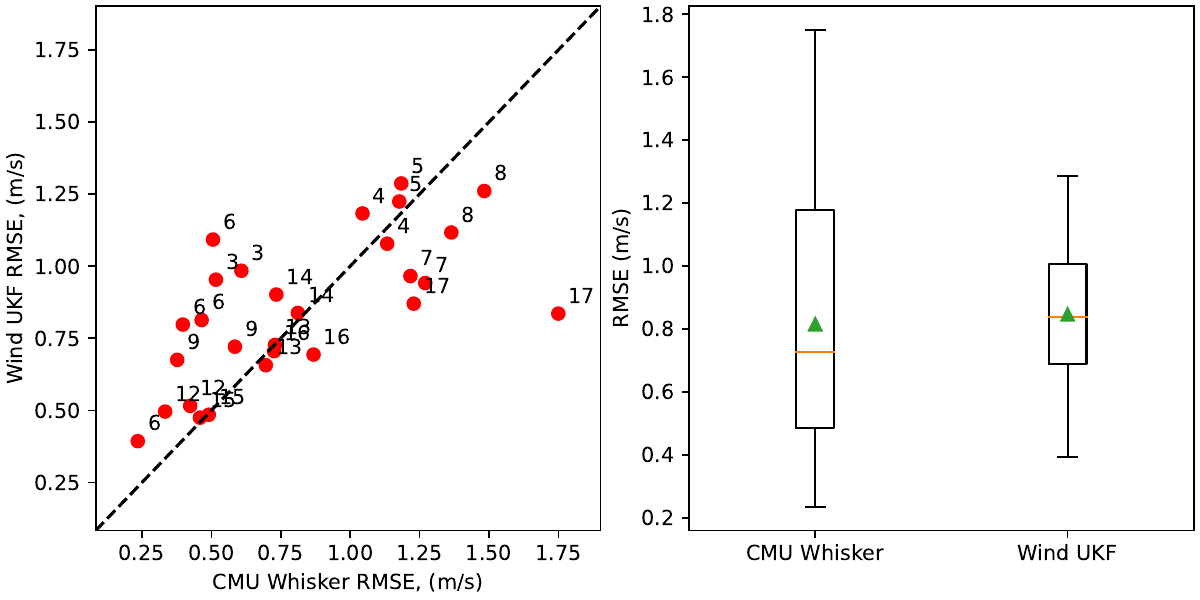}
    \caption{The root mean square error (RMSE) between the \textit{CMU Whisker} and \textit{Wind UKF} across all the experiments. In the left plot, the numbers correspond to the experiment IDs in  \autoref{tab:realworld:experiment_description}.}
    \label{fig:realworld:windrmsescatterboxplot}
\end{figure}

\autoref{fig:realworld:windrmsescatterboxplot} summarizes the performance of the \textit{Wind UKF} in comparison to the \textit{CMU Whisker} sensor. 
On the left, the RMSE corresponding to each method is plotted for all the \textbf{Flow Traversal} experiments. 
A dashed line marks parity between both methods--if the data point lies below this line, the \textit{Wind UKF} outperformed the \textit{CMU Whisker} for this trial, and vice versa. 
Both wind estimation schemes are comparable, with an average RMSE of roughly $0.8$ m/s which is in line with the simulation evaluations of the \textit{Wind UKF} found in \autoref{sec:filtering:simulationresults}.
Although there are slightly more trials where \textit{CMU Whisker} has a lower RMSE compared to the \textit{Wind UKF}, the \textit{Wind UKF} has a tighter spread of RMSE values across the all of the experiments indicating more consistent estimation behavior across all the scenarios.

\begin{figure}[b!]
    \centering
    \includegraphics[width=0.99\textwidth]{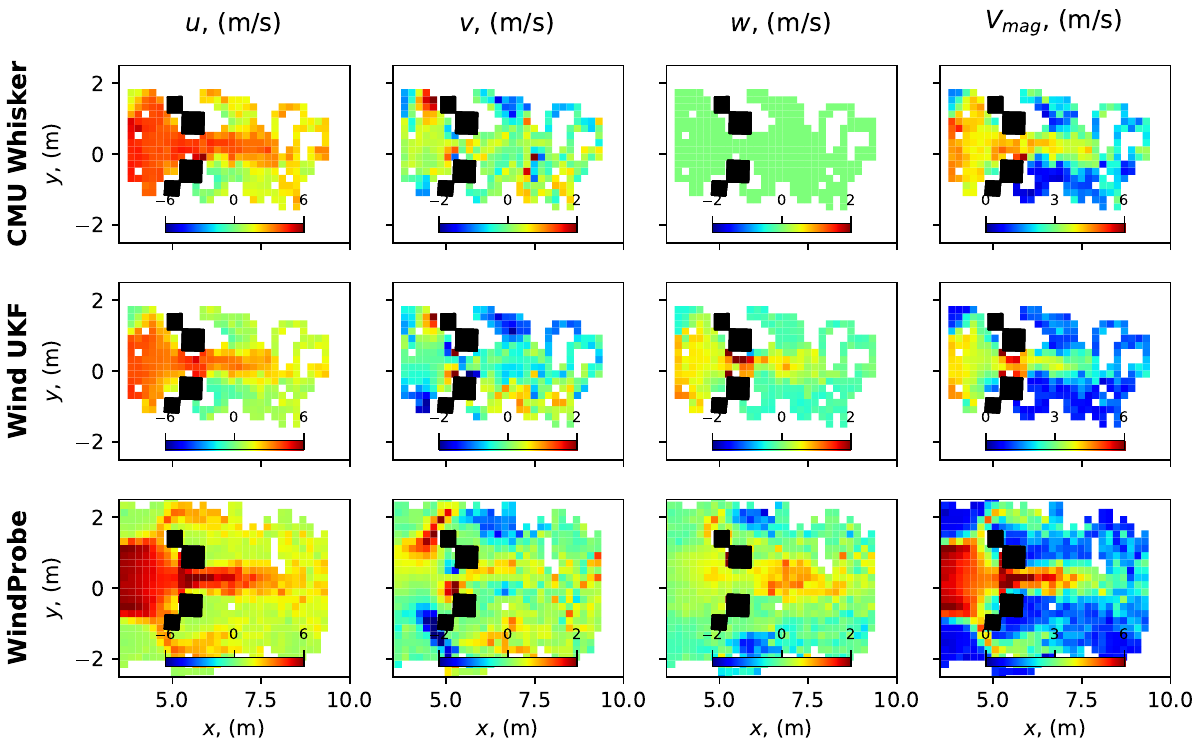}
    \caption{Time-averaged wind flow field reconstructions at $z=0.75$ for experiment 14 of the \textbf{Flow Traversal} series. The \textit{CMU Whisker} and \textit{Wind UKF} measurements were collected from the UAV, while the \textit{WindProbe} was collected during a separate sweep. }
    \label{fig:realworld:flowfieldreconstruction}
\end{figure}

In \autoref{fig:realworld:flowfieldreconstruction}, the time-averaged wind flow field from \textbf{Flow Traversal} experiment $14$ is reconstructed using wind measurements from the \revision{different wind sensing methods.} 
The flow field reconstructions are generated by binning the wind measurements based on their $x$ and $y$ location in the world frame, and then taking the mean of the data sampled in each bin. 
For the \textit{CMU Whisker} and \textit{Wind UKF}, wind measurements were only available along the UAV's trajectory, resulting in a somewhat sparse sampling of the flow field because of the limited battery life, whereas the \textit{WindProbe} was not subject to this limitation and therefore a denser sampling was possible.

For all intents and purposes, the \textit{WindProbe} can be considered a ``ground truth'' with which to compare the other two measurement sources. 
Qualitatively speaking, all three flow field reconstructions are in good agreement with one another. 
Both the \textit{Wind UKF} and \textit{CMU Whisker} capture the spatial structure of the strong column of air in the region [$x\in[5.0, 7.5]$, $y\in[-0.75, 0.75]$] that forms as wind accelerates between the two sets of obstacles. 
Both wind estimates also capture the corner accelerations located on the windward sides of the obstacles.  
One point of distinction between the two methods is in the $z$ wind estimate--the \textit{Wind UKF} is able to measure the upwards movement of the air located around $(6.5, 0)$, whereas the \textit{CMU Whisker} assumes $w_z=0$ by design. 
The reconstructed flow fields for the remaining experiments can be found in \autoref{appendix:complete_windshaper_experiments}, \autoref{appendix:additional_exp_analysis}.

\begin{figure}[b!]
    \centering
    \includegraphics[width=0.9\textwidth]{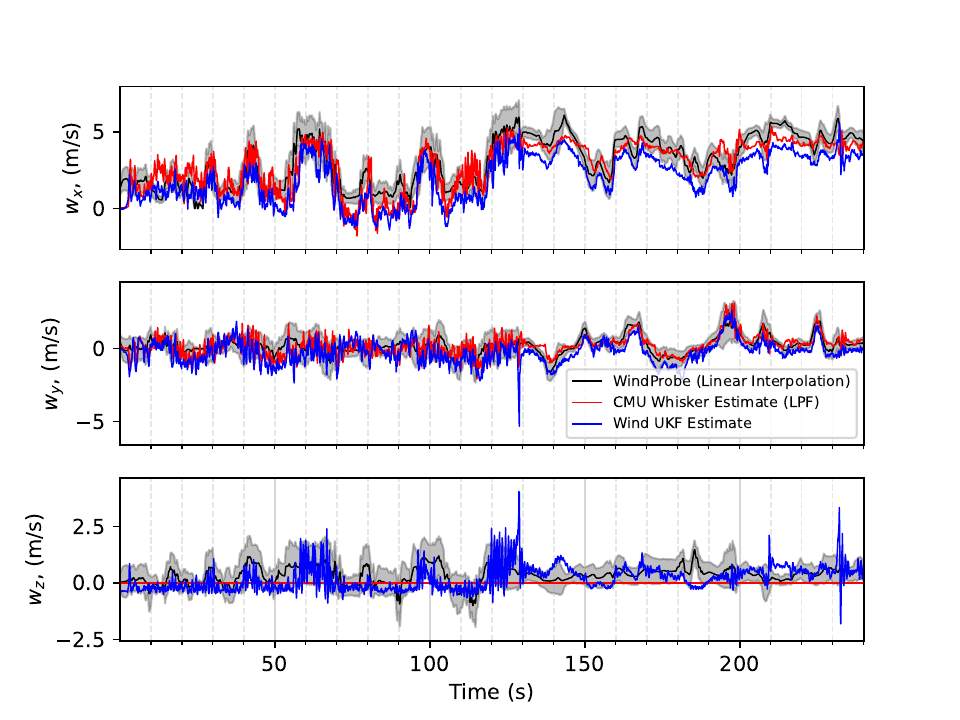}
    \caption{Comparisons of the instantaneous wind estimates between the UKF and \textit{CMU Whisker} for \textbf{Flow Traversal} experiment 14. The estimate provided by the \textit{WindProbe} represents a sampling of the mean and standard deviation of the flow field from the bins corresponding to the UAV's trajectory.}
    \label{fig:realworld:realworld_ukf_comparison}
\end{figure}

\autoref{fig:realworld:realworld_ukf_comparison} provides another perspective for comparison, showing the instantaneous wind estimates for each method during experiment $14$ as a time series. 
In these plots, the \textit{WindProbe} reference is generated by linearly interpolating the time-averaged flow field plotted in \autoref{fig:realworld:flowfieldreconstruction} along the UAV's trajectory. 
The solid black line represents the mean of the binned data while the shaded region indicates one standard deviation from the mean. 
The shaded region can be interpreted as a measure of the inherent unsteadiness in the flow, subject to the precision of the \textit{WindProbe}.
Once again there is good agreement in the general trends between all three wind measurements.
For instance, both the \textit{CMU Whisker} and \textit{Wind UKF} capture the sustained $5$ m/s increase in $w_x$ in the range $t\in[55, 70]$ or the rapid gusts in $w_y$ at $t=195$, $t=209$, and $t=225$.

\subsection{Wind Field Prediction from Learned Models}

In order to test the wind flow field decoder network, it was implemented on the Nvidia Jetson using \texttt{PyTorch}\footnote{\url{https://pytorch.org/}} and the on board GPU. 
The network was provided range measurements from the simulated LiDAR (\autoref{sec:realworld:rangesensing}) at a rate of $50$ Hz.
This version of the decoder network corresponded to that presented in \autoref{sec:prediction:model_selection}: a fully connected multilayer perceptron (MLP) with three hidden layers, each with $2{\small,}048$ neurons each.
The network was originally trained on \revision{full-scale} urban flows with a prediction region size of $25$$\times$$25$ m with a $1$ m grid resolution. 
To match this to the scaled environment, the output of the network was \revision{simply} scaled down by a factor of $10$ such that the region size was $2.5$$\times$$2.5$ m with $10$ cm grid resolution. 

\begin{figure}[t!]
    \centering
    \includegraphics[width=0.99\textwidth]{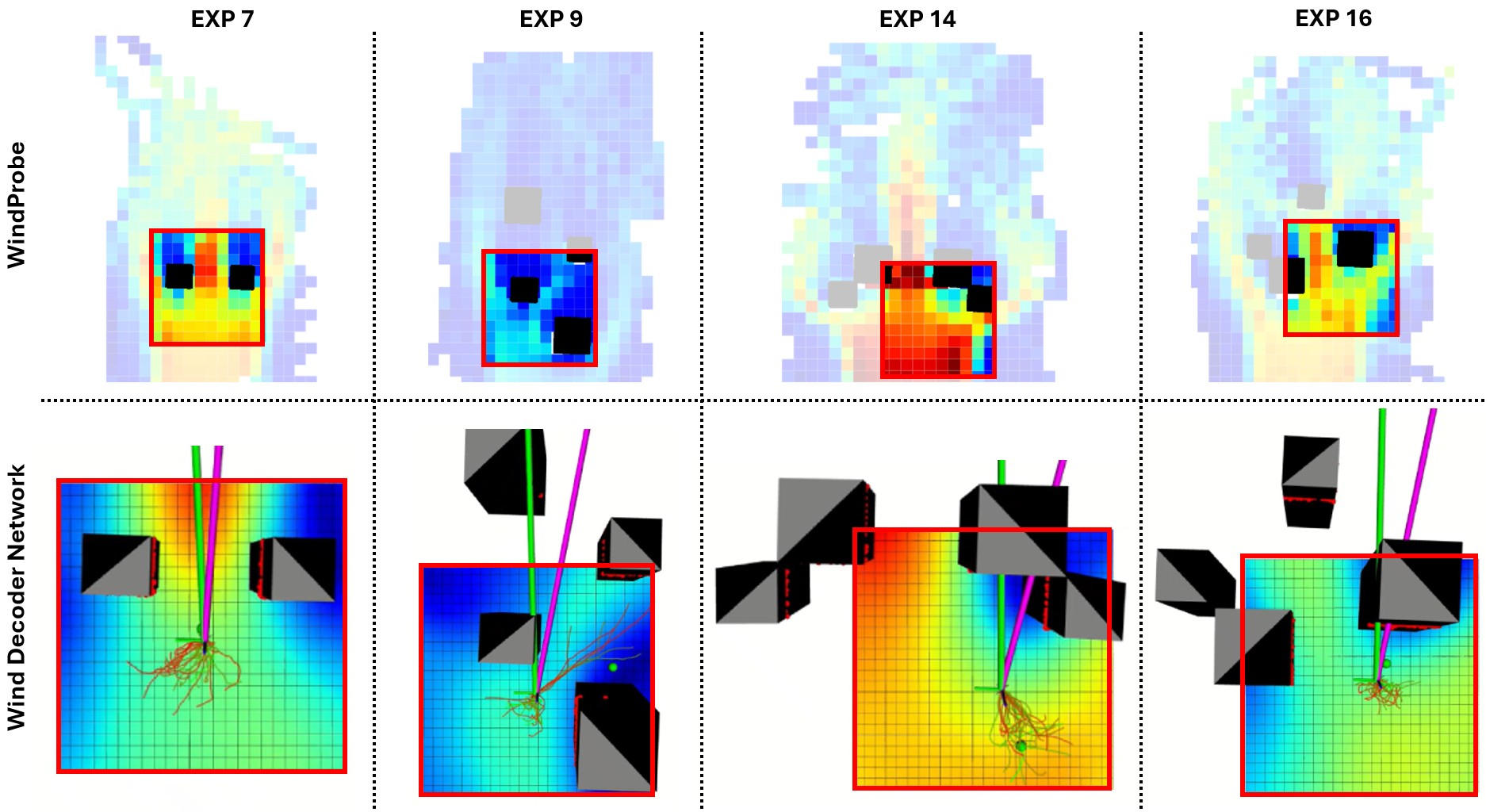}
    \caption{\revision{Local time-averaged velocity magnitude comparisons between reconstructions using the \textit{WindProbe} and the instantaneous predictions from the wind decoder network running on an Nvidia Jetson AGX Xavier.} The green sphere represents the desired position, the triad is the pose of the UAV, and the green and magenta arrows extending off screen are wind estimates from \textit{CMU Whisker} and \textit{Wind UKF}, respectively. \revision{The wind decoder network takes} as input the \textit{Wind UKF} wind estimate as well as the simulated point cloud data (red dots) \revision{to sense topography.} The top $25$ forecasted trajectories from the wind-aware motion planner \revision{can also be seen originating} the UAV's position.}
    \label{fig:realworld:fullstack_rviz}
\end{figure}

\autoref{fig:realworld:fullstack_rviz} visualizes the predictive capability of the \revision{neural} network at \revision{four} distinct instances during \revision{experiments $7$, $9$, $14$, and $16$.} 
The contour plots \revision{indicate the} velocity magnitude \revision{field}, with blue corresponding to low speed wind and red to high speed. 
\revision{In comparison to the ground truth flow fields obtained from the \textit{WindProbe},} the network's \revision{local flow} predictions were \revision{qualitatively} striking in that they capture key flow patterns--flow acceleration and subsequent diffusion in between two obstacles, low speed regions close to the buildings, and accelerations around the outer corners of the building arrangement\revision{--across experiments with both low and high speed winds and varying building arrangements.}
One concerning \revision{anecdotal} observation from \revision{these experiments} is that the predictions from one \revision{temporal} frame to the next were not as consistent as what was seen in simulation.
In other words, overlapping predictions \revision{during the same experimental run} would not ``mesh'' together consistently to provide a coherent view of the global flow field. 
The primary explanation for this is that the network was trained on time-averaged sampling of the wind at the UAV's location, but in hardware the noisy and unsteady wind estimates from the \textit{Wind UKF} were provided which could introduce variability in the network prediction. 
But also, it's important to note that the size of these obstacles, if scaled by $10$ times to the \revision{full-scale} environment, are smaller than the smallest buildings found in the training data which makes these predictions slightly out of distribution. 
Nevertheless, \revision{these} flow features are still captured by the network, and the inconsistency issue could be addressed with fine tuning on the smaller environment and filtering the wind measurement input to the network. 

\subsection{Wind-Aware Planning and Control}\label{sec:realworld:planning}

While the previous two result sections studied the predictive capabilities of the \textit{Wind UKF} and wind flow decoder network, the ultimate goal of this thesis is to close the loop by integrating local wind inference with decision making. 
The complete computational pipeline--model-based wind estimation, wind flow field \revision{prediction via} deep learning, and wind-aware motion planning on a receding horizon--was deployed on the hardware described in \autoref{fig:realworld:experimentdiagram}.

The objective of these experiments was to assess the system's ability to maintain stability \revision{and avoid collisions} while navigating the flow fields generated by the WindShaper. 
As noted in the introduction, the constrained nature of the motion capture space and the necessity of tracking a reference wand limited the planner's ability to exploit large-scale energy savings. 
However, these constraints provided a rigorous test bed for analyzing the computational feasibility of the architecture and the limits of control authority in turbulent urban wakes.

\begin{table}[b!]
\centering
\caption{Hyperparameters for the MPPI controller implemented on an Nvidia Jetson AGX Xavier. Note that because the actuator sampling scheme changed with the hardware implementation, the associated parameters below were not used.}
\label{tab:mppi_hardware_hyperparams}
\begin{tabularx}{0.5\columnwidth}{lXl}
\toprule
\textbf{Parameter} & \textbf{Value} & \textbf{Unit} \\ \midrule
$N_\tau$ & 1000 & - \\
$\Delta t$ & 0.04 & s \\
$N_k$ & 25 & - \\
$f_{plan}$ & 25 (target) & Hz \\
$\lambda$ & 5.0 & - \\
$\sigma_\mathrm{u}$ & N/A & m/s$^2$ \\
$\sigma_{gf}$ & N/A & m/s$^2$ \\
$Q_O$ & 1.0 & - \\ 
$D_{max}$ & 10 & m \\
$P_{hover}$ & 10.56 & W \\
\bottomrule
\end{tabularx}
\end{table}

\subsection{Computational Latency and System Integration}

A primary contribution of this chapter is the demonstration that deep-learning-based wind prediction and wind-aware motion planning can run in real time on flight-ready hardware.
The inference time of the wind prediction network was previously reported in \autoref{ch:prediction} to be $5$ ms which is well within the budget of the control loop.
However, whether or not the wind-aware motion planner can run fast enough remains up to this point an open question. 

To address this, the MPPI algorithm derived in \autoref{ch:planning} was implemented on the Nvidia Jetson AGX Xavier flight computer. 
The algorithm was implemented in Python, specifically leveraging the \texttt{PyTorch} library to interface with the Jetson's dedicated GPU. 
Further optimizations were achieved using \texttt{PyTorch}'s Just-In-Time (JIT) compilation library that compiles the dynamic Python code into a static graph representation known as \texttt{TorchScript}. 
More precisely, the dynamics integration and wind sampling routines were compiled into \texttt{TorchScript} because these represented the biggest bottlenecks during preliminary testing and development. 
\texttt{PyTorch} and \texttt{TorchScript} were crucial elements to this implementation because the MPPI algorithm is particularly suited for parallelized processing \cite{williams2016mppiintro}. 
In fact, MPPI performs better when more trajectories are generated ($N_\tau$ from \autoref{ch:planning}), and unlocking the GPU resource meant that many more trajectories could be generated with little impact on computational latency. 
The MPPI hyperparameters for the hardware implementation are listed in \autoref{tab:mppi_hardware_hyperparams}--of note, the integration horizon and time discretizations had to be adjusted to account for the faster UAV and wind dynamics. 
Also, in simulation a biased sampling scheme was used where control inputs were sampled from a normal distribution centered at the previous optimal control sequence.
For the flight computer implementation, control inputs were instead uniformly sampled in the minimum and maximum bounds for the acceleration vector. 

\begin{table}[b!]
\centering
\caption{Breakdown of computational latency for a single MPPI planning iteration on the Nvidia Jetson AGX Xavier. Data is averaged over 25 planning cycles, and the raw profiling log is available in \autoref{appendix:complete_windshaper_experiments}.}
\label{tab:realworld:profiling}
\begin{tabular}{@{}llcc@{}}
\toprule
\textbf{Category} & \textbf{Routine} & \textbf{Time (ms)} & \textbf{\% Contribution} \\ \midrule
\textbf{GPU Compute} & \texttt{update\_mppi} & \textbf{25.5} & \textbf{36.4\%} \\
\hspace{3mm} \textit{Cost Calculation} & \hspace{3mm} \texttt{compute\_cost} & 2.5 & - \\
\hspace{3mm} \textit{Rollouts} & \hspace{3mm} (remainder) & 23.0 & - \\ \addlinespace
\textbf{Memory Transfer} & \texttt{Tensor.cpu()} & \textbf{44.4} & \textbf{63.4\%} \\ \addlinespace
\textbf{Overhead} & \texttt{numpy/torch operations} & \textbf{0.1} & \textbf{0.2\%} \\ \midrule
\textbf{Total} & & \textbf{70.0 ms} & \textbf{100\%} \\ \bottomrule
\end{tabular}
\end{table}

\autoref{tab:realworld:profiling} summarizes the computational burden of the motion planner, obtained while running on the flight computer.
The table reports the average execution times obtained from Python's built-in \texttt{cProfile} module over $25$ calls to the motion planner's primary \texttt{update} method. 
The execution times are broken down into the critical components of the algorithm's implementation: running trajectory roll outs, computing the cost of trajectories, and memory IO. 
While the memory usage is not reported in these results, we would like to note that at no point did insufficient RAM become an issue during development and subsequent experiments. 

According to the data presented in \autoref{tab:realworld:profiling}, it takes a total of $70$ ms ($14$ Hz) to run a single MPPI update--this includes updating all the internal variables based on new measurements (e.g. LiDAR and motion capture), control sequence sampling and generating candidate trajectories through dynamics integration, computing the cost of each trajectory, and finally computing the next reference acceleration vector for the lower level controller to track. 
At $14$ Hz operation, the controller would be too slow to achieve stable flight on the Crazyflie. 
This can be mitigated by instead only running an MPPI solve iteration on 1 of every 4 controller updates. 
During the other $3$ controller updates in this cycle, the reference acceleration from \texttt{update\_mppi} is held constant and tracked by the lower level controller which has negligible computational latency. 
According to the full profile in \autoref{appendix:complete_windshaper_experiments}, the controller update loop takes just $19$ ms ($52$ Hz) on average (see \autoref{tab:realworld:profile_raw}) when this hierarchy is established. 

Turning attention once again back to \autoref{tab:realworld:profiling}, an important observation is that majority of the computational latency is not from \texttt{update\_mppi}, but rather in the \textit{memory transfer} between the CPU and GPU (\texttt{Tensor.cpu()}).  
At $44.4$ ms, memory transfer accounts for over $63\%$ of the total time spent updating the reference acceleration from MPPI. 
This represents a significant bottleneck, but one that would perhaps be surmounted by more deliberate integration of the MPPI planner into the autonomy stack. 
Assuming the memory transfer could be brought down to a more reasonable level ($5$-$10$ ms or less), we conjecture it would be possible to get the MPPI planning step alone to run at a much more comfortable $40$ Hz or higher. 

\subsection{Obstacle Avoidance}

During the \textbf{Flow Traversal} experiments, the UAV was tasked with tracking the $x$ and $y$ position of a wand operated by a human.  
To test the obstacle avoidance capabilities of the MPPI controller, occasionally the wand was positioned directly over the obstacles. 
Because the MPPI cost function has a specific term in the cost function for proximity to obstacles (\autoref{eq:cost_obs}), forecasted trajectories that go near points in the point cloud are attributed with high navigational cost. 
The expected behavior in this scenario is for the UAV to deviate from the wand's reference trajectory when that reference becomes unsafe. 

An example of this is seen in \autoref{fig:realworld:obstacleavoidance}, where the distance to the nearest obstacle is plotted versus time for both the UAV and the wand. 
When the wand is far away from any obstacles, the UAV tracks the wand's position accurately. 
However, in the opposite case when the wand gets too close to a building (e.g. $t\in[12, 16]$ and $t\in[21, 25]$), tracking degrades because the obstacle avoidance cost dominates the tracking cost.
In this example, the UAV would not get closer than $0.45$ m to the center of any of the obstacles. 
The qualitative behavior in these scenarios is for the UAV to hover in place, although if the wand continues to move away from the UAV the MPPI algorithm will attempt to find a way around the obstacle. 

\begin{figure}[t!]
    \centering
    \includegraphics[width=0.8\textwidth]{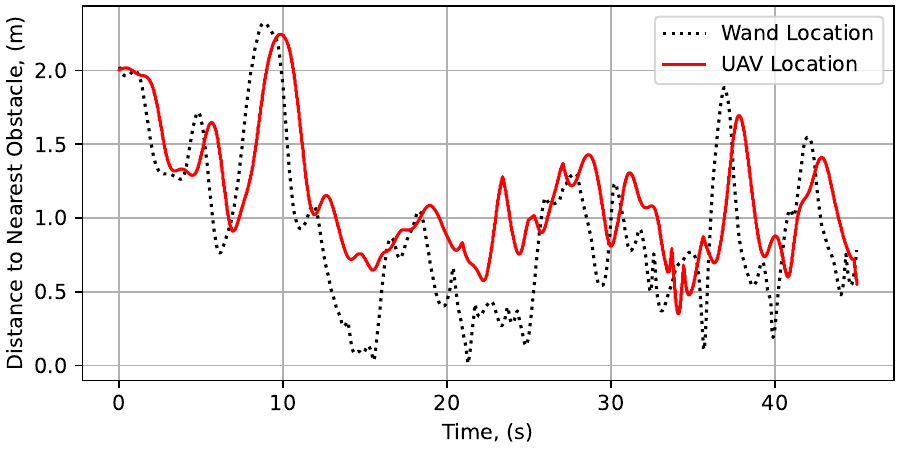}
    \caption{Time series measurements of the distance to the nearest obstacle for the wand and UAV while running the MPPI motion planner. The UAV tracks the wand adequately when it is far enough away from obstacles. However, the obstacle avoidance term in the MPPI cost function prevents the UAV from tracking the wand when it gets too close to an obstacle.}
    \label{fig:realworld:obstacleavoidance}
\end{figure}

\subsection{Unsteadiness Metrics}

Finally, we consider ways we might use the data from these experiments to characterize the aerodynamic challenges imposed by urban-like flow features on UAVs operating in the urban canyon. 
Understanding where these hazards occur is critical to formulating and validating risk-aware planning and control, and previous studies have provided simulated insights into UAV stability \cite{galway2011control} and tracking performance \cite{murray2014responseinenvironment, raza2015uavcontrolinwind, sutherland2016urbanquadrotorflightsim} around buildings in high winds. 
In this section, we present experimental data on two metrics related to controller effort and UAV stability that may connect with this literature towards identifying urban wind hazards.

In \autoref{fig:realworld:uam_metrics}, the standard deviation of the commanded lateral force and the average body rotation rates are plotted for experiment 11 of the \textbf{Flow Traversal} series.
The former metric captures how much additional control effort is required during tracking, while the latter is proportional to attitude variations due to gusts \cite{galway2011control}. 
From this data, an unsurprising trend emerges: control effort increases and stability decreases in the leeward (wake) regions of the obstacles, and this effect is magnified specifically at the corners of the obstacles. 

\begin{figure}[t!]
    \centering
    \includegraphics[width=0.9\textwidth]{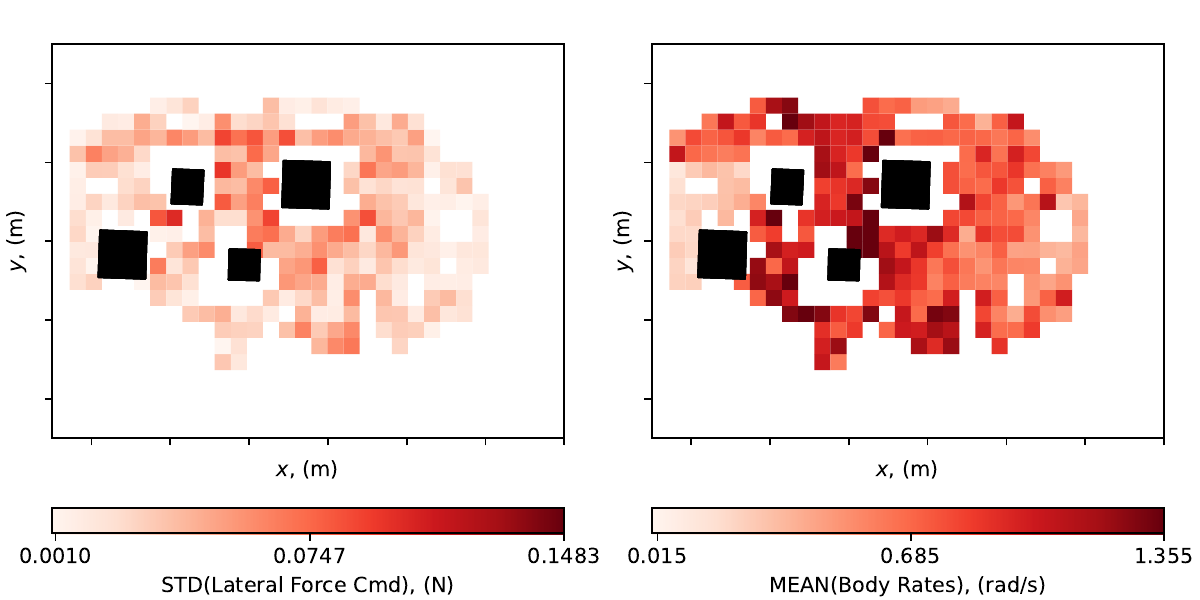}
    \caption{Heat maps of the standard deviation of the lateral force commands and the average body rotation rates of the UAV for experiment $11$, demonstrating regions of unsteadiness in urban-like winds.}
    \label{fig:realworld:uam_metrics}
\end{figure}

While this analysis represents a preliminary characterization, we include it here to highlight the potential of the dataset for further understanding of urban wind hazards and perhaps even validating future wind-aware control policies. 
While the aforementioned simulation studies have predicted these stability degradation regions, obtaining high-fidelity experimental data to validate these models has historically been difficult due to the risks of flying in turbulent conditions.
The sub-scale hardware experiments conducted in this thesis address a need for actual flight data in challenging urban winds, and we hope this data will inspire future experiments along this vein. 
To that end, the unsteadiness metrics presented here are also plotted for all of the other \textbf{Flow Traversal} experiments in \autoref{appendix:complete_windshaper_experiments}, \autoref{appendix:additional_exp_analysis}. 

\section{\revision{Summary}}

This chapter \revision{presents} the integration and experimental validation of \revision{the} novel wind-aware navigation framework on flight-ready hardware. 
These experiments involved a small UAV tracking the position of a human-controlled wand inside a motion capture space, while having to reason about obstacles and complex wind patterns generated by an open-air wind tunnel.

The contributions of this chapter are threefold. 
First, we provided a comprehensive characterization of the actuators, aerodynamics, and power consumption of the Crazyflie 2.1-Brushless via free flight testing, offering the research community the first detailed models of this widely used platform. 
These measurements were used to calculate the necessary aerodynamic parameters for bringing the algorithms in this thesis into the real world. 
Second, through comparisons with an on board whisker-based airflow sensor and a multi-probe pitot-static anemometer, we demonstrated the \textit{Wind UKF} \revision{mapping out} complex urban flow features like corner accelerations and channel flows using on board sensing. 
\revision{We also successfully transferred a wind flow decoder network trained in simulation into the real world with surprisingly accurate flow predictions.}
Lastly, we proved the computational feasibility of the proposed real time wind-aware navigation framework.
Code profiling confirmed that the full navigation stack--model-based indirect wind estimation, deep-learning-based wind field prediction, and wind-aware stochastic receding horizon control--can run in real time on an Nvidia Jetson at a sufficient frequency for stable flight.

While the geometric constraints of the laboratory prevented the realization of any measurable energy savings, the system demonstrated robust obstacle avoidance and importantly provided flight data which could serve as a preliminary dataset for future studies on the identification and characterization of hazardous zones in windy urban environments. 
\chapter{Conclusion}\label{ch:conclusion}

This thesis presents an exciting \revision{new} framework for the prediction and \revision{leveraging} of urban wind flow fields \revision{for robust autonomous navigation} using on board sensing and compute. 
Our approach to wind prediction \revision{builds} on the notion that winds are heavily constrained by the boundary conditions imposed by dense urban infrastructure.
By fusing sparse wind measurements from the UAV with a sensor that can observe these boundary conditions, such as a navigational LiDAR, we have shown that it is possible to reconstruct time-averaged flow fields in a \revision{local ego-centric region}. 
This concept is taken one step further by incorporating the wind predictions into a receding horizon motion planning framework, which continuously finds an energy-optimal trajectory through the environment while avoiding obstacles. 

\revision{The contributions of this thesis can be summarized as follows:
\begin{enumerate}
    \item \textbf{Data-Driven Inference of Local Wind Fields:} This thesis empirically demonstrates correlations between urban topography and local wind flow fields that can be learned provided sufficient training data. 
    Armed with this insight, we developed a scalable training data generation pipeline for urban wind flow fields and then trained a deep neural network capable of inferring complex flow structures solely from sparse wind and LiDAR measurements.
    \item \textbf{Proactive Wind-Aware Motion Planning is Tractable on a Horizon:} Existing energy-aware motion planners struggle with computational intractability, largely because they solve for an entire trajectory to global optimality. 
    In contrast, we developed a real-time autonomous wind-aware navigation stack made possible by decomposing the global motion planning problem into smaller local problems that can be solved on a horizon, while also using the wind predictions from the aforementioned neural network as a wind prior. 
    \item \textbf{Full-Stack Validation via Scaled Free Flight Testing:} Lastly, we designed a novel experimental procedure for sub-scale, closed-loop free flight wind tunnel testing of autonomous navigation algorithms in urban-like wind conditions.
    By validating our algorithms with real embedded hardware on a Froude-scaled platform in a state-of-the-art wind tunnel facility, we demonstrated new capabilities in sub-scale testing that can provide physical insight and act as a crucial intermediate step for de-risking urban autonomy algorithms before full-scale deployment.
\end{enumerate}
}
\revision{The paradigm presented herein} represents a departure from the current state of the art in wind-aware navigation, which \revision{largely} relies on detailed computational fluid dynamics simulations, wind tunnel studies, distributed sensing networks, significant exploration of the environment, or precise knowledge of the urban infrastructure. 
Our method complements these methods by enabling the UAV to partially observe the wind using visual cues from the environment. 

There are two fundamental themes \revision{that drove the entirety of work in} this thesis: on board sensing and computational tractability. 
Throughout our work, these themes were positioned front and center: \textit{how can UAVs reason about complex winds without needing specialized sensor payloads or access to supercomputers in the cloud?} 
To address this, a serious experimental campaign was undertaken to explore the applicability of the proposed algorithms in a real world setting. 
Through sub-scale experiments in an open-air wind tunnel with real winds, a real palm-sized UAV, and a real embedded flight computer, we demonstrated that on board sensors are sufficient for the task, and that the hardware and compute resources are mature enough to bring these concepts out of simulation. 

In our simulations, we gained crucial insight into the improvements to \revision{robustness} and energy efficiency when wind is predicted and accounted for in the core flight autonomy stack.  
These ideas were tested in the real world, providing valuable data of sub-scale flight in urban-like flows alongside useful aerodynamic and energetic characterization of a small UAV. 
We believe sub-scale free flight testing will become a necessary component to the thoughtful design, validation, and eventual approval of scaled autonomous aerial operations in highly populous regions. 

\section{Limitations and \revision{Path Forward}}

\revision{While} this thesis introduces many new concepts to the field, there is plenty of room for improvement in \revision{almost every aspect.}
\revision{In fact, there are many new research avenues that were revealed from these efforts.}
\revision{In this section, we will discuss some of the main shortcomings of the existing work and future directions that may address them.}

\subsubsection{Wind Prediction with Deep Learning}

The primary limitation of the existing efforts for the wind flow decoder is a restriction to \revision{2D} time-averaged flow fields and only relying on instantaneous sensor measurements rather than a history of observations.
The reason for this shortcoming is in one of practical limitations and yet another lesson on the curse of dimensionality. 
We believe our wind flow decoding methodology \revision{was initially} enabled by a scaling up in training data collection--\revision{sufficient} scaling was partly achieved through custom implementations of numerical fluid dynamics simulations \revision{(i.e., our Euler numerical solver),} but \revision{another factor} was limiting the dimensionality of the problem \revision{to just two spatial dimensions and ignoring time} in the prediction task. 
Limiting the dimensionality of the problem had two compounding benefits. 
First, the numerical fluid dynamics solver was made simple and fast enough to be massively parallelized using high performance computing clusters, which meant that hundreds of hours of fluid flows could be simulated in just a few hours. 
The second benefit is \revision{that} the size of the training dataset \revision{remained sufficiently small--even} with just two spatial dimensions, the dataset used to train the wind flow decoder networks was over $100$ GB in size which was \revision{approaching} the hardware limitations of the computing cluster used to train these models \revision{at the time.}
These two benefits meant that rapid prototyping was feasible, which was a crucial component to achieving initial success. 
\revision{However, a significant physical limitation of time-averaged prediction is that the mean flow field may fail to represent the instantaneous reality experienced by the UAV. 
For example, in regions of periodic vortex shedding like in the wake of buildings, the time-averaged cross-flow may approach zero, whereas the instantaneous flow consists of high-magnitude oscillations.
And while the quasi-2D flow assumption holds for very specific building geometries like the tightly-packed rectangular layouts investigated in this thesis, in practice the vertical wind components cannot be ignored for more natural urban footprints. 
}

In this light, a challenging yet necessary task will be scaling the method introduced here to predict \revision{3D} wind flow \revision{fields, alongside a dedicated treatment to the unsteadiness of the wind.} 
\revision{For instance, to improve the consistency of the network prediction, future architectures should incorporate temporal memory, such as Recurrent Neural Networks (RNNs) or Transformers. 
By looking backwards at the sequence of measurements and the topography already traversed, the model could utilize a Bayesian-like update step much like the \textit{WindUKF} to refine its forward-looking predictions. 
This would allow the UAV to build a latent belief state of the wind field that persists and improves as it moves through a canyon while explicitly handle sensor noise or missing measurements.}
\revision{Also, it may be possible to frame the problem as predicting the wind on a temporal horizon, using a historical buffer as a way to condition the network.}
However, introducing \revision{these additional dimensions into the learning problem, especially time,} exponentially increases how long it takes to generate data and how much space that data takes up.
\revision{This is not just} because of the curse of dimensionality, but also because other \revision{external} factors like temperature and pressure gradients will likely need to be modeled in the CFD solver as well \revision{because of their influence on the 3D structure of urban winds.} 
\revision{One very promising incremental next step to the work presented here that avoids some of these scaling issues would be to look more closely at predicting alternative metrics associated with the wind that are also relevant to navigation; for instance, rather than predicting the time-averaged flow field the network could predict unsteadiness metrics like the variance or a statistical uncertainty of the flow field.}
\revision{Crucially, flow unsteadiness is not entirely stochastic; rather, it is governed by predictable fluid dynamics.
Specific phenomena, such as the aforementioned vortex shedding behind high-rise structures, exhibit strong quasi-periodic behavior, typically occurring at a frequency defined by the Strouhal number.
This periodicity suggests that the time-varying component of the wind has predictable structure over a certain horizon, a structure that could be learned in the same way the time-averaged spatial features were learned by the wind flow decoder in this thesis. 
Beyond hand-engineered unsteadiness metrics, future work should also investigate the use of data-driven spectral analysis methods, such as Dynamic Mode Decomposition (DMD) or Koopman Operator theory, to automatically identify and forecast these dominant shedding modes.
However, whether or not the current model inputs--LiDAR scans coupled with sparse wind measurements--are sufficient for observing these shedding modes remains an open question. 
Explicit treatment of the temporal dynamics of the flow would pave the way towards more sophisticated motion planning, enabling the UAV to leverage vortex cores for navigation in the spirit of related work \cite{salzmann2024learning}.}

We anticipate that significant changes to the data representations and learning architecture--more modern network designs with better inductive biases and better parameter efficiency--will be necessary to ensure that training datasets can be generated in a reasonable time while remaining within existing computational limitations. 
In the age of large language models and internet-scale data distillation, we believe this challenge will be surmounted in a matter of time. 
\revision{For instance, just in the last year the \textit{UrbanTales} dataset \cite{nazarian2025urbantales} was released, containing state-of-the-art Large Eddy Simulations on over $538$ idealized and realistic 3D urban layouts.}
\revision{On network architecture improvements,} diffusion-based numerical weather prediction models \cite{price2023gencast} are coming online with very impressive predictive capability--perhaps lessons learned from these models could be transferred to meter-scale wind prediction \revision{alongside the recent architectural insights and advancements from the large language modeling communities.} 

\revision{Lastly, the concept of visual cues for wind should be expanded into a multi-modal fusion framework. 
While this work focused on LiDAR, we see this as just the beginning. 
Sister networks could be trained to infer flow from thermal imagery by detecting urban heat plumes and subsequent updrafts, or optical flow from vegetation movement could inform turbulence intensity. 
Integrating these diverse experts into a single inference architecture would mean various boundary conditions such as temperature and pressure gradients, which are missing in the current approach, could be factored into the prediction.
It is worth also mentioning that LiDAR is a very heavy and expensive technology, and while this is likely to improve in the years to come, investigating conventional (monocular) cameras as a lightweight and more cost-effective means of sensing urban topography will be a worthwhile endeavor, especially considering existing tools from the literature that provide ways to map environments even using monocular vision (e.g. \cite{simon2023mononav}).
}

\subsubsection{Wind-Aware Motion Planning}

The motion planner presented in this thesis is a promising approach to real time wind-aware motion planning. 
The choice to plan on a receding horizon not only simplified the motion planning problem to make it more tractable, but also made it possible to adapt to changing wind conditions \revision{which is a} favorable \revision{attribute} for when wind patterns shift unexpectedly. 
However, \revision{one limitation was made clear in our experiments:} the inherent myopic view of the receding horizon motion planner meant that often times suboptimal routes were \revision{taken, while other times} the algorithm settled in a local minimum \revision{commanding the UAV to hover and preventing} further progress to the goal. 
As far as the MPPI implementation itself, many simplifying assumptions with respect to the UAV dynamics and energetics were required to make the problem tractable in its current form--for instance, the power model assumed steady state level flight and neglected affects due to \revision{climb, descent, or rapid accelerations.} 
However, MPPI \revision{as an algorithm} is flexible enough to handle nonlinear dynamics as well as nonconvex cost functions and soft constraints. 

\revision{Future studies should attempt to solve these problems by embedding the MPPI as a middle-level controller in a hierarchical autonomy stack}. 
\revision{More concretely, there should be a monitoring system whose job it is to identify when the UAV is stuck in a local minimum and switch to a fail safe mechanism.}
\revision{Also, the stack could benefit from a higher level planner that has broader knowledge of the environment and task. 
The higher level planner could be used to plan intermediate goal waypoints for the MPPI planner to navigate to.}
\revision{The combination of these two mechanisms would likely help prevent the hovering failure conditions entirely.}

\revision{Obstacle avoidance is another critical component of this work that could be improved in future iterations.}
\revision{In its current form, obstacle avoidance is only handled using soft constraints as an additional cost term in the MPPI formulation.}
\revision{It will be necessary in future work to handle obstacle avoidance with hard constraints which the existing MPPI formulation does not support.}
\revision{One option is to consider adding control barrier functions \cite{ames2019controlbarrier} as an additional safety layer between the output of the MPPI controller and the command sent to the lower-level controller.}
\revision{The control barrier function(s) would add slight adjustments to the controller output to ensure obstacle avoidance is maintained, although the design of this mechanism would be nontrivial.}

\revision{Finally, a} thoughtful extension of the \revision{wind-aware motion planning stack presented herein} is to consider higher fidelity \revision{representations} of the aerodynamics and power consumption.
\revision{Better models that consider} the attitude and unsteady motions would help \revision{shrink the gap between the MPPI's model and the real world, increasing prediction accuracy and downstream performance.} 
\revision{However, the major compute bottleneck for MPPI is the time it takes to integrate the dynamics, so perhaps deep learning might provide yet another useful tool to increase the model fidelity without increasing the computational burden.} 
\revision{Similarly, other} phenomena \revision{beyond power consumption should be factored} into the cost function to improve the reasoning capabilities of the motion planner.
\revision{An unsteadiness metric based on the wind gradients, for instance, could enable the planner to reason more effectively about regions of high turbulence that would impact rider comfort.}


\subsubsection{\revision{Sub-Scale} Free Flight Testing}

Lastly, the hardware experiments presented in this thesis scratch the surface in terms of what is possible with free flight sub-scale testing with the WindShaper. 
This novel laboratory environment is highly poised to make considerable contributions in understanding how UAVs are affected by urban wind flow features in ways that simulation might not live up to. 
While the data presented here is good for preliminary analysis for these studies, a more concerted effort to achieve better dynamic similitude will be necessary to make sure the lessons learned in sub-scale environments appropriately transfer to full scale. 
We briefly remark on dynamic similitude in \autoref{sec:realworld:similitude}, but ultimately conclude that full dynamic similitude is impossible in the \revision{sub-scale} environment. 
This step is crucial to ensuring that any improvements observed in the lab are reliably transferred to full scale operations and regulatory approval in metropolitan areas.
Therefore, it is imperative that future work considers ways to achieve better \textit{partial} dynamic similitude through the design of the obstacles, UAV platform, and wind generated from the WindShaper. 

Improvements should be made to the facility to improve the instrumentation. 
Namely, during initial experimentation, a core issue was that obstacles occluded the UAV from view of the motion capture \revision{cameras} which often meant that pose tracking of the UAV failed.
\revision{Mitigating this} required more thoughtful yet \revision{constraining} building layouts that minimized this issue, \revision{as well as flying the UAV at higher altitudes which worsened the 3D flow effects from the buildings.} 
This issue could be solved with more cameras as well as making the obstacles translucent to IR \revision{using a different construction material to build them.} 
Another improvement to the facility could be the introduction of flow visualization methods like particle image velocimetry (PIV) to better understand the temporal dynamics of urban flow features. 
While PIV was tested during preliminary checkout procedures with the WindShaper, the large space coupled with fast moving air requires more specialized equipment that was beyond the scope of these initial experiments.
\revision{Ultimately,} ground truth probings of the wind field had to be conducted meticulously by hand using the \textit{WindProbe} \revision{which significantly slowed progress on experiments.}

\revision{Beyond instrumentation, the overall design of the sub-scale obstacles needs much more attention in future experiments to ensure that a specific critical Reynolds number is reached for Reynolds number independence by considering obstacle size, spacing, or perhaps even the surface roughness.}
\revision{A more careful treatment of this aspect of the experimental design will no doubt help minimize the gap between the sub- and full-scale environments.}
\revision{Also, the physical quality of the generated wind should be addressed to support higher-fidelity studies. 
The installation of flow straighteners like honeycomb structures at the fan array outlet would reduce background turbulence, allowing for further isolation of building-specific wake effects. 
Furthermore, the introduction of a secondary fan array to generate controlled lateral gusts would enable more dynamic wind conditions as well as the first repeatable benchmarks for cross-wind rejection in urban-scale models, a capability currently missing in existing free-flight literature.}

\section{\revision{Outlook}}

\revision{
The framework presented in this thesis represents a crucial step toward the ultimate goal of seamless, wind-aware autonomy for small aerial vehicles operating at scale in windy urban environments. 
By proving that complex urban flows can be inferred from on board sensors and utilized within a real-time planning horizon, we have addressed the primary bottlenecks of sensing and tractability, enabling a shift from reactive gust rejection toward proactive, wind-aware navigation. 
These advancements are essential for transitioning autonomous flight from controlled laboratory settings to the chaotic reality of the urban canopy.
With these limitations now addressed, we open up numerous doors in the near future that will lead us closer to unleashing the full potential of our urban airspaces. 
}

Broadly speaking, the \revision{possible research avenues} described above establish \revision{many} fruitful path for researchers with similar aspirations of improving the autonomy of UAVs in challenging windy environments. 
\revision{There are various opportunities for future research at the intersection of fluid mechanics, aerodynamics and flight mechanics, and robotics.}
\revision{The most promising avenues identified from this work are: extending the wind decoder paradigm to handle 3D unsteady flow fields and incorporate different sensing modalities (e.g. cameras); the incorporation of the MPPI planner as a mid-level controller in a hierarchical navigation architecture to avoid local minima; and finally a more thoughtful design of the sub-scale environment in the WindShaper facility with more instrumentation and explicit treatment of Reynolds similarity and Reynolds number independence.}
\revision{Beyond the algorithmic contributions, the methodologies for sub-scale validation developed here provide a blueprint for the rigorous safety testing required to certify autonomous operations in high-density population centers.}

\revision{To conclude, we} hope that the work presented in this thesis will inspire future researchers and engineers to think creatively and with ambition towards the goal of increasing aerial autonomy and keeping our skies safe for all. 

\end{mainf}




\clearpage 


\addtocontents{toc}{\protect\renewcommand{\protect\cftchappresnum}{APPENDIX }}
\addtocontents{toc}{\protect\setcounter{tocdepth}{0}}

\renewcommand{\thechapter}{\Alph{chapter}}
\setcounter{chapter}{0}


\titleformat{\chapter}[display]{\bfseries\center}
    {\huge APPENDIX \thechapter} 
    {0pt}
    {\Large\MakeUppercase} 
  \titlespacing*{\chapter}{0pt}{-33 pt}{6 pt}

\appendix
\renewcommand{\chapterautorefname}{Appendix}

\chapter{Wind Estimation Uncertainty Analysis}\label{appendix:uncertaintymodeling}

In this appendix, an uncertainty analysis of a model-based wind estimator for a UAV is presented. 
The goal is to understand how physical parameters of the UAV, such as mass and lumped parameter drag coefficients, combined with sensor uncertainty affect the theoretical downstream uncertainty of the wind estimate. 
The result of this analysis is the \textit{a priori} uncertainty associated with a single measurement, which is distinct from the \textit{a posteriori} uncertainty that would be obtained from successive incorporation of measurements in a filtering framework. 

To begin, consider the following measurement model
\begin{equation}\label{eq:appendix:uncertainty:measurementmodel}
    \boldsymbol{w} = \boldsymbol{h}(\boldsymbol{z})
\end{equation}
where $\boldsymbol{w}$ is the wind vector acting at the UAV's center of mass (CoM) and $\boldsymbol{h}(\cdot)$ is a (potentially nonlinear) function that maps measured quantities $\boldsymbol{z}$ to the wind vector.
Here, we consider the following ``measurements'' available to the wind estimation algorithm:
\begin{equation}
    \boldsymbol{\mathrm{z}} := \begin{bmatrix} \boldsymbol{a}^{IMU} \\ \boldsymbol{v} \\ \boldsymbol{b} \\ \boldsymbol{c}_{D} \\ T \\ m \end{bmatrix}
\end{equation}
which are the accelerometer measurement $\boldsymbol{a}^{IMU}$, ground velocity vector $\boldsymbol{v}$, accelerometer biases $\boldsymbol{b}$, lumped linear drag coefficients $\boldsymbol{c}_{D} = [c_{D,x}, c_{D,y}, c_{D,z}]$, control thrust $T$, and UAV mass $m$. 
Note that in this analysis, we use the word ``measurements'' loosely to mean quantities of interest that have uncertainty associated with them.
We assume the UAV has all rotors aligned with the body $z$ axis, and therefore the thrust vector can simply be considered as a scalar quantity. 

Each measurement has with it an associated uncertainty, which we can reasonably model as following a zero-mean Gaussian distribution with variance $\sigma_{(\cdot)}^2$. 
This variance can describe the noise associated with a measurement or the uncertainty of a parameter such as mass, or both. 
These variances define a covariance matrix describing the multivariate Gaussian distribution for the measurement space, $\boldsymbol{\Sigma}_{\boldsymbol{z}}$. 
Without loss of generality, we will assume that there are no cross correlations between measurements, and thus $\boldsymbol{\Sigma}_{\boldsymbol{z}}$ only contains diagonal elements.

For this analysis, the measurement function (\autoref{eq:appendix:uncertainty:measurementmodel}) will be linearized about a nominal operating point ($\bar{\boldsymbol{z}}$). 
\begin{equation}
    \boldsymbol{w} \approx \boldsymbol{h}(\bar{\boldsymbol{z}}) + \nabla \boldsymbol{h}(\boldsymbol{z} - \bar{\boldsymbol{z}})
\end{equation}
where $\nabla \boldsymbol{h}$ is the Jacobian of the measurement function. 

For a linear measurement function, the uncertainty covariance can be propagated in closed form to compute the uncertainty in the wind measurement:
\begin{equation}\label{eq:appendix:uncertainty:noisepropagation}
    \boldsymbol{\Sigma}_{\boldsymbol{w}} = (\nabla \boldsymbol{h})\boldsymbol{\Sigma}_{\boldsymbol{z}} (\nabla \boldsymbol{h})^\top
\end{equation}
We now have an equation that maps uncertainties in measured quantities to the corresponding wind vector. 
To further this analysis, we will consider a linear drag model but for illustrative purposes neglects the effect of the rotor speeds: 
\begin{equation}
    \boldsymbol{F}_{drag} = -\boldsymbol{C}(\boldsymbol{v} - \boldsymbol{w})
\end{equation}
where $\boldsymbol{C} = \texttt{diag}(\boldsymbol{c}_{D})$ is a matrix constructed from the lumped drag coefficients. 
From this drag model, we may construct a measurement function as follows: 
\begin{align}
    w_x &= \frac{m}{c_{D,x}}(a_x - b_x) + v_x \label{eq:appendix:uncertainty:scalarmodel1}  \\
    w_y &= \frac{m}{c_{D,y}}(a_y - b_y) + v_y \\ 
    w_z &= \frac{m}{c_{D,z}}(a_z - b_z) + v_z - \frac{T}{c_{D,z}} \label{eq:appendix:uncertainty:scalarmodel3}
\end{align}
These three equations define our measurement function. 
Notably, this function is nonlinear due to the ratio of mass and drag coefficients. 

Assuming no cross correlation between measured quantities, we can approximate the uncertainty by computing the Jacobian of the vector equation defined by \autoref{eq:appendix:uncertainty:scalarmodel1} through \autoref{eq:appendix:uncertainty:scalarmodel3}. 
The resulting diagonal elements of $\boldsymbol{\Sigma}_{\boldsymbol{w}}$ are:
\begin{align}
    \sigma_{w_x, w_x}^2 := \boldsymbol{\Sigma}_{\boldsymbol{w}}(1, 1)  &= \textcolor{red}{\frac{(\bar{a}_x - \bar{b}_x)^2 \bar{m}^2 \sigma_{c_{D,x}}^2}{\bar{c}_{D,x}^4}} + \frac{\bar{m}^2(\sigma_{a,x}^2 + \sigma_{b_x}^2) + \textcolor{red}{(\bar{a}_x - \bar{b}_x)^2\sigma_m^2}}{\bar{c}_{D,x}^2} + \sigma_{v_x}^2 \label{eq:appendix:uncertainty:winduncertainty1} \\ 
    \sigma_{w_y, w_y}^2 := \boldsymbol{\Sigma}_{\boldsymbol{w}}(2, 2) &= \textcolor{red}{\frac{(\bar{a}_y - \bar{b}_y)^2 \bar{m}^2 \sigma_{c_{D,y}}^2}{\bar{c}_{D,y}^4}} + \frac{\bar{m}^2(\sigma_{a,y}^2 + \sigma_{b_y}^2) + \textcolor{red}{(\bar{a}_y - \bar{b}_y)^2\sigma_m^2}}{\bar{c}_{D,y}^2} + \sigma_{v_y}^2  \label{eq:appendix:uncertainty:winduncertainty2} \\ 
    \sigma_{w_z, w_z}^2 := \boldsymbol{\Sigma}_{\boldsymbol{w}}(3, 3) &= \textcolor{red}{\frac{(\bar{T} - \bar{a}_z\bar{m} + \bar{b}_z \bar{m})^2 \bar{m}^2 \sigma_{c_{D,z}}^2}{\bar{c}_{D,z}^2}} + \frac{\bar{m}^2(\sigma_{a,z}^2 + \sigma_{b,z}^2) + (\bar{a}_z - \bar{b}_z)^2\sigma_m^2 + \sigma_T^2}{\bar{c}_{D,x}^2} + \sigma_{v_z}^2  \label{eq:appendix:uncertainty:winduncertainty3}
\end{align}
The portions in red indicate terms that we expect to be zero or near zero for nominal flight in hover with small accelerometer bias.

From the three equations above, we can glean insight into how the physical parameters affect the uncertainty of the wind estimate.
For instance, the uncertainties for the planar $x$ and $y$ wind velocities is highly sensitive to the ratio of the mass to the drag coefficients, $\bar{m}/\bar{c}_{D,x}$ and $\bar{m}/\bar{c}_{D,y}$. 
This follows the intuition that a small and light UAV ($\bar{m} \downarrow$) with high drag coefficients ($c_{D,x}, c_{D,y} \uparrow$) will be highly sensitive to the wind. 
Therefore, even small perturbations from the wind will be clearly identified from the accelerometer measurements. 

To explore this approach further, we consider approximate values for three very distinct UAVs and attempt to estimate the \textit{a priori} uncertainty if a wind estimator were to be applied. 
These results are summarized in \autoref{tab:appendix:uncertainty:robots}, values for the accelerometer uncertainty were selected based on a nominal MEMS accelerometer.
\begin{table}[b!]
\centering
\caption{Example wind uncertainties associated with three robots with varying size, mass, drag coefficients, and operating environments.}
\label{tab:appendix:uncertainty:robots}
\resizebox{\columnwidth}{!}{%
\begin{tabular}{@{}llllll@{}}
\toprule
\textbf{Robot} & \textbf{Nominal Mass ($kg$)}& \textbf{Rotor Diameter ($m$)} & \textbf{Acc Uncertainty ($m/s^2$)} & \textbf{Drag Coefficient ($Ns/m$)} & \textbf{X/Y Wind Uncertainty ($m/s$)} \\ \midrule
Falcon 4 & 4.2 & 0.381 & 0.0105$^{\dag}$ & 0.55* & 0.82 \\
Crazyflie 2.1& 0.030 & 0.045 & 0.0105$^{\dag}$ & 0.01* & 0.28 \\
Ingenuity & 1.61 & 1.61 & 0.0105$^{\dag}$ & 0.03** & 6.03 \\ \bottomrule
\end{tabular}%
}
\begin{minipage}{\columnwidth}
\small
\vspace{1mm}
\raggedright$^{\dag}$ Assuming 250 Hz sampling rate, using a noise density for an off the shelf IMU\footnote{\texttt{https://www.bosch-sensortec.com/products/motion-sensors/imus/bmi160/}}. \\
\raggedright*Identified from data. \\
\raggedright**Identified from \cite{hravard2020ingenuity}.
\end{minipage}
\end{table}
Robots with a lower ratio of mass to drag will be more sensitive to wind, and therefore the associated uncertainty will be lower compared to larger vehicles with similar drag coefficients. 

An important final remark is that the uncertainties described here are the point-wise \textit{a priori} uncertainties derived from the measurement model.
This means that the values from this analysis can be thought of as an approximate upper bound on the \textit{a posteriori} uncertainty after incorporating a sequence of observations.
For example, if the wind is constant (which it rarely ever is), the \textit{a posteriori} uncertainty would decrease by $\sqrt{N}$ where $N$ is the number of measurements taken.

\chapter{Supplementary WindShaper Experimental Data}\label{appendix:complete_windshaper_experiments}

In this appendix, supplemental data and analysis is provided to complement \revision{the hardware experiments} presented in of the thesis. 

\revision{
\section{Sub-Scale Building Design}\label{appendix:realworld:buildingdimensions}

A key novelty of the free flight experiments presented in Chapter 6 is the fact that \textit{real} obstacles are immersed in the wind tunnel in order to mimic ``urban-like'' flow features--vortex shedding, recirculation zones, corner accelerations, and more--in the scaled environment. 
To do this, obstacles were constructed with dimensions approximately matching a typical mid- or high-rise building in real cities. 
The four obstacles, seen in \autoref{fig:appendix:buildingdimensions}, were  constructed out of recycled $1$ inch plywood and rectangular in nature.
Two of the obstacles are ``slim'' and have a nominal width of $16$ inches, while the other two ``wide'' variants have a width of $24$ inches. 
For both widths, one of the obstacles stands at a height of $42$ inches while the other is $48$ inches high. 
The slightly taller $48$ inch obstacles have a $6$ inch tall cap that can be removed.
This allows one building to be rigidly stacked on top of the other resulting in a single $84$ inch tall obstacle. 
While this was helpful during preliminary experimentation, ultimately the stacked configurations were not used for the experiments presented in this thesis. 
This was because we wanted to prioritize configurations with four buildings to create more complex flow channeling. 

The obstacles are coated with a layer of gray paint to provide contrast against the floors and make the UAV more visible in media. 
Each obstacle has $7$ IR-reflective markers mounted on the surface so their location can be precisely tracked by the motion capture system--four markers are on the top corners of each obstacle, one marker is one of the bottom corners near the floor, and two markers are placed randomly on the side surfaces of each obstacle to create unique marker constellations so the motion capture software can disambiguate between each obstacle.
Lastly, the ``slim'' obstacles were fitted with tuft grids (\autoref{fig:appendix:buildingdimensions}, top picture) constructed by taping series of thin cassette tape onto the surfaces. 
These tuft grids were used during preliminary flow visualization experiments and were helpful in quickly identifying approximate flight altitudes where the 3D flow effects were minimized by seeing at what altitude the tufts were pointing neither up nor down.

\begin{figure}
    \centering
    \includegraphics[width=0.9\textwidth]{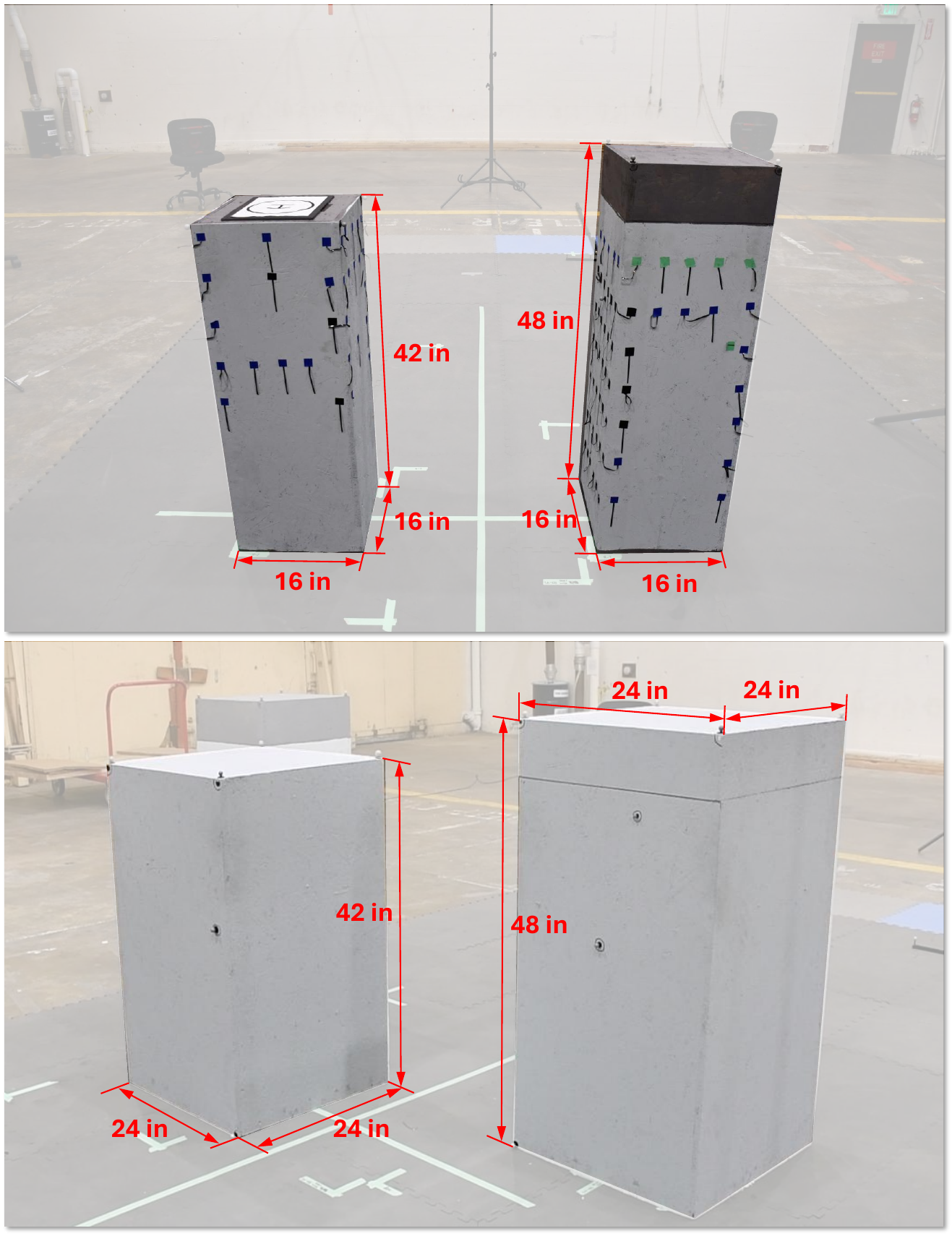}
    \caption{\revision{The four obstacles used during experiments in the WindShaper facility. The ``slim'' (top) and ``wide'' (bottom) obstacles were constructed from recycled $1$ inch thick plywood and fitted with motion capture marker arrays for pose tracking and tuft grids for flow visualization.}}
    \label{fig:appendix:buildingdimensions}
\end{figure}
}

\section{WindShaper Characterization}\label{appendix:windshaperwhisker}

Prior to any experiments, the WindShaper was characterized to provide mappings \revision{between the user commands (power level or fan speeds) and} the average \revision{wind} velocity \revision{produced by} the fans. 
This was important because \revision{the user commands} could be reported in real time \revision{but, due to network communication conflicts at the time, the \textit{WindProbe} could not be used simultaneously when the UAV was flying to provide real-time measurements of the freestream wind.} 
\revision{So these mappings could be used to estimate the} wind speed during experiments and preliminary testing \revision{without any postprocessing.} 
\revision{The} calibration data was used to provide the ground truth wind \revision{speeds} for the \textbf{Aerodynamic Characterization} \revision{experiments, which} were used for system identification of the UAV and subsequent \textit{Wind UKF} tuning during free flight testing. 

\begin{figure}[b!]
	\centering
	\includegraphics[width=0.95\textwidth]{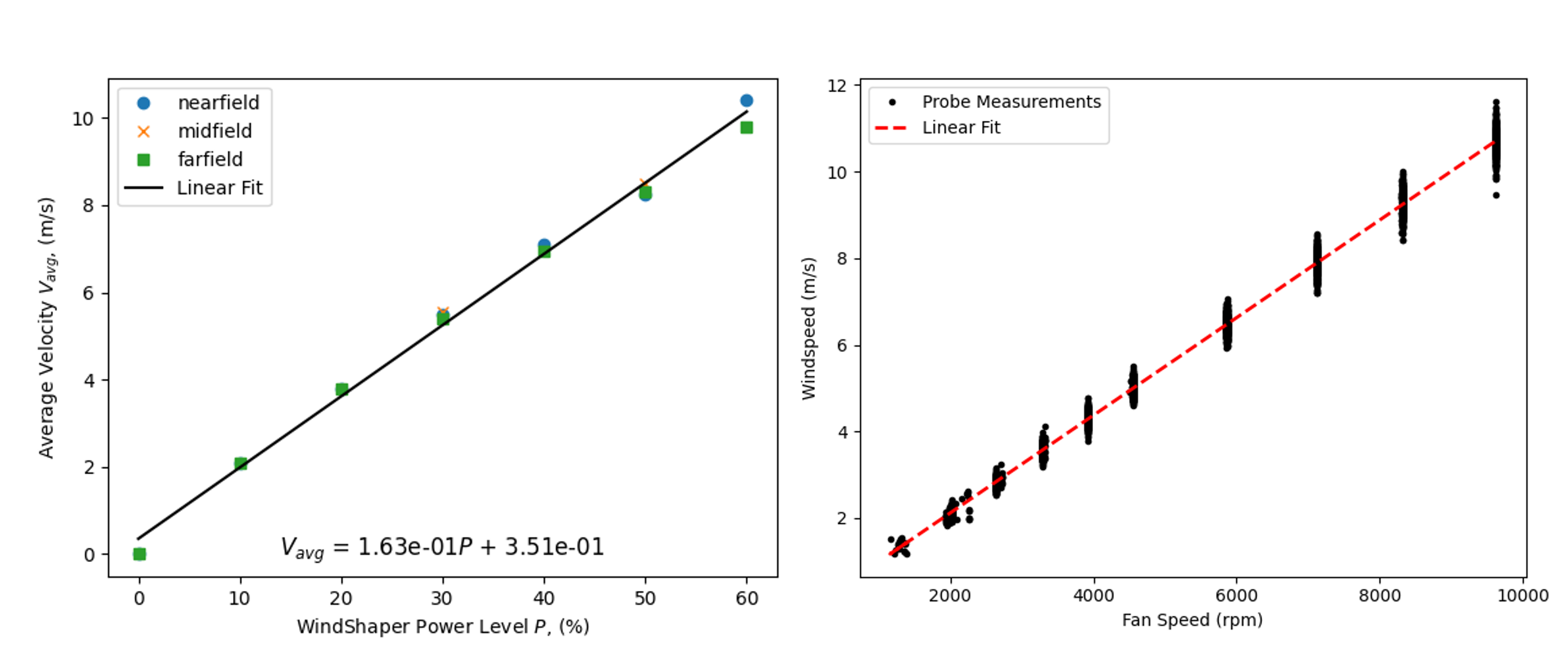}
	\caption{Characterization of the average velocity of the bulk wind flow generated by the WindShaper with varying fan power levels and fan speeds. Velocity measurements were taken at three downstream locations to confirm that the wind maintains a constant speed throughout the test section.}
	\label{fig:appendix:windshapercharacterization}
\end{figure}

In \autoref{fig:appendix:windshapercharacterization}, the average velocity from the WindShaper is plotted versus the fan power levels and fan speeds. 
On the left, the velocity was measured using the \revision{\textit{WindProbe}} at three downstream \revision{locations--\textit{near field}, \textit{mid field}, and \textit{far field}--approximately aligned} with the center of the fan array to see how the velocity varies with the downstream location. 
Two observations \revision{are} immediately apparent: 1) the wind speed is linear with the power command; and 2) the wind is roughly constant at all downstream locations, meaning it does not decelerate that much in the test section despite this being an open-air wind tunnel. 
The latter observation meant that during the \textbf{Aerodynamic Characterization} experiments the downstream location was not so important in terms of obtaining a ground truth wind speed from this data.
On the right side of the figure, the wind velocity is plotted against the fan speeds, which is useful because the power commands are open loop signals, whereas the fan speeds are measured and tracked with speed controllers and can also be logged at $100$ Hz.

\begin{figure}[t!]
	\centering
	\includegraphics[width=0.95\textwidth]{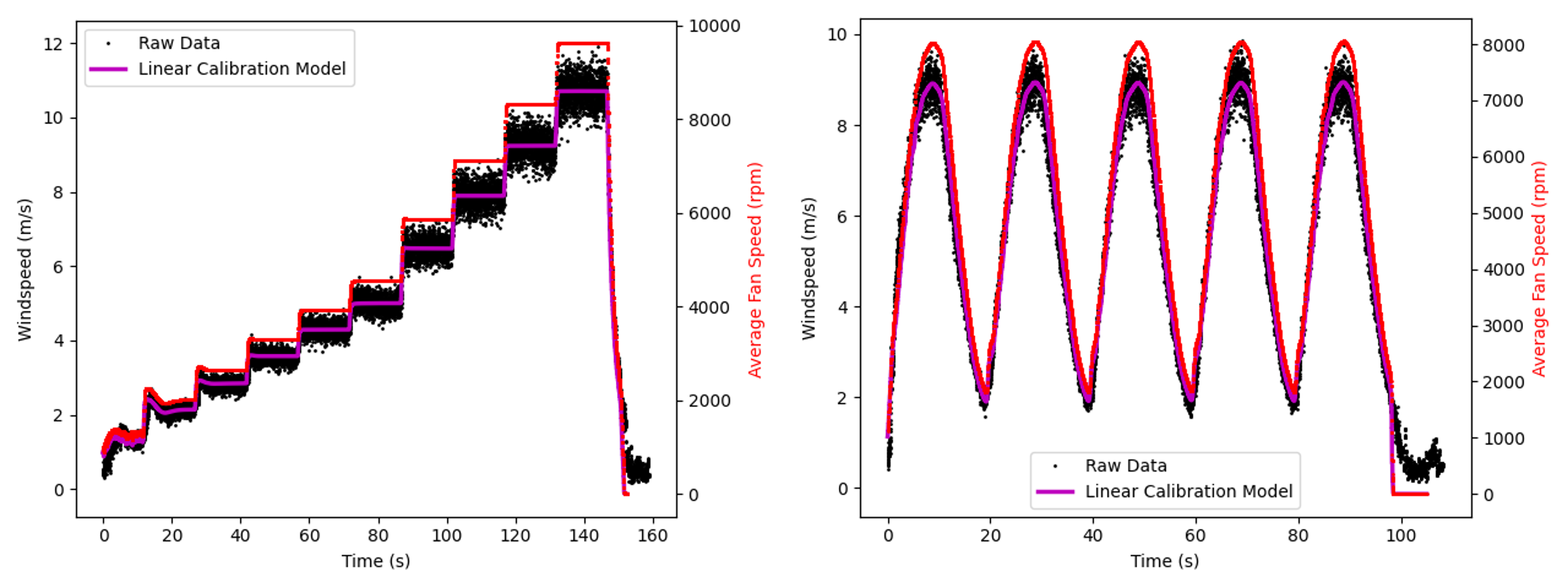}
	\caption{Validation of the WindShaper characterization on two experiments. On the left, the WindShaper receives a sequence of \revision{discrete} increasing power commands, whereas on the right the command is a \revision{continuous} sinusoid. The predicted wind from the fan speeds (magenta lines) \revision{is} in good agreement with the data collected from the \revision{\textit{WindProbe}.}}
	\label{fig:appendix:windshaper_calibrated}
\end{figure}

Because \revision{the fan speeds could be logged at $100$ Hz or more,} it \revision{was} possible to characterize the transient behavior of the WindShaper. 
Two tests were performed where the WindShaper was given transient power commands.
\revision{In} one experiment, the fans were ramped up in discrete increments and held for $15$ seconds, \revision{while} another experiment sent a sinusoid command with a period of $20$ seconds. 
The result from this testing is summarized in \autoref{fig:appendix:windshaper_calibrated}, where \revision{as expected} the wind predicted from the fan speeds match the average wind measured by the \revision{\textit{WindProbe}}. 

\section{CMU Whisker Characterization}

\begin{figure}[H]
	\centering
	\includegraphics[width=0.98\textwidth]{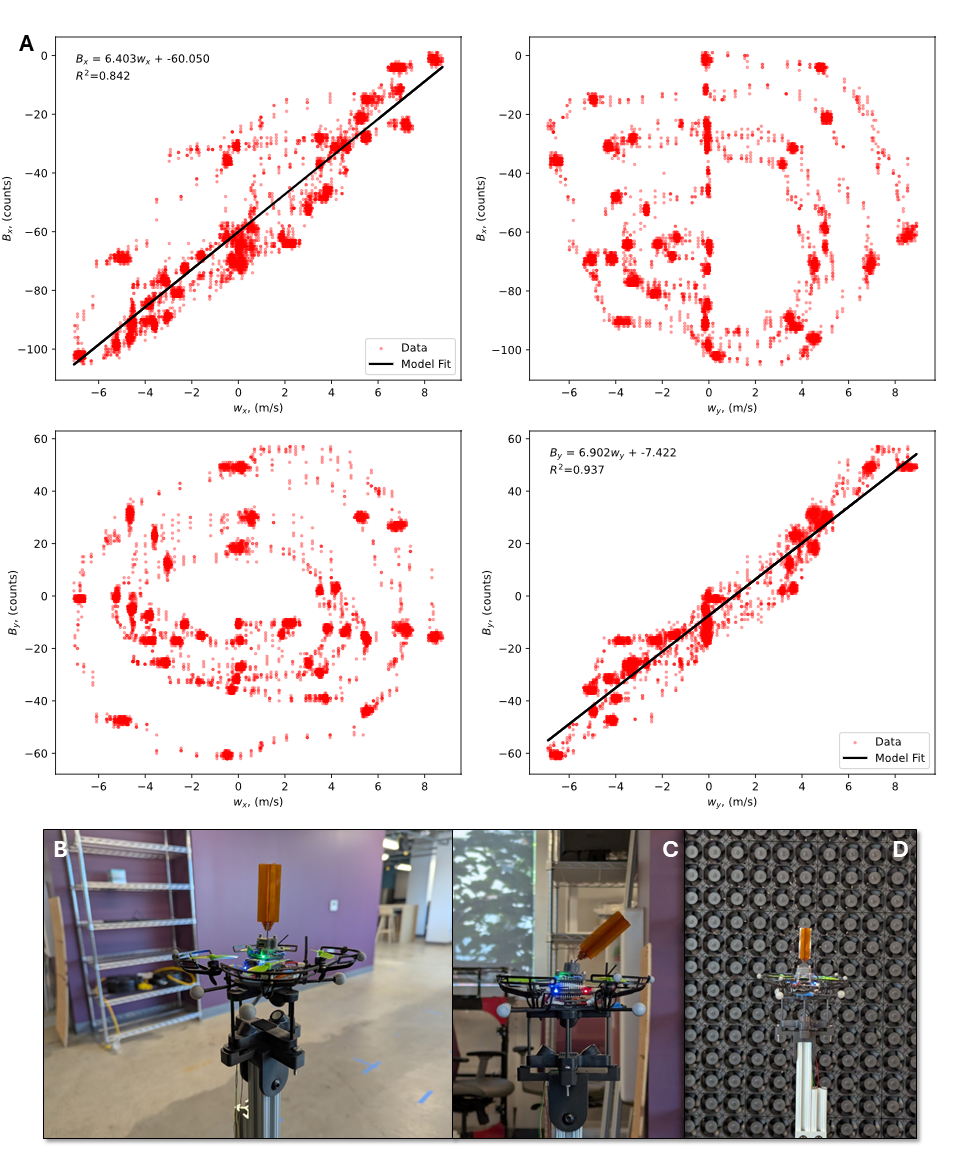}
	\caption{The whisker sensor calibration procedure: A) calibration curves for each of the sensor and wind axes; B) oblique view of the whisker sensor rigidly mounted on top of the anemometer; C) side view of the whisker experiencing maximum deflection due to the wind; D) side view of the whisker with a WindShaper fan array in the background.}
	\label{fig:appendix:whiskercalibration}
\end{figure}

The \textit{CMU Whisker} also had to be calibrated because the raw output of the sensor package was the magnetic strength $B_{(\cdot)}$ (in \revision{discrete} counts) measured by the Hall effect sensor on the $x$ and $y$ \revision{sensor} axes. 
To calibrate the whisker, it was mounted on the UAV and then \revision{the UAV was} strapped down on top of a Trisonica LI-$550$ sonic anemometer \revision{to measure} the wind speed and heading. 
As seen in \autoref{fig:appendix:whiskercalibration}, the entire whisker-UAV-anemometer system was placed in the center of the WindShaper fan array.
Because the anemometer was right below the whisker sensor and it was assumed that these two were measuring approximately the same wind speeds, \revision{and} the \revision{precise} placement of the system relative to the fan array did not matter. 
The WindShaper \revision{operated} at various wind speeds ranging from $0$ to approximately $8$ m/s, and the sensor tower was rotated a full $360$ degrees in increments of $45$ degrees.
\revision{Measurements} from both the Trisonica and the \textit{CMU Whisker} were recorded at $100$ Hz \revision{and later synchronized by hand in postprocessing using a calibrating signal.}

The data collected from this experiment is shown in the top four panels of in \autoref{fig:appendix:whiskercalibration}, where the velocity and magnetic strength are plotted against each other for the $x$ and $y$ sensor axes. 
From this data it is \revision{evident} that the whisker's $x$ and $y$ axes can be considered decoupled; that is, wind on the $x$ axis is uncorrelated with the $y$ axis of the whisker, and vice versa. 
The other notable observation is that the mapping from the magnetic strength to the wind speed can be captured quite accurately using a linear fit \revision{with R\textsuperscript{2} values of $0.842$ and $0.937$ for the $x$ and $y$ sensor axes, respectively.} 

The reader should be aware that the calibration data presented here is extremely sensitive to the whisker's construction. 
The calibration procedure had to be repeated on two \revision{additional} occasions during \revision{the full-stack} experiments \revision{when the whisker} had to be reattached to the base \revision{after crashes.}
\revision{While the linear trends persisted between crashes,} the slopes and intercepts of the linear fits were noticeably different between each calibration. 
\revision{For that reason, the specific slope and intercept values provided in this figure should not be used in other experiments. 
However, the calibration procedure remains valid.}

\newpage

\section{Experimental Logs}\label{appendix:experimentallogs}

This section provides the complete log of experimental trials conducted at the NASA Ames WindShaper facility, detailing the parameters used for each flight test, as summarized in \autoref{tab:realworld:experiment_description}.
There were a total of $40$ trials with the UAV operating in the WindShaper facility across three experimental series: \textbf{Aerodynamic Characterization}, \textbf{Flow Traversal}, and \textbf{Obstacle Avoidance}. 
The majority of the trials belonged to the \textbf{Flow Traversal} series, where the UAV was tracking a wand operated by a human while navigating around obstacles and handling wind generated from the WindShaper. 
However, there were also a limited number of trials where the WindShaper was off (\textbf{Obstacle Avoidance}) and only the obstacle avoidance feature of the MPPI controller was tested. 
For all the \textbf{Flow Traversal} experiments, at least one trial ran with the wind-aware MPPI controller while another ran with a standard implementation of the SE(3) controller for quadrotors.  

\begin{table}[htbp]
\centering
\caption{A complete list of the experiments conducted at the NASA Ames WindShaper facility. The building configurations are shown in \autoref{fig:realworld:facilities}. The number of trials per experiment are noted and broken down into which controller was running.}
\label{tab:realworld:experiment_description}
\begin{tabularx}{\textwidth}{@{} c X c c c @{}}
\toprule
\textbf{Exp. ID} & \textbf{Experiment Series} & \textbf{\# Trials (SE3/MPPI)} & \textbf{Building Config.} & \textbf{Fan Power (\%)} \\
\midrule
A1 & Aero. Characterization & 1 (1/0) & B0 & 0-50 \\
\midrule
1-2 & Flow Traversal & 4 (2/2) & B0 & 10, 30 \\
3-5 & Flow Traversal & 6 (3/3) & B1 & 10, 25, 35 \\
6-8 & Flow Traversal & 7 (4/3) & B2 & 10, 25, 35 \\
9-11 & Flow Traversal & 6 (3/3) & B3 & 10, 25, 35 \\
12-14 & Flow Traversal & 6 (3/3) & B4 & 10, 25, 35 \\
15-17 & Flow Traversal & 6 (3/3) & B5 & 10, 25, 35 \\
\midrule
100 & Obstacle Avoidance & 1 (0/1) & B3 & 0 \\
200 & Obstacle Avoidance & 1 (0/1) & B4 & 0 \\
300 & Obstacle Avoidance & 3 (0/3) & B5 & 0 \\
\bottomrule
\end{tabularx}
\end{table}

In all of the experiments, the following data was logged in ROS: the poses of the WindShaper, GoPro cameras, buildings, tracking wand, and UAV; the ``state'' of the WindShaper which includes the individual fan speeds and power commands; the UAV's battery voltage, supply current, power consumption, commanded motor speeds, IMU measurements, and high level controller commands (thrust and attitude); the simulated LiDAR point cloud; the raw sensor data from the \textit{CMU Whisker} as well as filtered estimates of the wind from these readings; and finally, the full state estimate from \textit{Wind UKF} which includes the wind estimate. 
All of \revision{this data} was recorded at rates varying from $10$ Hz to $120$ Hz depending on the sensor, and time synchronization was handled by ROS.
This means that each experiment listed in \autoref{tab:realworld:experiment_description} can be played \revision{back} in real time. 

These ROS data files can be made accessible upon request. 

\section{Additional Experiment Analysis}\label{appendix:additional_exp_analysis}

The figures presented in the main body of the thesis represent just a small portion of the total data collected at the NASA WindShaper facility. 
In this section, we provide plots that \revision{give} a more complete view into the measurements collected during these experiments. 

In \autoref{appendix:scalarfield_complete1} and \autoref{appendix:scalarfield_complete2}, the velocity magnitude of the flow field reconstructions from the \textit{WindProbe}, \textit{Wind UKF}, and \textit{CMU Whisker} are visualized for all of the \textbf{Flow Traversal} experiments. 
For these columns, the color limits are all set to the range $[0, 6]$ m/s.
The latter two columns are unsteadiness metrics--standard deviation in lateral force command and average body rotational rates. 
For these columns, the lower color limit represents the minimum value for the corresponding experiment, and the upper color limit represents the upper 90'th percentile. 

\newpage
\begin{figure}[H]
    \centering 
    \includegraphics[width=0.99\textwidth]{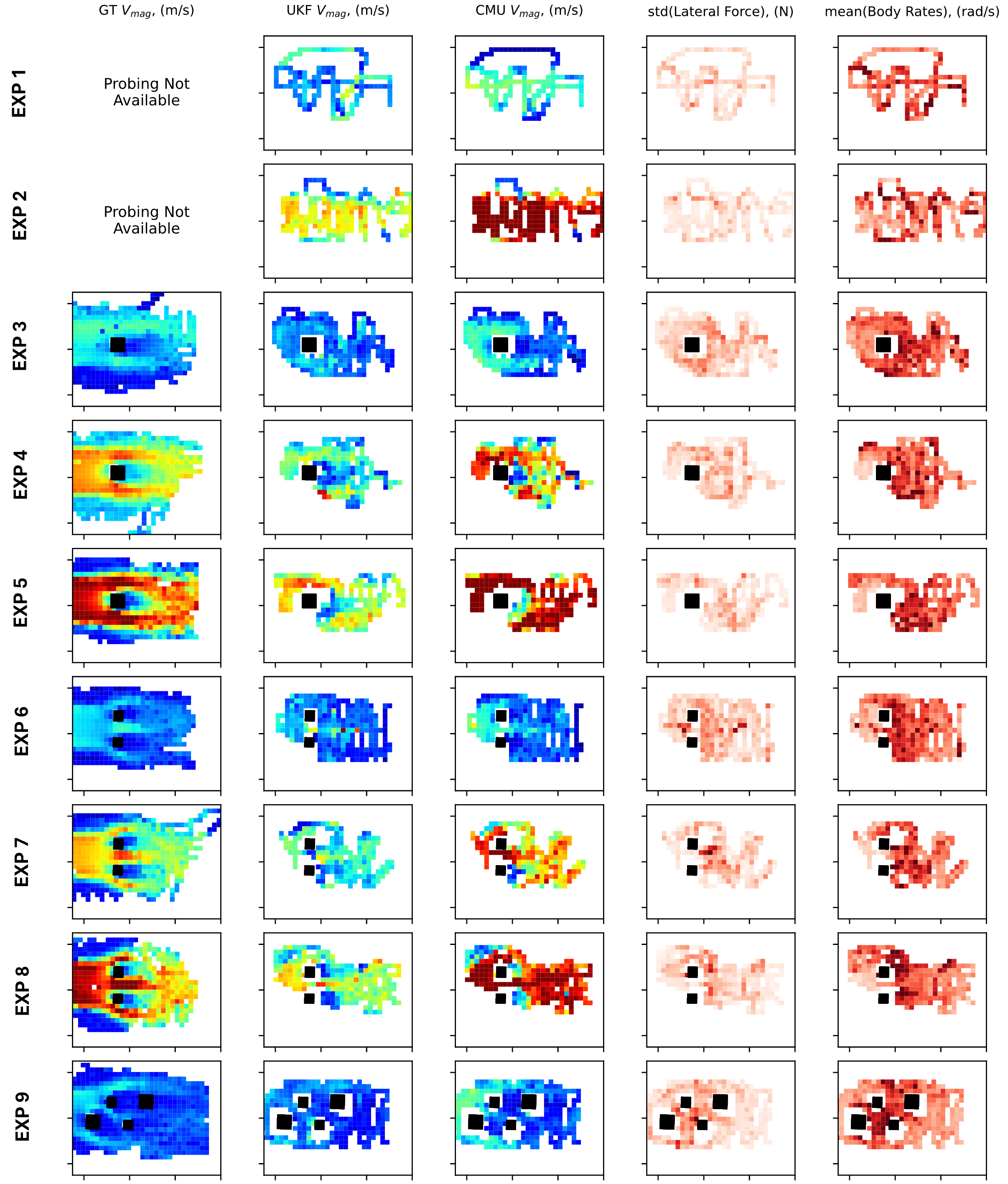}
    \caption{Part 1 of the flow field reconstructions and unsteadiness metrics for the \textbf{Flow Traversal} experimental series.}
    \label{appendix:scalarfield_complete1}
\end{figure}

\newpage
\begin{figure}[H]
    \centering 
    \includegraphics[width=0.99\textwidth]{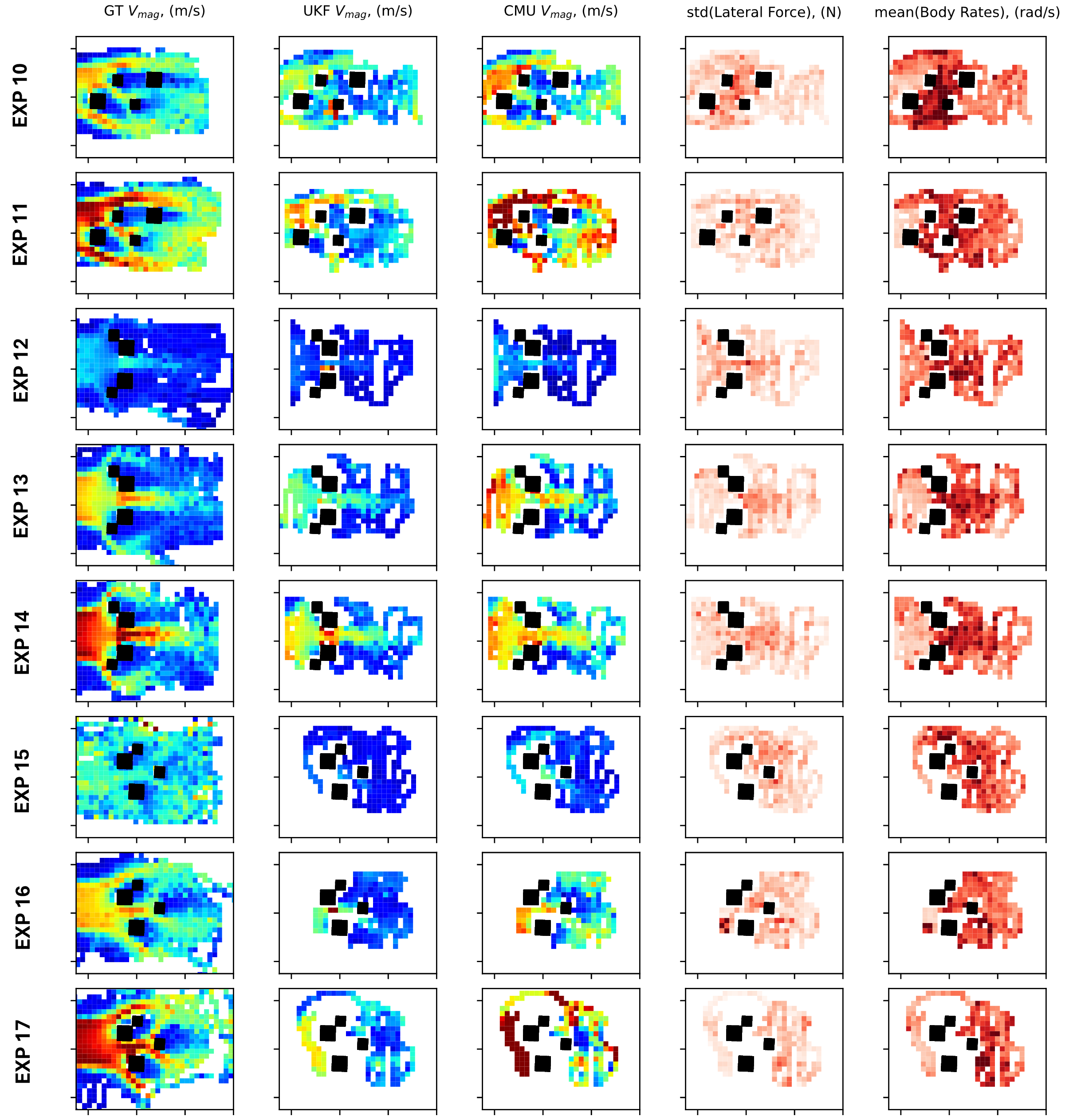}
    \caption{Part 2 of the flow field reconstructions and unsteadiness metrics for the \textbf{Flow Traversal} experimental series.}
    \label{appendix:scalarfield_complete2}
\end{figure}

\newpage
\section{MPPI Profiling}\label{app:profiling}

This section contains the raw computational profiling data for the \revision{wind-aware} Model Predictive Path Integral (MPPI) controller from the main text, running on an Nvidia Jetson AGX Xavier single-board computer.
The full data is provided to validate the summary statistics presented in Chapter 6, \autoref{tab:realworld:profiling}.

The profile was generated using the Python \texttt{cProfile} module. 
The log below captures a $1.86$ second window of operation, corresponding to $100$ total control loop cycles and $25$ updates from the MPPI controller.

\begin{table}[h!]
    \centering
    \caption{Raw \texttt{cProfile} output for the MPPI planner on Jetson AGX Xavier. The profile captures $100$ control loops and $25$ planning iterations.}
    \label{tab:realworld:profile_raw}
    \begin{lstlisting}[language=profilelog]
   ncalls  tottime  percall  cumtime  percall filename:lineno(function)
    100	0.018	0.000	1.863	0.019 	mppi_torch_control.py:457(update)
   	25	1.111	0.044	1.111	0.044 	{method 'cpu' of 'torch._C._TensorBase' objects}
   	25	0.544	0.022	0.637	0.025 	mppi_torch_control.py:397(update_mppi)
   	25	0.036	0.001	0.062	0.002	mppi_torch_control.py:353(compute_cost)
 	1325/625	0.007	0.000	0.061	0.000 	{built-in method numpy.core._multiarray_umath.implement_array_function}
  	200	0.000	0.000	0.049	0.000 	<__array_function__ internals>:177(cross)
  	200	0.017	0.000	0.047	0.000 	/usr/local/lib/python3.8/dist-packages/numpy/core/numeric.py:1486(cross)
  	125	0.032	0.000	0.032	0.000 	{method 'copy_' of 'torch._C._TensorBase' objects}
  	600	0.007	0.000	0.024	0.000 	/usr/local/lib/python3.8/dist-packages/numpy/core/numeric.py:1411(moveaxis)
   	25	0.004	0.000	0.020	0.001 	mppi_torch_control.py:426(update_goals)
  	150	0.001	0.000	0.018	0.000 	/usr/local/lib/python3.8/dist-packages/torch/_tensor.py:34(wrapped)
  	125	0.015	0.000	0.015	0.000 	{method 'pow' of 'torch._C._TensorBase' objects}
 	1200	0.009	0.000	0.012	0.000 	/usr/local/lib/python3.8/dist-packages/numpy/core/numeric.py:1348(normalize_axis_tuple)
  	100	0.000	0.000	0.007	0.000 	<__array_function__ internals>:177(stack)
  	200	0.001	0.000	0.007	0.000 	<__array_function__ internals>:177(norm)
  	200	0.003	0.000	0.005	0.000 	/usr/local/lib/python3.8/dist-packages/numpy/linalg/linalg.py:2342(norm)
  	100	0.002	0.000	0.005	0.000 	/usr/local/lib/python3.8/dist-packages/numpy/core/shape_base.py:383(stack)
  	100	0.005	0.000	0.005	0.000 	{method 'sum' of 'torch._C._TensorBase' objects}
  	100	0.004	0.000	0.004	0.000 	{built-in method from_matrix}
   	75	0.003	0.000	0.003	0.000 	{built-in method torch.sum}
 	1100	0.003	0.000	0.003	0.000 	{built-in method numpy.array}
   	50	0.003	0.000	0.003	0.000 	{method 'min' of 'torch._C._TensorBase' objects}
  	100	0.003	0.000	0.003	0.000 	{built-in method from_quat}
   	25	0.001	0.000	0.002	0.000 	/usr/local/lib/python3.8/dist-packages/torch/_tensor.py:838(__rdiv__)
  	100	0.000	0.000	0.002	0.000 	<__array_function__ internals>:177(concatenate)
 	1200	0.001	0.000	0.002	0.000 	/usr/local/lib/python3.8/dist-packages/numpy/core/numeric.py:1398(<listcomp>)
  	100	0.002	0.000	0.002	0.000 	{method 'as_matrix' of 'scipy.spatial.transform._rotation.Rotation' objects}
   	50	0.002	0.000	0.002	0.000 	{built-in method torch.sqrt}
    \end{lstlisting}
\end{table}



\begin{bibliof}
\bibliography{bibliography}

@book{leishman2006principles,
  title={Principles of helicopter aerodynamics with CD extra},
  author={Leishman, Gordon J.},
  year={2006},
  publisher={Cambridge university press}
}

@book{tomic2022modelbasedwindestimation,
  title={Model-based control of flying robots for robust interaction under wind influence},
  author={Tomi{\'c}, Teodor},
  volume={151},
  year={2022},
  publisher={Springer Nature}
}

@inproceedings{tagliabue2020touch,
  title={Touch the wind: Simultaneous airflow, drag and interaction sensing on a multirotor},
  author={Tagliabue, Andrea and Paris, Aleix and Kim, Suhan and Kubicek, Regan and Bergbreiter, Sarah and How, Jonathan P.},
  booktitle={2020 IEEE/RSJ international conference on intelligent robots and systems (IROS)},
  pages={1645--1652},
  year={2020},
  organization={IEEE}
}

@inproceedings{svacha2017improving,
  title={Improving quadrotor trajectory tracking by compensating for aerodynamic effects},
  author={Svacha, James and Mohta, Kartik and Kumar, Vijay},
  booktitle={2017 international conference on unmanned aircraft systems (ICUAS)},
  pages={860--866},
  year={2017},
  organization={IEEE}
}

@article{davoudi2020quadrotorflightsimulation,
author = {Davoudi, Behdad and Taheri, Ehsan and Duraisamy, Karthik and Jayaraman, Balaji and Kolmanovsky, Ilya},
title = {Quad-Rotor Flight Simulation in Realistic Atmospheric Conditions},
journal = {AIAA Journal},
volume = {58},
number = {5},
pages = {1992-2004},
year = {2020},
doi = {10.2514/1.J058327},

URL = { 
    
        https://doi.org/10.2514/1.J058327

},
eprint = { 
    
        https://doi.org/10.2514/1.J058327

}
}

@article{galway2011control,
author = {Galway, David and Etele, J. and Fusina, Giovanni},
title = {Modeling of Urban Wind Field Effects on Unmanned Rotorcraft Flight},
journal = {Journal of Aircraft},
volume = {48},
number = {5},
pages = {1613-1620},
year = {2011},
doi = {10.2514/1.C031325},
URL = {https://doi.org/10.2514/1.C031325},
eprint = {https://doi.org/10.2514/1.C031325}
}

@INPROCEEDINGS{orr2005fixedwingmodelinurbanenv,
  author={Orr, M.W. and Rasmussen, S.J. and Karni, E.D. and Blake, W.B.},
  booktitle={Proceedings of the 2005, American Control Conference, 2005.}, 
  title={Framework for developing and evaluating MAV control algorithms in a realistic urban setting}, 
  year={2005},
  volume={},
  number={},
  pages={4096-4101 vol. 6},
  doi={10.1109/ACC.2005.1470619},
  ISSN={2378-5861},
  month={June},}

@article{mir2018dynamicsoaring,
  title={Review of dynamic soaring: technical aspects, nonlinear modeling perspectives and future directions},
  author={Mir, Imran and Eisa, Sameh A. and Maqsood, Adnan},
  journal={Nonlinear Dynamics},
  volume={94},
  pages={3117--3144},
  year={2018},
  publisher={Springer}
}

@INPROCEEDINGS{patrikar2020particle,
  author={Patrikar, Jay and Moon, Brady G. and Scherer, Sebastian},
  booktitle={2020 IEEE/RSJ International Conference on Intelligent Robots and Systems (IROS)}, 
  title={Wind and the City: Utilizing {UAV}-Based In-Situ Measurements for Estimating Urban Wind Fields}, 
  year={2020},
  volume={},
  number={},
  pages={1254-1260},
  doi={10.1109/IROS45743.2020.9340812}}

@article{ebert2023gappy,
author = {Ebert, Carola and Weiss, Julien and Uijt de Haag, Maarten and Ruwisch, Christopher and Silvestre, Flavio J.},
title = {Trajectory Planning in Windy Urban Environment Using Gappy Proper Orthogonal Decomposition for Wind Estimates},
journal = {AIAA Journal},
volume = {61},
number = {6},
pages = {2640-2651},
year = {2023},
doi = {10.2514/1.J062049},
URL = {
        https://doi.org/10.2514/1.J062049
},
eprint = { 
    
        https://doi.org/10.2514/1.J062049
}
}

@inproceedings{ware2016canopy,
  author={Ware, John and Roy, Nicholas},
  booktitle={2016 IEEE International Conference on Robotics and Automation (ICRA)}, 
  title={An analysis of wind field estimation and exploitation for quadrotor flight in the urban canopy layer}, 
  year={2016},
  volume={},
  number={},
  pages={1507-1514},
  doi={10.1109/ICRA.2016.7487287}}

@mastersthesis{ware2016thesis,
    author = {John Ware},
    title = {An analysis of quadrotor flight in the urban canopy layer},
    school = {Massachusetts Institute of Technology},
    department = {Department of Aeronautics and Astronautics},
    year = {2019},
    url = {[https://dspace.mit.edu/handle/1721.1/107056](https://dspace.mit.edu/handle/1721.1/107056)}
}

@article{folk2023rotorpy,
  title={{R}otor{P}y: A {Python}-based Multirotor Simulator with Aerodynamics for Education and Research},
  author={Folk, Spencer and Paulos, James and Kumar, Vijay},
  journal={arXiv preprint arXiv:2306.04485},
  year={2023}
}

@article{bauranov2021airspace,
title = {Designing airspace for urban air mobility: A review of concepts and approaches},
journal = {Progress in Aerospace Sciences},
volume = {125},
pages = {100726},
year = {2021},
issn = {0376-0421},
doi = {https://doi.org/10.1016/j.paerosci.2021.100726},
url = {https://www.sciencedirect.com/science/article/pii/S0376042121000312},
author = {Bauranov, Aleksandar and Rakas, Jasenka},
}

@techreport{faa2023conceptofoperations,
  author      = {{Federal Aviation Administration}},
  title       = {Urban Air Mobility ({UAM}) Concept of Operations},
  institution = {U.S. Department of Transportation, Federal Aviation Administration},
  year        = {2023},
  type        = {Concept of Operations},
  number      = {v2.0},
  address     = {Washington, DC},
  url         = {https://www.faa.gov/air-taxis/uam_blueprint},
  note        = {Accessed October 25, 2023}
}

@misc{fortunebusinessinsights2020uam,
  author       = {{Fortune Business Insights}},
  title        = {Urban Air Mobility Market Size, Industry Share, Growth Report, 2030},
  year         = {2021},
  howpublished = {\url{https://www.fortunebusinessinsights.com/urban-air-mobility-uam-market-106344}},
  note         = {Accessed January 13, 2026}
}

@techreport{goyal2019uamstudy,
  author      = {Goyal, Rohit and Reiche, Colleen and Fernando, Chris and Serrao, Jacquie and Kimmel, Shawn and Cohen, Adam and Shaheen, Susan},
  title       = {Urban Air Mobility ({UAM}) Market Study},
  institution = {National Aeronautics and Space Administration (NASA)},
  year        = {2019},
  number      = {NASA/CR-2018-220037},
  address     = {Washington, DC},
  url         = {https://ntrs.nasa.gov/citations/20190001472},
  note        = {Accessed October 25, 2023}
}

@article{rienecker2023planning,
  author = {Rienecker, Hannes and Hildebrand, Veit and Pfifer, Harald},
  title = {Energy Optimal {3D} Flight Path Planning for Unmanned Aerial Vehicle in Urban Environments},
  journal = {CEAS Aeronautical Journal},
  volume = {14},
  number = {3},
  pages = {621-636},
  year = {2023},
  month = {July 01},
  doi = {10.1007/s13272-023-00666-x},
  url = {https://doi.org/10.1007/s13272-023-00666-x},
  issn = {1869-5590}
}

@incollection{stam2023stable,
  title={Stable fluids},
  author={Stam, Jos},
  booktitle={Seminal Graphics Papers: Pushing the Boundaries, Volume 2},
  pages={779--786},
  publisher={Association for Computing Machinery},
  year={2023}
}

@article{toparlar2017cfdreview,
title = {A review on the {CFD} analysis of urban microclimate},
journal = {Renewable and Sustainable Energy Reviews},
volume = {80},
pages = {1613-1640},
year = {2017},
issn = {1364-0321},
doi = {https://doi.org/10.1016/j.rser.2017.05.248},
url = {https://www.sciencedirect.com/science/article/pii/S1364032117308924},
author = {Toparlar, Yasin and Blocken, Bert J. E. and Maiheu, Bino and {van Heijst}, Gertjan Jan F.}
}

@article{achermann2024windseer,
  title={WindSeer: Real-time volumetric wind prediction over complex terrain aboard a small {UAV}},
  author={Achermann, Florian and Stastny, Thomas and Danciu, Bogdan and Kolobov, Andrey and Chung, Jen Jen and Siegwart, Roland and Lawrance, Nicholas},
  journal={arXiv preprint arXiv:2401.09944},
  year={2024}
}

@article{langelaan2009gust,
  title={Gust energy extraction for mini and micro uninhabited aerial vehicles},
  author={Langelaan, Jack W.},
  journal={Journal of guidance, control, and dynamics},
  volume={32},
  number={2},
  pages={464--473},
  year={2009}
}

@article{langelaan2011prediction,
  title={Wind field estimation for small unmanned aerial vehicles},
  author={Langelaan, Jack W. and Alley, Nicholas and Neidhoefer, James},
  journal={Journal of Guidance, Control, and Dynamics},
  volume={34},
  number={4},
  pages={1016--1030},
  year={2011}
}

@inproceedings{chakrabarty2013kinematictree,
  author={Chakrabarty, Anjan and Langelaan, Jack},
  booktitle={2013 American Control Conference}, 
  title={{UAV} flight path planning in time varying complex wind-fields}, 
  year={2013},
  volume={},
  number={},
  pages={2568-2574},
  doi={10.1109/ACC.2013.6580221}}

@article{lawrance2011soaring,
  title={Autonomous exploration of a wind field with a gliding aircraft},
  author={Lawrance, Nicholas R. J. and Sukkarieh, Salah},
  journal={Journal of guidance, control, and dynamics},
  volume={34},
  number={3},
  pages={719--733},
  year={2011}
}

@inproceedings{lawrance2010simultaneous,
author = {Nicholas Lawrance and Salah Sukkarieh},
title = {Simultaneous Exploration and Exploitation of a Wind Field for a Small Gliding {UAV}},
booktitle = {AIAA Guidance, Navigation, and Control Conference},
publisher = {AIAA},
chapter = {},
pages = {},
year={2010},
doi = {10.2514/6.2010-8032},
URL = {https://arc.aiaa.org/doi/abs/10.2514/6.2010-8032},
eprint = {https://arc.aiaa.org/doi/pdf/10.2514/6.2010-8032}
}

@inproceedings{lawrance2009planner,
author = {Nicholas Lawrance and Salah Sukkarieh},
title = {Wind Energy Based Path Planning for a Small Gliding Unmanned Aerial Vehicle},
booktitle = {AIAA Guidance, Navigation, and Control Conference},
publisher = {AIAA},
chapter = {},
pages = {},
year={2009},
doi = {10.2514/6.2009-6112},
URL = {https://arc.aiaa.org/doi/abs/10.2514/6.2009-6112},
eprint = {https://arc.aiaa.org/doi/pdf/10.2514/6.2009-6112}
}

@article{norberg1993strouhal,
title = {Flow around rectangular cylinders: Pressure forces and wake frequencies},
journal = {Journal of Wind Engineering and Industrial Aerodynamics},
volume = {49},
number = {1},
pages = {187-196},
year = {1993},
issn = {0167-6105},
doi = {https://doi.org/10.1016/0167-6105(93)90014-F},
url = {https://www.sciencedirect.com/science/article/pii/016761059390014F},
author = {Norberg, Christoffer}
}

@article{nakaguchi1968experimentalcylinder,
  title={An Experimental Study on Aerodynamic Drag of Rectangular Cylinders},
  author={Nakaguchi, Hiroshi and Hashimoto, Kikuhiro and Muto, Shinri},
  journal={The Journal of the Japan Society of Aeronautical Engineering},
  volume={16},
  number={168},
  pages={1-5},
  year={1968},
  doi={10.2322/jjsass1953.16.1}
}

@inproceedings{tomic2016flying,
  title={The flying anemometer: Unified estimation of wind velocity from aerodynamic power and wrenches},
  author={Tomi{\'c}, Teodor and Schmid, Korbinian and Lutz, Philipp and Mathers, Andrew and Haddadin, Sami},
  booktitle={2016 IEEE/RSJ international conference on intelligent robots and systems (IROS)},
  pages={1637--1644},
  year={2016},
  organization={IEEE}
}

@article{chao2021windestimationsurvey,
author = {Tian, Pengzhi and Chao, Haiyang and Rhudy, Matthew and Gross, Jason and Wu, Huixuan},
title = {Wind Sensing and Estimation Using Small Fixed-Wing Unmanned Aerial Vehicles: A Survey},
journal = {Journal of Aerospace Information Systems},
volume = {18},
number = {3},
pages = {132-143},
year = {2021},
doi = {10.2514/1.I010885},

URL = { 
    
        https://doi.org/10.2514/1.I010885
    
    

},
eprint = { 
    
        https://doi.org/10.2514/1.I010885
    
    

}

}

@INPROCEEDINGS{oettershagen2019realtimewindprediction,
  author={Oettershagen, Philipp and Müller, Benjamin and Achermann, Florian and Siegwart, Roland},
  booktitle={2019 IEEE Aerospace Conference}, 
  title={Real-time 3{D} wind field prediction onboard {UAVs} for safe flight in complex terrain}, 
  year={2019},
  volume={},
  number={},
  pages={1-10},
  doi={10.1109/AERO.2019.8742160}}

@article{fukami2023super,
  title={Super-resolution analysis via machine learning: A survey for fluid flows},
  author={Fukami, Kai and Fukagata, Koji and Taira, Kunihiko},
  journal={Theoretical and Computational Fluid Dynamics},
  volume={37},
  number={4},
  pages={421--444},
  year={2023},
  publisher={Springer}
}

@article{sharma2023quantifying,
  title={Quantifying the Effect of Weather on Advanced Air Mobility Operations},
  author={Sharma, Ashima and Patrikar, Jay and Moon, Brady and Scherer, Sebastian and Samaras, Constantine},
  journal={Findings},
  year={2023},
  publisher={Findings Press}
}

@article{blocken2014fiftyyears,
  title={50 years of computational wind engineering: past, present and future},
  author={Blocken, Bert},
  journal={Journal of Wind Engineering and Industrial Aerodynamics},
  volume={129},
  pages={69--102},
  year={2014},
  publisher={Elsevier}
}

@phdthesis{svacha2019thesis,
  title={{IMU}-based state estimation and control of quadrotors exploiting aerodynamic effects},
  author={Svacha, James},
  year={2019},
  school={University of Pennsylvania}
}

@ARTICLE{abichandani2020sensingreview,
  author={Abichandani, Pramod and Lobo, Deepan and Ford, Gabriel and Bucci, Donald and Kam, Moshe},
  journal={IEEE Access}, 
  title={Wind Measurement and Simulation Techniques in Multi-Rotor Small Unmanned Aerial Vehicles}, 
  year={2020},
  volume={8},
  number={},
  pages={54910-54927},
  doi={10.1109/ACCESS.2020.2977693}}

@article{gianfelice2022lookup,
title = {Real-time Wind Predictions for Safe Drone Flights in {Toronto}},
journal = {Results in Engineering},
volume = {15},
pages = {100534},
year = {2022},
issn = {2590-1230},
doi = {https://doi.org/10.1016/j.rineng.2022.100534},
url = {https://www.sciencedirect.com/science/article/pii/S2590123022002043},
author = {Gianfelice, Michael and Aboshosha, Haitham and Ghazal, Tarek}
}

@article{zhao2022windtunnelreview,
title = {Boundary layer wind tunnel tests of outdoor airflow field around urban buildings: A review of methods and status},
journal = {Renewable and Sustainable Energy Reviews},
volume = {167},
pages = {112717},
year = {2022},
issn = {1364-0321},
doi = {https://doi.org/10.1016/j.rser.2022.112717},
url = {https://www.sciencedirect.com/science/article/pii/S1364032122006062},
author = {Zhao, Yi and Li, Ruibin  and Feng, Lu and Wu, Yan and Niu, Jianlei and Gao, Naiping}
}

@article{shao2023pignncfd,
title = {{PIGNN-CFD}: A physics-informed graph neural network for rapid predicting urban wind field defined on unstructured mesh},
journal = {Building and Environment},
volume = {232},
pages = {110056},
year = {2023},
issn = {0360-1323},
doi = {https://doi.org/10.1016/j.buildenv.2023.110056},
url = {https://www.sciencedirect.com/science/article/pii/S0360132323000835},
author = {Shao, Xuqiang and Liu, Zhijian and Zhang, Siqi and  Zhao, Zijia and Hu, Chenxing}
}

@article{kastner2023gansurrogate,
title = {A {GAN}-Based Surrogate Model for Instantaneous Urban Wind Flow Prediction},
journal = {Building and Environment},
volume = {242},
pages = {110384},
year = {2023},
issn = {0360-1323},
doi = {https://doi.org/10.1016/j.buildenv.2023.110384},
url = {https://www.sciencedirect.com/science/article/pii/S0360132323004110},
author = {Kastner, Patrick and Dogan, Timur}
}

@article{gao2023GNNwindpredictor,
author = {Gao, Huanxiang and Hu, Gang and Zhang, Dongqin and Jiang, Wenjun and Tse, K. T. and Kwok, K. C. S. and Kareem, Ahsan},
title = {Urban wind field prediction based on sparse sensors and physics-informed graph-assisted auto-encoder},
journal = {Computer-Aided Civil and Infrastructure Engineering},
volume = {39},
number = {10},
pages = {1409-1430},
doi = {https://doi.org/10.1111/mice.13147},
url = {https://onlinelibrary.wiley.com/doi/abs/10.1111/mice.13147},
eprint = {https://onlinelibrary.wiley.com/doi/pdf/10.1111/mice.13147},
year = {2024}
}

@article{dimmig2023survey,
  title={Survey of Simulators for Aerial Robots},
  author={Dimmig, Cora A. and Silano, Giuseppe and McGuire, Kimberly and Gabellieri, Chiara and H{\"o}nig, Wolfgang and Moore, Joseph and Kobilarov, Marin},
  journal={arXiv preprint arXiv:2311.02296},
  year={2023}
}

@inproceedings{salzmann2024learning,
  title={Learning for {CasADi}: Data-driven models in numerical optimization},
  author={Salzmann, Tim and Arrizabalaga, Jon and Andersson, Joel and Pavone, Marco and Ryll, Markus},
  booktitle={6th Annual Learning for Dynamics \& Control Conference},
  pages={541--553},
  year={2024},
  organization={PMLR}
}

@article{salam2022learning,
  author={Salam, Tahiya and Edwards, Victoria and Hsieh, M. Ani},
  journal={IEEE Robotics and Automation Letters}, 
  title={Learning and Leveraging Features in Flow-Like Environments to Improve Situational Awareness}, 
  year={2022},
  volume={7},
  number={2},
  pages={2071-2078},
  doi={10.1109/LRA.2022.3141762}
}

@InProceedings{salam2023l4dc,
  title = 	 {Online Estimation of the {Koopman} Operator Using {Fourier} Features},
  author =       {Salam, Tahiya and Li, Alice K. and Hsieh, M. Ani},
  booktitle = 	 {Proceedings of The 5th Annual Learning for Dynamics and Control Conference},
  pages = 	 {1271--1283},
  year = 	 {2023},
  editor = 	 {Matni, Nikolai and Morari, Manfred and Pappas, George J.},
  volume = 	 {211},
  series = 	 {Proceedings of Machine Learning Research},
  month = 	 {15--16 Jun},
  publisher =    {PMLR},
  url = 	 {https://proceedings.mlr.press/v211/salam23a.html}
}

@article{biferale2019zermelo,
  title={Zermelo’s problem: optimal point-to-point navigation in {2D} turbulent flows using reinforcement learning},
  author={Biferale, Luca and Bonaccorso, Fabio and Buzzicotti, Michele and Clark Di Leoni, Patricio and Gustavsson, Kristian},
  journal={Chaos: An Interdisciplinary Journal of Nonlinear Science},
  volume={29},
  number={10},
  year={2019},
  publisher={AIP Publishing}
}

@INPROCEEDINGS{williams2016mppiintro,
  author={Williams, Grady and Drews, Paul and Goldfain, Brian and Rehg, James M. and Theodorou, Evangelos A.},
  booktitle={2016 IEEE International Conference on Robotics and Automation (ICRA)}, 
  title={Aggressive driving with model predictive path integral control}, 
  year={2016},
  volume={},
  number={},
  pages={1433-1440},
  doi={10.1109/ICRA.2016.7487277}}

@inproceedings{mohamed2020mppiquad,
  author={Mohamed, Ihab S. and Allibert, Guillaume and Martinet, Philippe},
  booktitle={2020 16th International Conference on Control, Automation, Robotics and Vision (ICARCV)}, 
  title={Model Predictive Path Integral Control Framework for Partially Observable Navigation: A Quadrotor Case Study}, 
  year={2020},
  volume={},
  number={},
  pages={196-203},
  doi={10.1109/ICARCV50220.2020.9305363}}

@article{rienecker2023windgraphplanning,
    author={Rienecker, Hannes and Hildebrand, Veit and Pfifer, Harald},
    year={2023},
    title={Energy optimal {3D} flight path planning for unmanned aerial vehicles in urban environments},
    journal={CAES Aeronautical Journal},
    volume={14},
    issue={3},
    pages={621-636},
    doi={10.1007/s13272-023-00666-x},
    eprint={https://doi.org/10.1007/s13272-023-00666-x}
}

@inproceedings{noca2019windshapeintro,
author = {Flavio Noca and Guillaume Catry and Nicolas Bosson and Luca Bardazzi and Sergio Marquez and Albéric Gros},
title = {Wind and Weather Facility for Testing Free-Flying Drones},
booktitle = {AIAA Aviation 2019 Forum},
publisher = {AIAA},
chapter = {},
year={2019},
pages = {},
doi = {10.2514/6.2019-2861},
URL = {https://arc.aiaa.org/doi/abs/10.2514/6.2019-2861},
eprint = {https://arc.aiaa.org/doi/pdf/10.2514/6.2019-2861}
}

@inproceedings{putzu2020aeroacoustic,
  title={Aeroacoustic measurements on a free-flying drone in a WindShaper wind tunnel},
  author={Putzu, Roberto and Boulandet, Romain and Rutschmann, Benjamin and Bujard, Thierry and Noca, Flavio and Guillaume, Catry and Bosson, Nicolas},
  booktitle={Proceedings of Quiet drones 2020-International e-Symposium on Noise of {UAV} and {UAS}, 19-21st October 2020, Paris, France},
  year={2020},
  organization={19-21st October 2020}
}

@inproceedings{catry2021windshapeicing,
author = {Guillaume Catry and Ozlem Ceyhan and Flavio Noca and Nicolas Bosson and Luca J. Bardazzi and Sergio Marquez and Pieter Jan Jordaens and Daniele Brandolisio},
title = {Performance Analysis of Rotorcraft Propulsion Units in a Combination of Wind and Icing Conditions},
booktitle = {AIAA AVIATION 2021 FORUM},
publisher = {AIAA},
chapter = {},
pages = {},
year = {2021},
doi = {10.2514/6.2021-2677},
URL = {https://arc.aiaa.org/doi/abs/10.2514/6.2021-2677},
eprint = {https://arc.aiaa.org/doi/pdf/10.2514/6.2021-2677}
}

@inproceedings{olejnik2022powerconsumption,
  title={An experimental study of wind resistance and power consumption in mavs with a low-speed multi-fan wind system},
  author={Olejnik, Diana A and Wang, Sunyi and Dupeyroux, Julien and Stroobants, Stein and Karasek, Matej and De Wagter, Christophe and de Croon, Guido},
  booktitle={2022 International Conference on Robotics and Automation (ICRA)},
  pages={2989--2994},
  year={2022},
  organization={IEEE}
}

@inproceedings{walpen2023windshapereplication,
  title={Real-scale atmospheric wind and turbulence replication using a fan-array for environmental testing and {UAV}/{AAM} validation},
  author={Walpen, Aur{\'e}lien and Catry, Guillaume and Noca, Flavio},
  booktitle={AIAA SCITECH 2023 Forum},
  pages={0812},
  year={2023}
}

@inproceedings{walpen2024windshapeautomation,
author = {Aurélien Walpen and Tony Govoni and Jonas Stirnemann and Matei Ionescu and Nicolas Bosson and Guillaume Catry and Flavio Noca},
title = {Automated control of complex aerodynamic flows generated by Windshaper fan arrays},
booktitle = {AIAA SCITECH 2024 Forum},
publisher={AIAA},
chapter = {},
pages = {},
year={2024},
doi = {10.2514/6.2024-2675},
URL = {https://arc.aiaa.org/doi/abs/10.2514/6.2024-2675},
eprint = {https://arc.aiaa.org/doi/pdf/10.2514/6.2024-2675}
}

@article{hravard2020ingenuity,
author = {Grip, H\r{a}vard Fj\ae{}r and Johnson, Wayne and Malpica, Carlos and Scharf, Daniel P. and Mandi\'{c}, Milan and Young, Larry and Allan, Brian and Mettler, B\'{e}r\'{e}nice and Martin, Miguel San and Lam, Johnny},
title = {Modeling and Identification of Hover Flight Dynamics for {NASA}’s {Mars} Helicopter},
journal = {Journal of Guidance, Control, and Dynamics},
volume = {43},
number = {2},
pages = {179-194},
year = {2020},
doi = {10.2514/1.G004228},
URL = { 
    
        https://doi.org/10.2514/1.G004228
},
eprint = { 
    
        https://doi.org/10.2514/1.G004228
}
}

@book{johnson2012helicopter,
  title={Helicopter theory},
  author={Johnson, Wayne},
  year={2012},
  publisher={Courier Corporation}
}

@article{mikhailuta2017urbanwindsstudy,
title = {Urban wind fields: Phenomena in transformation},
journal = {Urban Climate},
volume = {19},
pages = {122-140},
year = {2017},
issn = {2212-0955},
doi = {https://doi.org/10.1016/j.uclim.2016.12.005},
url = {https://www.sciencedirect.com/science/article/pii/S2212095516300621},
author = {Sergey V. Mikhailuta and Anatoly A. Lezhenin and Anne Pitt and Olga V. Taseiko}
}

@article{plate1999windphysicalmodels,
title = {Methods of investigating urban wind fields—physical models},
journal = {Atmospheric Environment},
volume = {33},
number = {24},
pages = {3981-3989},
year = {1999},
issn = {1352-2310},
doi = {https://doi.org/10.1016/S1352-2310(99)00140-5},
url = {https://www.sciencedirect.com/science/article/pii/S1352231099001405},
author = {Erich J. Plate}
}

@article{wang1996windtunnelscaling,
title = {Scale effects in wind tunnel modelling},
journal = {Journal of Wind Engineering and Industrial Aerodynamics},
volume = {61},
number = {2},
pages = {113-130},
year = {1996},
issn = {0167-6105},
doi = {https://doi.org/10.1016/0167-6105(96)00049-9},
url = {https://www.sciencedirect.com/science/article/pii/0167610596000499},
author = {Zhao Yin Wang and Erich J. Plate and Matthias Rau and Rolf Keiser}
}

@article{murakami1992les,
title = {Numerical study on velocity-pressure field and wind forces for bluff bodies by $\kappa$-$\epsilon$, {ASM} and {LES}},
journal = {Journal of Wind Engineering and Industrial Aerodynamics},
volume = {44},
number = {1},
pages = {2841-2852},
year = {1992},
note = {Special Issue 8th International Conference on Wind Engineering 1991},
issn = {0167-6105},
doi = {https://doi.org/10.1016/0167-6105(92)90079-P},
url = {https://www.sciencedirect.com/science/article/pii/016761059290079P},
author = {S. Murakami and A. Mochida and Y. Hayashi and S. Sakamoto}
}

@article{lobrano2011fieldmodeling,
title = {Quality of wind speed fitting distributions for the urban area of {Palermo}, {Italy}},
journal = {Renewable Energy},
volume = {36},
number = {3},
pages = {1026-1039},
year = {2011},
issn = {0960-1481},
doi = {https://doi.org/10.1016/j.renene.2010.09.009},
url = {https://www.sciencedirect.com/science/article/pii/S0960148110004210},
author = {Valerio {Lo Brano} and Aldo Orioli and Giuseppina Ciulla and Simona Culotta}
}

@article{gadian2004directionalpersistence,
title = {Directional persistence of low wind speed observations},
journal = {Journal of Wind Engineering and Industrial Aerodynamics},
volume = {92},
number = {12},
pages = {1061-1074},
year = {2004},
issn = {0167-6105},
doi = {https://doi.org/10.1016/j.jweia.2004.05.007},
url = {https://www.sciencedirect.com/science/article/pii/S0167610504000947},
author = {A. Gadian and J. Dewsbury and F. Featherstone and J. Levermore and K. Morris and C. Sanders}
}

@INPROCEEDINGS{simon2023flowdrone,
  author={Simon, Nathaniel and Ren, Allen Z. and Piqué, Alexander and Snyder, David and Barretto, Daphne and Hultmark, Marcus and Majumdar, Anirudha},
  booktitle={2023 IEEE International Conference on Robotics and Automation (ICRA)}, 
  title={FlowDrone: Wind Estimation and Gust Rejection on {UAVs} Using Fast-Response Hot-Wire Flow Sensors}, 
  year={2023},
  volume={},
  number={},
  pages={5393-5399},
  doi={10.1109/ICRA48891.2023.10160454},
  ISSN={},
  month={May},}

@inproceedings{kularatne2016time,
  title={Time and Energy Optimal Path Planning in General Flows.},
  author={Kularatne, Dhanushka and Bhattacharya, Subhrajit and Hsieh, M Ani},
  booktitle={Robotics: Science and Systems},
  pages={1-10},
  year={2016},
}

@article{karydis2017energetics,
  title={Energetics in robotic flight at small scales},
  author={Karydis, Konstantinos and Kumar, Vijay},
  journal={Interface focus},
  volume={7},
  number={1},
  pages={20160088},
  year={2017},
  publisher={The Royal Society}
}

@inproceedings{optuna2019,
    title={Optuna: A Next-generation Hyperparameter Optimization Framework},
    author={Akiba, Takuya and Sano, Shotaro and Yanase, Toshihiko and Ohta, Takeru and Koyama, Masanori},
    booktitle={Proceedings of the 25th {ACM} {SIGKDD} International Conference on Knowledge Discovery and Data Mining},
    year={2019}
}

@article{folk2024flowdecoding,
  author={Folk, Spencer and Melton, John and Margolis, Benjamin W. L. and Yim, Mark and Kumar, Vijay},
  journal={IEEE Robotics and Automation Letters}, 
  title={Learning Local Urban Wind Flow Fields From Range Sensing}, 
  year={2024},
  volume={9},
  number={9},
  pages={7413-7420},
  doi={10.1109/LRA.2024.3426209}}

@article{takemura2023energyperceptionawareplanning,
author = {Reiya Takemura and Nobuaki Aoki and Genya Ishigami},
title ={Energy-and-perception-aware planning and navigation framework for unmanned aerial vehicles},
journal = {Advances in Mechanical Engineering},
volume = {15},
number = {4},
pages = {16878132231169688},
year = {2023},
doi = {10.1177/16878132231169688},
URL = {  
        https://doi.org/10.1177/16878132231169688  
},
eprint = { 
        https://doi.org/10.1177/16878132231169688
}
}

@article{chan2023graphwindawarenavigation,
title = {Wind dynamic and energy-efficiency path planning for unmanned aerial vehicles in the lower-level airspace and urban air mobility context},
journal = {Sustainable Energy Technologies and Assessments},
volume = {57},
pages = {103202},
year = {2023},
issn = {2213-1388},
doi = {https://doi.org/10.1016/j.seta.2023.103202},
url = {https://www.sciencedirect.com/science/article/pii/S2213138823001959},
author = {Y.Y. Chan and Kam K.H. Ng and C.K.M. Lee and Li-Ta Hsu and K.L. Keung}
}

@article{aiello2022realtimeaware,
  title={Fixed-wing {UAV} energy efficient {3D} path planning in cluttered environments},
  author={Aiello, Giuseppe and Valavanis, Kimon P and Rizzo, Alessandro},
  journal={Journal of Intelligent \& Robotic Systems},
  volume={105},
  number={3},
  pages={60},
  year={2022},
  publisher={Springer}
}

@inproceedings{chakrabarty2010graphsearch,
author = {Anjan Chakrabarty and Jack Langelaan},
title = {Flight Path Planning for {UAV} Atmospheric Energy Harvesting Using Heuristic Search},
booktitle = {AIAA Guidance, Navigation, and Control Conference},
chapter = {},
pages = {},
publisher = {AIAA},
year={2010},
doi = {10.2514/6.2010-8033},
URL = {https://arc.aiaa.org/doi/abs/10.2514/6.2010-8033},
eprint = {https://arc.aiaa.org/doi/pdf/10.2514/6.2010-8033}
}

@article{hao2022bamdp,
  author={Xu, Hao and Pan, Jia},
  journal={IEEE Robotics and Automation Letters}, 
  title={{AUV} Motion Planning in Uncertain Flow Fields Using {Bayes} Adaptive MDPs}, 
  year={2022},
  volume={7},
  number={2},
  pages={5575-5582},
  doi={10.1109/LRA.2022.3157543}}

@inproceedings{yacef2020energytrajopt,
  author={Yacef, Fouad and Rizoug, Nassim and Degaa, Laid and Hamerlain, Mustapha},
  booktitle={2020 7th International Conference on Control, Decision and Information Technologies (CoDIT)}, 
  title={Energy-Efficiency Path Planning for Quadrotor {UAV} Under Wind Conditions}, 
  year={2020},
  volume={1},
  number={},
  pages={1133-1138},
  doi={10.1109/CoDIT49905.2020.9263968}}

@article{liu2024teevtol,
  title={{TEeVTOL}: Balancing Energy and Time Efficiency in {eVTOL} Aircraft Path Planning Across City-Scale Wind Fields},
  author={Liu, Songyang and Li, Shuai and Li, Haochen and Li, Weizi and Tan, Jindong},
  journal={arXiv preprint arXiv:2403.14877},
  year={2024}
}

@inproceedings{banerjee2024rlenergy,
author = {Portia Banerjee and Kevin Bradner},
title = {Energy-Optimized Path Planning for {UAS} in Varying Winds Via Reinforcement Learning},
booktitle = {AIAA AVIATION FORUM AND ASCEND 2024},
chapter = {},
pages = {},
publisher = {AIAA},
year = {2024},
doi = {10.2514/6.2024-4545},
URL = {https://arc.aiaa.org/doi/abs/10.2514/6.2024-4545},
eprint = {https://arc.aiaa.org/doi/pdf/10.2514/6.2024-4545},
}

@article{williams2017mppiref,
author = {Williams, Grady and Aldrich, Andrew and Theodorou, Evangelos A.},
title = {Model Predictive Path Integral Control: From Theory to Parallel Computation},
journal = {Journal of Guidance, Control, and Dynamics},
volume = {40},
number = {2},
pages = {344-357},
year = {2017},
doi = {10.2514/1.G001921},
URL = {https://doi.org/10.2514/1.G001921},
eprint = { 
        https://doi.org/10.2514/1.G001921}}

@inproceedings{higgins2023mppiquadrotor,
  title={A Model Predictive Path Integral Method for Fast, Proactive, and Uncertainty-Aware {UAV} Planning in Cluttered Environments},
  author={Higgins, Jacob and Mohammad, Nicholas and Bezzo, Nicola},
  booktitle={2023 IEEE/RSJ International Conference on Intelligent Robots and Systems (IROS)},
  pages={830--837},
  year={2023},
  organization={IEEE}
}

@INPROCEEDINGS{folk2025windplanning,
  author={Folk, Spencer and Melton, John and Margolis, Benjamin W. L. and Yim, Mark and Kumar, Vijay},
  booktitle={2025 IEEE International Conference on Robotics and Automation (ICRA)}, 
  title={Towards Safe and Energy-Efficient Real-Time Motion Planning in Windy Urban Environments}, 
  year={2025},
  volume={},
  number={},
  pages={2787-2793},
  doi={10.1109/ICRA55743.2025.11127986}}

@inproceedings{panerati2021pybullet,
  title={Learning to fly—a gym environment with {PyBullet} physics for reinforcement learning of multi-agent quadcopter control},
  author={Panerati, Jacopo and Zheng, Hehui and Zhou, SiQi and Xu, James and Prorok, Amanda and Schoellig, Angela P},
  booktitle={2021 IEEE/RSJ International Conference on Intelligent Robots and Systems (IROS)},
  pages={7512--7519},
  year={2021},
  organization={IEEE}
}

@article{bangura2012nonlinear,
  title={Nonlinear dynamic modeling for high performance control of a quadrotor},
  author={Bangura, Moses and Mahony, Robert and others},
  year={2012},
  journal={Australian Robotics and Automation Association}
}

@article{svacha2020inertia,
  author={Svacha, James and Paulos, James and Loianno, Giuseppe and Kumar, Vijay},
  journal={IEEE Robotics and Automation Letters}, 
  title={{IMU}-Based Inertia Estimation for a Quadrotor Using Newton-Euler Dynamics}, 
  year={2020},
  volume={5},
  number={3},
  pages={3861-3867},
  doi={10.1109/LRA.2020.2976308}}

@phdthesis{svacha2019imu,
  title={IMU-based state estimation and control of quadrotors exploiting aerodynamic effects},
  author={Svacha, James},
  year={2019},
  school={University of Pennsylvania}
}

@article{allen1946flapping,
author = {Allen, Ralph W.},
title = {Flapping Characteristics of Rigid Rotor Blades},
journal = {Journal of the Aeronautical Sciences},
volume = {13},
number = {4},
pages = {183-186},
year = {1946},
doi = {10.2514/8.11343},
URL = {https://doi.org/10.2514/8.11343},
eprint = {https://doi.org/10.2514/8.11343}
}

@INPROCEEDINGS{martin2010feedback,
  author={Martin, Philippe and Salaün, Erwan},
  booktitle={2010 IEEE International Conference on Robotics and Automation}, 
  title={The true role of accelerometer feedback in quadrotor control}, 
  year={2010},
  volume={},
  number={},
  pages={1623-1629},
  doi={10.1109/ROBOT.2010.5509980}
}

@article{faessler2018rotordragdiffflat,
  author={Faessler, Matthias and Franchi, Antonio and Scaramuzza, Davide},
  journal={IEEE Robotics and Automation Letters}, 
  title={Differential Flatness of Quadrotor Dynamics Subject to Rotor Drag for Accurate Tracking of High-Speed Trajectories}, 
  year={2018},
  volume={3},
  number={2},
  pages={620-626},
  doi={10.1109/LRA.2017.2776353},
  ISSN={2377-3766},
  month={April},}

@inproceedings{bristeau2009rotordrag,
  author={Bristeau, Pierre-Jean and Martin, Philippe and Salaün, Erwan and Petit, Nicolas},
  booktitle={2009 European Control Conference (ECC)}, 
  title={The role of propeller aerodynamics in the model of a quadrotor {UAV}}, 
  year={2009},
  volume={},
  number={},
  pages={683-688},
  doi={10.23919/ECC.2009.7074482}}

@inproceedings{loianno2016ufkonmanifold,
  author={Loianno, Giuseppe and Watterson, Michael and Kumar, Vijay},
  booktitle={2016 IEEE International Conference on Robotics and Automation (ICRA)}, 
  title={Visual inertial odometry for quadrotors on {SE(3)}}, 
  year={2016},
  volume={},
  number={},
  pages={1544-1551},
  doi={10.1109/ICRA.2016.7487292}}

@inproceedings{mulgaonkar2014power,
  title={Power and weight considerations in small, agile quadrotors},
  author={Mulgaonkar, Yash and Whitzer, Michael and Morgan, Brian and Kroninger, Christopher M and Harrington, Aaron M and Kumar, Vijay},
  booktitle={Micro-and Nanotechnology Sensors, Systems, and Applications VI},
  volume={9083},
  pages={376--391},
  year={2014},
  organization={SPIE}
}

@inproceedings{thomas2025whisker,
  author={Thomas, Lenworth and Bridges, Tjaden and Bergbreiter, Sarah},
  booktitle={2025 IEEE International Conference on Robotics and Automation (ICRA)}, 
  title={Airflow Source Seeking on Small Quadrotors Using a Single Flow Sensor}, 
  year={2025},
  volume={},
  number={},
  pages={12488-12494},
  doi={10.1109/ICRA55743.2025.11128722}}

@INPROCEEDINGS{kim2020whiskerprecursor,
  author={Kim, Suhan and Kubicek, Regan and Paris, Aleix and Tagliabue, Andrea and How, Jonathan P. and Bergbreiter, Sarah},
  booktitle={2020 IEEE/RSJ International Conference on Intelligent Robots and Systems (IROS)}, 
  title={A Whisker-inspired Fin Sensor for Multi-directional Airflow Sensing}, 
  year={2020},
  volume={},
  number={},
  pages={1330-1337},
  doi={10.1109/IROS45743.2020.9341723},
  ISSN={2153-0866},
  month={Oct},}

@mastersthesis{neely2015hummingbirdparams,
    author={William Neely},
    title={Design and Development of a High-Performance Quadrotor Control Architecture Based on Feedback Linearization},
    school={The University of New Mexico},
    year={2015}
}

@inproceedings{julier1997ukf,
  title={New extension of the {Kalman} filter to nonlinear systems},
  author={Julier, Simon J and Uhlmann, Jeffrey K},
  booktitle={Signal processing, sensor fusion, and target recognition VI},
  volume={3068},
  pages={182--193},
  year={1997},
  organization={Spie}
}

@INPROCEEDINGS{wan2000ukf,
  author={Wan, E.A. and Van Der Merwe, R.},
  booktitle={Proceedings of the IEEE 2000 Adaptive Systems for Signal Processing, Communications, and Control Symposium (Cat. No.00EX373)}, 
  title={The unscented {Kalman} filter for nonlinear estimation}, 
  year={2000},
  volume={},
  number={},
  pages={153-158},
  doi={10.1109/ASSPCC.2000.882463},
  ISSN={},
  month={Oct},}

@article{hermann2003nonlinearobservabilityanalysis,
  author={Hermann, R. and Krener, A.},
  journal={IEEE Transactions on Automatic Control}, 
  title={Nonlinear controllability and observability}, 
  year={1977},
  volume={22},
  number={5},
  pages={728-740},
  doi={10.1109/TAC.1977.1101601}}

@inproceedings{frey2024windtunnelurbanwind,
author = {Juergen Frey and Hannes Rienecker and Sebastian Schubert and Veit Hildebrand and Harald Pfifer},
title = {Wind Tunnel Measurement of the Urban Wind Field for Flight Path Planning of Unmanned Aerial Vehicles},
booktitle = {AIAA SCITECH 2024 Forum},
publisher = {AIAA},
year = {2024},
chapter = {},
pages = {},
doi = {10.2514/6.2024-2510},
URL = {https://arc.aiaa.org/doi/abs/10.2514/6.2024-2510},
eprint = {https://arc.aiaa.org/doi/pdf/10.2514/6.2024-2510}
}

@inproceedings{prudden2016flyinganemometer,
  title={A flying anemometer quadrotor: Part 1},
  author={Prudden, Samuel and Fisher, A and Mohamed, A and Watkins, S},
  booktitle={Proceedings of the International Micro Air Vehicle Conference (IMAV 2016), Beijing, China},
  pages={17--21},
  year={2016}
}

@inproceedings{foster2020freeflightuav,
  title={Recent {NASA} wind tunnel free-flight testing of a multirotor unmanned aircraft system},
  author={Foster, John V and Miller, Luke J and Busan, Ronald C and Langston, Sarah and Hartman, David},
  booktitle={AIAA Scitech 2020 Forum},
  pages={1504},
  year={2020}
}

@inproceedings{shivgan2020geneticpathplanning,
  author={Shivgan, Rutuja and Dong, Ziqian},
  booktitle={2020 IEEE 21st International Conference on High Performance Switching and Routing (HPSR)}, 
  title={Energy-Efficient Drone Coverage Path Planning using Genetic Algorithm}, 
  year={2020},
  volume={},
  number={},
  pages={1-6},
  doi={10.1109/HPSR48589.2020.9098989}}

@inproceedings{ding2018morepowermodels,
  author={Ding, Lige and Zhao, Dong and Ma, Huadong and Wang, Hao and Liu, Liang},
  booktitle={2018 IEEE 24th International Conference on Parallel and Distributed Systems (ICPADS)}, 
  title={Energy-Efficient Min-Max Planning of Heterogeneous Tasks with Multiple {UAVs}}, 
  year={2018},
  volume={},
  number={},
  pages={339-346},
  doi={10.1109/PADSW.2018.8644625},
  ISSN={1521-9097},
  month={Dec},}

@article{herz2025powervsattitude,
  title={Effects of rotor--rotor and rotor--body interactions on quadrotor vehicle performance for multiple flight configurations},
  author={Herz, Sage and Atte, Abraham and Seth, Dhuree and Rauleder, Juergen and McCrink, Matthew},
  journal={Aerospace Science and Technology},
  volume={158},
  pages={109873},
  year={2025},
  publisher={Elsevier}
}

@article{palmas2022jetsonexample,
  title={Deep Learning computer vision algorithms for real-time {UAVs} on-board camera image processing},
  author={Palmas, Alessandro and Andronico, Pietro},
  journal={arXiv preprint arXiv:2211.01037},
  year={2022}
}

@article{ayoub2021jetsonexample,
  title={Real-time on-board deep learning fault detection for autonomous {UAV} inspections},
  author={Ayoub, Naeem and Schneider-Kamp, Peter},
  journal={Electronics},
  volume={10},
  number={9},
  pages={1091},
  year={2021},
  publisher={MDPI}
}

@article{moller1997raycasting,
author = {Tomas M{\"o}ller and Ben Trumbore},
title = {Fast, Minimum Storage Ray-Triangle Intersection},
journal = {Journal of Graphics Tools},
volume = {2},
number = {1},
pages = {21--28},
year = {1997},
publisher = {Taylor \& Francis},
doi = {10.1080/10867651.1997.10487468},
URL = {https://doi.org/10.1080/10867651.1997.10487468},
eprint = { https://doi.org/10.1080/10867651.1997.10487468}
}

@mastersthesis{forster2015system,
  title={System identification of the crazyflie 2.0 nano quadrocopter},
  author={F{\"o}rster, Julian},
  type={{B.S.} thesis},
  year={2015},
  school={ETH Zurich}
}

@inproceedings{tagliabue2019modelfreeenergy,
  author={Tagliabue, Andrea and Wu, Xiangyu and Mueller, Mark W.},
  booktitle={2019 International Conference on Robotics and Automation (ICRA)}, 
  title={Model-free Online Motion Adaptation for Optimal Range and Endurance of Multicopters}, 
  year={2019},
  volume={},
  number={},
  pages={5650-5656},
  doi={10.1109/ICRA.2019.8793708},
  ISSN={2577-087X},
  month={May},}

@article{wu2022modelfreeenergy,
  title={Model-free online motion adaptation for energy-efficient flight of multicopters},
  author={Wu, Xiangyu and Zeng, Jun and Tagliabue, Andrea and Mueller, Mark W},
  journal={IEEE Access},
  volume={10},
  pages={65507--65519},
  year={2022},
  publisher={IEEE}
}

@article{rayleigh1883dynamicsoaring,
  author  = {Rayleigh, Lord},
  title   = {{The Soaring of Birds}},
  journal = {Nature},
  volume  = {27},
  issue   = {701},
  pages   = {534--535},
  year    = {1883},
  month   = {April},
  day     = {5},
}

@article{tao2025halo,
      title={{HALO}: High-Altitude Language-Conditioned Monocular Aerial Exploration and Navigation}, 
      author={Yuezhan Tao and Dexter Ong and Fernando Cladera and Jason Hughes and Camillo J. Taylor and Pratik Chaudhari and Vijay Kumar},
      year={2025},
      eprint={2511.17497},
      archivePrefix={arXiv},
      primaryClass={cs.RO},
      journal={arXiv preprint arXiv:2312.15796},
      url={https://arxiv.org/abs/2511.17497}, 
}

@inproceedings{baskar2020planning,
author = {Deepika Baskar and Alex Gorodetsky},
title = {A Simulated Wind-field Dataset for Testing Energy Efficient Path-Planning Algorithms for {UAVs} in Urban Environment},
booktitle = {AIAA AVIATION 2020 FORUM},
chapter = {},
pages = {},
publisher = {AIAA},
year = {2020},
doi = {10.2514/6.2020-2920},
URL = {https://arc.aiaa.org/doi/abs/10.2514/6.2020-2920},
eprint = {https://arc.aiaa.org/doi/pdf/10.2514/6.2020-2920}
}

@phdthesis{raza2015uavcontrolinwind,
  title={Autonomous {UAV} control for low-altitude flight in an urban gust environment},
  author={Raza, Syed Ali},
  year={2015},
  school={Carleton University}
}

@article{sutherland2016urbanquadrotorflightsim,
author = {Sutherland, M. and Etele, J. and Fusina, G.},
title = {Urban Wake-Field Generation Using Large-Eddy Simulation for Application to Quadrotor Flight},
journal = {Journal of Aircraft},
volume = {53},
number = {5},
pages = {1224-1236},
year = {2016},
doi = {10.2514/1.C033624},
URL = {https://doi.org/10.2514/1.C033624},
eprint = {https://doi.org/10.2514/1.C033624}
}

@inproceedings{murray2014responseinenvironment,
  title={On the response of an autonomous quadrotor operating in a turbulent urban environment},
  author={Murray, Craig WA and Ireland, Murray and Anderson, David},
  booktitle={AUVSI’s Unmanned Systems Conference},
  year={2014}
}

@article{kristner2024windtunnel,
author = {Kistner, J. and Neuhaus, L. and Wildmann, N.},
title = {High-resolution wind speed measurements with quadcopter uncrewed aerial systems: calibration and verification in a wind tunnel with an active grid},
journal = {Atmospheric Measurement Techniques},
volume = {17},
year = {2024},
number = {16},
pages = {4941--4955},
url = {https://amt.copernicus.org/articles/17/4941/2024/},
doi = {10.5194/amt-17-4941-2024}
}

@article{mohamed2016freeflight,
author = {Mohamed, A. and Abdulrahim, M. and Watkins, S. and Clothier, R.},
title = {Development and Flight Testing of a Turbulence Mitigation System for Micro Air Vehicles},
journal = {Journal of Field Robotics},
volume = {33},
number = {5},
pages = {639-660},
doi = {https://doi.org/10.1002/rob.21626},
url = {https://onlinelibrary.wiley.com/doi/abs/10.1002/rob.21626},
eprint = {https://onlinelibrary.wiley.com/doi/pdf/10.1002/rob.21626},
year = {2016}
}

@inproceedings{bannwarth2018freeflight,
  author={Bannwarth, J. X. J. and Chen, Z. J. and Stol, K. A. and MacDonald, B. A. and Richards, P. J.},
  booktitle={2018 IEEE/ASME International Conference on Advanced Intelligent Mechatronics (AIM)}, 
  title={Development of a Wind Tunnel Experimental Setup for Testing Multirotor Unmanned Aerial Vehicles in Turbulent Conditions}, 
  year={2018},
  volume={},
  number={},
  pages={724-729},
  doi={10.1109/AIM.2018.8452316},
  ISSN={2159-6255},
  month={July},}

@inproceedings{kubo2018freeflight,
author = {Daisuke Kubo},
title = {Gust Response Evaluation of small UAS via Free-Flight in Gust Wind Tunnel},
booktitle = {2018 AIAA Atmospheric Flight Mechanics Conference},
publisher = {AIAA},
year = {2018},
chapter = {},
pages = {},
doi = {10.2514/6.2018-0297},
URL = {https://arc.aiaa.org/doi/abs/10.2514/6.2018-0297},
eprint = {https://arc.aiaa.org/doi/pdf/10.2514/6.2018-0297}
}

@article{price2023gencast,
  title={Gencast: Diffusion-based ensemble forecasting for medium-range weather},
  author={Price, Ilan and Sanchez-Gonzalez, Alvaro and Alet, Ferran and Andersson, Tom R and El-Kadi, Andrew and Masters, Dominic and Ewalds, Timo and Stott, Jacklynn and Mohamed, Shakir and Battaglia, Peter and others},
  journal={arXiv preprint arXiv:2312.15796},
  year={2023}
}

@book{snyder1981guideline,
  title={Guideline for fluid modeling of atmospheric diffusion},
  author={Snyder, William H},
  volume={81},
  number={9},
  year={1981},
  publisher={Environmental Sciences Research Laboratory, Office of Research and Development, US Environmental Protection Agency}
}

@techreport{wolowicz1979similitude,
  title       = {Similitude requirements and scaling relationships as applied to model testing},
  author      = {Wolowicz, Chester H. and Brown, James S. and Gilbert, William P.},
  institution = {NASA Langley Research Center},
  year        = {1979},
  number      = {NASA-TP-1435},
  address     = {Hampton, VA, United States},
  url         = {https://ntrs.nasa.gov/citations/19790022005},
  type        = {Technical Publication}
}

@article{kushleyev2013towards,
  title={Towards a swarm of agile micro quadrotors},
  author={Kushleyev, Alex and Mellinger, Daniel and Powers, Caitlin and Kumar, Vijay},
  journal={Autonomous Robots},
  volume={35},
  number={4},
  pages={287--300},
  year={2013},
  publisher={Springer}
}

@inproceedings{kolmogorov1941equations,
  title={Equations of turbulent motion in an incompressible fluid},
  author={Kolmogorov, Andrej Nikolaevich},
  booktitle={Dokl. Akad. Nauk SSSR},
  volume={30},
  pages={299--303},
  year={1941}
}

@techreport{chambers2010modeling,
  title        = {Modeling Flight: The Role of Dynamically Scaled Free-Flight Models in Support of {NASA}'s Aerospace Programs},
  author       = {Chambers, Joseph R.},
  year         = 2010,
  institution  = {National Aeronautics and Space Administration},
  address      = {Washington, DC},
  number       = {NASA SP-2009-575},
  url          = {https://ntrs.nasa.gov/citations/20110012492}
}

@inproceedings{simon2023mononav,
  title = {{MonoNav: MAV Navigation via Monocular Depth Estimation and Reconstruction}},
  author = {Simon, Nathaniel and Majumdar, Anirudha},
  booktitle = {International Symposium on Experimental Robotics (ISER)},
  year = {2023}
}

@article{zhu2025reynoldsindependence,
title = {Flows across {3D} urban street canyons: Revisiting {Reynolds} number independence by large eddy simulation},
journal = {Building and Environment},
volume = {285},
pages = {113635},
year = {2025},
issn = {0360-1323},
doi = {https://doi.org/10.1016/j.buildenv.2025.113635},
url = {https://www.sciencedirect.com/science/article/pii/S0360132325011060},
author = {Sirui Zhu and Lup Wai Chew}
}

@article{uehara2003reynoldsindependence,
title = {Studies on critical {Reynolds} number indices for wind-tunnel experiments on flow within urban areas},
journal = {Boundary-Layer Meteorology},
volume = {107},
pages = {353-370},
year = {2003},
doi = {https://doi.org/10.1023/A:1022162807729},
author = {Kiyoshi Uehara and Shinji Wakamatsu and Ryozo Ooka},
}

@article{cui2014reynoldsindependence,
title = {Investigation of {Re}-independence of turbulent flow and pollutant dispersion in urban street canyon using numerical wind tunnel (NWT) models},
journal = {International Journal of Heat and Mass Transfer},
volume = {79},
pages = {176-188},
year = {2014},
issn = {0017-9310},
doi = {https://doi.org/10.1016/j.ijheatmasstransfer.2014.07.096},
url = {https://www.sciencedirect.com/science/article/pii/S0017931014006851},
author = {Peng-Yi Cui and Zhuo Li and Wen-Quan Tao}
}

@article{chew2018reynoldsindependence,
  author  = {Chew, Lup Wai and Aliabadi, Amir A. and Norford, Leslie K.},
  title   = {Flows across high aspect ratio street canyons: {Reynolds} number independence revisited},
  journal = {Environmental Fluid Mechanics},
  year    = {2018},
  volume  = {18},
  number  = {5},
  pages   = {1275--1291},
  month   = {Oct},
  doi     = {10.1007/s10652-018-9601-0},
  url     = {https://doi.org/10.1007/s10652-018-9601-0},
}

@article{shu2020reynoldsindependence,
title = {Dimensional analysis of {Reynolds} independence and regional critical {Reynolds} numbers for urban aerodynamics},
journal = {Journal of Wind Engineering and Industrial Aerodynamics},
volume = {203},
pages = {104232},
year = {2020},
issn = {0167-6105},
doi = {https://doi.org/10.1016/j.jweia.2020.104232},
url = {https://www.sciencedirect.com/science/article/pii/S0167610520301422},
author = {Chang Shu and Liangzhu (Leon) Wang and Mohammad Mortezazadeh}
}

@book{townsend1976structure,
  title={The structure of turbulent shear flow},
  author={Townsend, A.A.R.},
  year={1976},
  publisher={Cambridge University Press}
}

@article{nazarian2025urbantales,
  title={Urbantales: A comprehensive {LES} dataset for urban canopy layer turbulence analyses and parameterization development},
  author={Nazarian, Negin and Lu, Jiachen and Lipson, Mathew and Liu, Sijie and Hart, Melissa Anne and Krayenhoff, Scott and Blunn, Lewis and Martilli, Alberto},
  year={2025},
  publisher={EarthArXiv},
  journal={EarthArXiv},
}

@article{oke1988streetcanopy,
title = {Street design and urban canopy layer climate},
journal = {Energy and Buildings},
volume = {11},
number = {1},
pages = {103-113},
year = {1988},
issn = {0378-7788},
doi = {https://doi.org/10.1016/0378-7788(88)90026-6},
url = {https://www.sciencedirect.com/science/article/pii/0378778888900266},
author = {T.R. Oke}
}

@article{martinuzzi1993pristmaticfluidstudy,
    author = {Martinuzzi, R. and Tropea, C.},
    title = {The Flow Around Surface-Mounted, Prismatic Obstacles Placed in a Fully Developed Channel Flow (Data Bank Contribution)},
    journal = {Journal of Fluids Engineering},
    volume = {115},
    number = {1},
    pages = {85-92},
    year = {1993},
    month = {03},
    issn = {0098-2202},
    doi = {10.1115/1.2910118},
    url = {https://doi.org/10.1115/1.2910118},
    eprint = {https://asmedigitalcollection.asme.org/fluidsengineering/article-pdf/115/1/85/5566737/85_1.pdf},
}

@article{li2024enkode,
  author={Li, Alice K. and Silva, Thales C. and Hsieh, M. Ani},
  journal={IEEE Robotics and Automation Letters}, 
  title={{EnKode}: Active Learning of Unknown Flows With {Koopman} Operators}, 
  year={2024},
  volume={9},
  number={12},
  pages={11282-11289},
  doi={10.1109/LRA.2024.3486217}}

@article{raissi2019pinns,
title = {Physics-informed neural networks: A deep learning framework for solving forward and inverse problems involving nonlinear partial differential equations},
journal = {Journal of Computational Physics},
volume = {378},
pages = {686-707},
year = {2019},
issn = {0021-9991},
doi = {https://doi.org/10.1016/j.jcp.2018.10.045},
url = {https://www.sciencedirect.com/science/article/pii/S0021999118307125},
author = {M. Raissi and P. Perdikaris and G.E. Karniadakis}
}

@inproceedings{lee2019gaussianprocessoceans,
  author={Lee, Ki Myung Brian and Yoo, Chanyeol and Hollings, Ben and Anstee, Stuart and Huang, Shoudong and Fitch, Robert},
  booktitle={2019 International Conference on Robotics and Automation (ICRA)}, 
  title={Online Estimation of Ocean Current from Sparse {GPS} Data for Underwater Vehicles}, 
  year={2019},
  volume={},
  number={},
  pages={3443-3449},
  doi={10.1109/ICRA.2019.8794308}
}

@inproceedings{berlinghieri2023gaussianprocessoceans,
  author    = {Berlinghieri, Renato and Trippe, Brian L. and Burt, David R. and Giordano, Ryan and Srinivasan, Kaushik and {\"O}zg{\"o}kmen, Tamay and Xia, Junfei and Broderick, Tamara},
  title     = {Gaussian Processes at the {H}elm(holtz): A More Fluid Model for Ocean Currents},
  booktitle = {Proceedings of the 40th International Conference on Machine Learning},
  year      = {2023},
  volume    = {202},
  pages     = {2113--2163},
  publisher = {PMLR},
  doi       = {10.48550/arXiv.2302.10364}
}

@article{goncalves2019gaussianprocessoceans,
      author = "Rafael C. Gonçalves and Mohamed Iskandarani and Tamay Özgökmen and W. Carlisle Thacker",
      title = "Reconstruction of Submesoscale Velocity Field from Surface Drifters",
      journal = "Journal of Physical Oceanography",
      year = "2019",
      publisher = "American Meteorological Society",
      address = "Boston MA, USA",
      volume = "49",
      number = "4",
      doi = "10.1175/JPO-D-18-0025.1",
      pages=      "941 - 958",
      url = "https://journals.ametsoc.org/view/journals/phoc/49/4/jpo-d-18-0025.1.xml"
}

@inproceedings{hansen2018gaussianprocessesoceans,
  author={Hansen, Johanna and Dudek, Gregory},
  booktitle={2018 IEEE/RSJ International Conference on Intelligent Robots and Systems (IROS)}, 
  title={Coverage Optimization with Non-Actuated, Floating Mobile Sensors using Iterative Trajectory Planning in Marine Flow Fields}, 
  year={2018},
  volume={},
  number={},
  pages={1906-1912},
  doi={10.1109/IROS.2018.8594281}}

@article{mezic2013koopmanforfluidflows,
   author = "Mezić, Igor",
   title = "Analysis of Fluid Flows via Spectral Properties of the {Koopman} Operator", 
   journal= "Annual Review of Fluid Mechanics",
   year = "2013",
   volume = "45",
   number = "Volume 45, 2013",
   pages = "357-378",
   doi = "https://doi.org/10.1146/annurev-fluid-011212-140652",
   url = "https://www.annualreviews.org/content/journals/10.1146/annurev-fluid-011212-140652",
   publisher = "Annual Reviews",
   issn = "1545-4479",
   type = "Journal Article",
  }

@article{koopman1931koopmantheory,
author = {B. O. Koopman },
title = {Hamiltonian Systems and Transformation in Hilbert Space},
journal = {Proceedings of the National Academy of Sciences},
volume = {17},
number = {5},
pages = {315-318},
year = {1931},
doi = {10.1073/pnas.17.5.315},
URL = {https://www.pnas.org/doi/abs/10.1073/pnas.17.5.315},
eprint = {https://www.pnas.org/doi/pdf/10.1073/pnas.17.5.315}
}

@INPROCEEDINGS{ames2019controlbarrier,
  author={Ames, Aaron D. and Coogan, Samuel and Egerstedt, Magnus and Notomista, Gennaro and Sreenath, Koushil and Tabuada, Paulo},
  booktitle={2019 18th European Control Conference (ECC)}, 
  title={Control Barrier Functions: Theory and Applications}, 
  year={2019},
  volume={},
  number={},
  pages={3420-3431},
  doi={10.23919/ECC.2019.8796030}}
\end{bibliof}
\end{document}